\documentclass[journal]{IEEEtran}
\usepackage{cite}
\ifCLASSINFOpdf
   \usepackage{subfig}
   \usepackage[pdftex]{graphicx}
\else
   \usepackage[dvips]{graphicx}
\fi
\usepackage{amsmath}
\usepackage{array}

\usepackage{xspace}
\makeatletter
\newcommand\onedot{\futurelet\@let@token\@onedot}
\def\@onedot{\ifx\@let@token.\else.\null\fi\xspace}
\def\eg{\emph{e.g}\onedot} 
\def\ie{\emph{i.e}\onedot} 
 
\def\etal{\emph{et al}\onedot}
\makeatother

\usepackage{color}
\usepackage{amsmath}
\usepackage{amssymb}
\usepackage{mathrsfs}
\usepackage{adjustbox}
\usepackage[cal=boondox,scr=boondoxo]{mathalfa}
\usepackage[capitalize]{cleveref}
\usepackage{pifont}% http://ctan.org/pkg/pifont
\newcommand{\cmark}{\ding{51}}%
\newcommand{\xmark}{\ding{55}}%
\usepackage[utf8x]{inputenc}

\begin{document}
% paper title
% Titles are generally capitalized except for words such as a, an, and, as,
% at, but, by, for, in, nor, of, on, or, the, to and up, which are usually
% not capitalized unless they are the first or last word of the title.
% Linebreaks \\ can be used within to get better formatting as desired.
% Do not put math or special symbols in the title.
\title{Non-Local Representation based Mutual Affine-Transfer Network for Photorealistic Stylization}
%
%
% author names and IEEE memberships
% note positions of commas and nonbreaking spaces ( ~ ) LaTeX will not break
% a structure at a ~ so this keeps an author's name from being broken across
% two lines.
% use \thanks{} to gain access to the first footnote area
% a separate \thanks must be used for each paragraph as LaTeX2e's \thanks
% was not built to handle multiple paragraphs
%
%\author{Ying Qu\textsuperscript{1}, \IEEEmembership{Member,~IEEE,},\qquad Zhenzhou Shao\textsuperscript{2}*, \IEEEmembership{Member,~IEEE,} \qquad Hairong Qi\textsuperscript{1}, \IEEEmembership{Fellow,~IEEE}\\
%	\textsuperscript{1}The University of Tennessee, Knoxville, TN \qquad \textsuperscript{2} Capital Normal University, Beijing, China\\
%	{\tt\small yqu3@vols.utk.edu \qquad zshao@cnu.edu.cn \qquad hqi@utk.edu}}
\author{Ying Qu, \IEEEmembership{Member,~IEEE,},\qquad Zhenzhou Shao*, \IEEEmembership{Member,~IEEE,} \qquad Hairong Qi, \IEEEmembership{Fellow,~IEEE}\\

\IEEEcompsocitemizethanks{\IEEEcompsocthanksitem Ying Qu, and Hairong Qi are with the Advanced Imaging and Collaborative Information Processing Group, Department of Electrical Engineering and Computer Science, University of Tennessee, Knoxville, TN 37996 USA (e-mail: yqu3@vols.utk.edu; hqi@utk.edu).%\protect\\
% note need leading \protect in front of \\ to get a newline within \thanks as
% \\ is fragile and will error, could use \hfil\break instead.
\IEEEcompsocthanksitem Zhenzhou Shao is with the Beijing Key Laboratory of Light-weight Industrial Robot and Safety Verification, College of Information Engineering, Capital Normal University, Beijing 100048 China. (e-mail: zshao@cnu.edu.cn).}% <-this % stops an unwanted space
%\thanks{Manuscript received March 26, 2020; revised August 26, 2015.}}
\thanks{Corresponding author: Zhenzhou Shao.}}
\markboth{Accepted by IEEE TRANSACTIONS ON PATTERN ANALYSIS AND MACHINE INTELLIGENCE}%
{Shell \MakeLowercase{\textit{et al.}}: Bare Demo of IEEEtran.cls for IEEE Journals}
% The only time the second header will appear is for the odd numbered pages
% after the title page when using the twoside option.
% 
% *** Note that you probably will NOT want to include the author's ***
% *** name in the headers of peer review papers.                   ***
% You can use \ifCLASSOPTIONpeerreview for conditional compilation here if
% you desire.

% If you want to put a publisher's ID mark on the page you can do it like
% this:
%\IEEEpubid{0000--0000/00\$00.00~\copyright~2015 IEEE}
% Remember, if you use this you must call \IEEEpubidadjcol in the second
% column for its text to clear the IEEEpubid mark.

% use for special paper notices
%\IEEEspecialpapernotice{(Invited Paper)}

% make the title area
\maketitle

% As a general rule, do not put math, special symbols or citations
% in the abstract or keywords.
\begin{abstract}
Photorealistic stylization aims to transfer the style of a reference photo onto a content photo in a natural fashion, such that the stylized image looks like a real photo taken by a camera. State-of-the-art methods stylize the image locally within each matched semantic region and are prone to global color inconsistency across semantic objects/parts, making the stylized image less photorealistic. To tackle the challenging issues, we propose a non-local representation scheme, constrained  with a mutual affine-transfer network (NL-MAT). Through a dictionary-based decomposition, NL-MAT is able to  successfully decouple matched non-local representations and color information of the image pair, such that the context correspondence between the image pair is incorporated naturally, which largely facilitates local style transfer in a global-consistent fashion. To the best of our knowledge, this is the first attempt to address the photorealistic stylization problem with a non-local representation scheme, such that no additional models or steps for semantic matching are required during stylization. Experimental results demonstrate that, the proposed method is able to generate photorealistic results with local style transfer while preserving both the spatial structure and global color consistency of the content image.\textcolor{blue}{~Please find the final version from IEEE Transactions on Pattern Analysis and Machine Intelligence on IEEE Xplore. The code will be released on https://github.com/yingutk/NL-MAT.}

\end{abstract}

%TODO: fix the abstract

% Note that keywords are not normally used for peerreview papers.

% For peer review papers, you can put extra information on the cover
% page as needed:
% \ifCLASSOPTIONpeerreview
% \begin{center} \bfseries EDICS Category: 3-BBND \end{center}
% \fi
%
% For peerreview papers, this IEEEtran command inserts a page break and
% creates the second title. It will be ignored for other modes.
\IEEEpeerreviewmaketitle

\section{Introduction}
\label{sec:intro}
\begin{figure}[htbp]
\vspace{-6mm}
\setlength{\abovecaptionskip}{0.cm}
\setlength{\belowcaptionskip}{-0.cm}
\begin{center}
		\centering
		\begin{minipage}{1\linewidth}
			\subfloat[Style]{\includegraphics[width=0.32\linewidth]{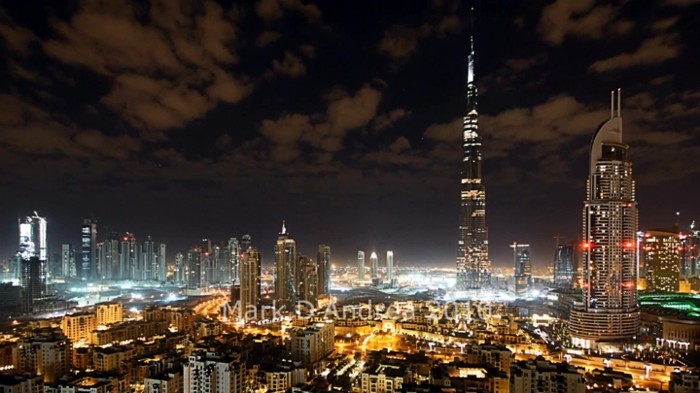}
				\label{fig:demo:tar7}}
			\subfloat[Content]{\includegraphics[width=0.32\linewidth]{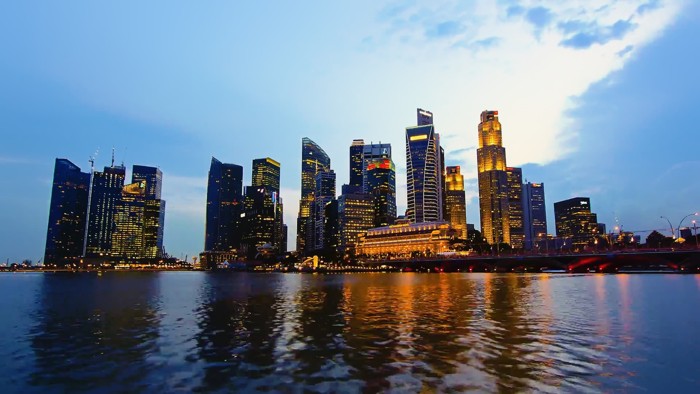}
				\label{fig:demo:in7}}
			\subfloat[Global]{\includegraphics[width=0.32\linewidth]{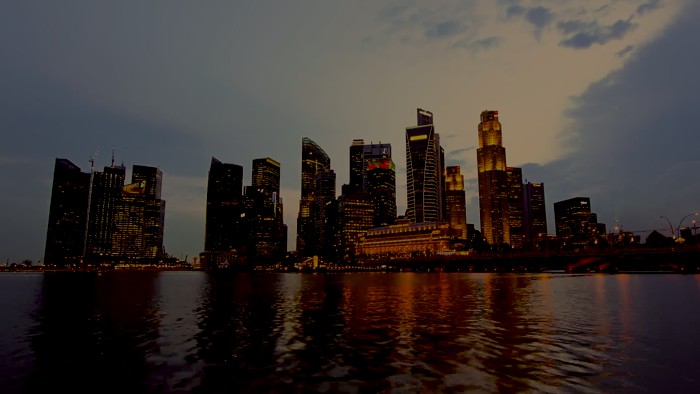}
				\label{fig:demo:global}}\\\vspace{-3mm}
			\subfloat[PhotoWCT]{\includegraphics[width=0.32\linewidth]{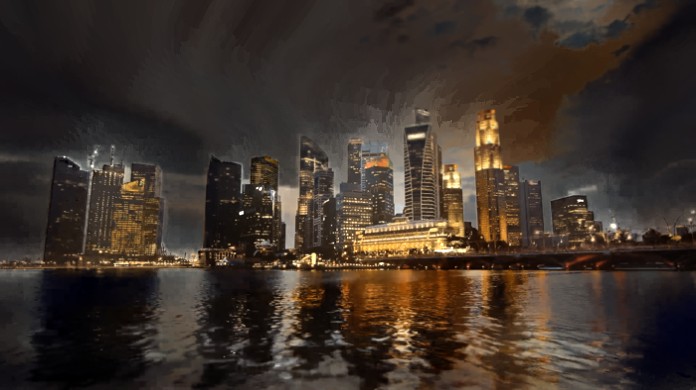}
				\label{fig:demo:wct}}
			\subfloat[WCT$^2$]{\includegraphics[width=0.32\linewidth]{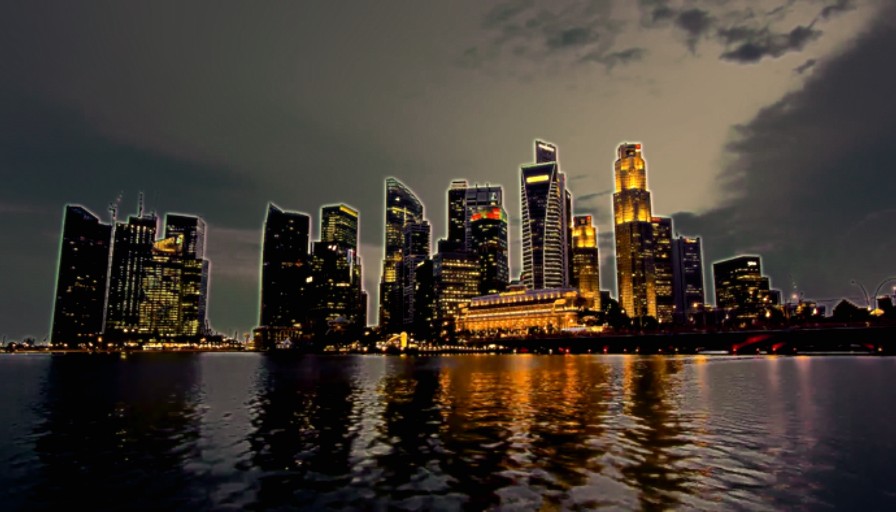}
				\label{fig:demo:wct2}}
			\subfloat[Ours]{\includegraphics[width=0.32\linewidth]{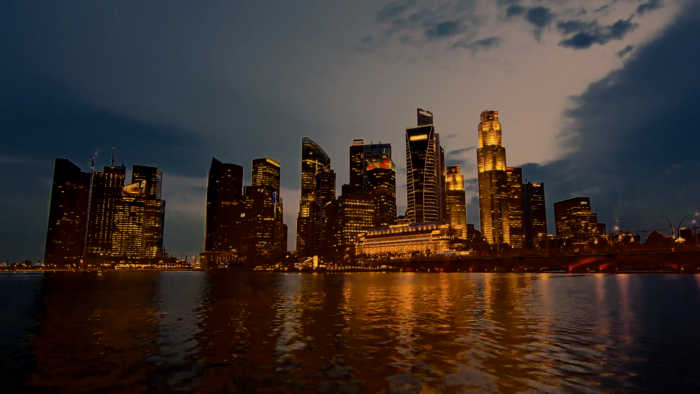}
				\label{fig:demo:ours}}\\%	\hspace{-2mm}
			\centering
			\begin{minipage}{1\linewidth}
					\subfloat[Content]{\adjincludegraphics[width=0.185\linewidth, trim={{0.58\width} {0.232\height} {0.27\width} {0.2675\height}},clip]{fig/demo/in7}}\hspace{0.001mm}
					\subfloat[Global]{\adjincludegraphics[width=0.185\linewidth, trim={{0.58\width} {0.232\height} {0.27\width} {0.2675\height}},clip]{fig/demo/7_rc}}\hspace{0.001mm}
					\subfloat[PhotoWCT]{\adjincludegraphics[width=0.185\linewidth, trim={{0.58\width} {0.232\height} {0.27\width} {0.2675\height}},clip]{fig/demo/7_wct}}\hspace{0.001mm}
					\subfloat[WCT$^2$]{\adjincludegraphics[width=0.185\linewidth, trim={{0.58\width} {0.232\height} {0.2655\width} {0.26\height}},clip]{fig/demo/7_wct2}}\hspace{0.001mm}
					\subfloat[Ours]{\adjincludegraphics[width=0.185\linewidth, trim={{0.58\width} {0.232\height} {0.27\width} {0.2675\height}},clip]{fig/demo/7_ours}}
			\end{minipage}\label{fig:demo:trim}
		\end{minipage}
		\caption{\textcolor{black}{Given a reference style photo taken at night, the content image is stylized as if it was taken at night. (a) Style image. (b) Content image. Photorealistic-stylized images with (c) global-based method~\cite{reinhard2001color} without local color changes, (d) PhotoWCT~\cite{li2018closed} with structure distortion, (e) WCT$^2$~\cite{yoo2019photorealistic} with abrupt color changes between sky and buildings, and (f) our method (NL-MAT) with global consistency. (g-k) Sub-images of (b)-(f), respectively.}}
		\label{fig:demo}
		\vspace{-4mm}
\end{center}
\end{figure}

\IEEEPARstart{T}{he} objective of photorealistic style transfer is to change the style of a content photo to that of a reference photo as shown in Fig.~\ref{fig:demo}. 
By choosing different reference photos, one could make the content photo look as if, for example, it was taken under different illuminations, at different time of the day or season of the year~\cite{luan2017deep,mechrez2017photorealistic,li2018closed}. \textcolor{black}{It is worth mentioning that}
photorealistic style transfer is different from the general stylization approaches~\cite{gatys2016image,li2017universal}, \textcolor{black}{which
 %where the content photos are converted to the style of the reference photos like paintings with artistic texture. These approaches 
tend to generate the painting-like photos with artistic textures and distorted structures}. As emphasized in previous works~\cite{luan2017deep,mechrez2017photorealistic,li2018closed,yoo2019photorealistic}, a successful photorealistic stylization method should be able to transfer sophisticated \textit{matched} styles with \textit{local} color changes while at the same time \textit{preserve the spatial} (or structural) information as well as \textit{global color consistency} of the content photo naturally, such that the resulting image looks like a real photo taken by a camera.

Existing approaches generally perform \textcolor{black}{photorealistic} stylization either in a global or a local way. Global-based methods~\cite{reinhard2001color,pitie2005n} transfer the style of a photo with spatially invariant functions. Although they perform well for global color shifting, they usually \textcolor{black}{could not stylize images effectively within local areas. For example, to stylize the content photo in Fig.~\ref{fig:demo:in7} according to the style photo in Fig.~\ref{fig:demo:tar7}, the bright day sky of Fig.~\ref{fig:demo:in7} should become dark night sky, while the light of the buildings should be preserved. However, as shown in Fig.~\ref{fig:demo:global}, the global-based method turns both light and the day sky to dark. Local-based methods ~\cite{luan2017deep,mechrez2017photorealistic,li2018closed,li2018learning,yoo2019photorealistic,liao2017visual, he2019progressive} are  generally carried out in three major steps, including 1) extraction of high-level features from the content and style image pair with \textcolor{black}{pre-trained models on large datasets}, 2) semantic context matching of the image pair, and 3) local context-based stylization on the extracted features. For most approaches, post-processing has to be performed to preserve the structure of the stylized image, \textcolor{black}{due to the information loss in high-level features~\cite{yoo2019photorealistic}}. \textcolor{black}{However, as shown in Fig.~\ref{fig:demo:wct}, post-processing may not be effective in some scenarios.} By way of the ~\textit{context correspondence} between the content and style images, the local-based approaches could transfer local styles successfully. \textit{Nonetheless}, \textcolor{black}{the extracted features with pre-trained models generally could not indicate the context matching information of the image pair directly. Thus,} the stability of the style transfer relies heavily on the context correspondence estimated from additional supervised segmentation or classification models. Such supervised models may fail if the given image has complex or unknown objects, or has more than three channels. The failure of the models may lead to the failure of  stylization. \textit{More importantly}, these approaches perform stylization locally within each context area, thus tending to ignore the global consistency within or across objects. \textcolor{black}{As shown in Fig.~\ref{fig:demo:wct2}, abrupt color changes across context regions can be easily introduced} with the resulting image being less photorealistic.} %Since the distributions of the representations in different regions may very well be different, abrupt color changes across context regions can be easily introduced with the resulting image being less photorealistic, } 

%TODO: The segmentation itself is a non-trivial task

\textit{There are, in general, ~\textcolor{black}{three key challenging issues with the current stylization approaches, %\textcolor{black}{the current photorealistic representation scheme}, 
namely, 1) how to match the context correspondence \textcolor{black}{of the image pairs} without additional segmentation or classification models? 2) how to perform  \textcolor{black}{context-based} local style transfer in a globally consistent fashion, i.e., without introducing abrupt changes/artifacts within or across semantic regions? and 3) how to preserve the structure information of the stylized image naturally?}}  

%TODO: is there a good example for how to write a new scheme based on totally new directions?

%The representation scheme is the fundamental problem, it loss structure information. It can not identify the context. It break the local-global consistency. 

\begin{figure}[h]
\setlength{\abovecaptionskip}{1pt}
\setlength{\belowcaptionskip}{1pt}
\centering\includegraphics[width=0.5\linewidth]{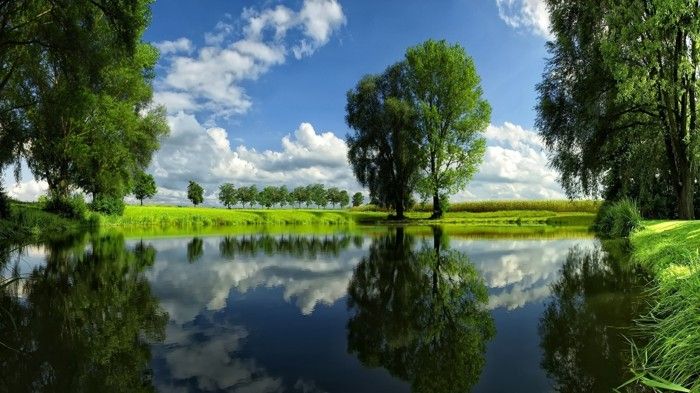}
\caption{The non-local nature of context. For example, the similar context regions, the trees, although possess similar color, are scattered at disjoint regions across the image. }
\label{fig:nonlocal}
\end{figure}

\textcolor{black}{To addressing the challenges,} we exploit the potential of ``non-local'' features that would effectively \textcolor{black}{indicate the context information of images} and break the barrier between local style transfer and global consistency. We argue that ``context'' should not be kept as a local feature, as shown in Fig.~\ref{fig:nonlocal}, where the similar context regions, \eg, the \textcolor{black}{trees}, usually share similar color but may scatter at disjoint locations across the image. \textcolor{black}{Since such non-local similarities could not be readily captured by the current approaches, \textit{a new representation scheme} should be developed to overcome the challenges above.}

\textcolor{black}{In this paper, we propose a non-local representation scheme through dictionary-based image decomposition to address the challenges of photorealistic style transfer. In this scheme, each pixel (as a 3-D vector) can be decoupled into a linear combination of a set of color bases with the corresponding coefficients serving as the ``representation'' of the pixel. The physical implication of this representation is %is that it indicates the proportions of the color bases, ~\ie, 
how much each color \textcolor{black}{basis} contributes to the color formation of that pixel. This representation thus needs to %each color basis has in constructing the color of a certain pixel, thus it should
satisfy two physical constraints,  \ie, the sum-to-one constraint and the \textcolor{black}{non-negative constraint}. Since the entire image shares the same set of color bases,  the representations (\ie, the coefficients of color bases) for similar context would be similar regardless of their spatial location in the image. Thus, the proposed representation scheme can successfully capture the non-local similarities over the entire image, and has the potential to distinguish context according to the coefficients/proportions of the color bases. \textcolor{black}{From this perspective,} we consider the extracted representation to be \textit{context-correlated}. Since such decoupling procedure is done at the pixel level, %would not loss any spatial information of the images, 
it \textit{preserves the structure information} of the images well. In addition, %The ultimate goal of the stylization is to transfer the colors of the content image to that of the style image without structure distortion, which can be achieved by performing affine-transfer on the color bases of the content image. That is, 
the color bases of the content and style images hold an affine relationship such that the colors of the content image can be transferred to those of the style image \textit{without structure distortion}. By enforcing the extracted representations to be sparse, for each context region, only the dominant color bases with non-zero representations would take effect in the global-affine transfer. %Since the dominant color bases are generally different for different contexts, it allows for diverse local transfer depending on the representations of the contexts. 
Since the dominant color bases are generally different for different contexts, it allows for \textit{diverse local transfer in a global-consistent fashion}. Finally, to transfer the correct color for each context, %~\eg the color of the sky in the content image to that in the style image, 
mutual information is employed to \textit{match the context correspondence} of the representations for the image pair. %Unlike the existing representation scheme, where the extracted representations of the image pair have to be matched with additional models and the transfer has to be performed within each matched semantic region, the proposed scheme is able to extract matched representations of the same context that might scatter at disjoint locations across the image,  %incorporated with context information, thus the local transfer can be conducted in a global-consistent manner. %  globally but also locally according to the representations of the contexts.
}

%To match the context correspondence between the content and style images for a correct color transfer, the mutual information between the extracted representations and their own input are maximized, such that the context with similar distributions in both images
%Based on this principle, we enforce the color bases of the image pair to follow an affine-relationship, such that it has the potential to extract matched representations according to the context correspondence of the images which have different colors.}

The proposed method is referred to as the non-local representation based mutual affine-transfer network, or NL-MAT. The contribution of this work is three-fold, pertaining to the three major components of the proposed NL-MAT network:
\begin{itemize}
	\item 
	First, a non-local representation scheme is realized \textcolor{black}{which projects} the images from the three-dimensional RGB color space to a $k$-dimensional representation space with each of the $k$ elements reflecting the proportion of a certain color basis in making up the RGB color. This way, the context information is embedded naturally without adopting any \textcolor{black}{additional models}. The representation is extracted \textcolor{black}{with} a stick-breaking encoder to enforce the two physical constraints of the \textcolor{black}{proportions}, without losing images' spatial information. %\textcolor{black}{of the coefficients of color bases and decoders as a shared dictionaries}.  %spatial structure can be preserved while performing style transfer. 

%	matching the statistics of the content representations to that of the style representations with whitening and coloring transformation and the 
%	even similar objects in the content and style images may have different colors, their distributions of the representations can still be similar

	\item  
	Second, an affine-transfer decoder is constructed that embeds the shared color bases of the content and style images, \textcolor{black}{such that the potential transfer relationship between the image pair can be learned without structure distortion.} By enforcing the sparsity of the non-local representations, we are able to perform  local style transfer using the decoder, while preserving global consistency. % Such transfer is performed in a globally consistent fashion, without abrupt color changes. 

	\item Third, in order to \textcolor{black}{match the colors of similar objects (or parts) in the image pairs} for affine style transfer, we design a mutual discriminative network to \textcolor{black}{extract the context-correspondence representations from the image pairs} by maximizing the mutual information (MI) between the representations and their own RGB inputs. The statistic transformation is adopted to enforce the matched representations to have similar statistical characteristics, which  further improves the photorealistic-stylization capacity of the proposed method.
\end{itemize}

The proposed \textcolor{black}{representation scheme can naturally match the context correspondence between the image pair and realize diverse local style transfer in a global-consistent fashion. } %method can effectively capture the characteristics of the given content-style image pair, thus no additional information is needed for stylization. It 
This is fundamentally different from existing \textcolor{black}{representation schemes} that require additional  \textcolor{black}{segmentation models or steps to match the correspondence.} To the best of our knowledge, this work is the first effort that performs photorealistic style transfer ~\textcolor{black}{with a non-local representation model}.

\begin{table*}[htbp]
\setlength{\abovecaptionskip}{1pt}
\setlength{\belowcaptionskip}{1pt}
	\caption{\textcolor{black}{Capabilities of the State-of-the-art Photorealistic Stylization Approaches}}
	\label{tab:procon}
	\centering
%	\resizebox{\textwidth}{15mm}{
	\setlength{\tabcolsep}{1mm}{
	\begin{tabular}{l|c c |c c c c|c c |c|c}
		\hline
		{}&\multicolumn{2}{c|}{Global-based}&\multicolumn{4}{c|}{Context-based}&\multicolumn{2}{c|}{Patch-based}&\multicolumn{1}{c|}{SOTA}&{Proposed}\\		\hline
		Capabilities&Reinhard\cite{reinhard2001color}&Pitie\cite{pitie2005n}&Luan\cite{luan2017deep}&Li\cite{li2018closed}&LST~\cite{li2018learning}&WCT$^2$~\cite{yoo2019photorealistic}&Liao\cite{liao2017visual}&He\cite{he2019progressive}&PhotoNAS\cite{an2020ultrafast}&NL-MAT\\
		\hline
		Structure preservation &\cmark &\cmark &\xmark &\cmark &\xmark &\cmark &\xmark &\cmark &\cmark &\cmark\\
		\hline
		Local style transfer &\xmark &\xmark &\cmark &\cmark &\cmark &\cmark &\cmark &\cmark &\xmark &\cmark\\
		\hline
		Context-correspondence transfer &\xmark &\xmark &\cmark &\cmark &\cmark &\cmark &\cmark &\cmark &\xmark &\cmark\\
		\hline
		Global consistency &\cmark &\xmark &\xmark &\xmark &\xmark &\xmark &\xmark &\xmark &\cmark &\cmark\\
		\hline
		\shortstack{No additional models}&\cmark &\cmark &\xmark &\xmark &\xmark &\xmark &\xmark &\xmark &\xmark &\cmark\\
		\hline
	\end{tabular}}
\vspace{-5mm}
\end{table*}

The rest of the paper is organized as follows. Sec.~\ref{sec:related} provides an overview of the state-of-the-art photorealistic style transfer approaches. Sec.~\ref{sec:formulate} formulates the photorealistic style transfer problem. Sec.~\ref{sec:proposed} elaborates and analyzes the proposed NL-MAT method. Sec.~\ref{sec:results} performs comprehensive evaluations of the proposed approach. Conclusions are drawn in Sec.~\ref{sec:conclusion}.

\section{Related Work}
\label{sec:related}
Classical style transfer methods stylize an image in a global fashion with spatial-invariant transfer functions~\cite{reinhard2001color,pitie2005n,bae2006two,pitie2007automated, freedman2010object, luan2017deep}. These methods can handle global color shifts well, but they are limited in matching sophisticated styles with color changes~\cite{luan2017deep,li2018closed}, as shown in Fig.~\ref{fig:demo:global}.

\textcolor{black}{Recent photorealistic approaches can be generally categorized into patch-based and context-based, which perform the stylization in a local fashion.} Patch-based approaches stylize images according to the patch similarity on high-level features extracted with the supervised pre-trained CNN. Liao~\etal~\cite{liao2017visual} transferred images by finding dense correspondences between the high-level patch features of the content and style images, with nearest-neighbor field search. A weighted least squares filter (WLS)~\cite{farbman2008edge} was adopted as a post-processing step to refine the structures of the resulting images. He~\etal~\cite{he2019progressive} further improved the stylization results by generating a guidance image based on the strategy of~\cite{liao2017visual}, where local style transfer can be guided in the image domain according to the guidance image, to avoid structure distortion. %However, patch-based methods mainly focus on the patch-similarities while ignoring the global consistency within or across objects. 
%matched the dense semantic correspondence with nearest-neighbor field in deep feature domain, resulting in the faithful local color transfer in the image domain. 

Context-based approaches perform the style transfer \textcolor{black}{based on the high-level features extracted from the supervised pre-trained CNN model and semantic matching obtained from the supervised pre-trained segmentation model}. Luan~\etal~\cite{luan2017deep} preserved the structure of the content image by adopting a color-affine-transfer constraint and color transfer is performed according to the semantic regions generated using the pre-trained DeepLab segmentation model ~\cite{chen2017deeplab}. Mechrez~\etal~\cite{mechrez2017photorealistic} proposed to maintain the fidelity of the stylized image with a post-processing step based on the screened poisson equation (SPE). Li~\etal~\cite{li2018closed} improved the spatial consistency of the output image by adopting the manifold ranking algorithm as the post-processing step. % However, the generated results from context-based approaches easily suffer from abrupt color changes with noticeable artifacts especially between adjacent regions/segments.
\textcolor{black}{LST~\cite{li2018learning} concatenated a linear propagation module after the stylization network to preserve the structure information of the resulting image. %For these approaches, although post-processing steps could preserve the structure information well, it may introduce blurry artifact~\cite{yoo2019photorealistic}.
To address the blurry artifact caused by post-processing, WCT$^2$~\cite{yoo2019photorealistic} was proposed, which introduces a wavelet module in the network to preserve the structure information of the stylized image without post-processing.}

%\textcolor{black}{For most approaches, a series of post-processing steps have to be performed to prevent structure distortion, which may generate stylized images with blurry artifact~\cite{yoo2019photorealistic}.}
%However, the generated results from context-based approaches easily suffer from abrupt color changes with noticeable artifacts especially between adjacent regions/segments.

\textcolor{black}{Although these methods could transfer the local styles well, the effectiveness of the stylization relies heavily on the semantic correspondence estimated from additional supervised segmentation models pre-trained on a different dataset. Segmentation itself is a non-trivial task, and the failure of which may lead to the failure of stylization. Moreover, since stylization is performed within each context region,}  %preserve the spatial structure well with post-processing or color transfer constraints, 
the light and color changes of different parts and materials across the entire image may not be smooth or natural. See Figs.~\ref{fig:patch},~\ref{fig:local_contex} and~\ref{fig:local_contex2}  for a comparison later. %Aside from image quality, these methods need additional priors from pre-trained models which were trained with large amount of parameters on large datasets.
\textcolor{black}{Recently, PhotoNAS~\cite{an2020ultrafast} was proposed to perform smooth stylization by applying WCT~\cite{li2017universal} on the stacked multi-level features extracted from pre-trained models and the normalized skip links. However, since the context information is not considered during the optimization, the stylization is not dramatic within local areas. See Fig.~\ref{fig:compare_nas} for a comparison later.}

%TODO: add figures for the three methods for comparison.

%This is because the methods focus more on local color transfer, and the global color transitions are not considered during the stylization.
%Luan~\etal~\cite{luan2017deep} proposed to preserve the spatial distortion of the content with locally affine-color transfer constraint and segmentation information. The performance of photorealistic stylization is further improved by a few methods~\cite{mechrez2017photorealistic,li2018closed} with different post-processing steps according to the content image.

Based on the discussions above, Table~\ref{tab:procon} summarizes, from five aspects, the pros and cons of some state-of-the-art photorealistic stylization approaches, including preserving the spatial structure of the content image, realizing local style transfer while maintaining global consistency, transferring styles based on context or semantic correspondence between the content and the style images, and needing no ~\textcolor{black}{additional models from supervised classification or segmentation}. In the following, we elaborate on how the proposed approach tackles the challenges brought from each aspect.

\section{Problem Formulation}
\label{sec:formulate}
As discussed in Sec.~\ref{sec:intro}, the key issue in obtaining a context-correspondence and high-quality photorealistic style transfer is \textcolor{black}{the realization of a new representation scheme that extracts the matched non-local representations from the  image pair. Such scheme is designed
%the decoupling of spatial structure and color information for both the content and style images.
%We solve this problem from a unique angle, 
according to the theory of dictionary-based image decomposition, where natural images can be represented by a set of color bases with its coefficient vectors (\ie, representations)~\cite{omer2004color,laffont2012coherent,2010Non}. In this paper, the decomposition is learned through neural network by minimizing the reconstruction error of the image pair.} 
%To facilitate the subsequent processing,  the images are unfolded into 2D matrices. 

Given a content image, $I_c \in \mathbb{R}^{m\times n\times l}$, where  $m$, $n$, and $l$ denote its width, height, and number of channels, respectively, and a style image, $I_s \in \mathbb{R}^{M\times N \times l}$, where $M$, $N$, and $l$ denote its width, height, and number of channels, respectively, the goal is to generate the image $I_{cs} \in \mathbb{R}^{m \times n \times l}$ with its content coming from $I_c$ but using the style from $I_s$. \textcolor{black}{For each image, to capture the non-local similarities, the entire image is enforced to share the same set of color bases. That is, a single pixel in the content and style images can be expressed as Eqs.~\eqref{equ:content} and~\eqref{equ:style}, respectively.} 
\begin{align}
\begin{split}\label{equ:content}
&\mathbcal{i}_c = \mathbf{s}_c {D}_c
\end{split}\\
\begin{split}\label{equ:style}
&\mathbcal{i}_s = \mathbf{s}_s {D}_s
\end{split}
\end{align}
where $\mathbcal{i}_c\in\mathbb{R}^{1 \times l}$ and $\mathbcal{i}_s\in\mathbb{R}^{1 \times l}$ denote a single pixel of the content image and the style image, respectively. $D_c\in\mathbb{R}^{k \times l}$ and $D_s\in\mathbb{R}^{k \times l}$ are two matrices with each row of which denoting the color basis that preserves the color information of the entire content and style images, respectively.  $\mathbf{s}_c\in\mathbb{R}^{1\times k}$ and $\mathbf{s}_s\in\mathbb{R}^{1\times k}$ denote the corresponding coefficients for each of the $k$ color bases of the content and style image, respectively. Note that, in our case, we have $k\gg l$. That is, the number of color bases is much larger than the dimension of the input pixels. %Since each row vector of $S_C$ (or $S_S$) indicates how the color bases are combined at a specific location, it only carries the color information of an individual pixel. Applying transfer on such vector would not affect the spatial structure of the content image. 
\begin{figure}
\setlength{\abovecaptionskip}{1pt}
\setlength{\belowcaptionskip}{1pt}
	\begin{center}
		\includegraphics[width=0.7\linewidth]{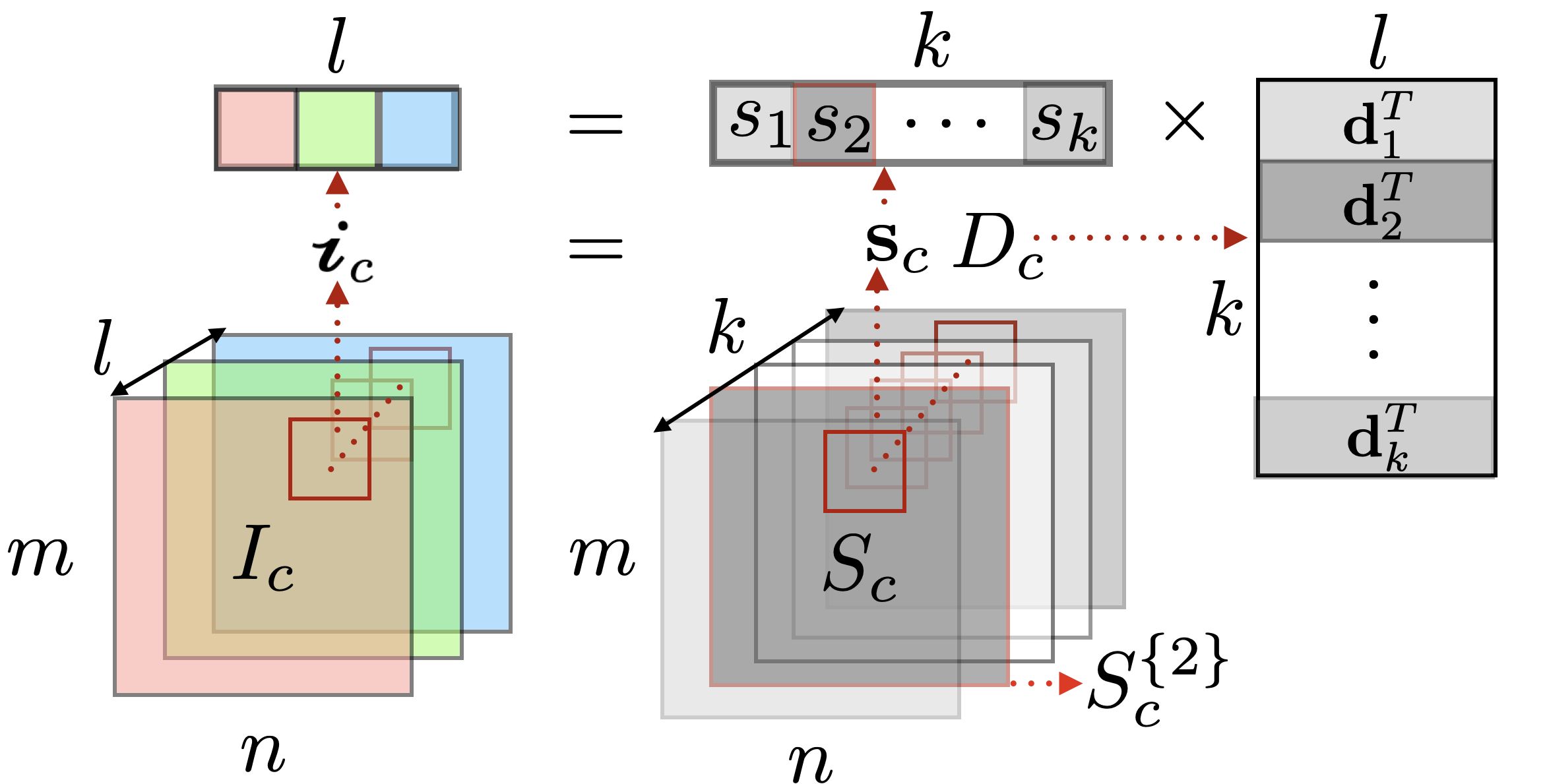}
	\end{center}
	\caption{\textcolor{black}{Dictionary-based image decomposition. Note: The darker the shades, the larger the proportions.}}
	\label{fig:decompose}
\vspace{-6mm}
\end{figure}
Taking the content image as an example, \textcolor{black}{the decomposition is illustrated in Fig.~\ref{fig:decompose}}. For an individual pixel \textcolor{black}{$\mathbcal{i}_c$}, Eq.~\eqref{equ:content} can be written as 
\[ 
\mathbcal{i}_c = \mathbf{s}_c {D}_c = [s_1,\cdots,s_i,\cdots,s_k][{{\mathbf{d}_1}},\cdots,{\mathbf{d}_i},\cdots, {\mathbf{d}_k}]^T,
\]
%\[ 
%\mathbcal{i}_c = \mathbf{s}_c {D}_c = [s_1,\cdots,s_i,\cdots,s_k][{{\mathbf{d}_1}_\rightarrow},\cdots,{\mathbf{d}_i}_\rightarrow,\cdots, {\mathbf{d}_k}_\rightarrow]^T,
%\]
where \textcolor{black}{the transpose of} %${{\mathbf{d}_i}_\rightarrow} \in \mathbb{R}^{1\times l}$ 
${{\mathbf{d}_i}} \in \mathbb{R}^{l\times 1}$ 
denotes a row vector of $D_c$ carrying one color basis,  and $s_i$, the $i$th component of $\mathbf{s}_c$, indicates the proportion (representation) of the color basis, ${\mathbf{d}_i}$, %${\mathbf{d}_i}_\rightarrow$, 
in making up the color of the given pixel, and $i=1,\cdots,k$. With all the $\mathbf{s}_c$'s for each pixel, we obtain the representation $S_c \in \mathbb{R}^{m\times n\times k}$ for the entire image, where the $i$th slice (or plane) of the representation, $S_c^{\{i\}} \in \mathbb{R}^{m\times n}$, indicates the proportions of the $i$th color basis, ${\mathbf{d}_i}$, %${\mathbf{d}_i}_\rightarrow$ 
for all the pixels in the image. \textcolor{black}{Based on the physical properties of the proportions, the non-negative and sum-to-one constraints are enforced on the representation $S_c$ to extract the desired color bases. With such settings, similar contexts, although scatter across the image in disjoint locations,  would have similar representations, thus the scheme is able to capture the non-local context-correlated information by nature.}

\textcolor{black}{The goal of the photorealistic style transfer is to transfer the colors of the content image to that of the style image. To find the potential transfer without structure distortion, we enforce the color bases of the image pair $D_c$ and $D_s$ to have an affine relationship. As analyzed in Sec.~\ref{sec:intro}, by further enforcing the representations to be sparse, only the dominant color bases (\eg, $\mathbf{d}_2$ in Fig.~\ref{fig:decompose}) with larger representations (\eg, $s_2$ in Fig.~\ref{fig:decompose}) would be transferred  effectively in the global affine transfer. %, which facilitates the local transfer in a global-consistent fashion. 
On the other hand, since different objects or parts consist of different dominant color bases, the extracted representations are \textit{context-sensitive} with discriminative capacity, \ie, the objects or parts with different colors can be easily identified by their representations. %Therefore, in the following sections, we are able to evaluate the effect of the local color transfer through the \textit{discriminative capacity} of the representations. 
To transfer the correct colors for different contexts, the representations between the content and style images are matched through mutual information. 
%\textcolor{black}{With the above constraints, the decomposition is solved by reconstructing both the content and style images. 
In the next section, we elaborate on how these constraints are enforced to realize the proposed photorealistic style transfer.}

%\begin{align}
%\begin{split}\label{equ:content_obj}
%&min\sum\Vert \mathbcal{i}_c - \mathbf{s}_c {D}_c\Vert
%\end{split}\\
%\begin{split}\label{equ:style_obj}
%&\mathbcal{i}_s = \mathbf{s}_s {D}_s
%\end{split}
%\end{align}
%Since $S_c$ (or S_s$) indicates how the color bases are combined at each specific spatial location, it carries the spatial information (of color mixture) of an individual pixel at a specific spatial location, and is decoupled from the color bases of the image.
%\color{blue}(To do, will they work on images of different type?)\color{black}

\section{Proposed Method}
\label{sec:proposed}
\begin{figure}
\setlength{\abovecaptionskip}{0pt}
\setlength{\belowcaptionskip}{0pt}
	\begin{center}
		\includegraphics[width=1\linewidth]{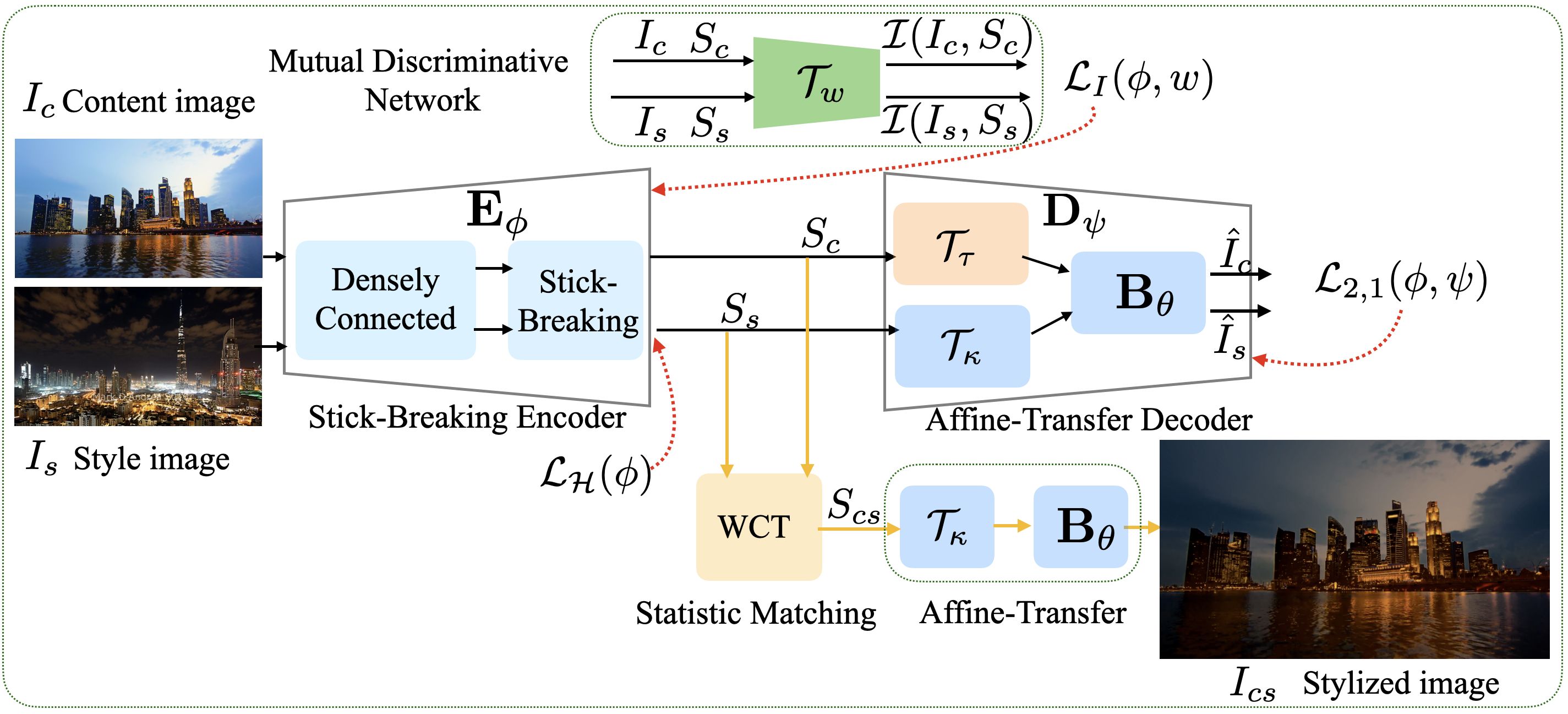}
	\end{center}
	\caption{Flowchart of the proposed NL-MAT.}
	\label{fig:flow}
\vspace{-5mm}
\end{figure}

We propose a \textcolor{black}{non-local representation based} mutual affine-transfer network (NL-MAT) architecture that % perform photorealistic style transfer through two procedures,
%~\ie, representation extraction and style transfer. The network 
mainly consists of three unique structures:  1) a shared stick-breaking encoder %and an affine-transfer decoder 
for the decoupling of \textcolor{black}{non-local} representations and color information of both the content and style images, 2) a sparse entropy function with affine-transfer decoder for the local style transfer in a global-consistent fashion, and 3) a mutual discriminative network to enforce the correspondence of context-sensitive representations with statistical matching between the content-style image pair. %for the faithful color transfer. 4) A stylization network with statistic matching of the representations and the affine-transfer decoder for the color transfer without structure distortion. 
\textcolor{black}{The stylized image is generated by performing both the color transfer with bases, as well as statistic matching on the corresponding representations with WCT.}
The unique architecture is shown in Fig.~\ref{fig:flow}.

%The network mainly consists of three unique structures, a shared sparse Dirichlet encoder for the extraction of feature vectors with both representative and discriminative capacity, an affine-transfer decoder for global color transfer, and a simplified mutual discriminative network to enforce the correspondence between multi-modal representations. The stylization is done by whitening and coloring the context-sensitive representation vectors and transfer them according to the global color bases of the style image. 

\subsection{Overview of Network Architecture}
\label{sec:proposed:arch}
As shown in Fig.~\ref{fig:flow}, \textcolor{black}{the inputs of the network are the content and style images $\mathcal{G} =\{I_c, I_s\}$ with $l$ channels ($l=3$ for RGB color images), and the outputs of the network are their reconstructed images $\hat{\mathcal{G}} = \{\hat{I}_c, \hat{I}_s\}$. The network decomposes both the content image $I_c$ and the style image $I_s$ by learning a shared encoder structure, $\mathbf{E}_{\phi}$, and an affine-transfer decoder structure, $\mathbf{D}_{\psi}$. The representation domain in the hidden layer is denoted as $\mathcal{S} = \{S_c, S_s\}$.} %Let us define the input domain as $\mathcal{G} =\{I_c, I_s\}$, output domain as $\hat{\mathcal{G}} = \{\hat{I}_c, \hat{I}_s\}$, and the representation domain as $\mathcal{S} = \{S_c, S_s\}$. 
The encoder of the network, $\mathbf{E}_{\phi}:\mathcal{G}\rightarrow\mathcal{S}$, maps the input data to high-dimensional representations (latent variables on the hidden layer), \ie, $p_{\phi}(\mathcal{S}\vert {\mathcal{G}})$, and the affine transfer decoder,  $\mathbf{D}_{\psi}:\mathcal{S}\rightarrow\hat{\mathcal{G}}$, reconstructs the images from the representations, \ie, $p_{\psi}(\hat{\mathcal{G}} \vert \mathcal{S})$. %Note that $\mathbf{D}_{\psi}$ is constructed with fully-connected layers with only identity activation functions. 
The representation $\mathcal{S}$ contains the coefficients that reflect the local contribution of different color bases, and the weights of the decoders $\mathcal{T}_\tau(\mathbf{B}_\theta)$ and $\mathcal{T}_\kappa(\mathbf{B}_\theta)$ [to be explained in Eqs.~\eqref{equ:bc} and \eqref{equ:bs}] serve as color bases $D_c$ and $D_s$ in Eqs.~\eqref{equ:content} and~\eqref{equ:style}, respectively. \textcolor{black}{The representation layer is built with the stick-breaking structure to naturally enforce the non-negative and sum-to-one physical properties of the proportions. This will be further elaborated in Sec.~\ref{sec:proposed:decouple}.}%This correspondence is further elaborated below.

To encourage local style transfer, the sparsity constraint defined by the entropy function $\mathcal{L}_\mathcal{H}(\phi)$ is applied to the representation domain. Both the inputs, $I_c, I_s$, and the representations, $S_c, S_s$, are fed into the mutual discriminator \textcolor{black}{$\mathcal{T}_w$ with weights $w$} to enforce the context correspondence between $S_c$ and $S_s$ for a context-correspondence style transfer. \textcolor{black}{The network is constructed with only fully connected layers, which are optimized according to the reconstruction error and regularized by the physical constraints incorporated. This will be further elaborated in Sec.~\ref{sec:proposed:context}}.

In the stylization procedure, as shown in the lower-right part of Fig.~\ref{fig:flow}, the distribution of $S_c$ is matched with that of $S_s$ using the whitening and coloring transformation (WCT)~\cite{li2017universal}. The transferred  $S_{sc}$ is then fed into the style's affine-transfer decoder $\mathcal{T}_\kappa(\mathbf{B}_\theta)$, to generate the stylized image $I_{cs}$. Note that the dashed lines in Fig.~\ref{fig:flow} show the path of back-propagation which will be further elaborated in Sec.~\ref{sec:proposed:opt}. 

%\subsection{Spatial Structure and Color Information Decoupling}
\subsection{\textcolor{black}{Non-local Representation and Color Information Decoupling}}
\label{sec:proposed:decouple}
\begin{figure}
\setlength{\abovecaptionskip}{1pt}
\setlength{\belowcaptionskip}{1pt}
	\begin{center}
		\includegraphics[width=1\linewidth]{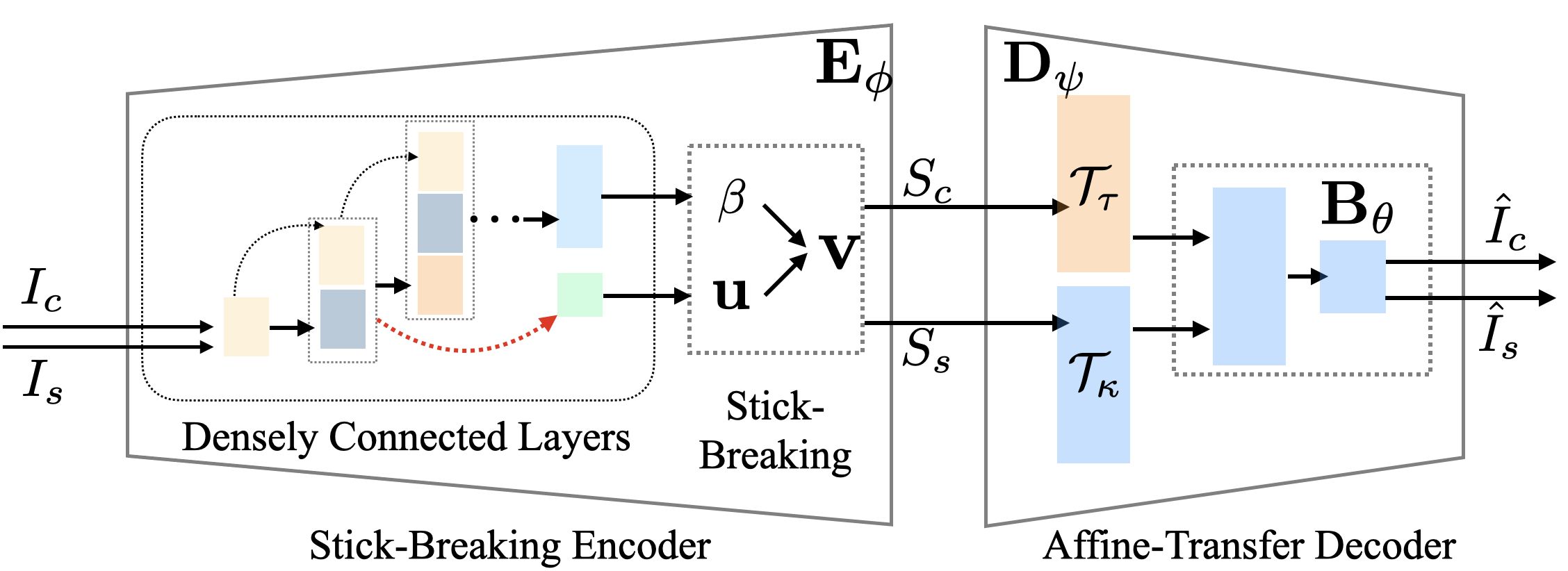}
	\end{center}
	\caption{Network structure of the shared stick-breaking encoder and the affine-transfer decoder.}
	\label{fig:flow_feature}
\vspace{-5mm}
\end{figure}

As elaborated in Sec.~\ref{sec:formulate}, both the content and style images can be decoupled to the representations (indicating proportion coefficients ) and color bases (indicating color composition). However, for different images, such representations or color bases have diverse statistic distributions. To facilitate the style transfer, we construct a network with a shared stick-breaking encoder for the extraction of representations, %satisfying physical constraints, 
and an affine-transfer decoder carrying transferred color information. Since the arrangement of adjacent pixels are untouched, this decoupling mechanism effectively preserves spatial/structural distribution of the content image while performing color transfer based on the style image. %which decouples both images simultaneously. 
The detailed structure is shown in Fig.~\ref{fig:flow_feature}. 

\subsubsection{Representation Extraction with Shared Stick-Breaking Encoder}
\label{sec:proposed:decouple:stick}
As discussed in Sec.~\ref{sec:formulate}, pixels in natural images can be represented as a linear combination of a set of color bases with the coefficients satisfying two %The content and style images may have similar spatial structures, \eg, sky, cloud, tree canopy and tree trunk, etc. In addition, such spatial structures should naturally meet coefficients' 
physical constraints, \ie, sum-to-one and non-negativity. In order to guarantee the constraints are met, in the network design, we adopt a shared stick-breaking encoder to naturally incorporate the physical constraints. The detailed network structure of the shared stick-breaking encoder is shown in the left part of Fig.~\ref{fig:flow_feature}, \textcolor{black}{where each rectangle block denotes a fully-connected layer with neurons}.

The stick-breaking process can be illustrated as breaking a unit-length stick into $k$ pieces, where the length of each piece follows the Dirichlet distribution~\cite{sethuraman1994constructive}. Samples collected from the Dirichlet distribution naturally satisfy the sum-to-one and non-negativity constraints. Here, we follow the work of \cite{nalisnick2016deep,qu2018unsupervised}, which draw the samples of representation $\mathcal{S}$ from the Kumaraswamy distribution \cite{kumaraswamy1980generalized}. Assuming that the row vector of representations for a single pixel is denoted as $\mathbf{s}_\rightarrow = \{s_{i}\}_{1 \leq i \leq k}$, we have $0\leq s_{i}\leq1$, and $\sum_{i=1}^{k}{s_{i}}=1$, where $k$ is the number of color bases.  Each variable $s_{i}$ can be defined as
\begin{equation}
s_{i} =\left\{
\begin{array}{ll}
v_{1} \quad & \text{for} \quad i = 1\\
v_{i}\prod_{t<i}(1-v_{t}) \quad &\text{for} \quad i>1 ,
\end{array}\right.
\label{equ:stick}
\end{equation}
where  $v_{i}\sim 1-(1-u_{i}^\frac{1}{\beta_{i}})$ is drawn from the inverse transform of the Kumaraswamy distribution. Both parameters $u_{i}$ and $\beta_{i}$ are learned through the network for each row vector, as illustrated in Fig.~\ref{fig:flow_feature}.  Since $\beta_i>0$, a softplus is adopted as the activation function \cite{dugas2001incorporating} at the ${\boldsymbol\beta}$ layer. Similarly, a sigmoid \cite{han1995influence} is used to map ${u}$ into the $(0,1)$ range at the $\mathbf{u}$ layer. The input of the encoder has three neurons carrying the color information of the RGB channels of each pixel in the images, and it is densely connected to all the subsequent layers by stacking the layers together \textcolor{black}{to increase the representation power of the network, as shown in the left block of Fig.~\ref{fig:flow_feature}. More details are described in Sec.1 of the supplementary file.}  

%\textcolor{black}{The input of the encoder has three neurons carrying the color information of the RGB channels of each pixel in the images, and it is densely connected to all the subsequent layers by stacking the layers together. As shown in the left block of Fig.~\ref{fig:flow_feature}, the first layer has three neurons, it is passed to the second layer by stacking the first layer on the second layer, then the stacked layer with six neurons is passed on to the third layer by stacking on top of the third layer. Similarly, all the subsequent layers in the left block of the encoder in Fig.~\ref{fig:flow_feature} are densely connected. In this way, the information is easily passed on to the representation layer, which strengthens the feature propagation and increases the representative power of the network.}

\subsubsection{Color Information Extraction with Affine-Transfer Decoder}
\label{sec:proposed:decouple:decoder}

\begin{figure*}
\setlength{\abovecaptionskip}{1pt}
\setlength{\belowcaptionskip}{1pt}
    \centering
		\begin{minipage}{0.9\linewidth}
			\subfloat[Content $I_c$]{
				\includegraphics[width=0.174\linewidth]{fig/pair50_rp/50_0in}
				\label{fig:context:a}}
			\subfloat[$S_c^{\{1\}}$ Sky]{
				\includegraphics[width=0.19\linewidth]{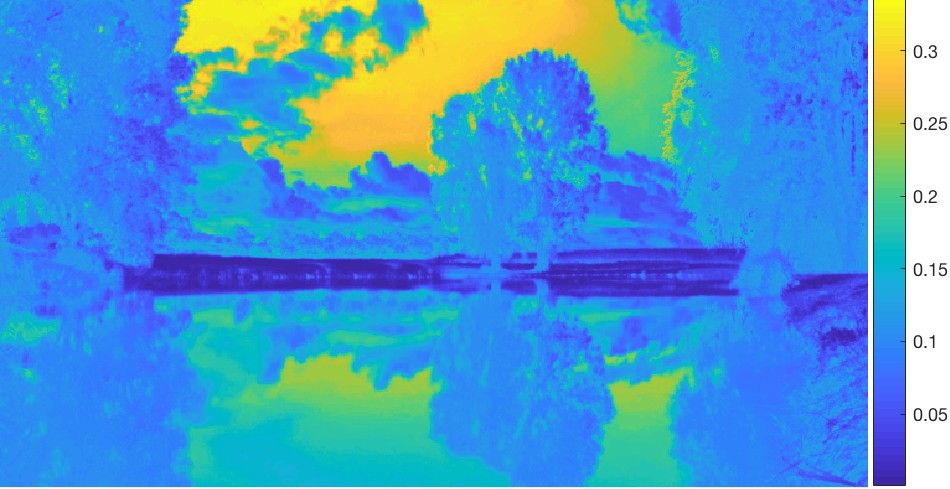}
				\label{fig:context:b}}
			\subfloat[$S_c^{\{2\}}$ Cloud]{
				\includegraphics[width=0.19\linewidth]{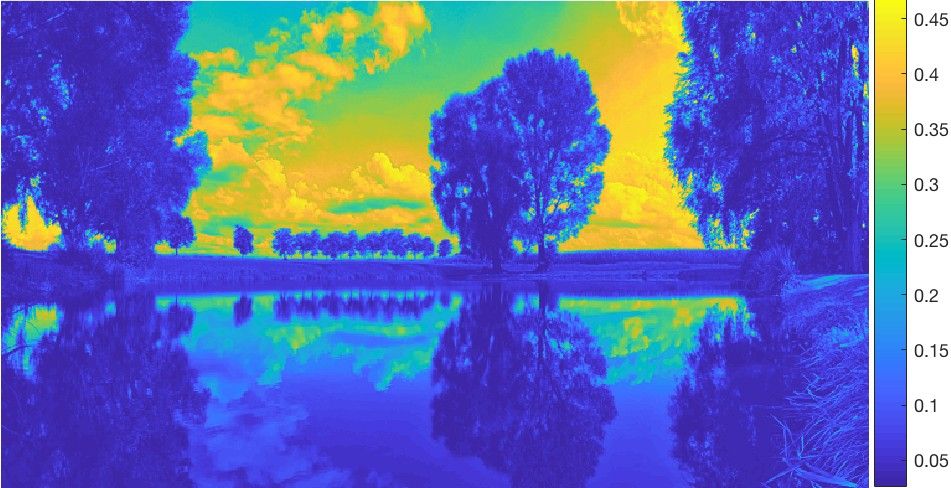}
				\label{fig:context:c}}
			\subfloat[$S_c^{\{3\}}$ Tree]{
				\includegraphics[width=0.19\linewidth]{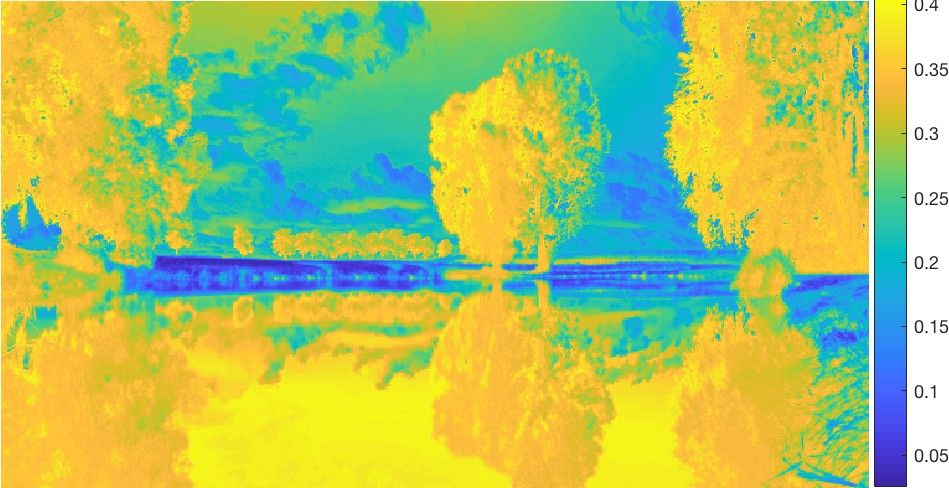}
				\label{fig:context:d}}
			\subfloat[$S_c^{\{4\}}$ Grass]{
				\includegraphics[width=0.19\linewidth]{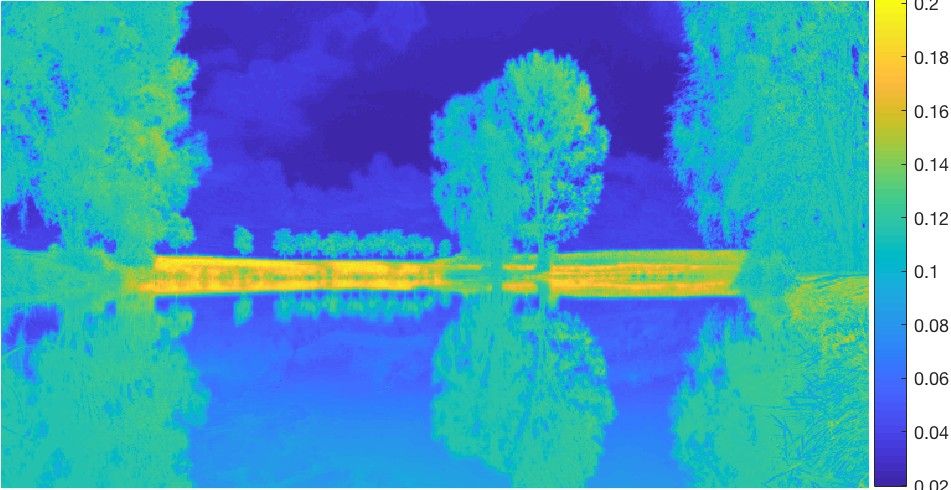}
				\label{fig:context:e}}\\
			\subfloat[Style $I_s$]{
				\includegraphics[width=0.174\linewidth]{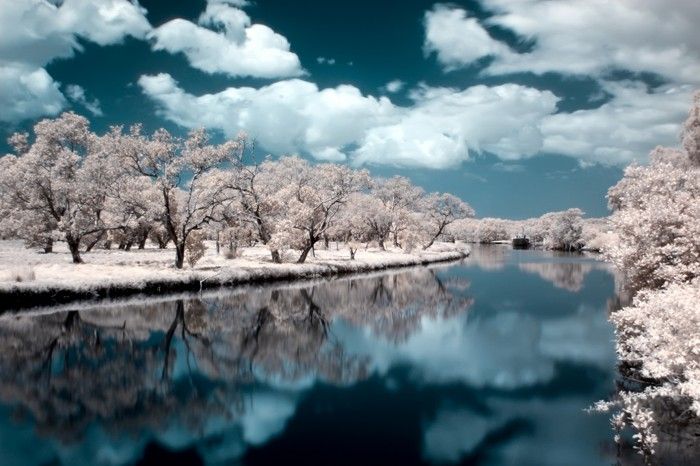}
				\label{fig:context:f}}
			\subfloat[$S_s^{\{1\}}$ Sky]{
				\includegraphics[width=0.19\linewidth]{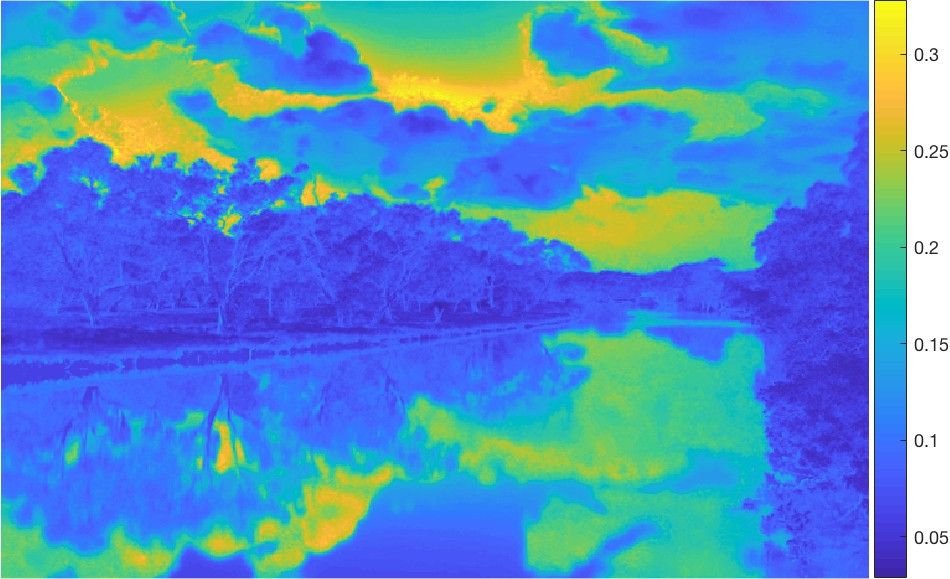}
				\label{fig:context:g}}
			\subfloat[$S_c^{\{2\}}$ Cloud]{
				\includegraphics[width=0.19\linewidth]{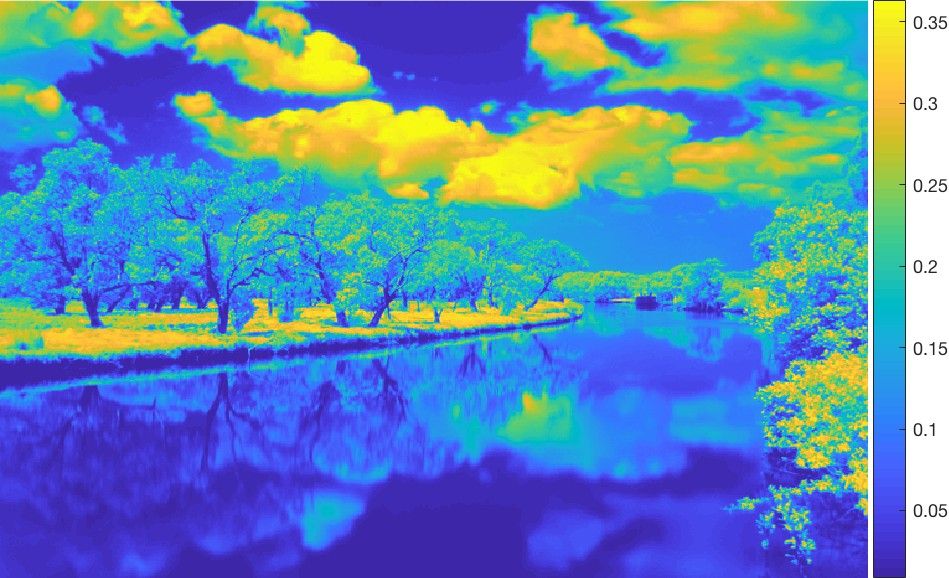}
				\label{fig:context:h}}
			\subfloat[$S_c^{\{3\}}$ Tree]{
				\includegraphics[width=0.19\linewidth]{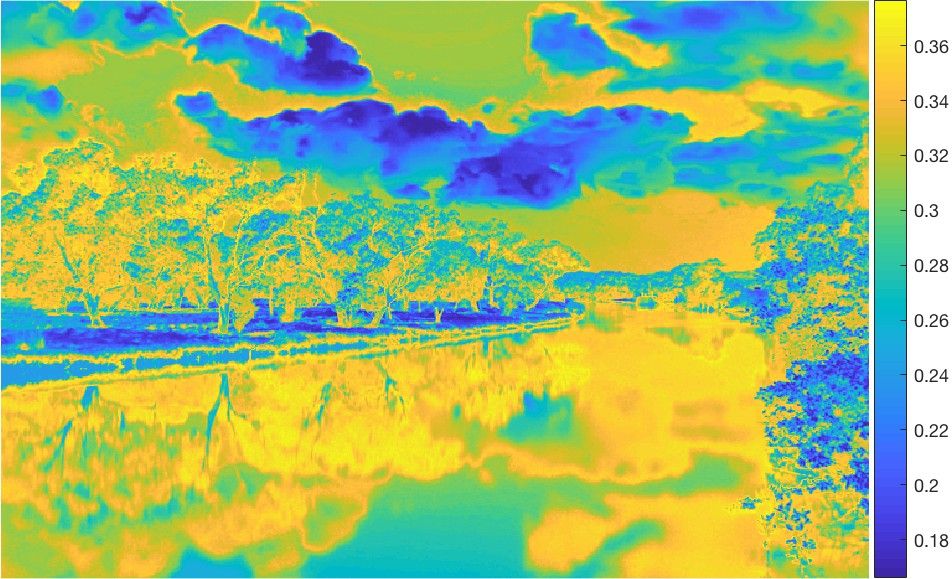}
				\label{fig:context:i}}
		    \subfloat[$S_c^{\{4\}}$ Grass]{
				\includegraphics[width=0.19\linewidth]{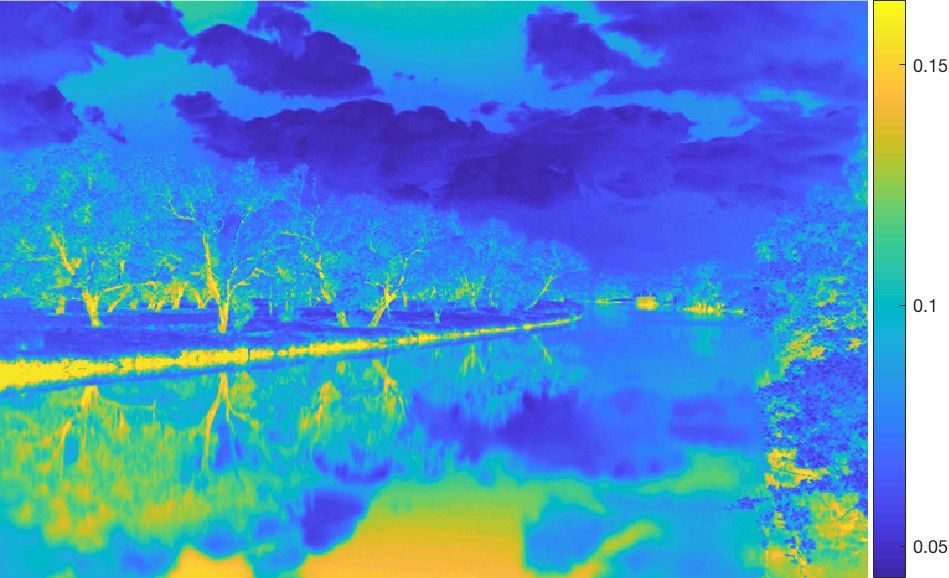}
				\label{fig:context:j}}
		\end{minipage}
	\caption{The representations extracted with NL-MAT from the content image (top) and style image (bottom). Ten color bases are assumed and the representations of the top four most contributing color bases are shown. Brighter color indicates higher proportion (or coefficient) value. Columns 2--5 show the  representation slices of the content (top) and style (bottom) images corresponding to the proportions of their corresponding color bases. With the sparsity constraint, different objects have different dominant color bases, thus the representation is context-sensitive. With the mutual discriminative network, the extracted representations of the content and style images are encouraged to be matched with each other,~\ie, \textcolor{black}{tree-to-tree}, sky-to-sky.}%sky-to-sky, cloud-to-cloud.}
	\label{fig:context}
\vspace{-5mm}
\end{figure*}

\textcolor{black}{As analyzed in Secs.~\ref{sec:intro} and \ref{sec:formulate}, to transfer the colors of the content image to those of the style image without structure distortion, we enforce the color bases of the content image, $D_c$, and the style image, $D_s$, to hold an affine relationship with the proposed affine transfer decoder.} Such decoder not only carries the color information of both images but also their transfer information. The network structure of the affine-transfer decoder is shown in the right part of Fig.~\ref{fig:flow_feature}.

The transfer between \textcolor{black}{the color bases} could be modeled as $D_s = \mathbf{a}D_c+\mathbf{b}$. 
To improve the flexibility and the representative power of the network so as to facilitate decoupling, instead of relating $D_c$ and $D_s$ with an affine transformation modeled by $\mathbf{a}$ and $\mathbf{b}$, we express $D_c$ and $D_s$ as affine transformation of a shared basis weights $\mathbf{B}_\theta$, with $\mathcal{T}_\tau(\mathbf{B}_\theta)$ and $\mathcal{T}_\kappa(\mathbf{B}_\theta)$, respectively, as
%introduce another basis $\mathbf{B}_\theta$ to link the content and style bases, then the affine-transfer decoder is defined as 
\begin{align}
\begin{split}\label{equ:bc}
&\mathcal{T}_\tau(\mathbf{B}_\theta) = \mathbf{a}_{\tau}\mathbf{B}_\theta+ \mathbf{b}_\tau
\end{split}\\
\begin{split}\label{equ:bs}
&\mathcal{T}_\kappa(\mathbf{B}_\theta) = \mathbf{a}_{\kappa}\mathbf{B}_\theta + \mathbf{b}_\kappa,
\end{split}
\end{align}
% \begin{align}
% \begin{split}\label{equ:bc}
% & \mathbf{D}_{\tau,\theta} = \mathcal{T}_\tau(\mathbf{B}_\theta) = \mathbf{a}_{\tau}\mathbf{B}_\theta+ \mathbf{b}_\tau
% \end{split}\\
% \begin{split}\label{equ:bs}
% & \mathbf{D}_{\kappa,\theta} = \mathcal{T}_\kappa(\mathbf{B}_\theta) = \mathbf{a}_{\kappa}\mathbf{B}_\theta + \mathbf{b}_\kappa,
% \end{split}
% \end{align}
where $\mathbf{B}_\theta$, $\mathbf{a}_{\tau}$, $\mathbf{b}_\tau$ and $\mathbf{a}_{\kappa}$, $\mathbf{b}_\kappa$ are the network structure consisting of weights $\{\theta,\tau,\kappa\}$. $\mathcal{T}_\tau(\mathbf{B}_\theta)$ and $\mathcal{T}_\kappa(\mathbf{B}_\theta)$ correspond to $D_c$ and $D_s$ in Eq.~\eqref{equ:content} and Eq.~\eqref{equ:style}, respectively, and share the same basis weights $\mathbf{B}_\theta$.  The bases of the content and style images still hold an affine relationship $\mathcal{T}_\tau(\mathbf{B}_\theta) = \mathbf{a}\mathcal{T}_\kappa(\mathbf{B}_\theta)+\mathbf{b}$, where
\begin{align}
\begin{split}\label{equ:transfa}
& \mathbf{a}  = \mathbf{a}_{\kappa}\mathbf{a}_{\tau}^{-1}
\end{split}\\
\begin{split}\label{equ:transfb}
&\mathbf{b}  = \mathbf{b}_\kappa -  \mathbf{a}_{\kappa}\mathbf{a}_{\tau}^{-1}\mathbf{b}_\tau,
\end{split}
\end{align}
In this way, both the color information of the content $D_c$ and style images $D_s$ and their transfer information $\{\mathbf{a}, \mathbf{b}\}$ are encoded into the network. %Then the next problem would be how to perform faithful color transfer in a local manner based on the context information of the image, \eg, the color of the sky in the style image should be transferred to that in the content image. 

\textcolor{black}{With the stick-breaking encoder and the affine-transfer decoder, the proposed scheme is able to decouple the non-local representations and their color bases of both the content and style images successfully. Since the color bases are shared for the entire image, the representations are context-correlated, enabling the capture of the  non-local similarities in the image. }

\subsection{Context-Correspondence Local Style Transfer}
\label{sec:proposed:context}
\subsubsection{Local Style Transfer with Entropy Function}
\label{sec:proposed:context:sparse}
%The affine-transfer decoder carries the color information of the entire content and style images. 
%TODO? should we describe this with context instead of pixels. 
For each \textcolor{black}{context}, its representation, extracted by the encoder of the network, indicates the proportion of each color basis in making up the given \textcolor{black}{context. %Since the entire image shares a common set of color bases, \eg, $D_s$ or $D_c$, the representation itself is discriminative. That is, the contexts with different colors should have different representations. From this perspective, we say that the proposed non-local representation scheme is context-sensitive that would enable local style transfer. To further enhance the discriminative power of the representation, we apply a sparsity constraint 
To enforce the local style transfer, we encourage the representations to be sparse, such that each context is mainly constructed by a few dominant color bases determined by the corresponding representations. From color transformation perspective, the color transfer for each context is mainly related to a few dominant color bases with large representations. % proportions. %The color bases with small proportions are also transferred, but they have little effect on color changes. 
In this way, we are able to perform local style transfer smoothly in a global consistent fashion due to the non-local characteristic of the representation scheme. With sparse constraint, the representation itself is more discriminative. That is, the contexts with different colors are easier to be distinguished. From this perspective, we say that the proposed non-local representation scheme is context-sensitive that would enable local style transfer.}

%for each context region, only the color bases with non-zero representations would take effect in the affine transfer. Since the locations of the non-zero representations are different for different contexts, it allows diverse local transfer in a global consistent fashion.

%TODO: ? should I remove this? this is not relevant.

~\textcolor{black}{The traditional widely used $l_1$ regularization or Kullback-Leibler divergence \cite{Goodfellow-et-al-2016} regularizes the sparsity of the network by reducing the summation of the representations. However, they cannot be used here to measure the sparsity as the representations are equal to one almost surely, due to the stick-breaking structure.} Instead, we adopt the normalized entropy function~\cite{huang2017sparse}, defined in Eq.~\eqref{equ:entropyfun}, which decreases monotonically when the data become sparse. For example, if the representation for the $i$th pixel has two dimensions (\ie, two color bases) with $s_{1}+s_{2}=1$, the local minimum only occurs at the boundaries of the quadrants,~\ie, either $s_{1}$ or $s_{2}$ is zero. This nice property guarantees the sparsity of arbitrary data even under the condition that the data need to sum-to-one.

\begin{equation}
\mathcal{H}_{p}(\mathbf{s_\rightarrow}) = -\sum_{i=1}^{num}\frac{{\Vert \mathbf{s}_i}_\rightarrow\Vert^p}{\Vert {\mathbf{s}_i}_\rightarrow \Vert_p^p}
\log\frac{{\Vert \mathbf{s}_i}_\rightarrow \Vert^p}{\Vert {{\mathbf{s}_i}_\rightarrow} \Vert_p^p},
\label{equ:entropyfun}
\end{equation}
In Eq.~\eqref{equ:entropyfun}, $num$ denotes the total number of pixels of the image. For the content image $num=m\times n$, and for the style image, $num=M\times N$. We choose $p=1$ for efficiency. The objective function for sparse loss can then be defined as 
\begin{equation}
\mathcal{L}_\mathcal{H}(\phi) = \mathcal{H}_{1}(\mathbf{E}_\phi(I_c)) + \mathcal{H}_{1}(\mathbf{E}_\phi(I_s)).
\end{equation}

%So far, we have learned that the proposed representation scheme is able to decouple the spatial structure and style information and perform local style transfer in a globally consistent fashion with sparse reinforcement. What we need to explore further is if this representation scheme is also context-sensitive that carries the potential benefit of performing faithful color transfer.

%\textcolor{black}{With the sum-to-one and the sparse constraint, the representations are context-sensitive with discriminative capacity. That is, the contexts can be reliably  distinguished, since each context consists of only a few dominant color bases with relatively large representation values. The sensitivity level of the representations to the contexts would influence the affine transformation matrix, thus is the key to realizing local style transfer. }%On the other hand, we are able to evaluate the effectiveness of the local style transfer by checking whether the representations are context-sensitive.}

%To examine
Let's examine a toy example \textcolor{black}{for an intuitive illustration of the effectiveness of the proposed representation scheme.} Fig.~\ref{fig:context} shows a pair of content-style images and the representation slices of the content and the style images, respectively. Note that each representation slice corresponds to the coefficients of a certain color basis in constructing the original content/style image at each spatial location; therefore, the representation slice related to a color basis is an image itself with pixel values ranging from 0 to 1. Take the content image as an example, Figs.~\ref{fig:context:b},~\ref{fig:context:c}, \textcolor{black}{~\ref{fig:context:d}} and ~\ref{fig:context:e} show the representation slices of the first \textcolor{black}{four} % $S_c^{\{1\}}$, the second $S_c^{\{2\}}$ and the third $S_c^{\{3\}}$
color bases, respectively. We can observe that, the dominant color basis of the sky is the first color basis because the sky is mostly highlighted in the first representation slice $S_c^{\{1\}}$, \ie, the \textcolor{black}{proportion} value of the first color basis is higher for the ``sky'' object than for other objects in the image, as shown in Fig.~\ref{fig:context:b}. Similarly, the dominant color basis of the tree is the third color basis, as shown in Fig.~\ref{fig:context:d}. This simple example clearly illustrates \textcolor{black}{that the proposed scheme is able to capture the non-local and context-sensitive representations}, where objects (or parts) with different colors are able to be differentiated by such representations. ~\textcolor{black}{When we transfer the representations of the entire image, different components of the transfer function would be activated according to the representations of the context, which means the network is able to perform diverse local style transfer in a global consistency fashion. That is, contexts that share the same color will be transferred in the same way even though they are not spatially adjacent.}
%In addition, the sparsity constraint described in Sec.~\ref{sec:proposed:context:sparse} also enhances the discriminative power of the representation. 

\subsubsection{Context-Sensitive Transfer through Mutual Discriminative Network}
\label{sec:proposed:context:mi}
%Context information is important for conducting faithful color transfer.
In addition to \textcolor{black}{the ability of extracting context-sensitive representations and performing local transfer in a global-consistent fashion}, for a context-correspondence (or semantically accurate) color transfer, the proposed scheme also needs to find the correct color matching according to the context \textcolor{black}{correspondence} of the objects. This will be achieved by the representation matching. 

%TODO:?do we need to add the discussion of how the other method match the context?
%Previous researchers either find the context correspondence based on the pre-trained supervised segmentation model~\cite{luan2017deep,li2018closed,yoo2019photorealistic} or though patch similarity measured with the high-level features extracted from pre-trained CNN~\cite{li2016combining,he2019progressive}. However, for the former one, since the segmentation is not a trivial task, the failure of it may lead to the failure of the context matching. The latter one only focuses on the local similarity, thus the matched context may not be global consistency.  

%may generate results such that the color of many patches are transferred from the same style patch~\cite{luan2017deep}

%In addition to carrying the context information from semantic perspective, for faithful (or semantically accurate) color transfer, we also need to find the correct color matching according to the context information of the objects. This will be achieved by %In the network structure, such color match is determined by 
%the representation match.

\begin{figure}[htbp]
\setlength{\abovecaptionskip}{1pt}
\setlength{\belowcaptionskip}{1pt}
	\begin{center}
		\includegraphics[width=0.7\linewidth]{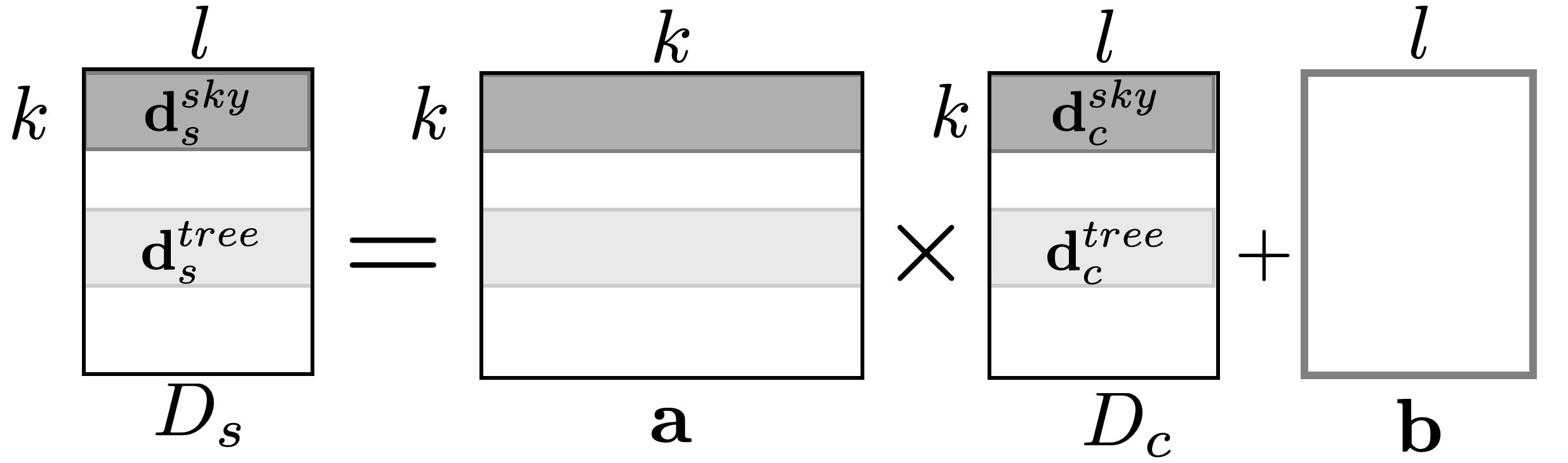}
	\end{center}
	\caption{\textcolor{black}{The affine relationship between the bases of the content image and the style image with representation matching,~\ie, $D_s = \mathbf{a}D_c+\mathbf{b}$.}}
	\label{fig:affine_dict}
\vspace{-5mm}
\end{figure}

\textcolor{black}{With the affine-transfer decoder, the affine relationship is learnt between the color bases of the content image, $D_c$, and the color bases of the style image, $D_s$, as shown in Fig.~\ref{fig:affine_dict}. As analyzed in the previous section, with the sparse constraint, each context is mainly constructed by a few dominant color bases with large representation values. %Thus, for a given context, its transfer is mainly determined by its transferred dominant color bases. 
Let us go back to the same toy example as shown in Figs.~\ref{fig:context} and~\ref{fig:affine_dict}. For the content image, the representation indicates the \textcolor{black}{third} color basis is the dominant basis of the \textcolor{black}{tree}, denoted as \textcolor{black}{$\mathbf{d}_{c}^{tree}$}. \textcolor{black}{$\mathbf{d}_{c}^{tree}$} is transferred to the \textcolor{black}{third} color basis of the style image as shown in Fig.~\ref{fig:affine_dict}. For a context-correspondence transfer, %If the transfer is correct, 
the \textcolor{black}{third} color basis of the style image should also be the dominant basis of the \textcolor{black}{tree}, \textcolor{black}{$\mathbf{d}_{s}^{tree}$}, ~\ie, the proportion of \textcolor{black}{$\mathbf{d}_{s}^{tree}$} is higher for the \textcolor{black}{``tree''} object than for other objects in the \textcolor{black}{third} representation slice \textcolor{black}{$S_s^{\{3\}}$} as shown in Fig.\textcolor{black}{~\ref{fig:context:i}}. This would then imply that  the representations of the content and style image should be matched. Let's take another more intuitive example: %To understand it from another perspective, let's assume one object consists of four different materials with different proportions. The color of the object would be a linear combination of the colors of the four materials (color bases). 
Say we take two pictures of the same object under two different lighting conditions, the color of the object would then look different, so do the corresponding color bases. However, how the color bases are combined to form the color of the object,~\ie, the proportions, would remain the same for the object.}

\textcolor{black}{In essence, to perform semantically-accurate color transfer, the distributions of the representations extracted from the same context should be similar in both the content and style images.} Such correspondence can be encouraged by maximizing the dependency between $S_c$ and $S_s$. Since our encoder is non-linear, traditional constraints like correlation may not catch such dependency. Instead, we maximize the dependency by maximizing their mutual information.

\begin{figure}[htbp]
\setlength{\abovecaptionskip}{1pt}
\setlength{\belowcaptionskip}{1pt}
	\begin{center}
		\includegraphics[width=0.6\linewidth]{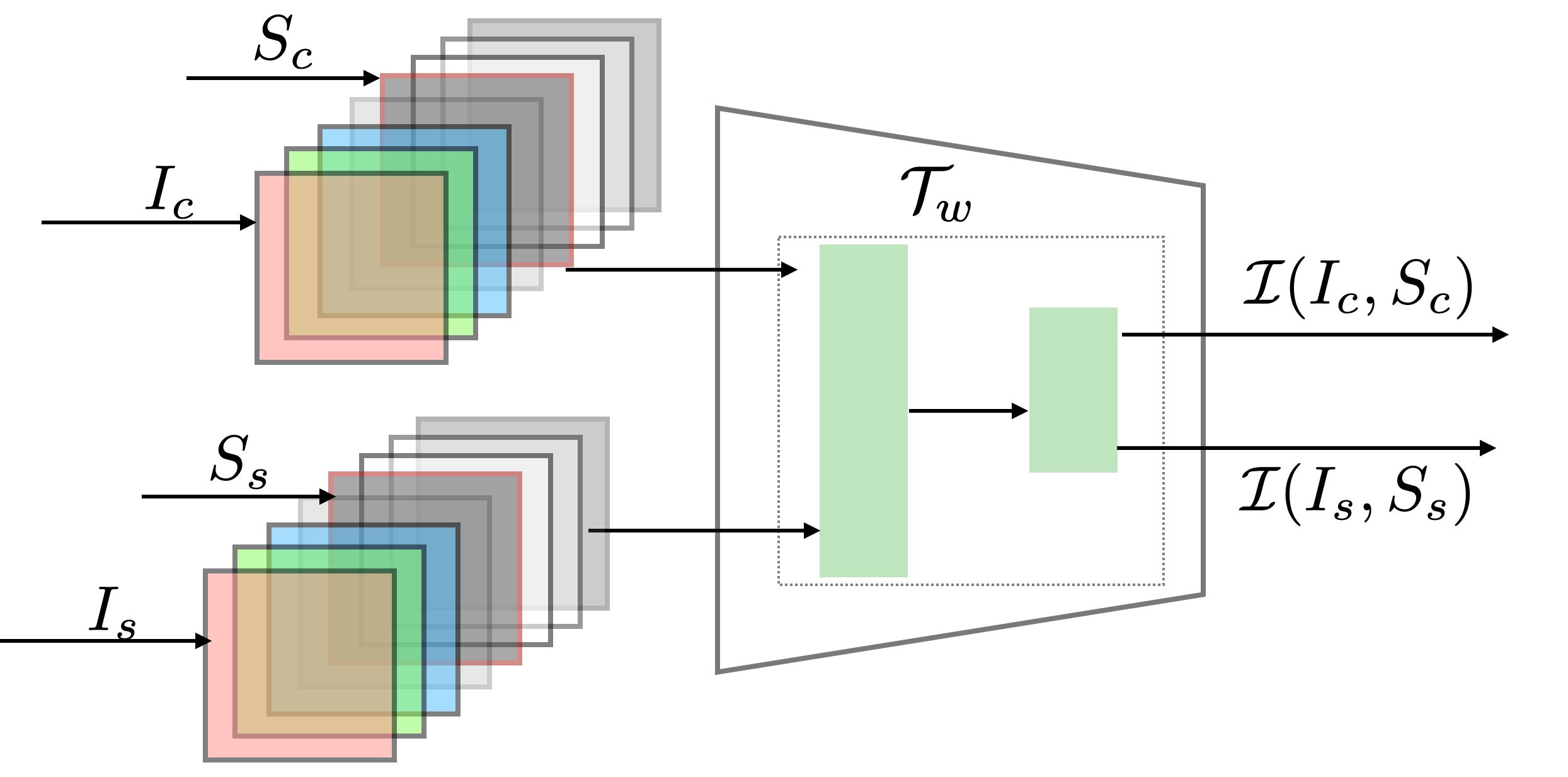}
	\end{center}
	\caption{\textcolor{black}{Structure of the mutual discriminative network.}}
	\label{fig:flow_mi}
\end{figure}

%In the proposed network structure, 
We \textcolor{black}{propose a mutual discriminative network based on mutual information to enforce the correspondence of the extracted representations. The network structure is shown in Fig.~\ref{fig:flow_mi}.}

%For similar objects in the content and style images, when the distributions of their representations are similar. 
%We go back to the same toy example as shown in Fig.~\ref{fig:context}. For the content image, the representation indicates the first content color basis is the dominant basis of the sky. Similarly, the first style color basis is the dominant basis of the sky as well. If the representations of the object in the content image match those of the style image, we could easily swap the color bases of the style image to those of the content image for semantically-accurate color transfer. Such correspondence can be encouraged by maximizing the dependency of $S_c$ and $S_s$. Since our encoder is non-linear, traditional constraints like correlation may not catch such dependency. Thus we maximize the dependency by maximizing their mutual information. 

%In the proposed network structure, we  extract representations carrying context information and such representations are matched and correspondence enforced using a mutual discriminative network based on mutual information. %between the multiple input modalities (i.e., the content and style images in our case) for faithful (or semantically accurate) color transfer. Such context correspondence is  

Mutual information (MI) has been widely used for multi-modality registrations~\cite{zitova2003image,woo2015multimodal}. It is a Shannon-entropy based measurement of mutual independence between two random variables, \eg, $S_c$ and $S_s$. The mutual information $\mathcal{I}( S_c; S_s)$ measures how much uncertainty of one variable  ($S_c$ or  $S_s$) is reduced given the other variable ($S_s$ or $S_c$). Mathematically, it is defined as 
\begin{equation}
\begin{array}{ll}
\mathcal{I}(S_c; S_s) &= H(S_c) - H(S_c \vert S_s)\\
&= \int_{\mathcal{S}_c\times\mathcal{S}_s}\log\frac{\mathbb{P}_{S_c S_s}}
{\mathbb{P}_{S_c}\otimes \mathbb{P}_{S_s}}d\mathbb{P}_{S_c S_s},
\end{array}
\end{equation}
where $H$ indicates the Shannon entropy, $H(S_c\vert S_s)$ is the conditional entropy of $S_c$ given $S_s $. $\mathbb{P}_{S_c S_s}$ is the joint probability distribution, and $\mathbb{P}_{S_c}\otimes\mathbb{P}_{S_s}$ denotes the product of marginals. %Belghazi \etal ~\cite{belghazi2018mine} introduced an MI estimator, which allows to estimate MI through neural network.

In our problem, since $S_c = \mathbf{E}_{\phi}(I_c)$ and $S_s = \mathbf{E}_{\phi}(I_s)$, their MI can also be expressed as $\mathcal{I}(\mathbf{E}_{\phi}(I_c); \mathbf{E}_{\phi}(I_s))$. However, it is difficult to maximize their dependency by maximizing their MI through MI estimator directly, because the resolution of the content and style images  might be different. Instead, we maximize the average MI between the representations and their own inputs, \ie, $\mathcal{I}(I_c, \mathbf{E}_{\phi}(I_c))$ and $\mathcal{I}(I_s, \mathbf{E}_{\phi}(I_s))$,  simultaneously, through the same discriminative network $\mathcal{T}_w$. Note that, %$S_c$ and $S_s$ are achieved by projecting the $I_c$ and $I_s$ images into the abundance space. 
in the \textcolor{black}{representation} space, $S_c$ and $S_s$ are context-sensitive, so their distributions are related to the distributions of objects in their images, not their color information. When we maximize the MI with the same $\mathcal{T}_w$,  the dependency between $I_c$ and $\mathbf{E}_{\phi}(I_c)$ would be similar to that of $I_s$ and $\mathbf{E}_{\phi}(I_s)$,~\ie, the slices of $\mathbf{E}_{\phi}(I_c)$ and $\mathbf{E}_{\phi}(I_s)$, which carry the context distribution information, are encouraged to be similar if they possess similar objects. 

Let's take $\mathcal{I}(I_c, \mathbf{E}_{\phi}(I_c))$ as an example. It is equivalent to Kullback-Leibler (KL) divergence~\cite{belghazi2018mine} between the joint distribution $\mathbb{P}_{I_c\mathbf{E}_{\phi}(I_c)}$ and the product of the marginals $\mathbb{P}_{I_c}\otimes \mathbb{P}_{\mathbf{E}_{\phi}(I_c)}$. Such MI can be maximized by maximizing the KL-divergence's lower bound based on the Donsker-Varadhan (DV) representation~\cite{donsker1983asymptotic}. Since we do not need to calculate the exact MI, we introduce an alternative lower bound based on Jensen-Shannon which works more stable than DV-based objective function~\cite{hjelm2018learning}.

\textcolor{black}{The mutual discriminative network, $\mathcal{I}_w:\mathcal{G}\times\mathcal{S}\rightarrow\mathbb{R}$, is constructed by fully-connected layers with weights $w$. The raw image and its extracted representations are stacked and fed into the network as shown in Fig.~\ref{fig:flow_mi}.}
%In the network design, the mutual discriminative network 
% is built with parameter $w$. 
The MI estimator can be defined as
\begin{equation}
\label{equ:dvmi}
\begin{array}{ll}
\mathcal{I}_{\phi,w}(I_c, \mathbf{E}_{\phi}(I_c)) &=
\mathbb{E}_{\mathbb{P}}[-sp(-\mathcal{T}_{\phi,w}(I_c,\mathbf{E}_{\phi}(I_c)))]\\
&- \mathbb{E}_{\mathbb{P}\times\tilde{\mathbb{P}}}[sp(\mathcal{T}_{\phi,w}(I_c',\mathbf{E}_{\phi}(I_c)))],\\
\end{array}
\end{equation}
where $sp(x) =\log(1+e^x) $ and $I_c'$ is an input sampled from $\tilde{\mathbb{P}}=\mathbb{P}$ by randomly shuffling the input data. The term carrying the shuffling data is called the negative sample. 
%Since our input only has two images, it is unstable to train the network with random shifting input data. Thus, we simplify the network without taking the negative samples in Eq.~\eqref{equ:dvmi}.
Combined with the MI of $I_s$, our objective function is defined as
\begin{equation}
\label{optmiall}
\mathcal{L}_{\mathcal{I}}(\phi,w) = \mathcal{I}_{\phi,w}(I_c, \mathbf{E}_{\phi}(I_c)) + \mathcal{I}_{\phi,w}(I_s, \mathbf{E}_{\phi}(I_s))
%\begin{array}{ll}
%\mathcal{L}_{\mathcal{I}}(\phi,w) &= (\mathbb{E}_{\mathbb{P}}[-sp(-\mathcal{T}_{w,\phi}(I_c,\mathbf{E}_{\phi}(I_c))])\\
%&+ \mathbb{E}_{\mathbb{P}}[-sp(-\mathcal{T}_{w,\phi}(I_s,\mathbf{E}_{\phi}(I_s))]).
%\end{array}
\end{equation}

By maximizing $\mathcal{L}_{\mathcal{I}}(\phi,w)$, we could extract optimized representations $S_c$ and $S_s$ that can best represent $I_c$ and $I_s$, and the slices of $S_c$ and $S_s$ are ordered in a similar way as shown in Fig.~\ref{fig:context}. For example, the \textcolor{black}{third} slice of $S_c$ carries the spatial distribution information of the \textcolor{black}{tree}, and the \textcolor{black}{third} slice of $S_s$ also carries the distribution of the \textcolor{black}{tree} object in the style image. Hence $S_c$ and $S_s$ have been encouraged to be matched semantically, achieving context-sensitive representation. 

\begin{figure}
\setlength{\abovecaptionskip}{1pt}
\setlength{\belowcaptionskip}{1pt}
	\begin{center}
		\includegraphics[width=1\linewidth]{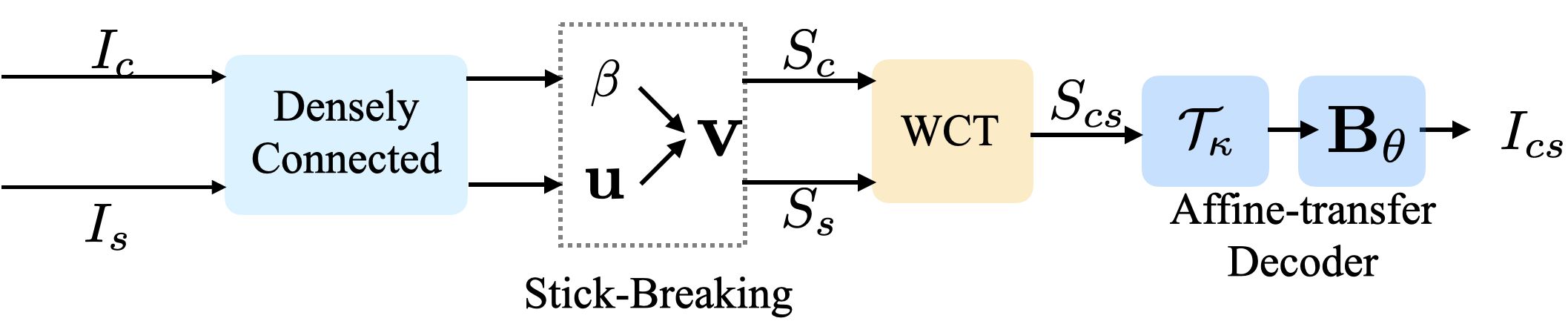}
	\end{center}
	\caption{Flowchart of the style transfer procedure.}
	\label{fig:flow_trasnfer}
\vspace{-5mm}
\end{figure}

\subsection{Style Transfer and Implementation Details}
\label{sec:proposed:opt}

\subsubsection{Style Transfer with WCT and Affine-Transfer Decoder}
\label{sec:proposed:opt:tr}
%Previous researchers either transfer color within a region according to its semantic labels~\cite{luan2017deep,li2018closed,yoo2019photorealistic}, or between the most likely patches~\cite{li2016combining,he2019progressive}. However, the former is prone to introducing abrupt color changes especially at the border between adjacent regions; and the latter only focuses on the local similarity which may generate results such that the color of many patches are transferred from the same style patch~\cite{luan2017deep}. 

%With the structure shown at the upper part of Fig.~\ref{fig:flow}, the network learns to extract context-sensitive spatial representations and transferred color bases during spatial and color information decoupling. 
As analyzed in Sec.~\ref{sec:proposed:context}, the learned color bases embed the transfer information between the content and style images, thus we can achieve preliminary style transfer results by feeding the representation of the content image to the affine-transfer decoder of the style image. 

In the ideal case, the distributions of the matched representations $S_c$ and $S_s$ would be similar for the same type of context. Depending on the similarity between the content and the style images, the matching, however, might not be perfect due to content differences between the content and the style images. In order to further match the statistic characteristics  of $S_c$ to that of $S_s$, we adopt the classical signal whitening and coloring transforms (WCTs) approach~\cite{li2017universal}, which changes the covariance of $S_c$ to that of $S_s$. The toy example in Fig.~\ref{fig:reason} demonstrates that WCT could move the representations in $S_c$ to the matched representations in $S_s$ effectively. \textcolor{black}{Then the transferred representations} $S_{cs}$ is fed into the style decoder to generate the stylization image $I_{cs}$. The style transfer procedure is illustrated in Fig.~\ref{fig:flow_trasnfer}, \textcolor{black}{and it will be further demonstrated} using a proposed visualization mechanism in Sec.~\ref{sec:evaluationdecoupling}.

\subsubsection{ Implementation Details}
\label{sec:proposed:opt:dt}
In order to extract better color bases, we adopt the $l_{2,1}$ norm \cite{nie2010efficient} instead of the traditional $l_2$ norm for reconstruction loss. The objective function for $l_{2,1}$ loss is defined as 
\begin{equation}
\begin{array}{ll}
\mathcal{L}_{2,1}(\phi,\psi)  &=  \Vert D_{\psi}(E_\phi(I_c))- I_c\Vert_{2,1}\\
&+\Vert D_{\psi}(E_\phi(I_s))- I_s\Vert_{2,1},
\end{array}
\end{equation}
where $\Vert X \Vert_{2,1} = \sum_{i=1}^{m}\sqrt{\sum_{j=1}^{n}X_{i,j}^2}$. $l_{2,1}$ \textcolor{black}{can be treated as first applying $l_2$ norm on each pixel, then applying $l_1$ norm to enforce the reconstruction errors on the entire image to be sparse. In this way, it will enforce most of the reconstruction errors of individual pixels to be zero.} That is, the network is designed to learn individual pixels as accurate as possible, which would extract better color bases to further facilitate  the style transfer.

The objective function of the proposed network architecture can then be expressed as:
\begin{equation}
\label{equ:optall}
\mathcal{L}(\phi,\psi,w) = \mathcal{L}_{2,1}(\phi,\psi) + \alpha\mathcal{L}_\mathcal{H}(\phi)- \lambda\mathcal{L}_{\mathcal{I}}(\phi,w),
\end{equation}
where $\alpha$ and $\lambda$ are the parameters that balance the trade-off among the reconstruction error, the sparse loss, and the negative of mutual information. The network is optimized with back-propagation as illustrated in Fig.~\ref{fig:flow} with red-dashed lines. \textcolor{black}{More details of the WCT transfer and network structure are described in Sec.1 of the supplementary file.}

%$l_2$ norm is applied on the decoder weights $\psi$ to prevent over-fitting.
%Note that $\mathcal{L}_{\mathcal{I}}(\phi,w)$  is achieved by first reshaping the representations $S_C$ and $S_S$ to 3D domain according to their image size, then stacking  them on top of $I_C$ and $I_S$, respectively, before feeding into the network $\mathcal{T}_w$.

%Before style transfer, the content and style images can be down-sampled to a smaller size for efficiency and transformed to zero-mean vectors by subtracting from their own means. Because the down-sampled image pair still have similar distributions as the input image pair, the learned weights can be used to generate the stylized image with original resolution. 

%The network consists of a few fully-connected layers. The number of nodes for different layers are shown in Table~\ref{tab:layers}. It is optimized with back-propagation as illustrated in Fig.~\ref{fig:flow} with red-dashed lines. 

%\begin{table}[htb]
%	\caption{The number of layers and nodes in the proposed network.}
%	\label{tab:layers}
%	\begin{center}
%		\begin{tabular}{c|cccc}
%			\hline
%			&$\mathbf{E}_\phi$/$\mathbf{u}$/${\beta}$&$\mathcal{T}_w$&$\mathcal{T}_\tau/$$\mathcal{T}_\kappa$&$\mathbf{B_\theta}$\\
%			\hline
%			$\#$layers &4/1/1&2&1/1&2\\
%			$\#$nodes &[3,3,3,3]/10/1&[13,1]&10/10&[10,10]\\
%			\hline
%		\end{tabular}
%	\end{center}
%\end{table}

\section{Experimental Results}
%TODO:should focus more on the context and global consistency
\label{sec:results}
The stylization results of the proposed NL-MAT on various types of photos \textcolor{black}{in two datasets from ~\cite{luan2017deep} and \cite{an2020ultrafast}} are compared with those from the state-of-the-art methods, including two patch-similarity based methods~\cite{liao2017visual,he2019progressive}, four context-based methods~\cite{luan2017deep,li2018closed,li2018learning,yoo2019photorealistic}, and PhotoNAS~\cite{an2020ultrafast}. For the methods we compare to, we either use the published results provided by the authors or generate the results from published pre-trained models. 
Both visual comparisons (Sec.~\ref{sec:viscomp}) and user study results (Sec.~\ref{sec:userstudy}) are provided to evaluate the effectiveness of the proposed method both qualitatively and quantitatively. Experiments are also conducted to show the important role \textcolor{black}{of the non-local representation scheme (Sec.~\ref{sec:evaluationdecoupling})} %that the \textcolor{black}{non-local}-based representation plays (Sec.~\ref{sec:evaluationdecoupling}) 
as well as \textcolor{black}{its} contributing components, \ie, the affine-transfer decoder (Sec.~\ref{sec:evaluate_affine}) and the mutual-discriminative network (Sec.~\ref{sec:ablation2}). The effect of the number of color bases is discussed in Sec.~\ref{sec:numofbs}.
\textcolor{black}{Computational efficiency and some failure cases are  discussed and analyzed in Secs.2 and 3 of the supplementary file, respectively.}%is compared in Sec.~\ref{sec:efficiency}.

\subsection{Visual Comparison}
\label{sec:viscomp}
\begin{figure*}
\setlength{\abovecaptionskip}{1pt}
\setlength{\belowcaptionskip}{1pt}
	\centering
	\begin{minipage}{1\linewidth}
		{\includegraphics[width=0.16\linewidth]{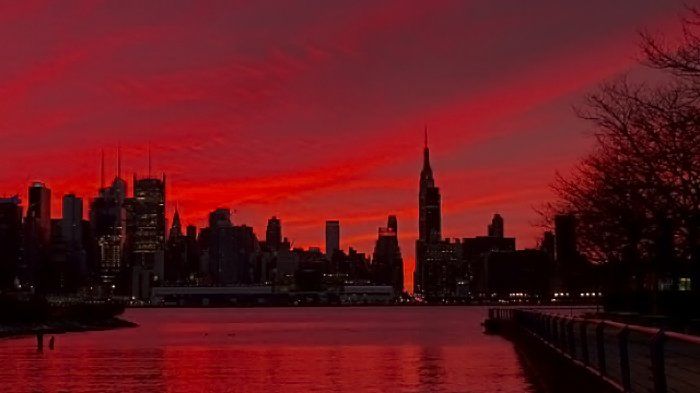}
			\label{fig:patch:a1}}
		{\includegraphics[width=0.16\linewidth]{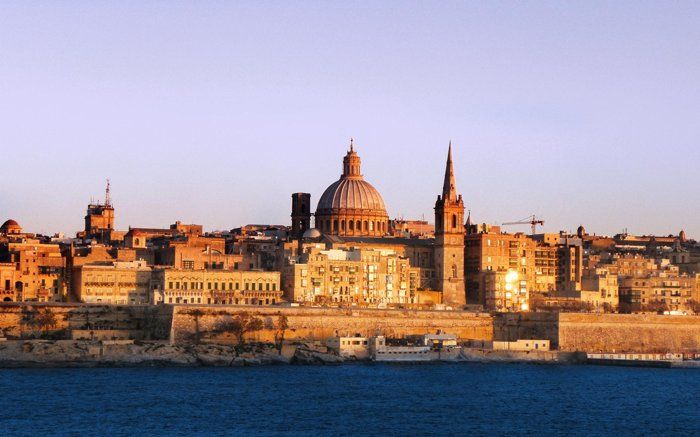}
			\label{fig:patch:b1}}\hspace{-2.4mm}
		\adjincludegraphics[width=0.04\linewidth, trim={{.4\width} {.225\width} {.5\width} {.15\width}},clip]{fig/compare_patch/1_0in}
		{\includegraphics[width=0.16\linewidth]{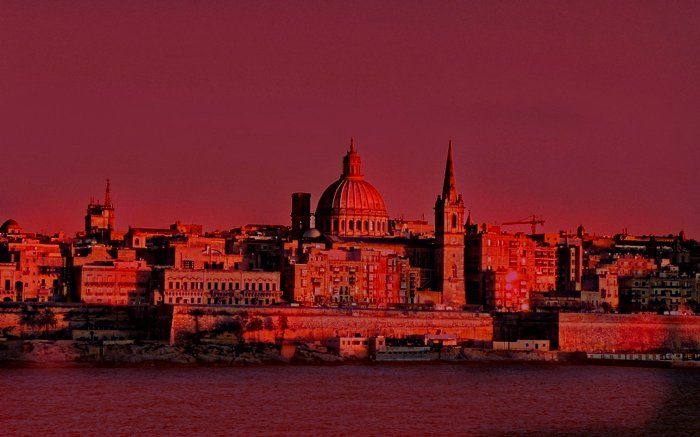}
			\label{fig:patch:c1}}\hspace{-2.4mm}
		\adjincludegraphics[width=0.04\linewidth, trim={{.4\width} {.225\width} {.5\width} {.15\width}},clip]{fig/compare_patch/1_liao}
		{\includegraphics[width=0.16\linewidth]{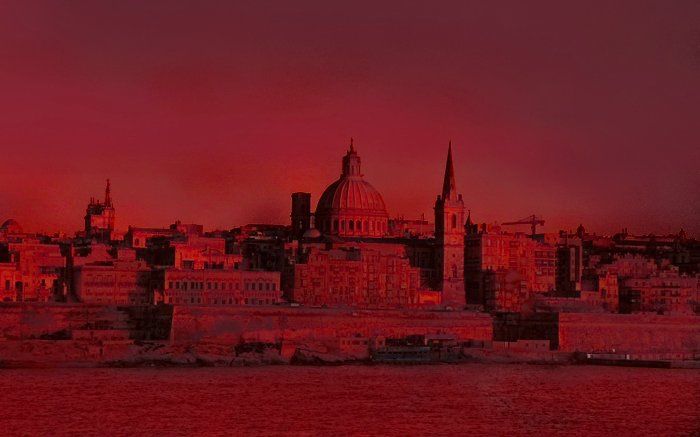}
			\label{fig:patch:d1}}\hspace{-2.4mm}
		\adjincludegraphics[width=0.04\linewidth, trim={{.4\width} {.225\width} {.5\width} {.15\width}},clip]{fig/compare_patch/1_he}
		{\includegraphics[width=0.16\linewidth]{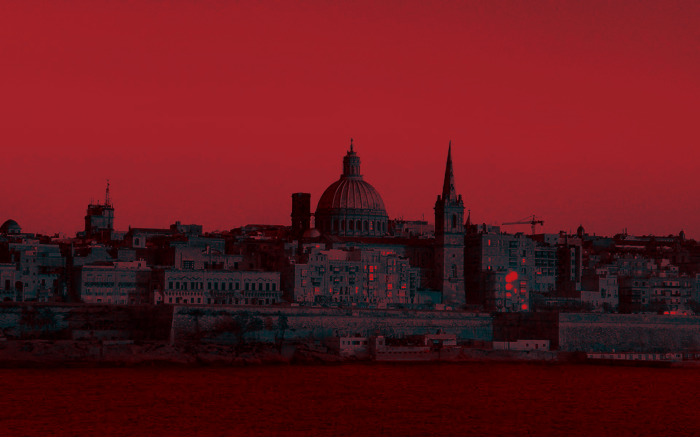}
			\label{fig:patch:e1}}\hspace{-2.4mm}
		\adjincludegraphics[width=0.04\linewidth, trim={{.4\width} {.225\width} {.5\width} {.15\width}},clip]{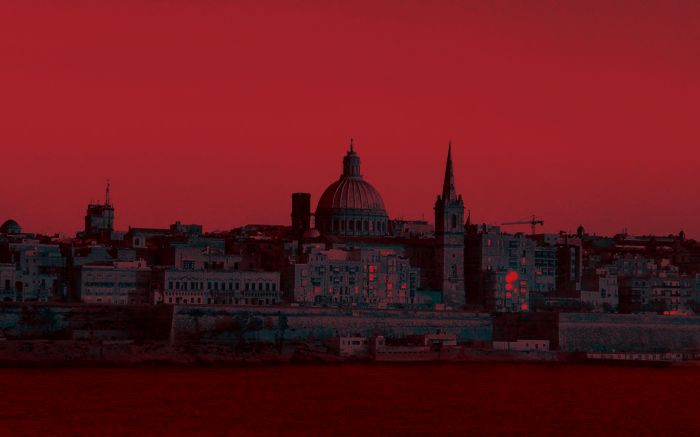}\hfill
	\end{minipage}\\
	\begin{minipage}{1\linewidth}
		{\includegraphics[width=0.16\linewidth]{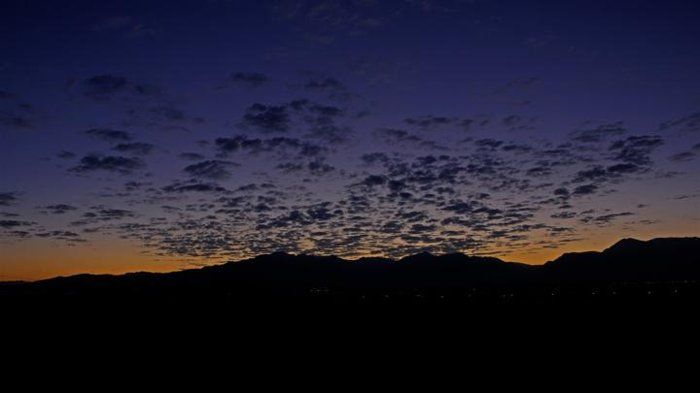}
			\label{fig:patch:a2}}
		{\includegraphics[width=0.16\linewidth]{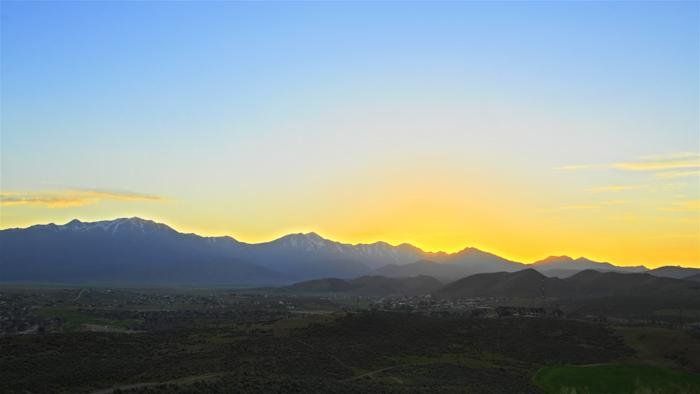}
			\label{fig:patch:b2}}\hspace{-2.4mm}
		\adjincludegraphics[width=0.04\linewidth, trim={0 {0.05\width} {.839\width} {0.15\width}},clip]{fig/compare_patch/2_0in}
		{\includegraphics[width=0.16\linewidth]{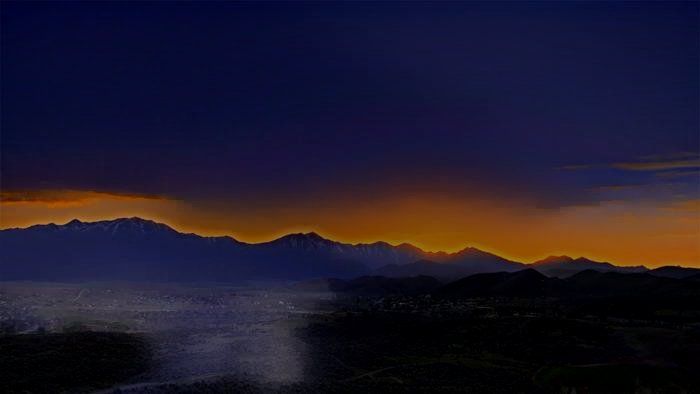}
			\label{fig:patch:c2}}\hspace{-2.4mm}
		\adjincludegraphics[width=0.04\linewidth, trim={0 {0.05\width} {.839\width} {0.15\width}},clip]{fig/compare_patch/2_liao}
		{\includegraphics[width=0.16\linewidth]{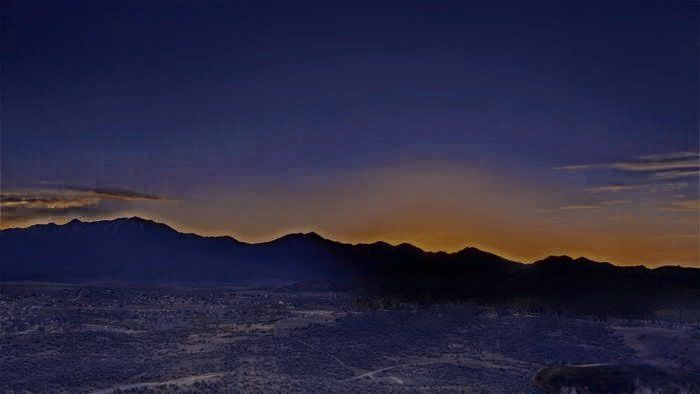}
			\label{fig:patch:d2}}\hspace{-2.4mm}
		\adjincludegraphics[width=0.04\linewidth, trim={0 {0.05\width} {.839\width} {0.15\width}},clip]{fig/compare_patch/2_he}
		{\includegraphics[width=0.16\linewidth]{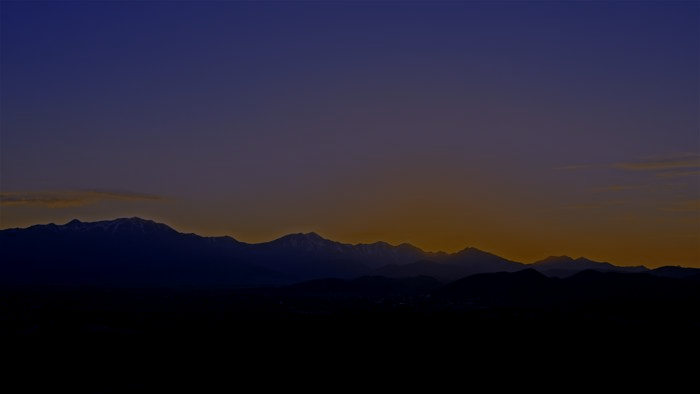}
			\label{fig:patch:e2}}\hspace{-2.4mm}
		\adjincludegraphics[width=0.04\linewidth, trim={0 {0.05\width} {.839\width} {0.15\width}},clip]{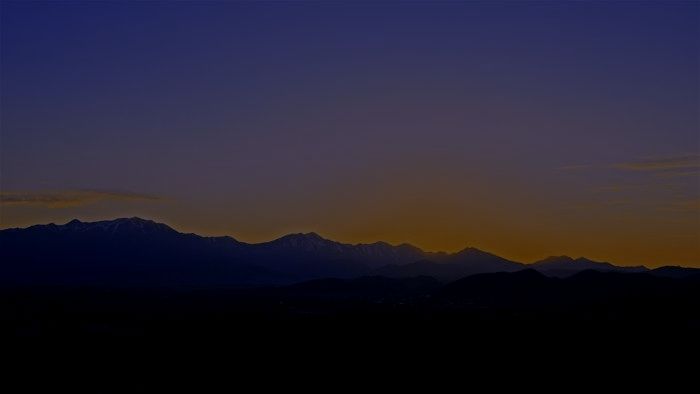}
	\end{minipage}\\
	\begin{minipage}{1\linewidth}
		{\includegraphics[width=0.16\linewidth]{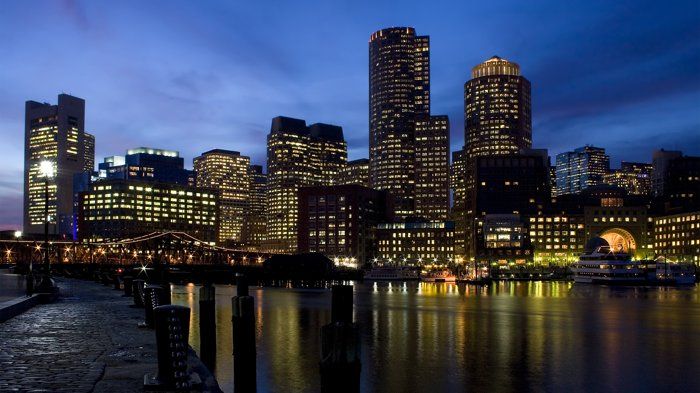}
			\label{fig:patch:a3}}
		{\includegraphics[width=0.16\linewidth]{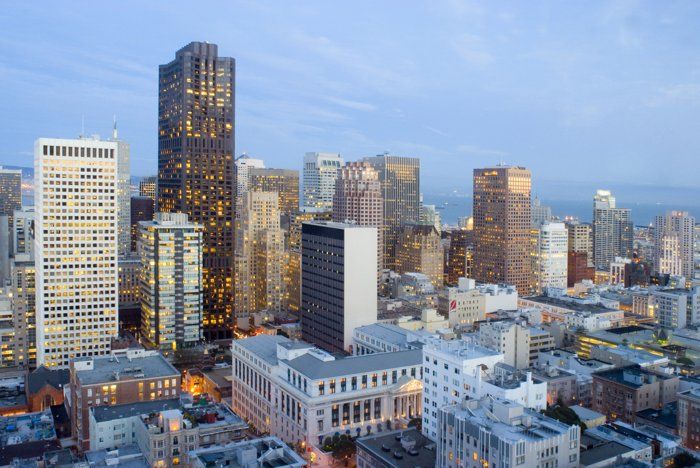}
			\label{fig:patch:b3}}\hspace{-2.4mm}
		\adjincludegraphics[width=0.04\linewidth, trim={{0.05\width} {.2\width} {.8\width} {.07\width}},clip]{fig/compare_patch/3_0in}
		{\includegraphics[width=0.16\linewidth]{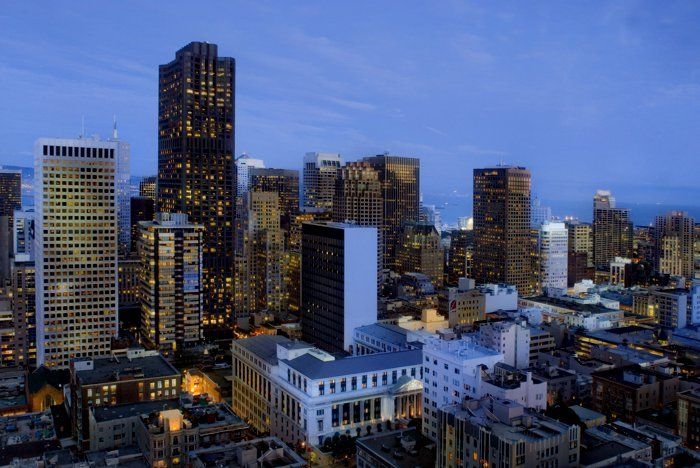}
			\label{fig:patch:c3}}\hspace{-2.4mm}
		\adjincludegraphics[width=0.04\linewidth, trim={{0.05\width} {.2\width} {.8\width} {.07\width}},clip]{fig/compare_patch/3_liao}
		{\includegraphics[width=0.157\linewidth]{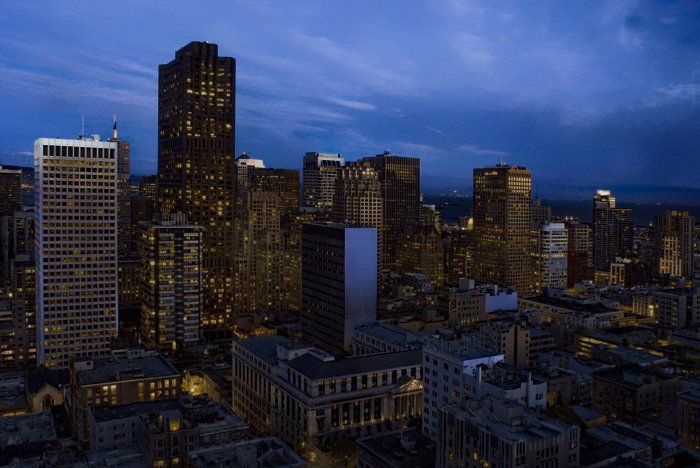}
			\label{fig:patch:d3}}\hspace{-2.4mm}
		\adjincludegraphics[width=0.0392\linewidth, trim={{0.05\width} {.2\width} {.8\width} {.07\width}},clip]{fig/compare_patch/3_he}
		{\includegraphics[width=0.16\linewidth]{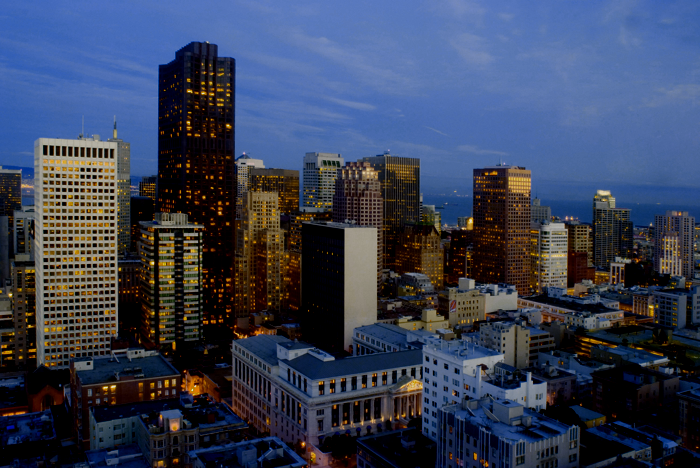}
			\label{fig:patch:e3}}\hspace{-2.4mm}
		\adjincludegraphics[width=0.04\linewidth, trim={{0.05\width} {.2\width} {.8\width} {.07\width}},clip]{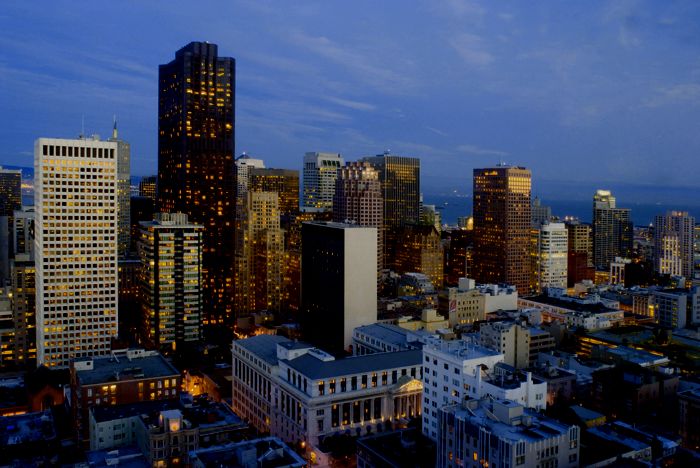}
	\end{minipage}
	\begin{minipage}{1\linewidth}
		{\includegraphics[width=0.16\linewidth]{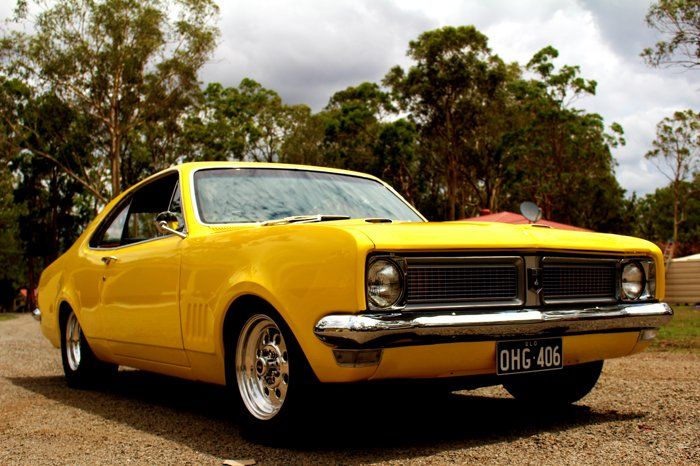}
			\label{fig:patch:a4}}
		{\includegraphics[width=0.16\linewidth]{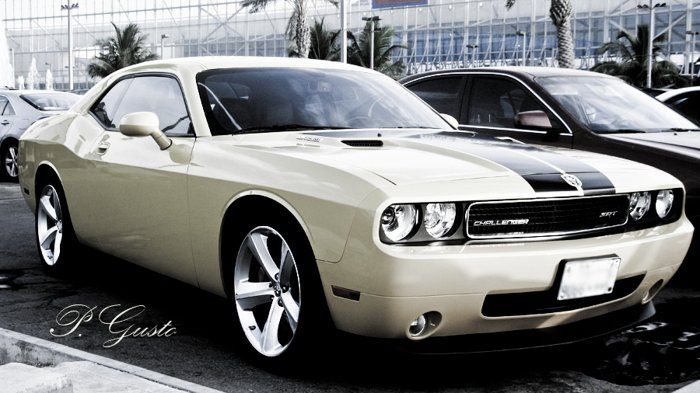}
			\label{fig:patch:b4}}\hspace{-2.4mm}
		\adjincludegraphics[width=0.04\linewidth, trim={{0.15\width} {.21\width} {.71\width} {.04\width}},clip]{fig/compare_patch/29_0in}
		{\includegraphics[width=0.16\linewidth]{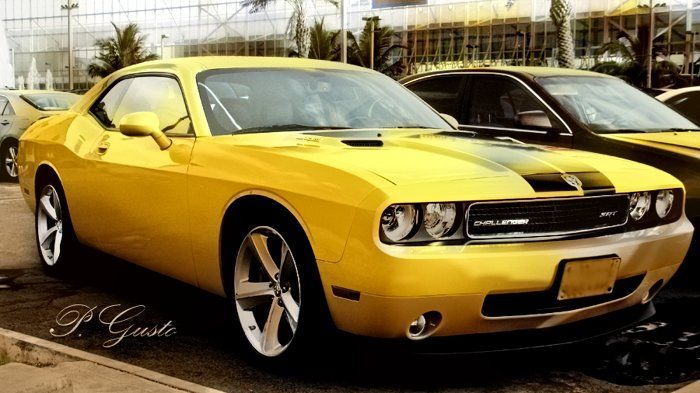}
			\label{fig:patch:c4}}\hspace{-2.4mm}
		\adjincludegraphics[width=0.04\linewidth, trim={{0.15\width} {.21\width} {.71\width} {.04\width}},clip]{fig/compare_patch/29_liao}
		{\includegraphics[width=0.16\linewidth]{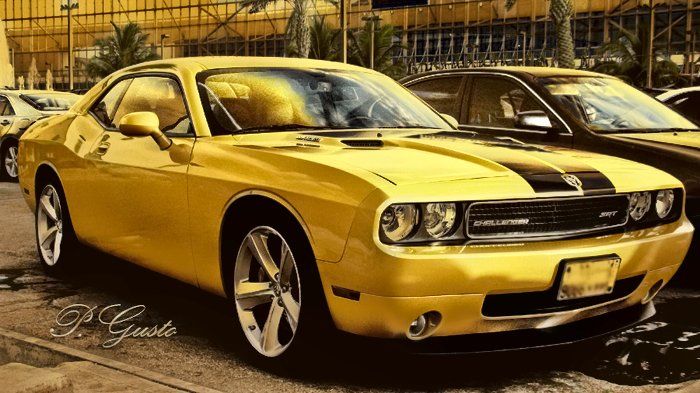}
			\label{fig:patch:d4}}\hspace{-2.4mm}
		\adjincludegraphics[width=0.04\linewidth, trim={{0.15\width} {.21\width} {.71\width} {.04\width}},clip]{fig/compare_patch/29_he}
		{\includegraphics[width=0.16\linewidth]{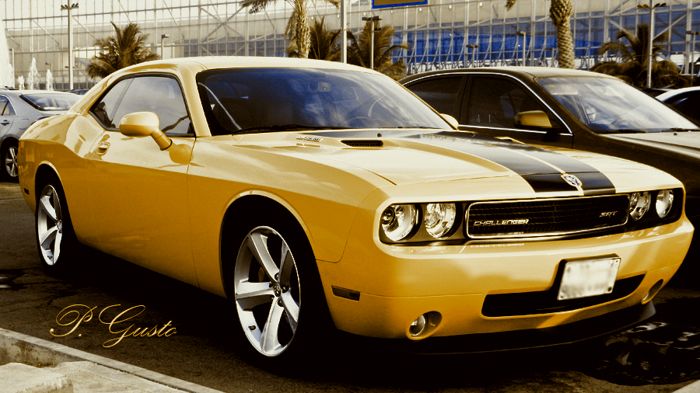}
			\label{fig:patch:e4}}\hspace{-2.4mm}
		\adjincludegraphics[width=0.04\linewidth, trim={{0.15\width} {.21\width} {.71\width} {.04\width}},clip]{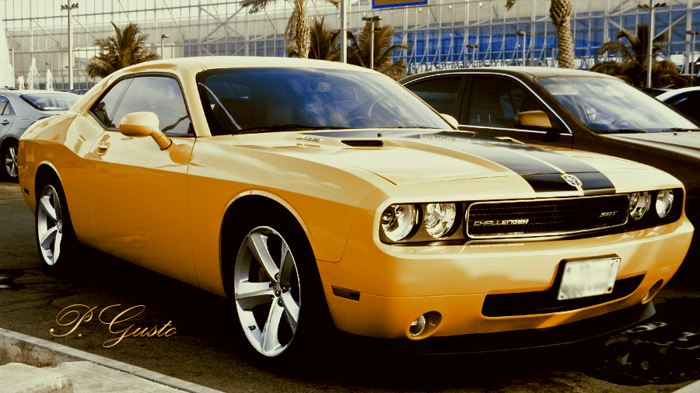}
	\end{minipage}
	\begin{minipage}{1\linewidth}
		{\includegraphics[width=0.16\linewidth]{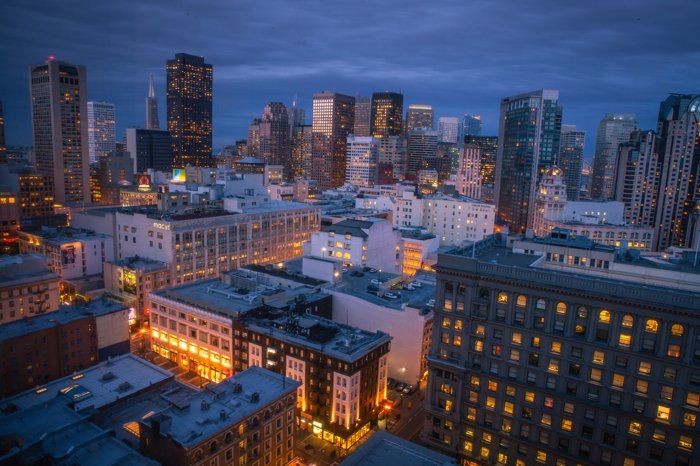}
			\label{fig:patch:a5}}
		{\includegraphics[width=0.16\linewidth]{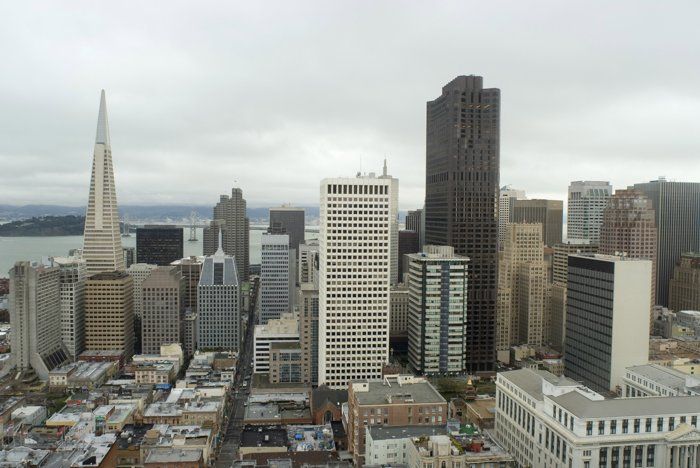}
			\label{fig:patch:b5}}\hspace{-2.4mm}
		\adjincludegraphics[width=0.04\linewidth, trim={{0.1\width} {.1\width} {.7\width} {.03\width}},clip]{fig/compare_patch/31_0in}
		{\includegraphics[width=0.16\linewidth]{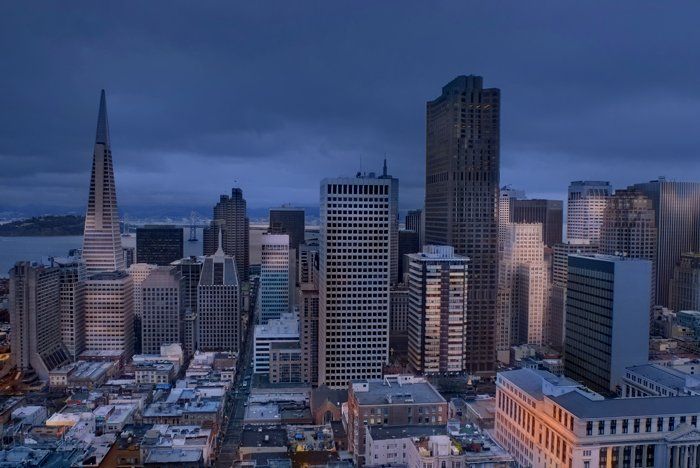}
			\label{fig:patch:c5}}\hspace{-2.4mm}
		\adjincludegraphics[width=0.04\linewidth, trim={{0.1\width} {.1\width} {.7\width} {.035\width}},clip]{fig/compare_patch/31_liao}
		{\includegraphics[width=0.16\linewidth]{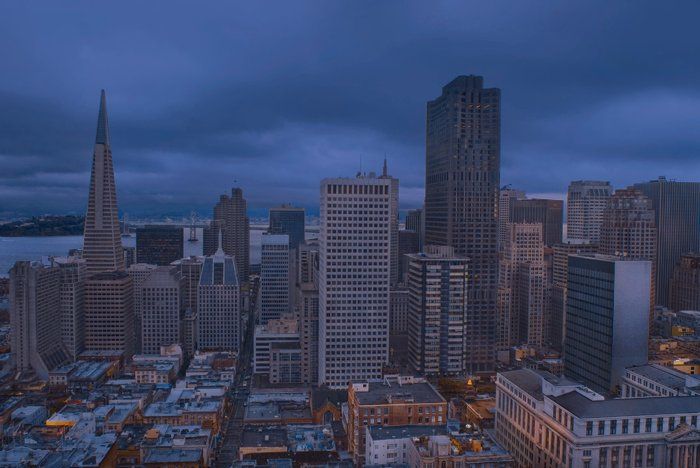}
			\label{fig:patch:d5}}\hspace{-2.4mm}
		\adjincludegraphics[width=0.04\linewidth, trim={{0.1\width} {.1\width} {.7\width} {.035\width}},clip]{fig/compare_patch/31_he}
		{\includegraphics[width=0.16\linewidth]{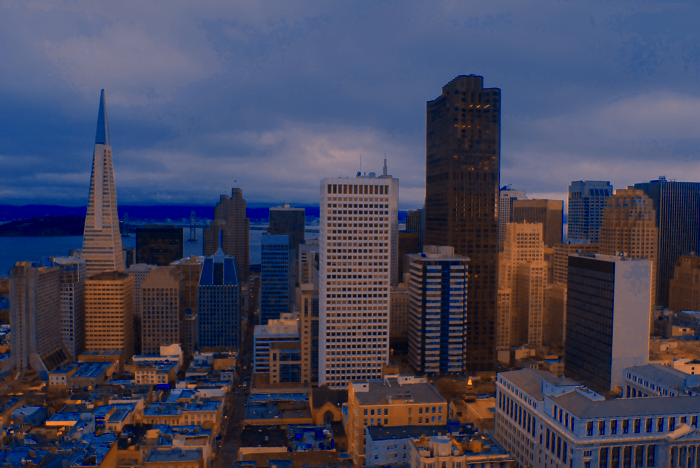}
			\label{fig:patch:e5}}\hspace{-2.4mm}
		\adjincludegraphics[width=0.04\linewidth, trim={{0.1\width} {.1\width} {.7\width} {.035\width}},clip]{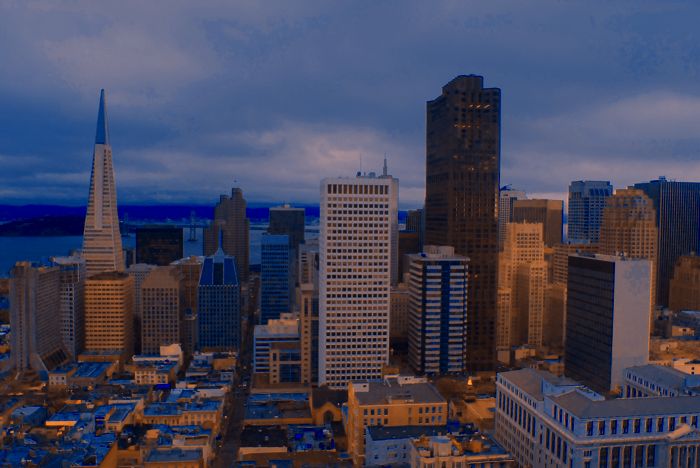}
	\end{minipage}
	\begin{minipage}{1\linewidth}
		{\includegraphics[width=0.16\linewidth]{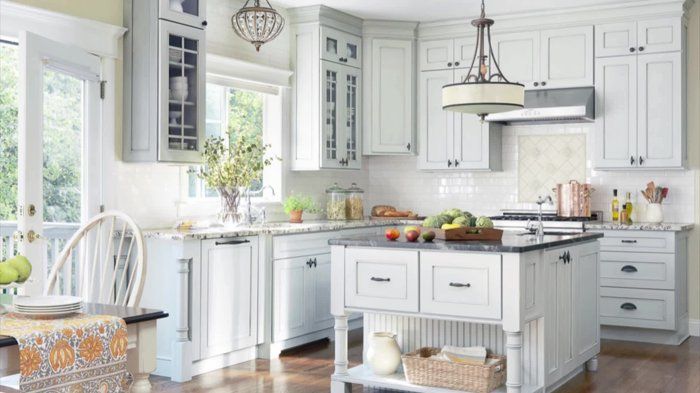}
			\label{fig:patch:a6}}
		{\includegraphics[width=0.16\linewidth]{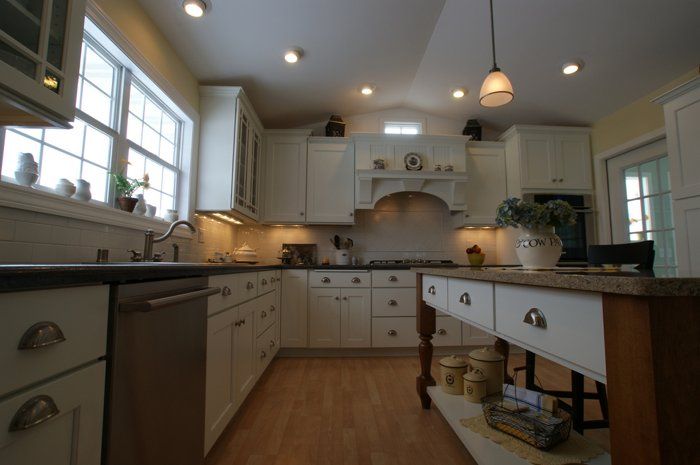}
			\label{fig:patch:b6}}\hspace{-2.4mm}
		\adjincludegraphics[width=0.04\linewidth, trim={{0.67\width} {.25\width} {.18\width} {.015\width}},clip]{fig/compare_patch/40_0in}
		{\includegraphics[width=0.16\linewidth]{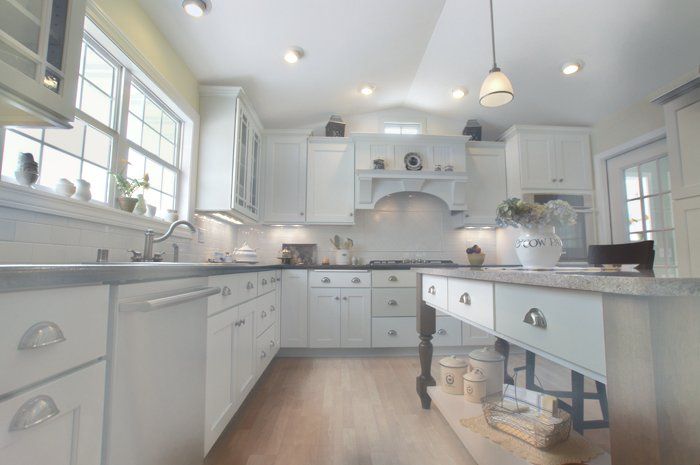}
			\label{fig:patch:c6}}\hspace{-2.4mm}
		\adjincludegraphics[width=0.04\linewidth, trim={{0.67\width} {.25\width} {.18\width} {.015\width}},clip]{fig/compare_patch/40_liao}
		{\includegraphics[width=0.16\linewidth]{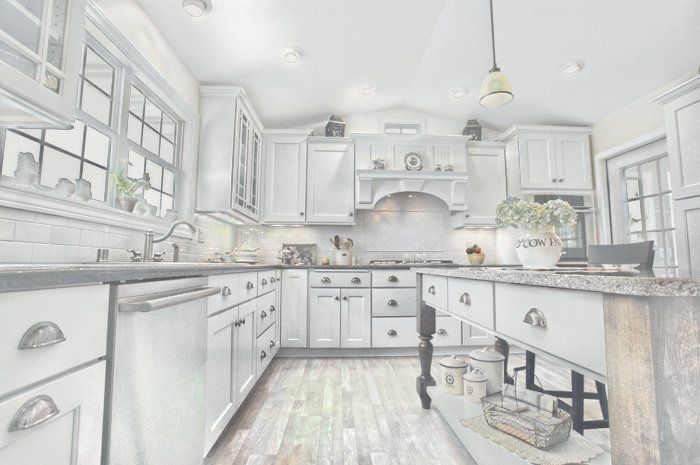}
			\label{fig:patch:d6}}\hspace{-2.4mm}
		\adjincludegraphics[width=0.04\linewidth, trim={{0.67\width} {.25\width} {.18\width} {.015\width}},clip]{fig/compare_patch/40_he}
		{\includegraphics[width=0.16\linewidth]{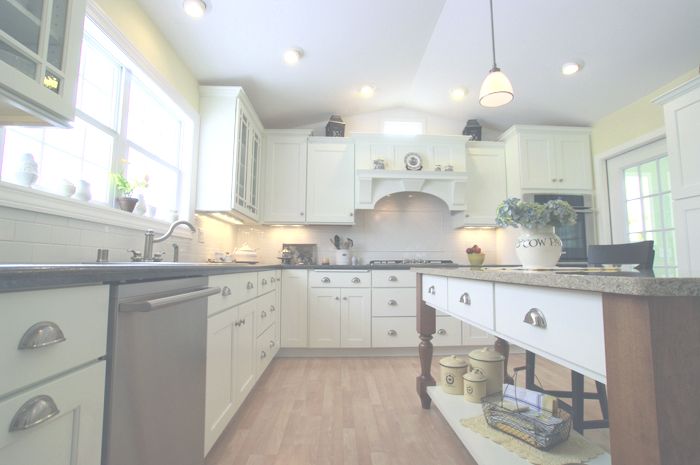}
			\label{fig:patch:e6}}\hspace{-2.4mm}
		\adjincludegraphics[width=0.04\linewidth, trim={{0.67\width} {.25\width} {.18\width} {.015\width}},clip]{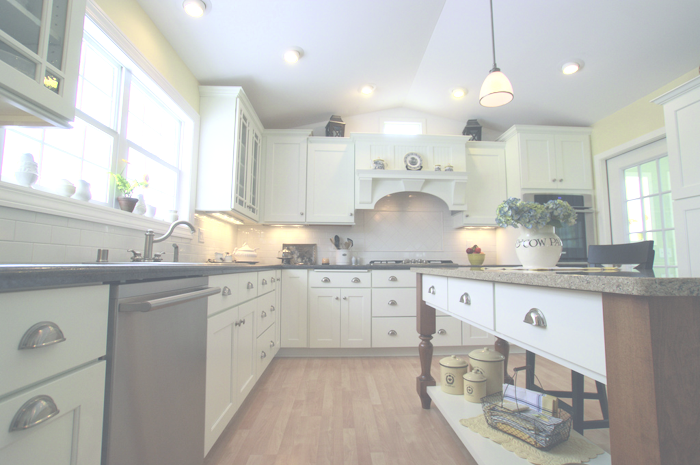}
	\end{minipage}
	\begin{minipage}{1\linewidth}
		{\includegraphics[width=0.16\linewidth]{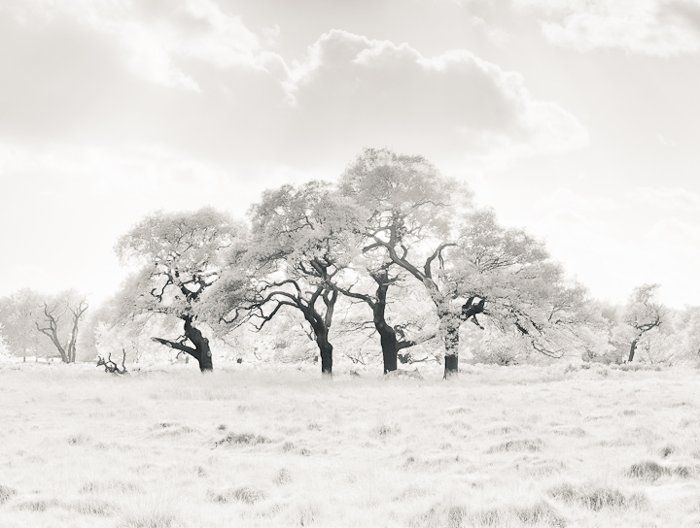}
			\label{fig:patch:a8}}
		{\includegraphics[width=0.16\linewidth]{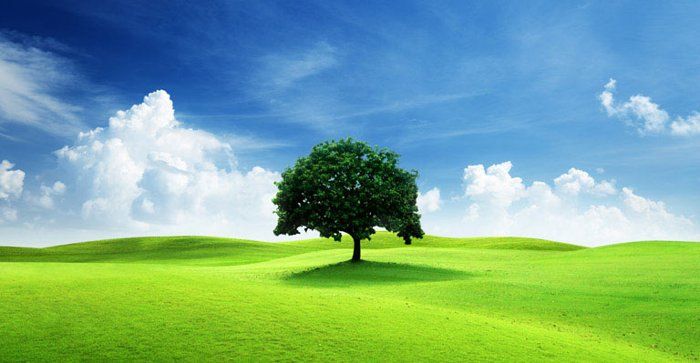}
			\label{fig:patch:b8}}\hspace{-2.4mm}
		\adjincludegraphics[width=0.04\linewidth, trim={{0.4\width} {.1\width} {.445\width} {.1\width}},clip]{fig/compare_patch/48_0in}
		{\includegraphics[width=0.16\linewidth]{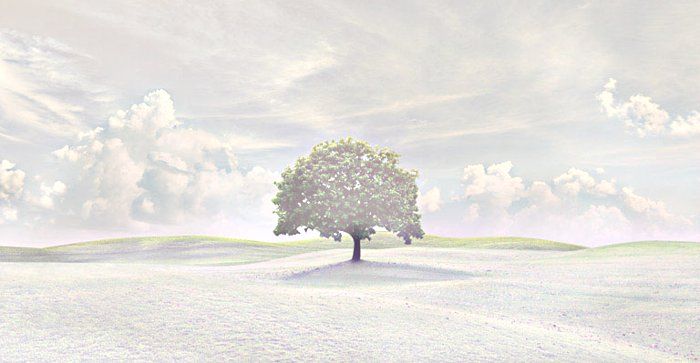}
			\label{fig:patch:c8}}\hspace{-2.4mm}
		\adjincludegraphics[width=0.04\linewidth, trim={{0.4\width} {.1\width} {.445\width} {.1\width}},clip]{fig/compare_patch/48_liao}
		{\includegraphics[width=0.16\linewidth]{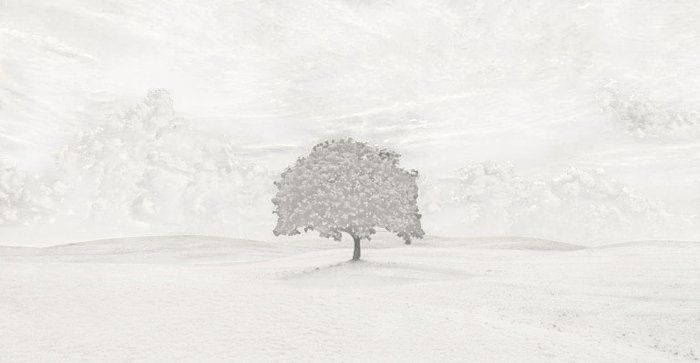}
			\label{fig:patch:d8}}\hspace{-2.4mm}
		\adjincludegraphics[width=0.04\linewidth, trim={{0.4\width} {.1\width} {.445\width} {.1\width}},clip]{fig/compare_patch/48_he}
		{\includegraphics[width=0.16\linewidth]{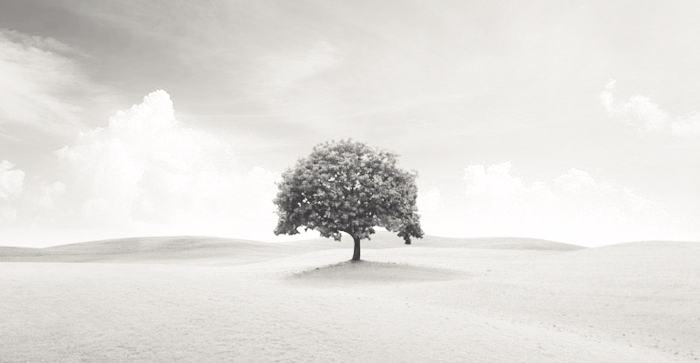}
			\label{fig:patch:e8}}\hspace{-2.4mm}
		\adjincludegraphics[width=0.04\linewidth, trim={{0.4\width} {.1\width} {.445\width} {.1\width}},clip]{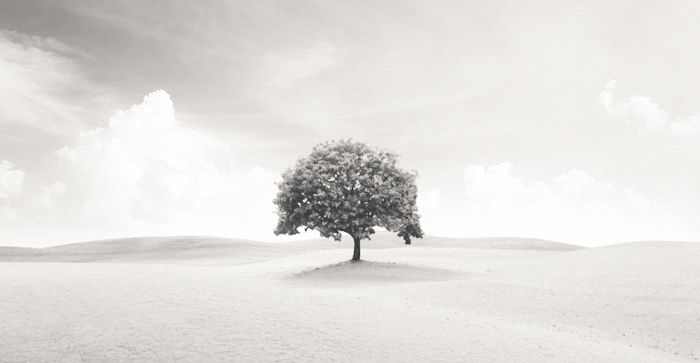}
	\end{minipage}
	\begin{minipage}{1\linewidth}
		{\includegraphics[width=0.16\linewidth]{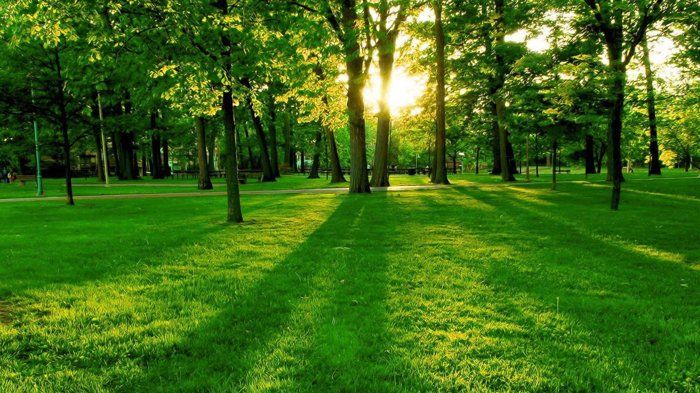}
			\label{fig:patch:a9}}
		{\includegraphics[width=0.16\linewidth]{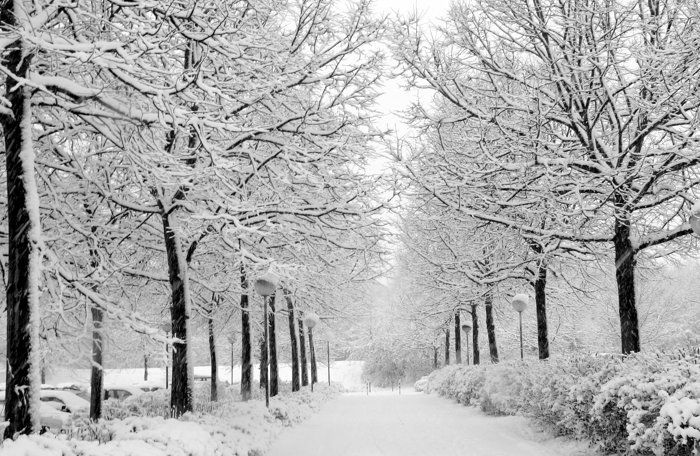}
			\label{fig:patch:b9}}\hspace{-2.4mm}
		\adjincludegraphics[width=0.04\linewidth, trim={{0.45\width} {.13\width} {.35\width} {.0\width}},clip]{fig/compare_patch/56_0in}
		{\includegraphics[width=0.16\linewidth]{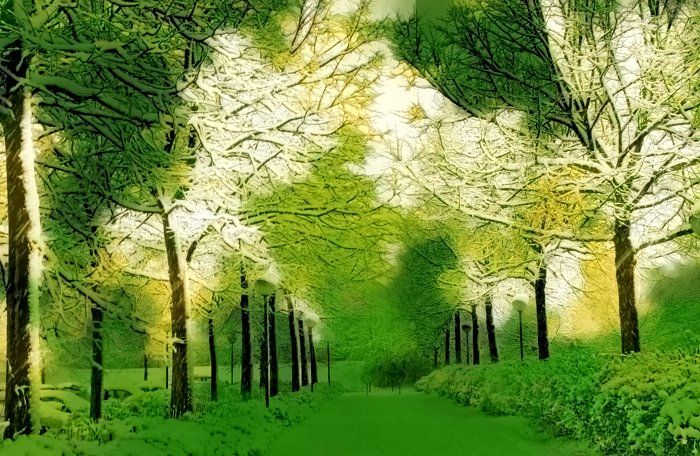}
			\label{fig:patch:c9}}\hspace{-2.4mm}
		\adjincludegraphics[width=0.04\linewidth, trim={{0.45\width} {.13\width} {.35\width} {.0\width}},clip]{fig/compare_patch/56_liao}
		{\includegraphics[width=0.16\linewidth]{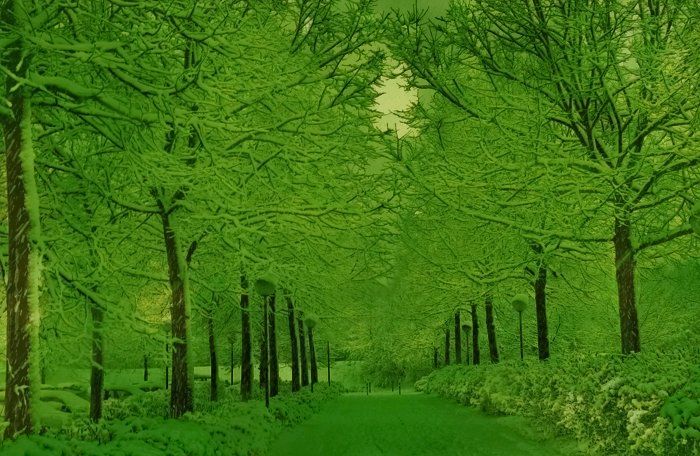}
			\label{fig:patch:d9}}\hspace{-2.4mm}
		\adjincludegraphics[width=0.04\linewidth, trim={{0.45\width} {.13\width} {.35\width} {.0\width}},clip]{fig/compare_patch/56_he}
		{\includegraphics[width=0.16\linewidth]{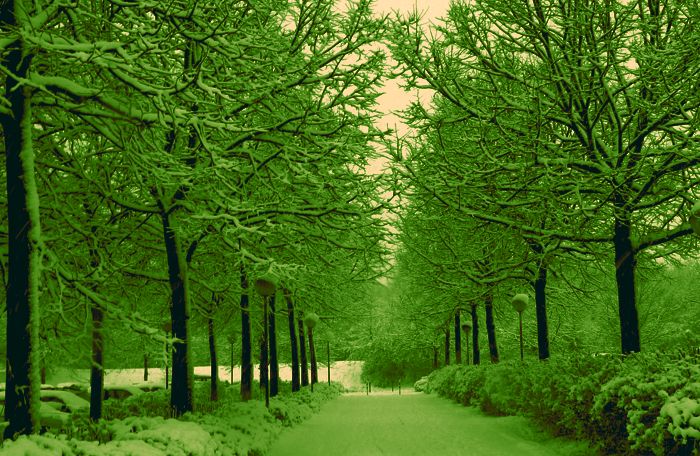}
			\label{fig:patch:e9}}\hspace{-2.4mm}
		\adjincludegraphics[width=0.04\linewidth, trim={{0.45\width} {.13\width} {.35\width} {.0\width}},clip]{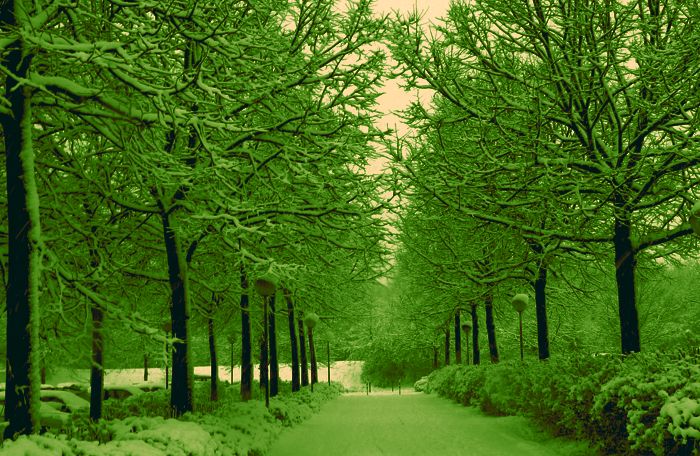}
	\end{minipage}
	\caption{Visual comparison with patch-based photorealistic methods. 1st column: reference style image. 2nd column: content image. 3rd column: Liao~\etal\cite{liao2017visual}. 4th column: He~\etal~\cite{he2019progressive}. 5th column: proposed NL-MAT.}
	\label{fig:patch}
\vspace{-5mm}
\end{figure*}\textcolor{black}
Fig. ~\ref{fig:patch} shows visual results of the proposed method as compared to the patch-based photorealistic stylization methods,~\ie, \textcolor{black}{Liao~\etal\cite{liao2017visual} and He~\etal~\cite{he2019progressive}}. We can observe that patch-based methods \textcolor{black}{are able to perform local style transfer on the content image successfully, because they could find the correspondence between the image pairs according to the patch similarity measured with the help of the pre-trained VGG-net.} However, although post-processing is applied to smooth the reconstructed result, Liao~\etal\cite{liao2017visual} still suffers from abrupt color changes within or across objects, as shown in Fig.~\ref{fig:patch}. For example, the mountain in the 2nd row, the building in the 3rd row, the background in the 4th four, the floor in the 6th row, and the tree in the 7th row do not have smooth color transitions. This is because patch-based methods mainly focus on style transfer in the local area while neglecting the global consistency within or across objects. He~\etal~\cite{he2019progressive} transfers style better than Liao~\etal\cite{liao2017visual} because it optimizes a local linear model for color transfer satisfying both local and global constraints. However, due to patch-similarity, the style of one patch from one object may be matched to similar patches belonging to other objects. For example, in the 4th row, the background of the car is transferred with the same color as the car, and its windshield has abrupt color changes. In the 6th row, the floor is transferred with the same color of the furniture. In the 8th row, the sky is transferred with the color of the tree. As a comparison, the proposed method is able to generate photorealistic results without color inconsistency caused by patch mismatch. \textcolor{black}{This is largely due to the realization of the proposed representations scheme, which can capture the non-local representations with matched context information that facilitates the local style transfer with global consistency.}

\begin{figure*}[htbp]
\centering
	\begin{minipage}{1\linewidth}
		{\includegraphics[width=0.16\linewidth]{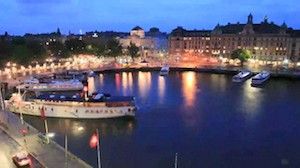}
			\label{fig:local_contex:a2}}
		{\includegraphics[width=0.16\linewidth]{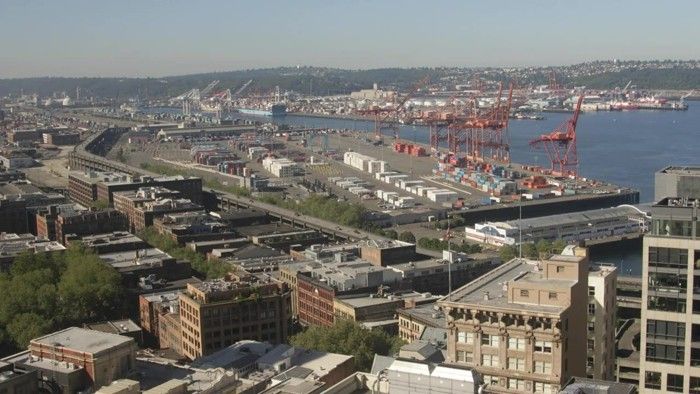}
			\label{fig:local_contex:b2}}\hspace{-2.4mm}
		\adjincludegraphics[width=0.04\linewidth, trim={ {.5\width} {0.3\width} {.4\width} {0.036\width}},clip]{fig/compare_context/6_0in}
		{\includegraphics[width=0.16\linewidth]{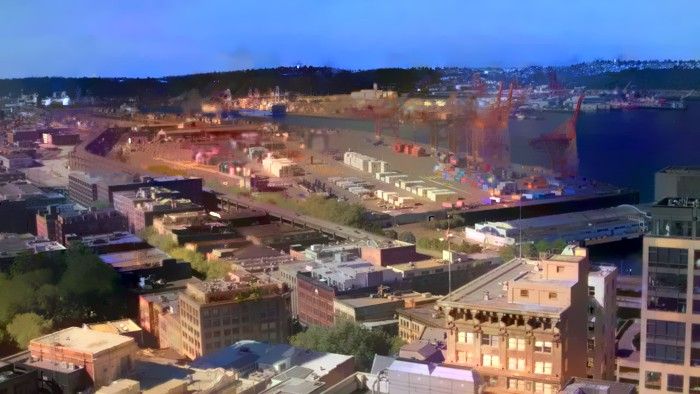}
			\label{fig:local_contex:c2}}\hspace{-2.4mm}
		\adjincludegraphics[width=0.04\linewidth, trim={ {.5\width} {0.3\width} {.4\width} {0.036\width}},clip]{fig/compare_context/6_luan}
		{\includegraphics[width=0.16\linewidth]{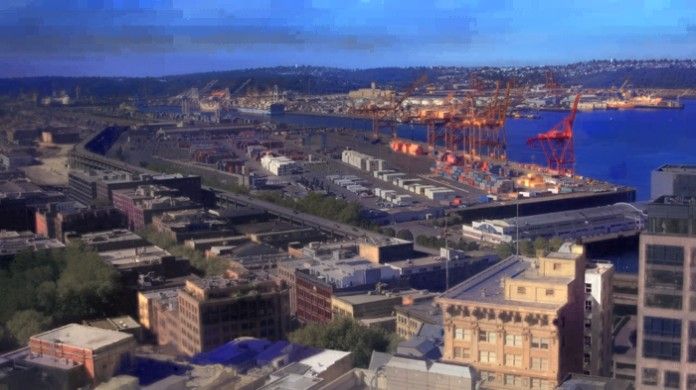}
			\label{fig:local_contex:d2}}\hspace{-2.4mm}
		\adjincludegraphics[width=0.04\linewidth, trim={ {.5\width} {0.3\width} {.4\width} {0.036\width}},clip]{fig/compare_context/6_li}
		{\includegraphics[width=0.16\linewidth]{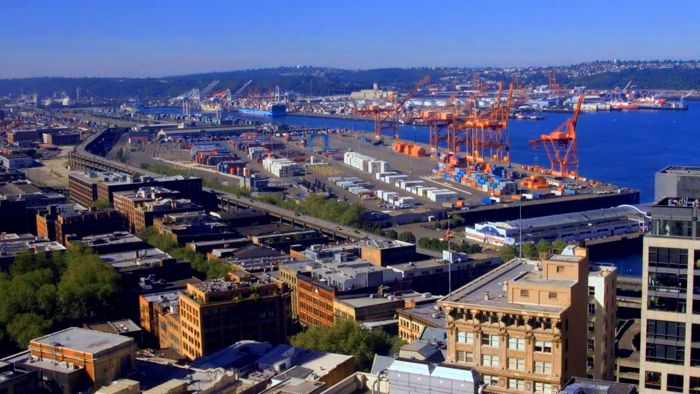}
			\label{fig:local_contex:e2}}\hspace{-2.4mm}
		\adjincludegraphics[width=0.04\linewidth, trim={ {.5\width} {0.3\width} {.4\width} {0.036\width}},clip]{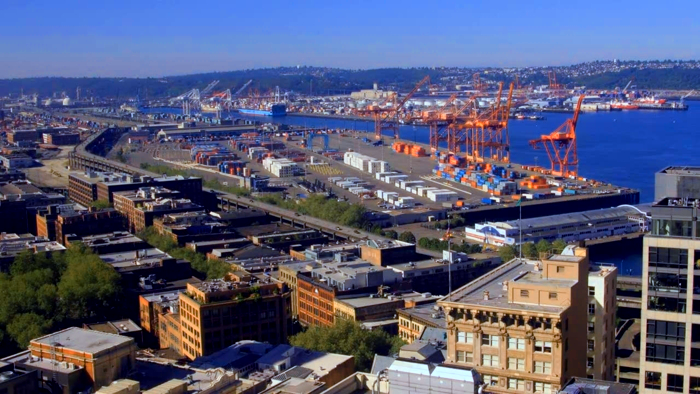}
	\end{minipage}\\
	\begin{minipage}{1\linewidth}
		{\includegraphics[width=0.16\linewidth]{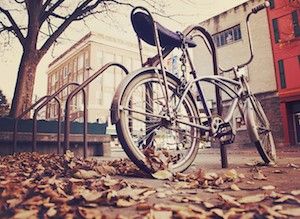}
			\label{fig:local_contex:a3}}
		{\includegraphics[width=0.16\linewidth]{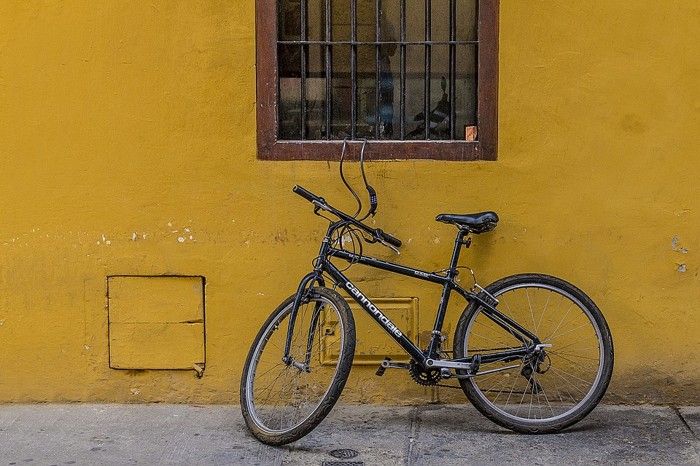}
			\label{fig:local_contex:b3}}\hspace{-2.4mm}
		\adjincludegraphics[width=0.04\linewidth, trim={{0.2\width} {.1\width} {.59\width} {.005\width}},clip]{fig/compare_context/43_0in}
		{\includegraphics[width=0.16\linewidth]{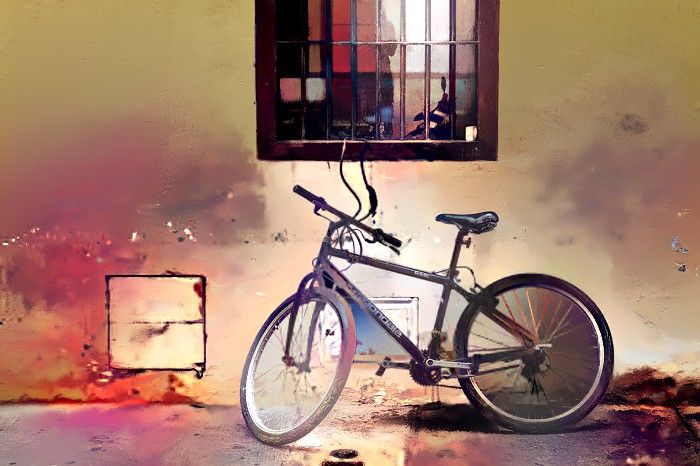}
			\label{fig:local_contex:c3}}\hspace{-2.4mm}
		\adjincludegraphics[width=0.04\linewidth, trim={{0.2\width} {.1\width} {.59\width} {.005\width}},clip]{fig/compare_context/43_luan}
		{\includegraphics[width=0.16\linewidth]{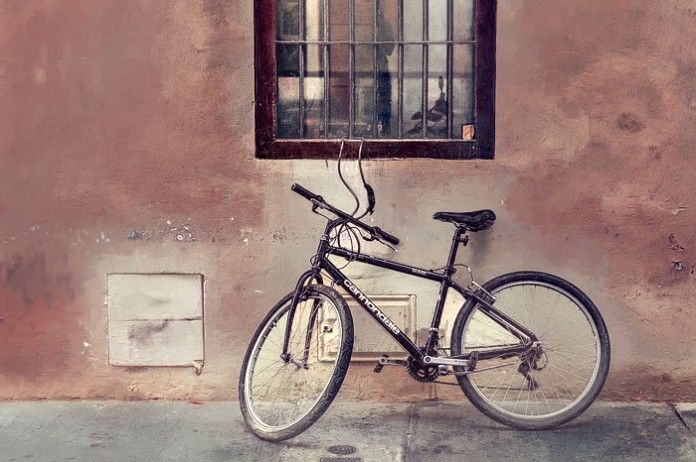}
			\label{fig:local_contex:d3}}\hspace{-2.4mm}
		\adjincludegraphics[width=0.04\linewidth, trim={{0.2\width} {.1\width} {.59\width} {.005\width}},clip]{fig/compare_context/43_li}
		{\includegraphics[width=0.16\linewidth]{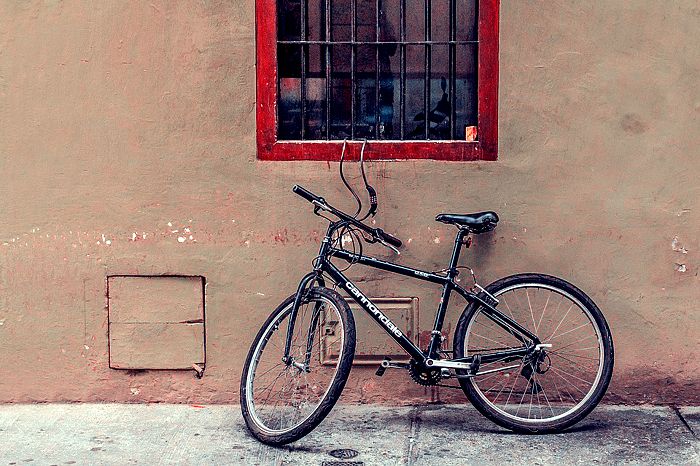}
			\label{fig:local_contex:e3}}\hspace{-2.4mm}
		\adjincludegraphics[width=0.04\linewidth, trim={{0.2\width} {.1\width} {.59\width} {.005\width}},clip]{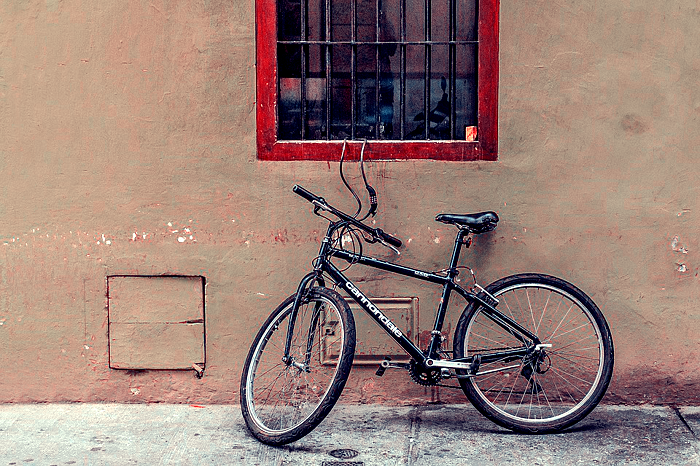}
	\end{minipage}
	\begin{minipage}{1\linewidth}
		{\includegraphics[width=0.16\linewidth]{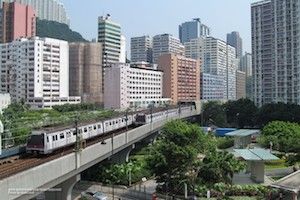}
			\label{fig:local_contex:a4}}
		{\includegraphics[width=0.16\linewidth]{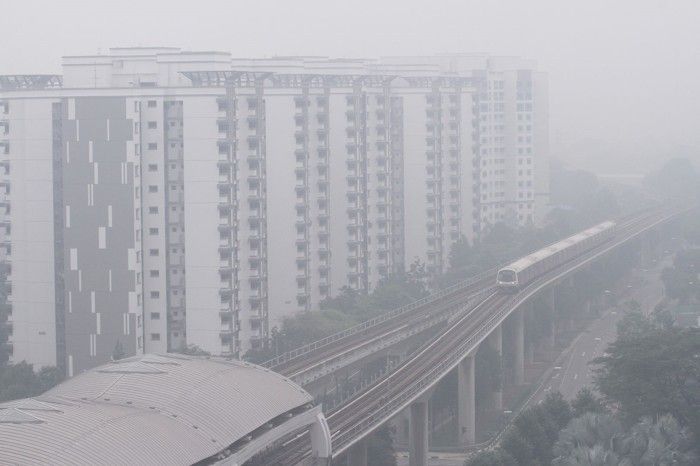}
			\label{fig:local_contex:b4}}\hspace{-2.4mm}
		\adjincludegraphics[width=0.04\linewidth, trim={{0.2\width} {.11\width} {.59\width} {.0\width}},clip]{fig/compare_context/44_0in}
		{\includegraphics[width=0.16\linewidth]{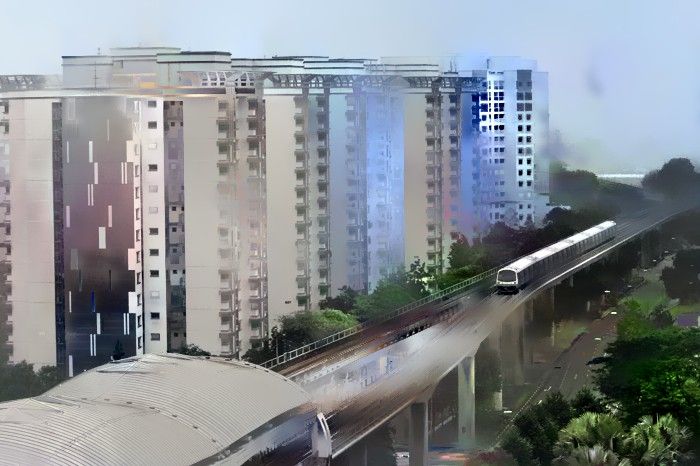}
			\label{fig:local_contex:c4}}\hspace{-2.4mm}
		\adjincludegraphics[width=0.04\linewidth, trim={{0.2\width} {.11\width} {.59\width} {.0\width}},clip]{fig/compare_context/44_luan}
		{\includegraphics[width=0.16\linewidth]{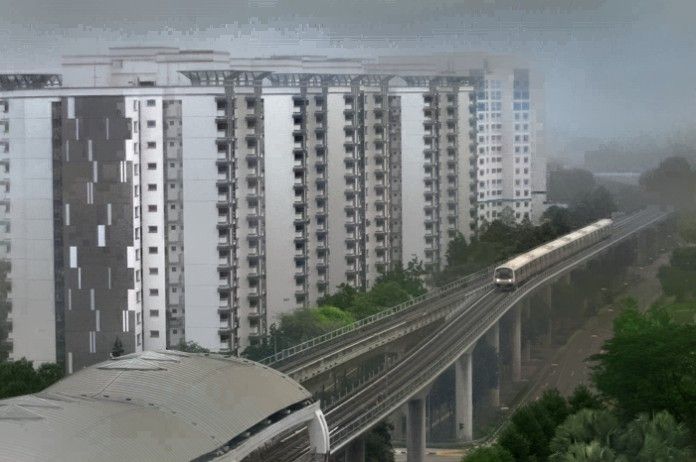}
			\label{fig:local_contex:d4}}\hspace{-2.4mm}
		\adjincludegraphics[width=0.04\linewidth, trim={{0.2\width} {.11\width} {.59\width} {.0\width}},clip]{fig/compare_context/44_li}
		{\includegraphics[width=0.16\linewidth]{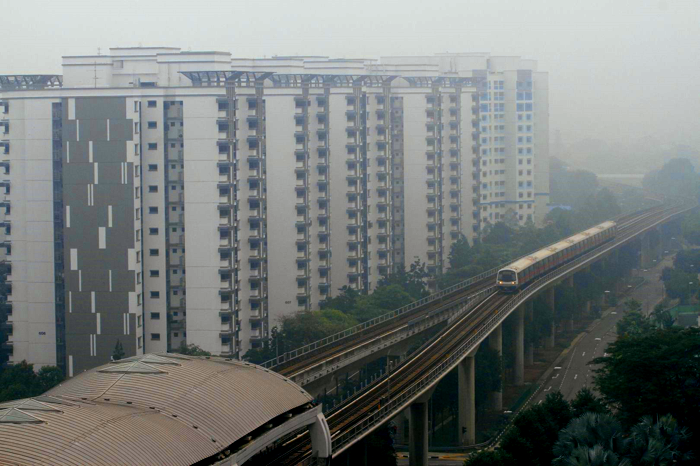}
			\label{fig:local_contex:e4}}\hspace{-2.4mm}
		\adjincludegraphics[width=0.04\linewidth, trim={{0.2\width} {.11\width} {.59\width} {.0\width}},clip]{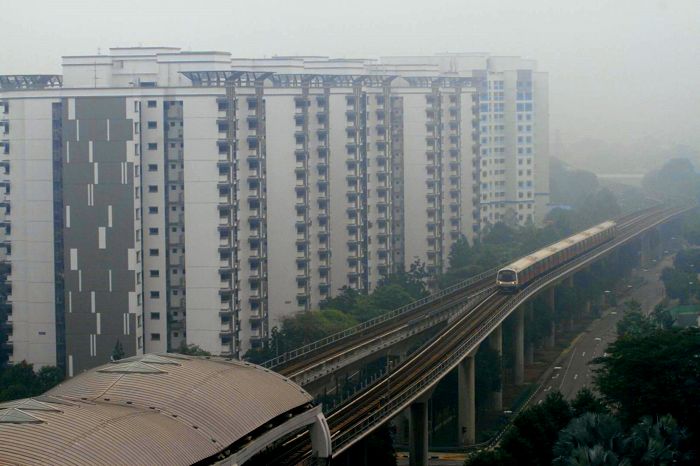}
	\end{minipage}
	\begin{minipage}{1\linewidth}
		{\includegraphics[width=0.16\linewidth]{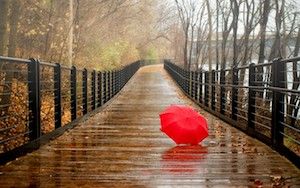}
			\label{fig:local_contex:a5}}
		{\includegraphics[width=0.16\linewidth]{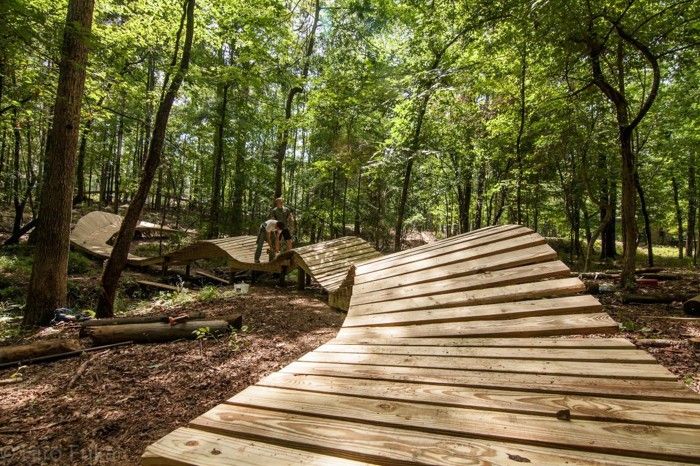}
			\label{fig:local_contex:b5}}\hspace{-2.4mm}
		\adjincludegraphics[width=0.04\linewidth, trim={{0.45\width} {.12\width} {.4\width} {.145\width}},clip]{fig/compare_context/54_0in}
		{\includegraphics[width=0.16\linewidth]{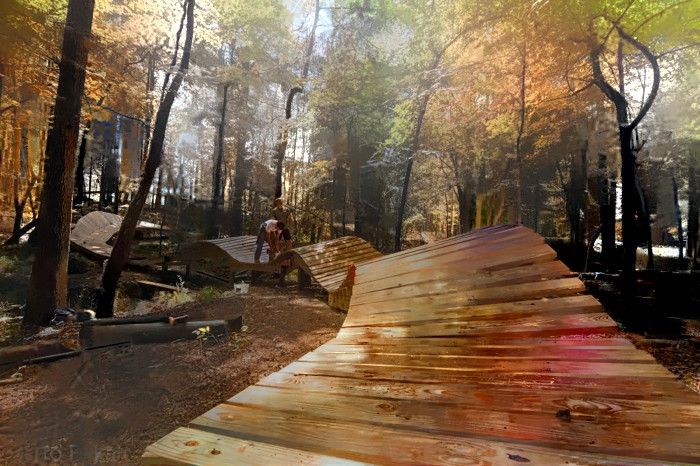}
			\label{fig:local_contex:c5}}\hspace{-2.4mm}
		\adjincludegraphics[width=0.04\linewidth, trim={{0.45\width} {.12\width} {.4\width} {.145\width}},clip]{fig/compare_context/54_luan}
		{\includegraphics[width=0.16\linewidth]{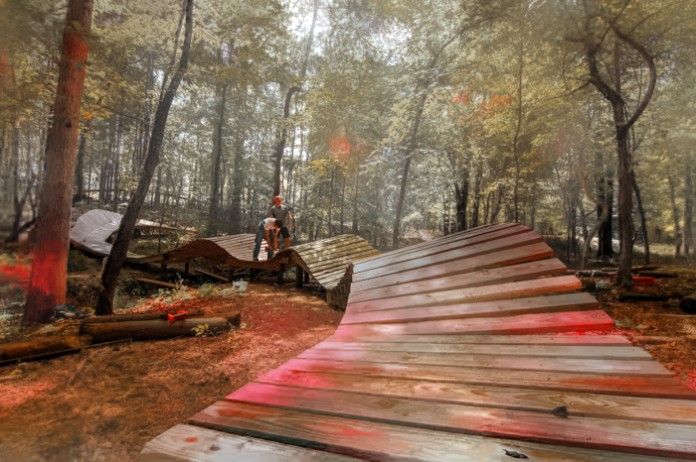}
			\label{fig:local_contex:d5}}\hspace{-2.4mm}
		\adjincludegraphics[width=0.04\linewidth, trim={{0.45\width} {.12\width} {.4\width} {.145\width}},clip]{fig/compare_context/54_li}
		{\includegraphics[width=0.16\linewidth]{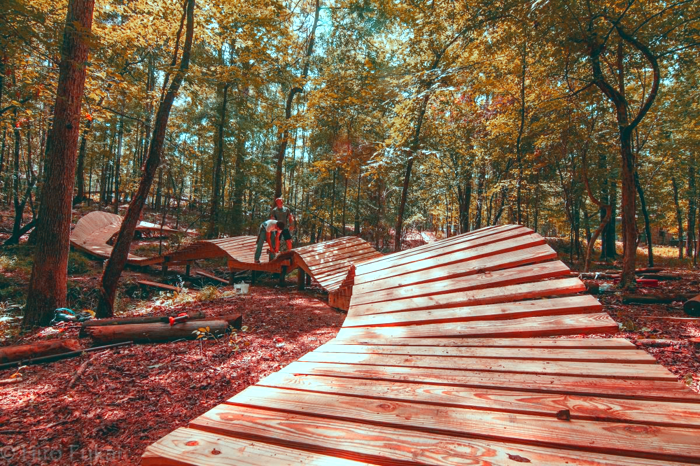}
			\label{fig:local_contex:e5}}\hspace{-2.4mm}
		\adjincludegraphics[width=0.04\linewidth, trim={{0.45\width} {.12\width} {.4\width} {.145\width}},clip]{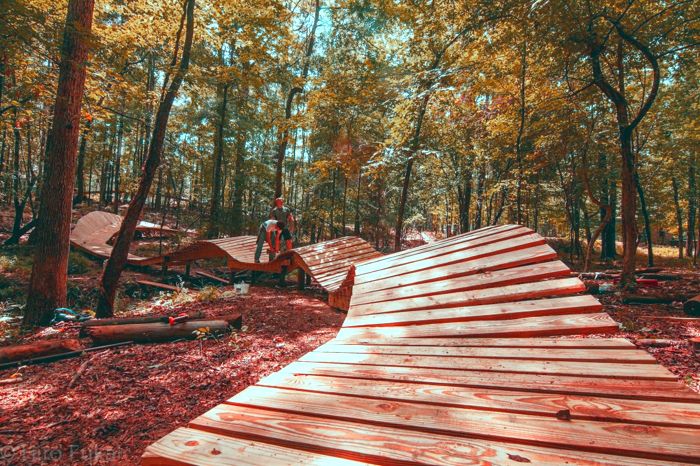}
	\end{minipage}
	\begin{minipage}{1\linewidth}
		{\includegraphics[width=0.16\linewidth]{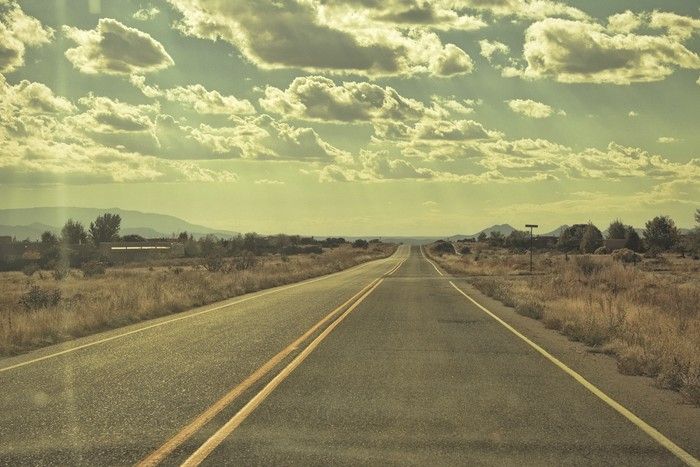}
			\label{fig:local_contex:a6}}
		{\includegraphics[width=0.16\linewidth]{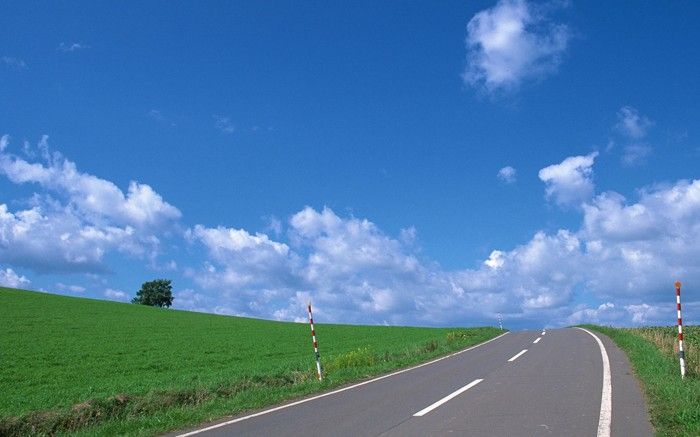}
			\label{fig:local_contex:b6}}\hspace{-2.4mm}
		\adjincludegraphics[width=0.04\linewidth, trim={{0.2\width} {.0\width} {.59\width} {.1\width}},clip]{fig/compare_context/32_0in}
		{\includegraphics[width=0.16\linewidth]{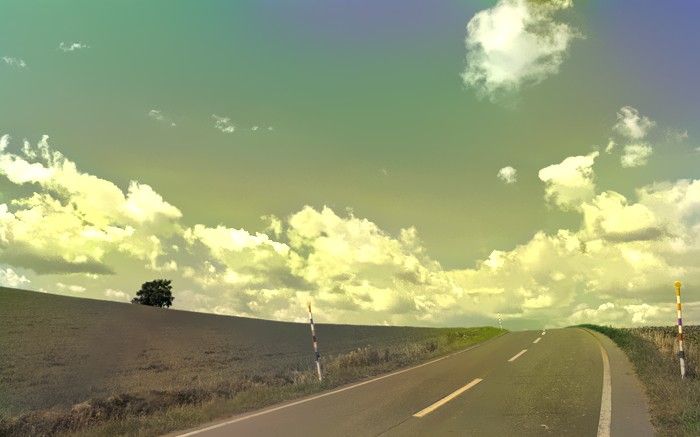}
			\label{fig:local_contex:c6}}\hspace{-2.4mm}
		\adjincludegraphics[width=0.04\linewidth, trim={{0.2\width} {.0\width} {.59\width} {.1\width}},clip]{fig/compare_context/32_luan}
		{\includegraphics[width=0.16\linewidth]{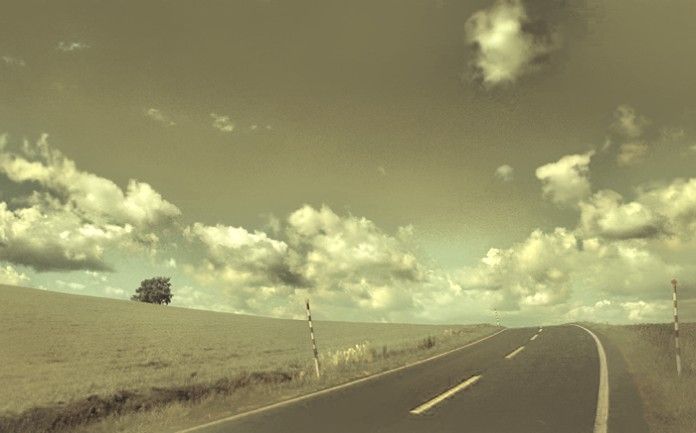}
			\label{fig:local_contex:d6}}\hspace{-2.4mm}
		\adjincludegraphics[width=0.04\linewidth, trim={{0.2\width} {.0\width} {.59\width} {.1\width}},clip]{fig/compare_context/32_li}
		{\includegraphics[width=0.16\linewidth]{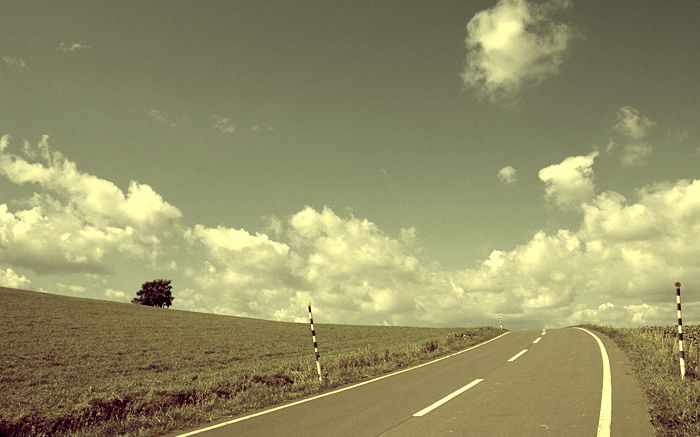}
			\label{fig:local_contex:e6}}\hspace{-2.4mm}
		\adjincludegraphics[width=0.04\linewidth, trim={{0.2\width} {.0\width} {.59\width} {.1\width}},clip]{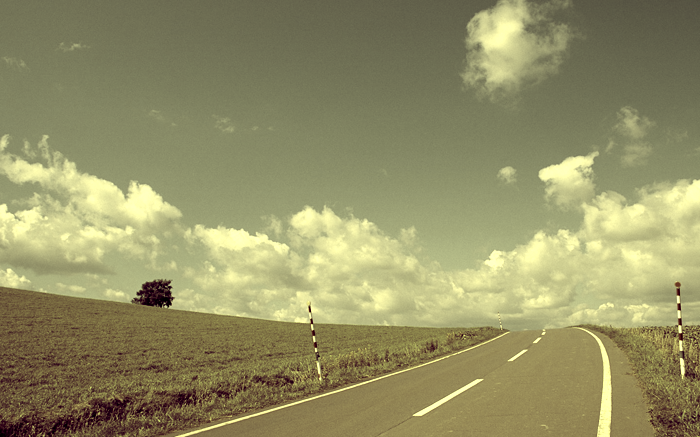}
	\end{minipage}
	\begin{minipage}{1\linewidth}
		{\includegraphics[width=0.16\linewidth]{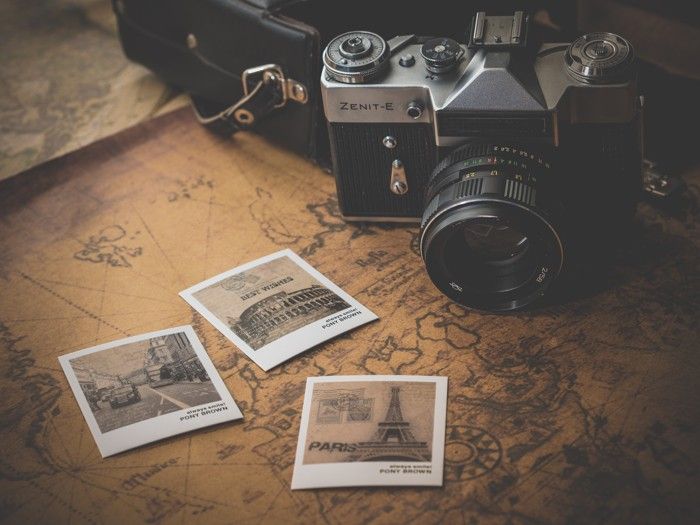}
			\label{fig:local_contex:a7}}
		{\includegraphics[width=0.16\linewidth]{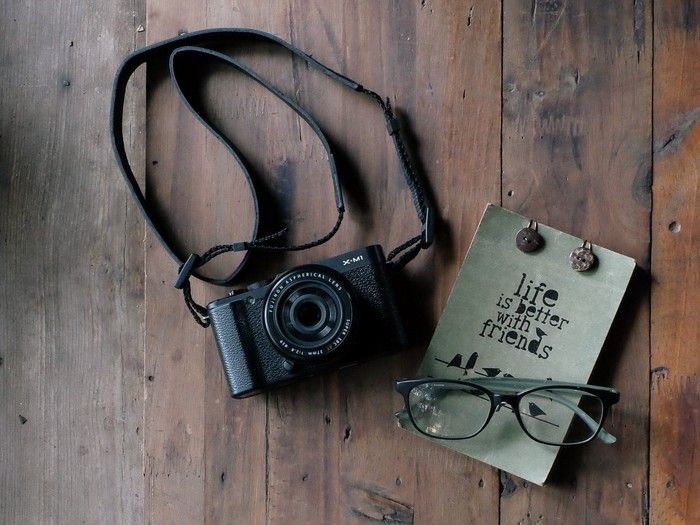}
			\label{fig:local_contex:b7}}\hspace{-2.4mm}
		\adjincludegraphics[width=0.04\linewidth, trim={{0.2\width} {.0\width} {.59\width} {.1\width}},clip]{fig/compare_context/41_0in}
		{\includegraphics[width=0.16\linewidth]{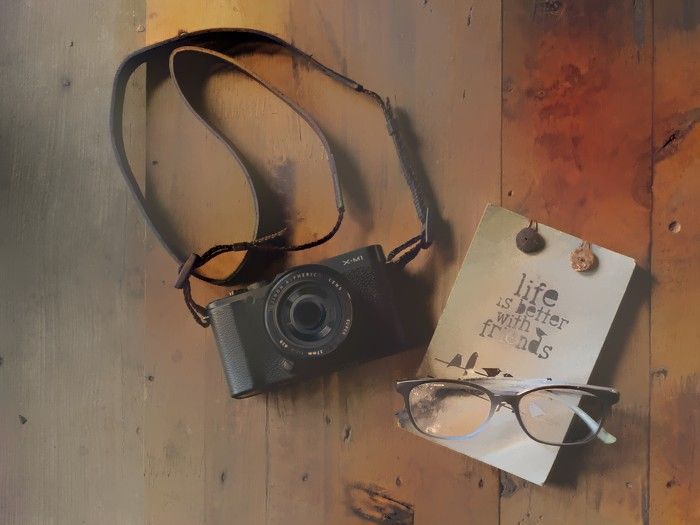}
			\label{fig:local_contex:c7}}\hspace{-2.4mm}
		\adjincludegraphics[width=0.04\linewidth, trim={{0.2\width} {.0\width} {.59\width} {.1\width}},clip]{fig/compare_context/41_luan}
		{\includegraphics[width=0.16\linewidth]{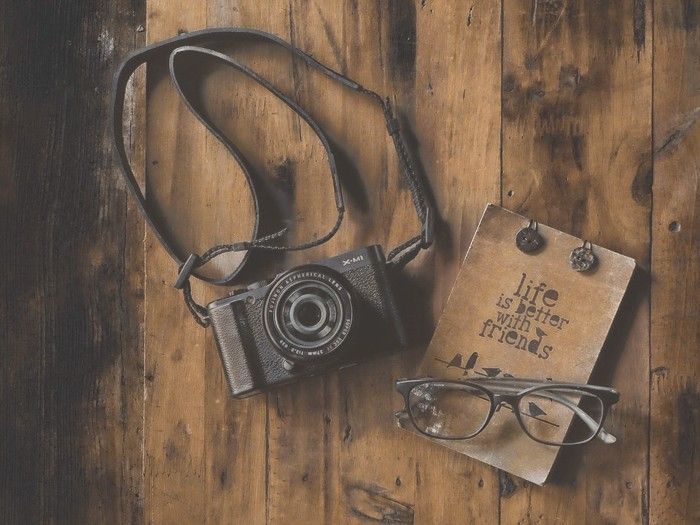}
			\label{fig:local_contex:d7}}\hspace{-2.4mm}
		\adjincludegraphics[width=0.04\linewidth, trim={{0.2\width} {.0\width} {.59\width} {.1\width}},clip]{fig/compare_context/41_li}
		{\includegraphics[width=0.16\linewidth]{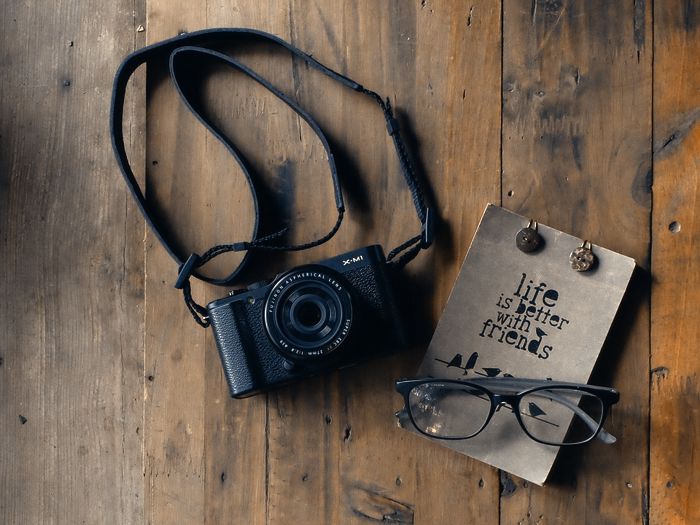}
			\label{fig:local_contex:e7}}\hspace{-2.4mm}
		\adjincludegraphics[width=0.04\linewidth, trim={{0.2\width} {.0\width} {.59\width} {.1\width}},clip]{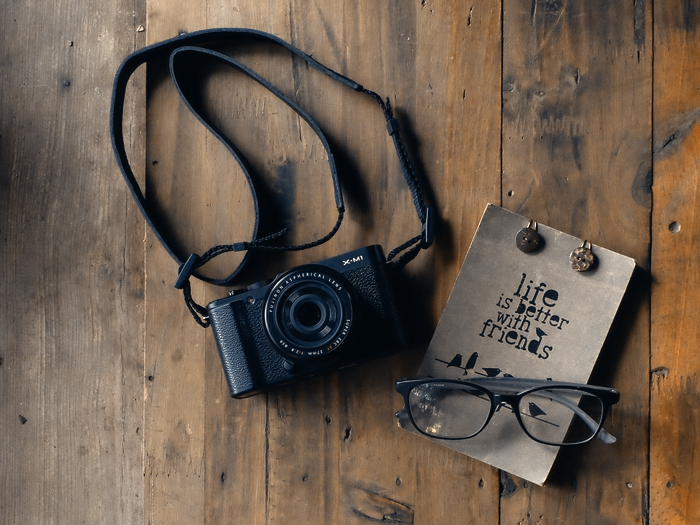}
	\end{minipage}
	\caption{Visual comparison with context-based photorealistic methods. 1st column: reference style image. 2nd column: content image. 3rd column: Luan~\etal\cite{luan2017deep}. 4th column: Li~\etal~\cite{li2018closed}. 5th column: proposed NL-MAT.}
	\label{fig:local_contex}
	\begin{minipage}{1\linewidth}
		{\includegraphics[width=0.16\linewidth]{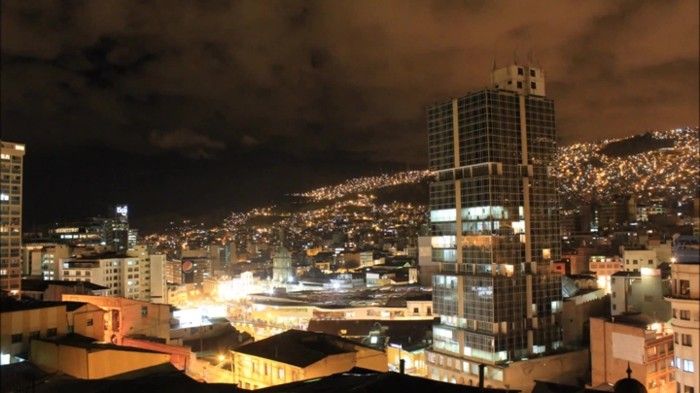}
			\label{fig:wct2:tar4}}
		{\includegraphics[width=0.16\linewidth]{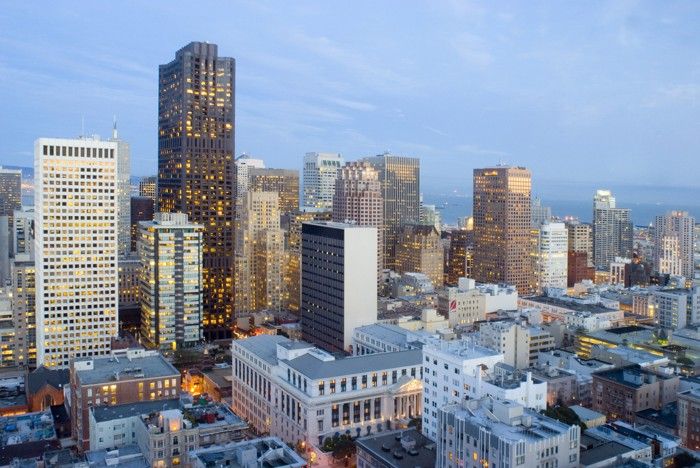}
			\label{fig:wct2:in4}}\hspace{-2.4mm}
		\adjincludegraphics[width=0.04\linewidth, trim={{0.21\width} {0.61\height} {0.71\width} {0.07\height}},clip]{fig/compare_wct2/in4}
		{\includegraphics[width=0.16\linewidth]{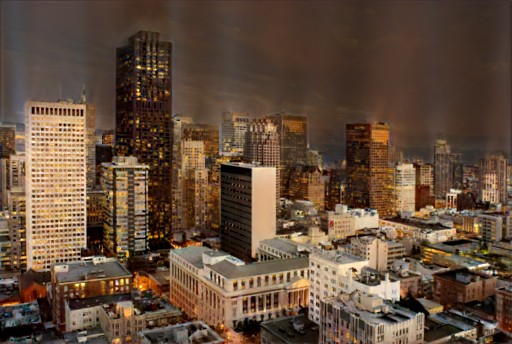}
			\label{fig:lst:res4}}\hspace{-2.4mm}
		\adjincludegraphics[width=0.04\linewidth, trim={{0.21\width} {0.61\height} {0.71\width} {0.07\height}},clip]{fig/compare_lst/in4_filtered}
		{\includegraphics[width=0.16\linewidth]{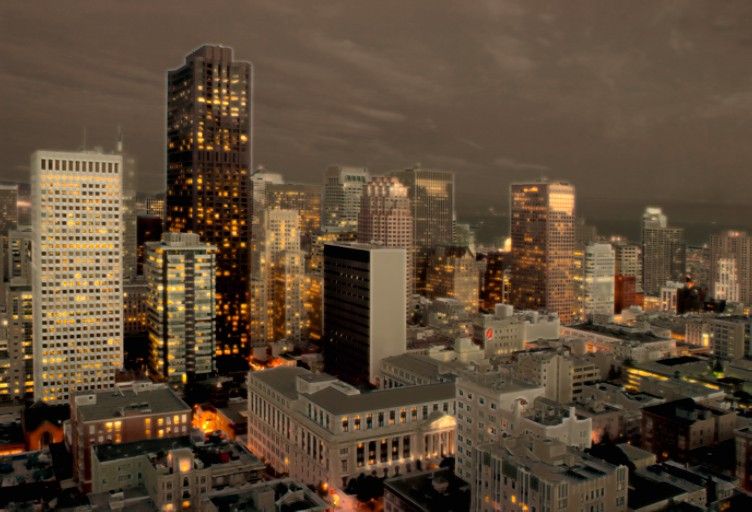}
			\label{fig:wct2:res4}}\hspace{-2.4mm}
		\adjincludegraphics[width=0.04\linewidth, trim={{0.21\width} {0.61\height} {0.71\width} {0.07\height}},clip]{fig/compare_wct2/in4_cat5_encoder_skip}
		{\includegraphics[width=0.16\linewidth]{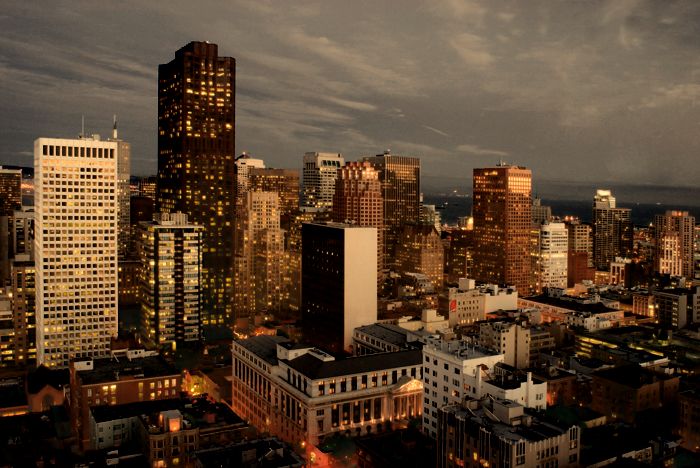}
			\label{fig:ours_4}}\hspace{-2.4mm}
		\adjincludegraphics[width=0.04\linewidth, trim={{0.21\width} {0.61\height} {0.71\width} {0.07\height}},clip]{fig/compare_wct2/4_ours}
	\end{minipage}\\
	\begin{minipage}{1\linewidth}
		{\includegraphics[width=0.16\linewidth]{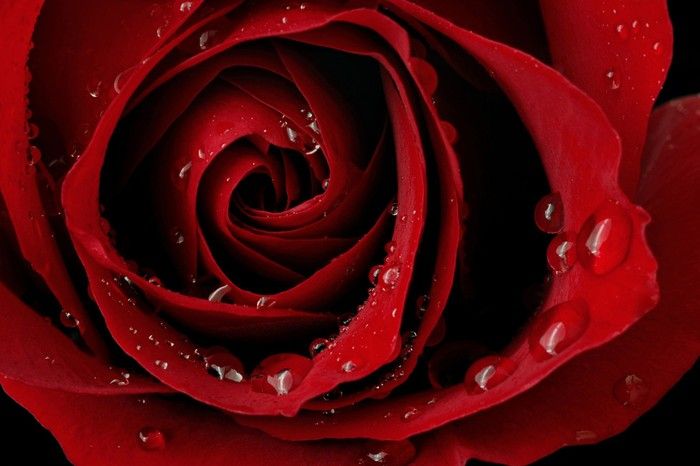}
			\label{fig:wct2:tar9}}
		{\includegraphics[width=0.16\linewidth]{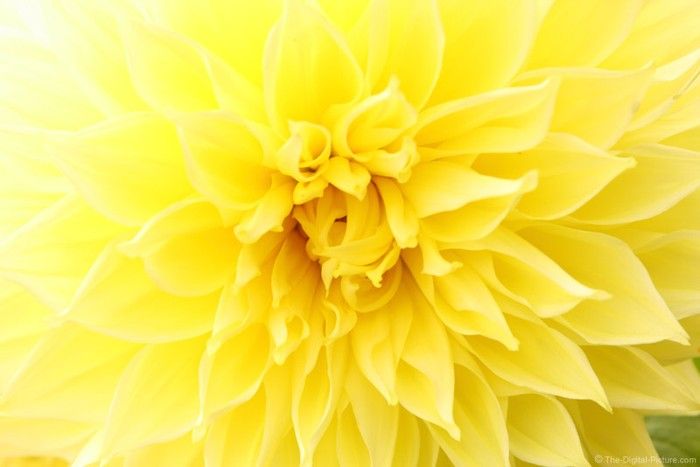}
			\label{fig:wct2:in9}}\hspace{-2.4mm}
		\adjincludegraphics[width=0.04\linewidth, trim={{0.1\width} 0 {0.75\width} {0.4\height}},clip]{fig/compare_wct2/in9}
		{\includegraphics[width=0.16\linewidth]{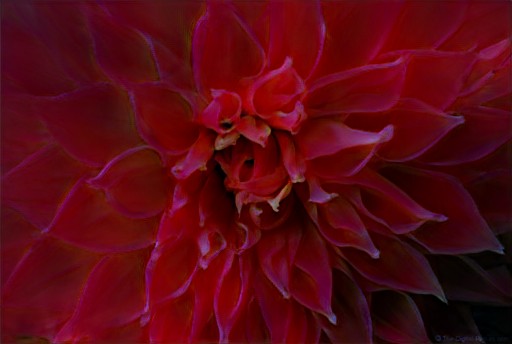}
			\label{fig:lst:res9}}\hspace{-2.4mm}
		\adjincludegraphics[width=0.04\linewidth, trim={{0.1\width} 0 {0.75\width} {0.4\height}},clip]{fig/compare_lst/in9_filtered}
		{\includegraphics[width=0.16\linewidth]{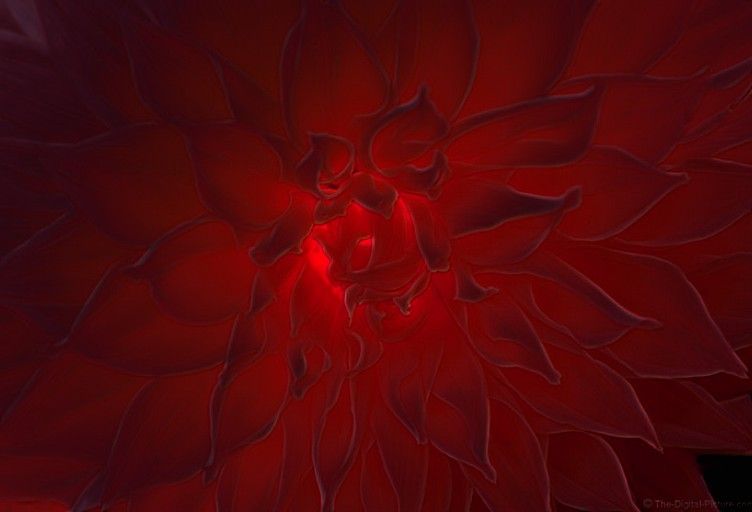}
			\label{fig:wct2:res9}}\hspace{-2.4mm}
		\adjincludegraphics[width=0.04\linewidth, trim={{0.1\width} 0 {0.75\width} {0.4\height}},clip]{fig/compare_wct2/in9_cat5_encoder_skip}
		{\includegraphics[width=0.16\linewidth]{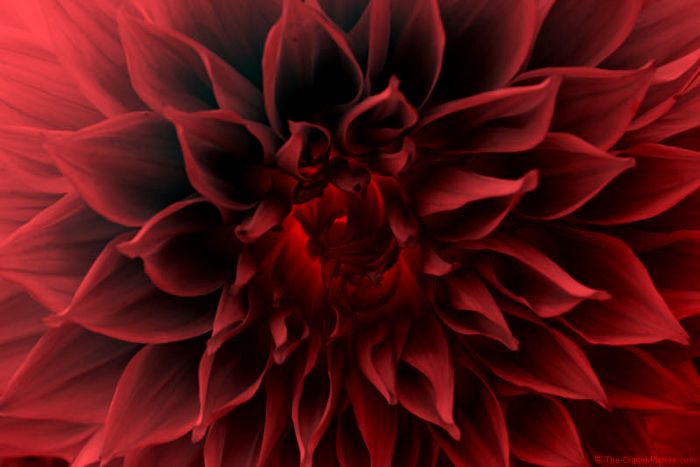}
			\label{fig:ours_9}}\hspace{-2.4mm}
		\adjincludegraphics[width=0.04\linewidth, trim={{0.1\width} 0 {0.75\width} {0.4\height}},clip]{fig/compare_wct2/9_ours}
	\end{minipage}
	\begin{minipage}{1\linewidth}
		{\includegraphics[width=0.16\linewidth]{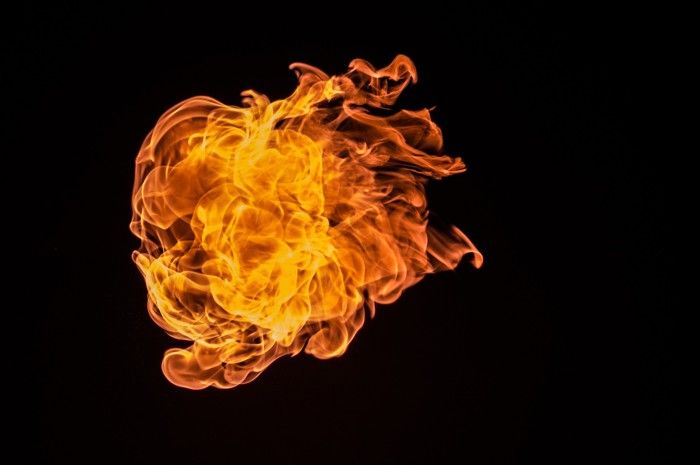}
			\label{fig:wct2:tar16}}
		{\includegraphics[width=0.16\linewidth]{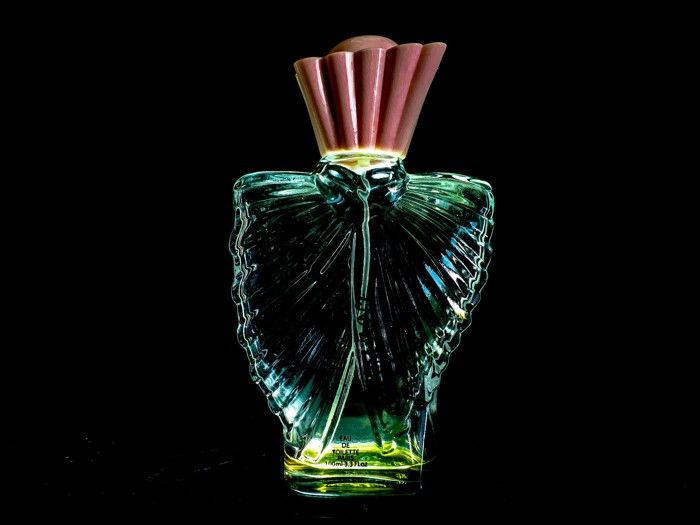}
			\label{fig:wct2:in16}}\hspace{-2.4mm}
		\adjincludegraphics[width=0.04\linewidth, trim={{0.6\width} {0.24\height} {0.28\width} {0.28\height}},clip]{fig/compare_wct2/in16}
		{\includegraphics[width=0.16\linewidth]{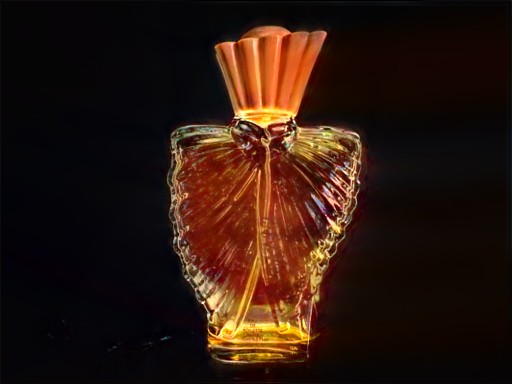}
			\label{fig:lst:res16}}\hspace{-2.4mm}
		\adjincludegraphics[width=0.04\linewidth, trim={{0.6\width} {0.24\height} {0.28\width} {0.28\height}},clip]{fig/compare_lst/in16_filtered}
		{\includegraphics[width=0.16\linewidth]{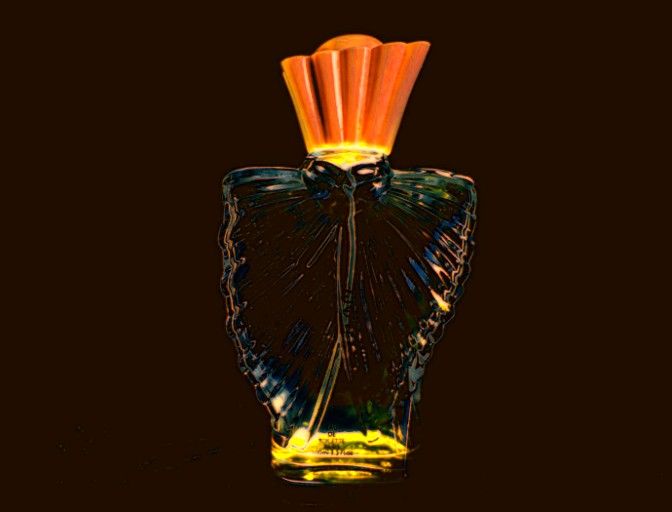}
			\label{fig:wct2:res16}}\hspace{-2.4mm}
		\adjincludegraphics[width=0.04\linewidth, trim={{0.6\width} {0.24\height} {0.28\width} {0.28\height}},clip]{fig/compare_wct2/in16_cat5_encoder_skip}
		{\includegraphics[width=0.16\linewidth]{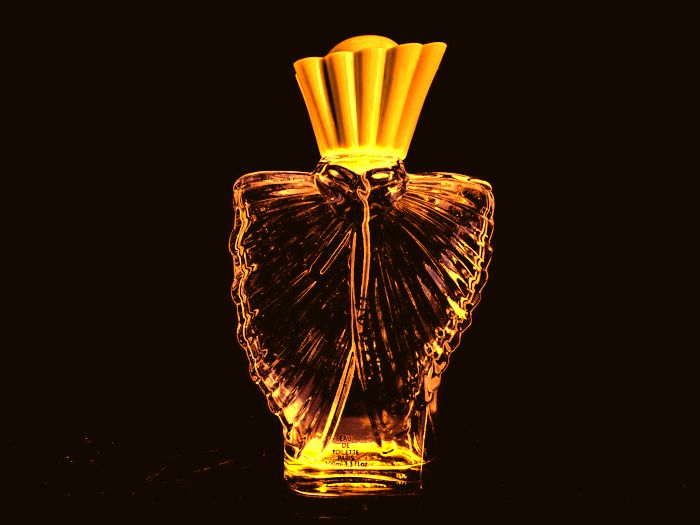}
			\label{fig:ours_16}}\hspace{-2.4mm}
		\adjincludegraphics[width=0.04\linewidth, trim={{0.6\width} {0.24\height} {0.28\width} {0.28\height}},clip]{fig/compare_wct2/16_ours}
	\end{minipage}
	\begin{minipage}{1\linewidth}
		{\includegraphics[width=0.16\linewidth]{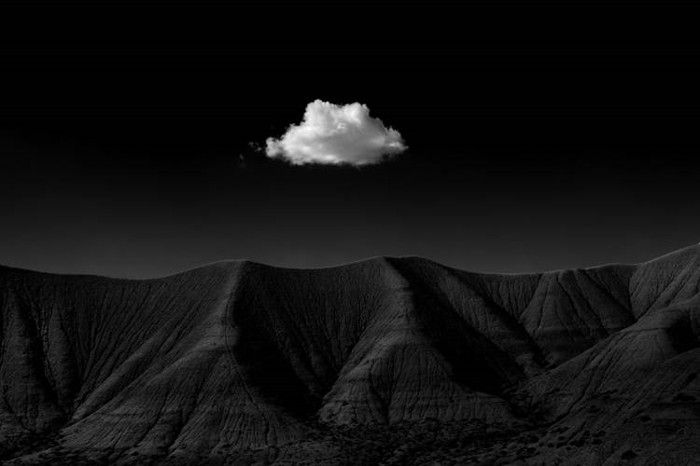}
			\label{fig:wct2:tar46}}
		{\includegraphics[width=0.16\linewidth]{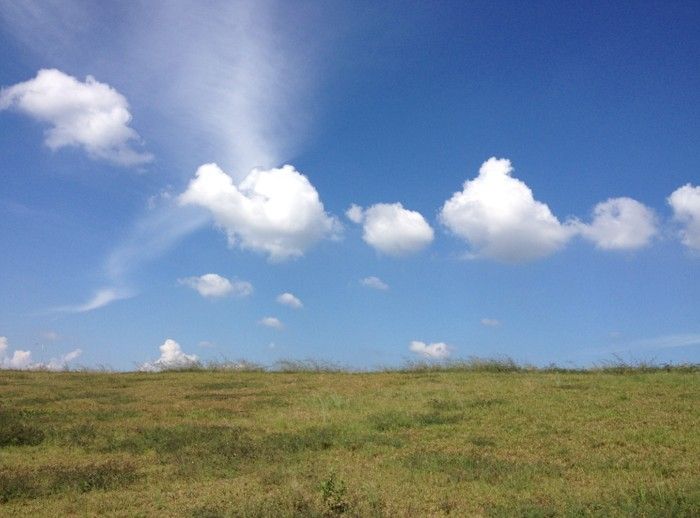}
			\label{fig:wct2:in46}}\hspace{-2.4mm}
		\adjincludegraphics[width=0.04\linewidth, trim={{0.2\width} {0.16\height} {0.59\width} {0.0\height}},clip]{fig/compare_wct2/in46}
		{\includegraphics[width=0.16\linewidth]{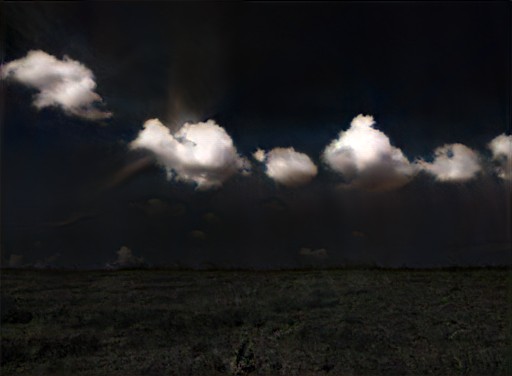}
			\label{fig:lst:res46}}\hspace{-2.4mm}
		\adjincludegraphics[width=0.04\linewidth, trim={{0.2\width} {0.16\height} {0.59\width} {0.0\height}},clip]{fig/compare_lst/in46_filtered}
		{\includegraphics[width=0.16\linewidth]{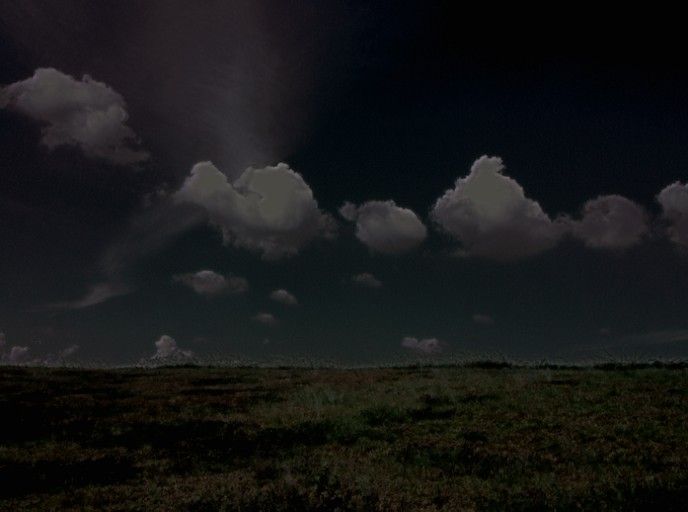}
			\label{fig:wct2:res46}}\hspace{-2.4mm}
		\adjincludegraphics[width=0.04\linewidth, trim={{0.2\width} {0.16\height} {0.59\width} {0.0\height}},clip]{fig/compare_wct2/in46_cat5_encoder_skip}
		{\includegraphics[width=0.16\linewidth]{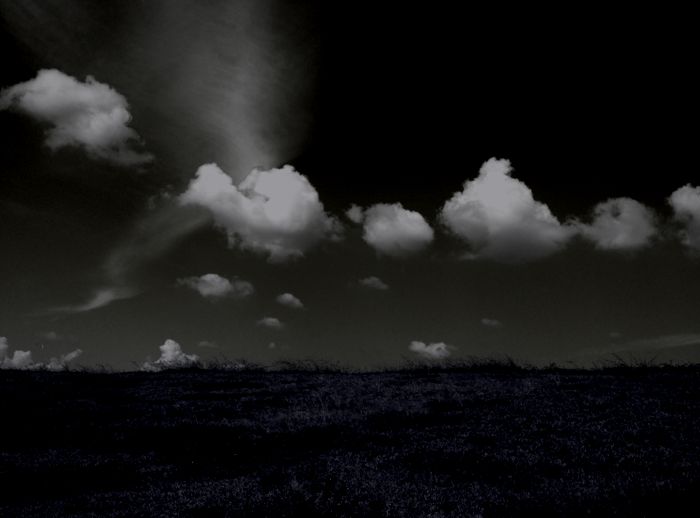}
			\label{fig:ours_46}}\hspace{-2.4mm}
		\adjincludegraphics[width=0.04\linewidth, trim={{0.2\width} {0.16\height} {0.59\width} {0.0\height}},clip]{fig/compare_wct2/46_ours}
	\end{minipage}
	\begin{minipage}{1\linewidth}
		{\includegraphics[width=0.16\linewidth]{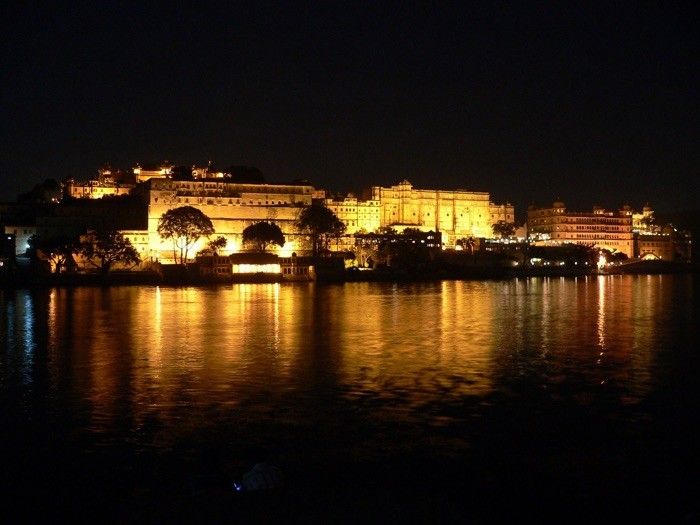}
		\label{fig:wct2:tar24}}
		{\includegraphics[width=0.16\linewidth]{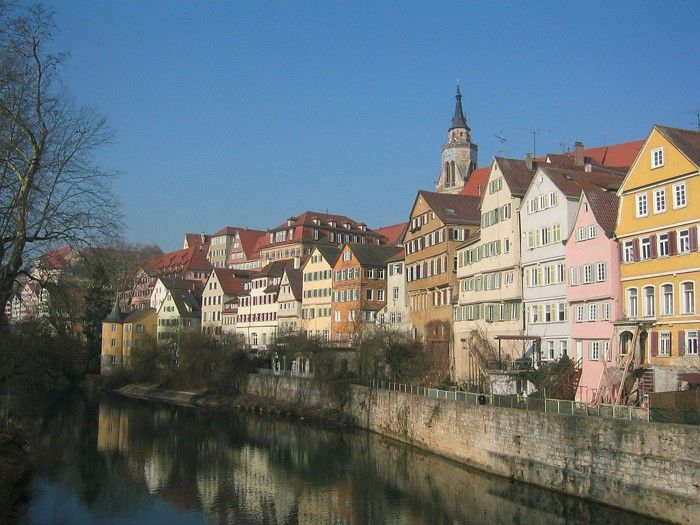}
		\label{fig:wct2:in24}}\hspace{-2.4mm}
		\adjincludegraphics[width=0.04\linewidth, trim={{0.62\width} {0.68\height} {0.305\width} {0.02\height}},clip]{fig/compare_wct2/in24}
		{\includegraphics[width=0.16\linewidth]{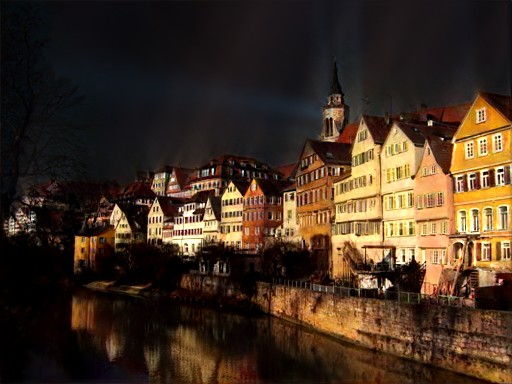}
		\label{fig:lst:res24}}\hspace{-2.4mm}
		\adjincludegraphics[width=0.04\linewidth, trim={{0.62\width} {0.68\height} {0.305\width} {0.02\height}},clip]{fig/compare_lst/in24_filtered}
		{\includegraphics[width=0.16\linewidth]{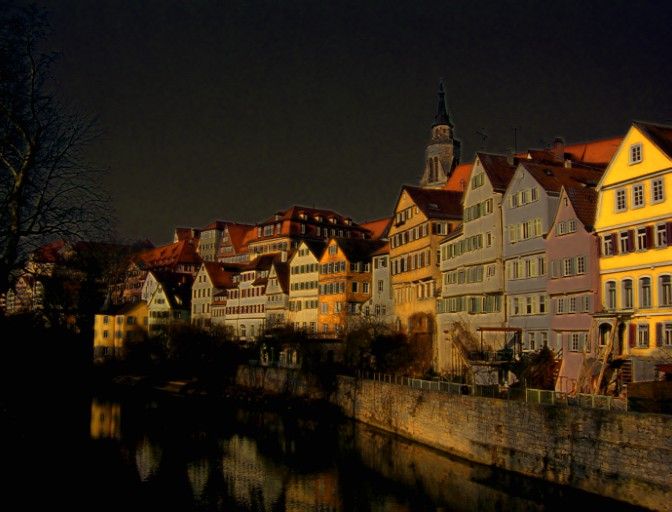}
		\label{fig:wct2:res24}}\hspace{-2.4mm}
		\adjincludegraphics[width=0.04\linewidth, trim={{0.62\width} {0.68\height} {0.305\width} {0.02\height}},clip]{fig/compare_wct2/in24_cat5_encoder_skip}
		{\includegraphics[width=0.16\linewidth]{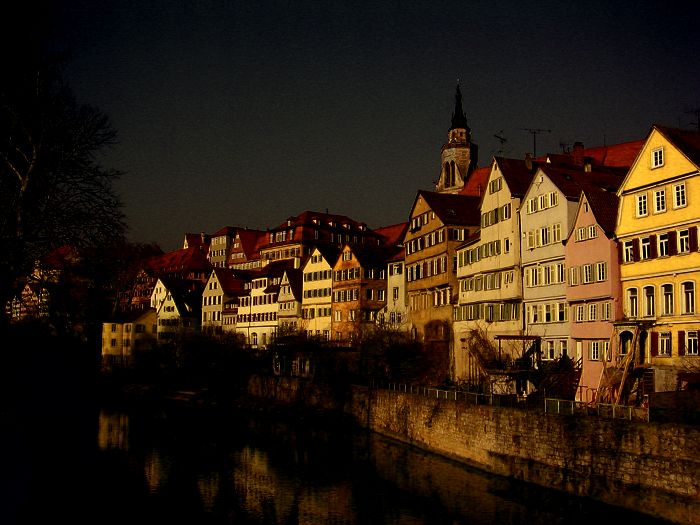}
		\label{fig:ours_24}}\hspace{-2.4mm}
		\adjincludegraphics[width=0.04\linewidth, trim={{0.62\width} {0.68\height} {0.305\width} {0.02\height}},clip]{fig/compare_wct2/24_ours}
	\end{minipage}
	\caption{Visual comparison with context-based photorealistic methods. 1st column: reference style image. 2nd column: content image. 3rd column: LST~\etal\cite{li2018learning}. 4th column: WCT$^2$~\etal~\cite{yoo2019photorealistic}. 5th column: proposed NL-MAT.}
	\label{fig:local_contex2}
\end{figure*}

Figs. ~\ref{fig:local_contex} and~\ref{fig:local_contex2} show visual results of the proposed method as compared to the context-based photorealistic stylization methods\textcolor{black}{, \ie, Luan~\etal~\cite{luan2017deep}, Li~\etal~\cite{li2018closed}, LST~\cite{li2018learning}, and WCT$^2$~\cite{yoo2019photorealistic}.} The generated results from Luan~\etal~\cite{luan2017deep} and LST~\cite{li2018learning} can preserve the spatial structure well with the local color affine transfer constraint/filter. However, they tend to cause color inconsistency, especially in homogeneous areas, as shown in the 3rd column of Figs.~\ref{fig:local_contex} and~\ref{fig:local_contex2}. With the post-processing smoothing step, Li~\etal~\cite{li2018closed} generates better results. \textcolor{black}{However, the results have some blurry artifacts introduced by the post-processing. WCT$^2$~\cite{yoo2019photorealistic} produces smoother results with less artifacts due to the adopted wavelet module, as shown in the 4th column of Fig.~\ref{fig:local_contex2}.} These methods can successfully transfer the color style to the content image \textcolor{black}{in most scenarios, based on the pre-trained segmentation model. However, they tend to fail when the contexts are mismatched or not recognized by the pre-trained segmentation model.} 
% We also observe that they tend to match the color of the content image to that of the style image at the same position when the semantic labels of the style and content images do not match.
For example, observe the style-content image pair in the \textcolor{black}{3rd} to the last row of Fig.~\ref{fig:local_contex}, where the red umbrella is in the semantic label of the style image but not in that of the content image, the red color is transferred, by both Luan~\etal~\cite{luan2017deep} and Li~\etal~\cite{li2018closed} methods, to the content image at around the same spatial location as it appears in the style image even though semantically, there is no additional object at that location in the content image. \textcolor{black}{Another example is shown in the 2nd row of Fig.~\ref{fig:local_contex2}, where the shadow of the flowers cannot be recognized by the pre-trained model, thus the methods fail to transfer the correct style. In addition to the problem caused by pre-trained segmentation model,} all these methods still exhibit abrupt color changes between different semantic regions. This is because they perform region-based transfer within the semantic regions of the content and style images without adequately considering the global consistency. 

On the contrary, the images generated by the proposed method can not only transfer the color style correctly but also preserve the natural color transitions among neighborhood pixels, especially the transitions between different contexts. The key contribution to the performance gain is that, instead of relying on additional segmentation models, the proposed \textcolor{black}{representation scheme is able to extract matched context-sensitive non-local representations based on the characteristics of the context with the sparse constraint and mutual-discriminative network. %with the sparse constraint and mutual-discriminative network. 
With this scheme}, we are able to transfer the style locally with WCT and affine-transfer decoder in a globally consistent fashion to produce more photorealistic photos with desired styles. %\textcolor{black}{without any pre-trained models}. 

\begin{figure*}[htbp]
\setlength{\abovecaptionskip}{1pt}
\setlength{\belowcaptionskip}{1pt}
	\centering
	\begin{minipage}{1\linewidth}
		{\includegraphics[width=0.21\linewidth]{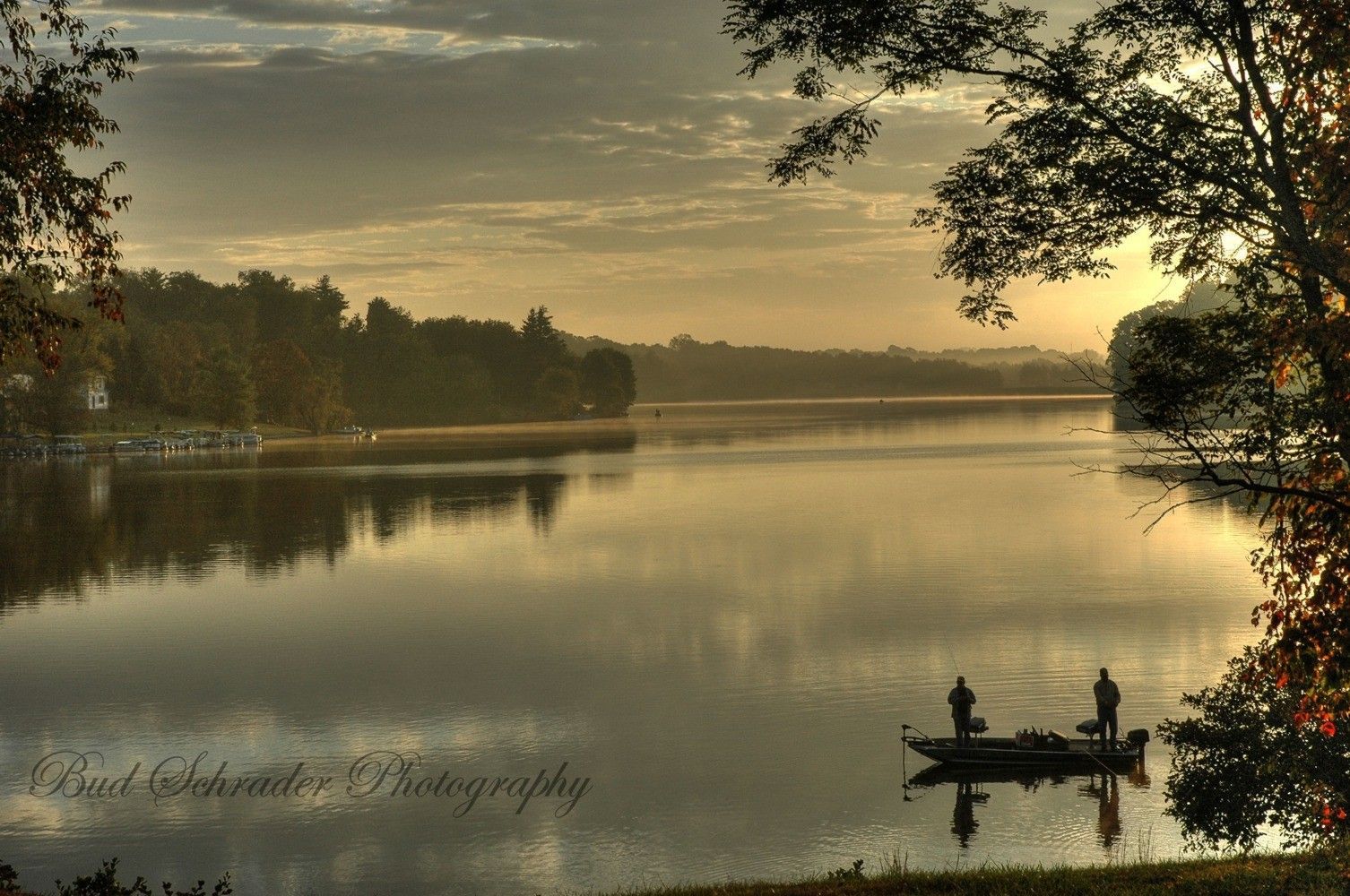}
			\label{fig:nas:tar9}}
		{\includegraphics[width=0.216\linewidth]{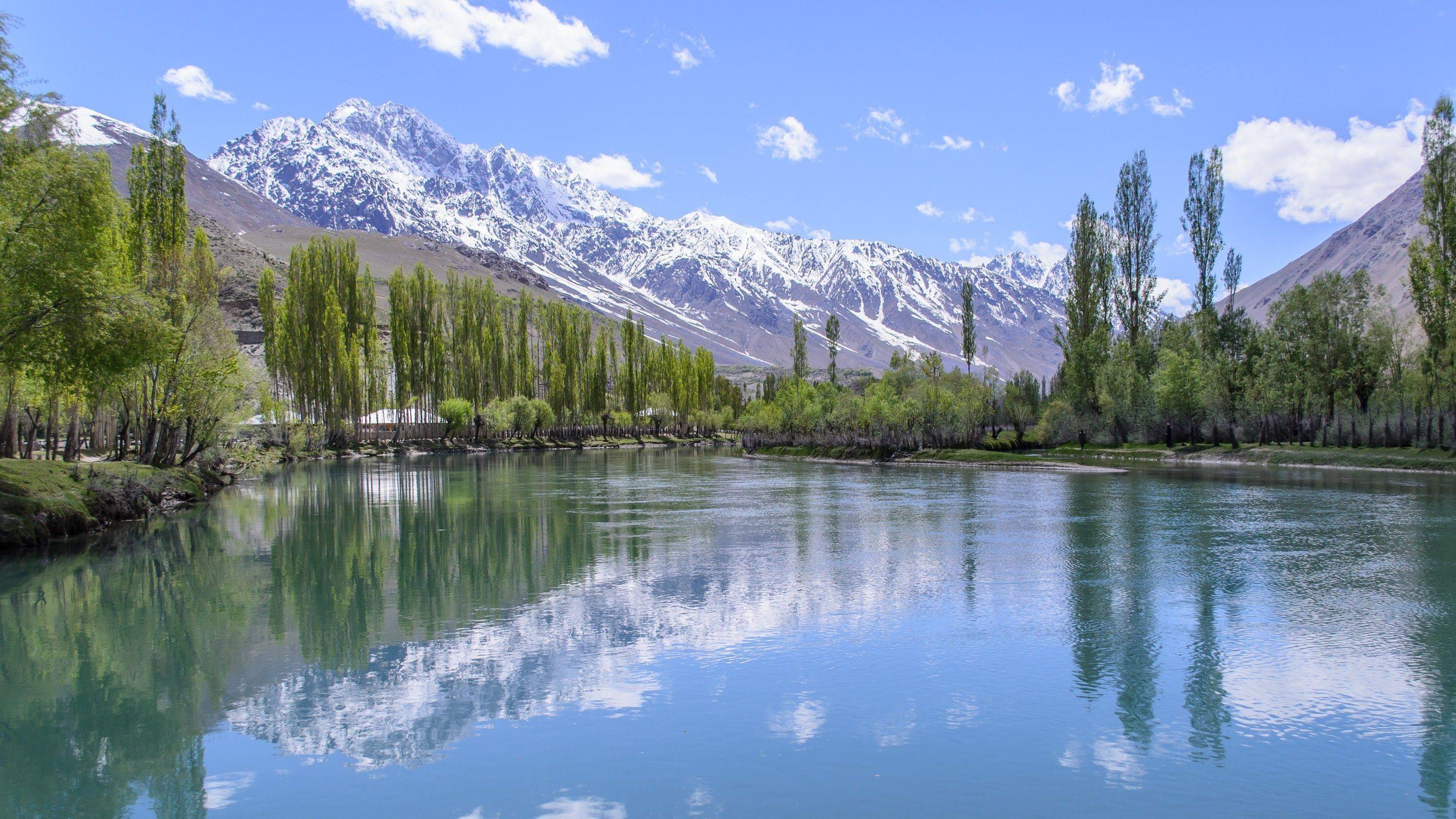}
			\label{fig:nas:in9}}\hspace{-2.4mm}
		\adjincludegraphics[width=0.04\linewidth, trim={{0.17\width} {0.168\height} {0.73\width} {0.292\height}},clip]{fig/compare_nas/input/9}
		{\includegraphics[width=0.216\linewidth]{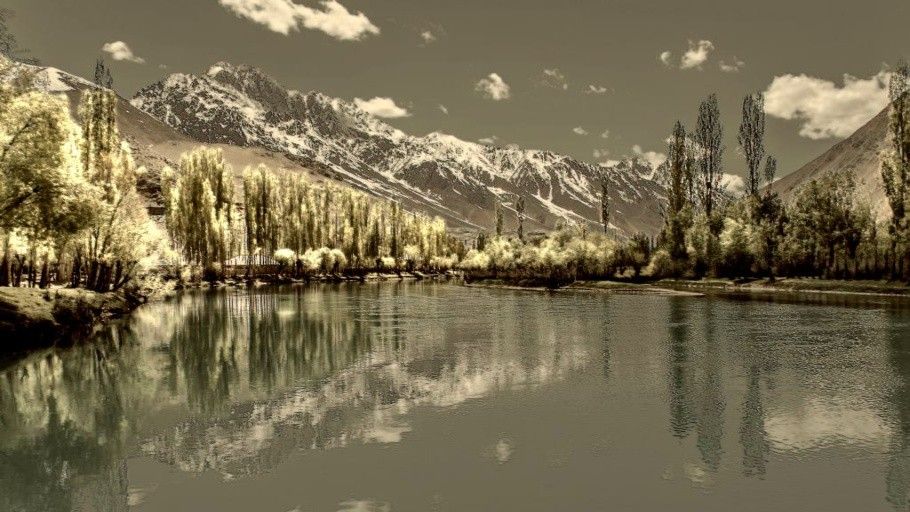}
			\label{fig:nas:res9}}\hspace{-2.4mm}
		\adjincludegraphics[width=0.04\linewidth, trim={{0.17\width} {0.168\height} {0.73\width} {0.292\height}},clip]{fig/compare_nas/nas/9}
		{\includegraphics[width=0.216\linewidth]{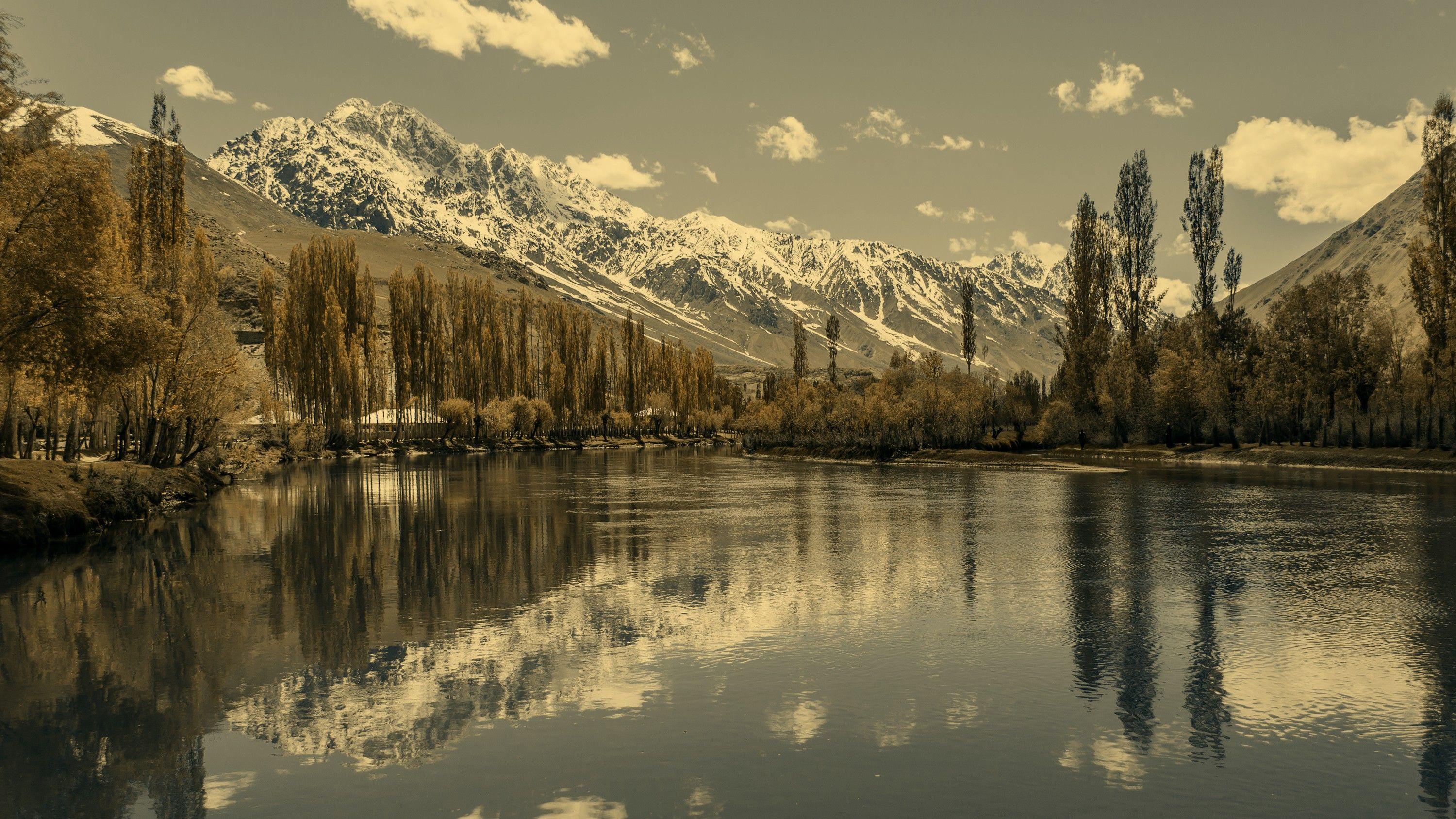}
			\label{fig:ours9}}\hspace{-2.4mm}
		\adjincludegraphics[width=0.04\linewidth, trim={{0.17\width} {0.168\height} {0.73\width} {0.292\height}},clip]{fig/compare_nas/ours/data3_9}
	\end{minipage}\\
%	\begin{minipage}{1\linewidth}
%		{\includegraphics[width=0.21\linewidth]{fig/compare_nas/style/13}
%			\label{fig:nas:tar13}}
%		{\includegraphics[width=0.216\linewidth]{fig/compare_nas/input/13}
%			\label{fig:nas:in13}}\hspace{-2.4mm}
%		\adjincludegraphics[width=0.04\linewidth, trim={{0.559\width} {0.3195\height} {0.402\width} {0.47\height}},clip]{fig/compare_nas/input/13}
%		{\includegraphics[width=0.216\linewidth]{fig/compare_nas/nas/13}
%			\label{fig:nas:res13}}\hspace{-2.4mm}
%		\adjincludegraphics[width=0.04\linewidth, trim={{0.559\width} {0.3195\height} {0.402\width} {0.47\height}},clip]{fig/compare_nas/nas/13}
%		{\includegraphics[width=0.216\linewidth]{fig/compare_nas/ours/data3_13}
%			\label{fig:ours13}}\hspace{-2.4mm}
%		\adjincludegraphics[width=0.04\linewidth, trim={{0.559\width} {0.3195\height} {0.402\width} {0.47\height}},clip]{fig/compare_nas/ours/data3_13}
%	\end{minipage}\\
	\begin{minipage}{1\linewidth}
		{\includegraphics[width=0.21\linewidth]{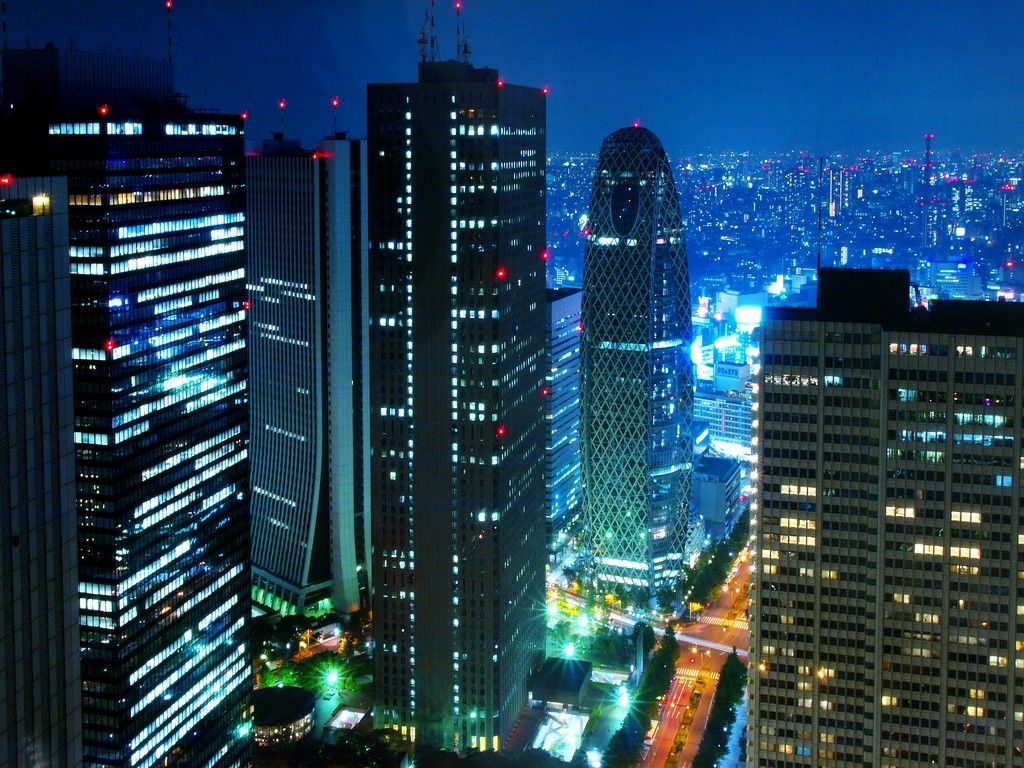}
			\label{fig:nas:tar14}}
		{\includegraphics[width=0.216\linewidth]{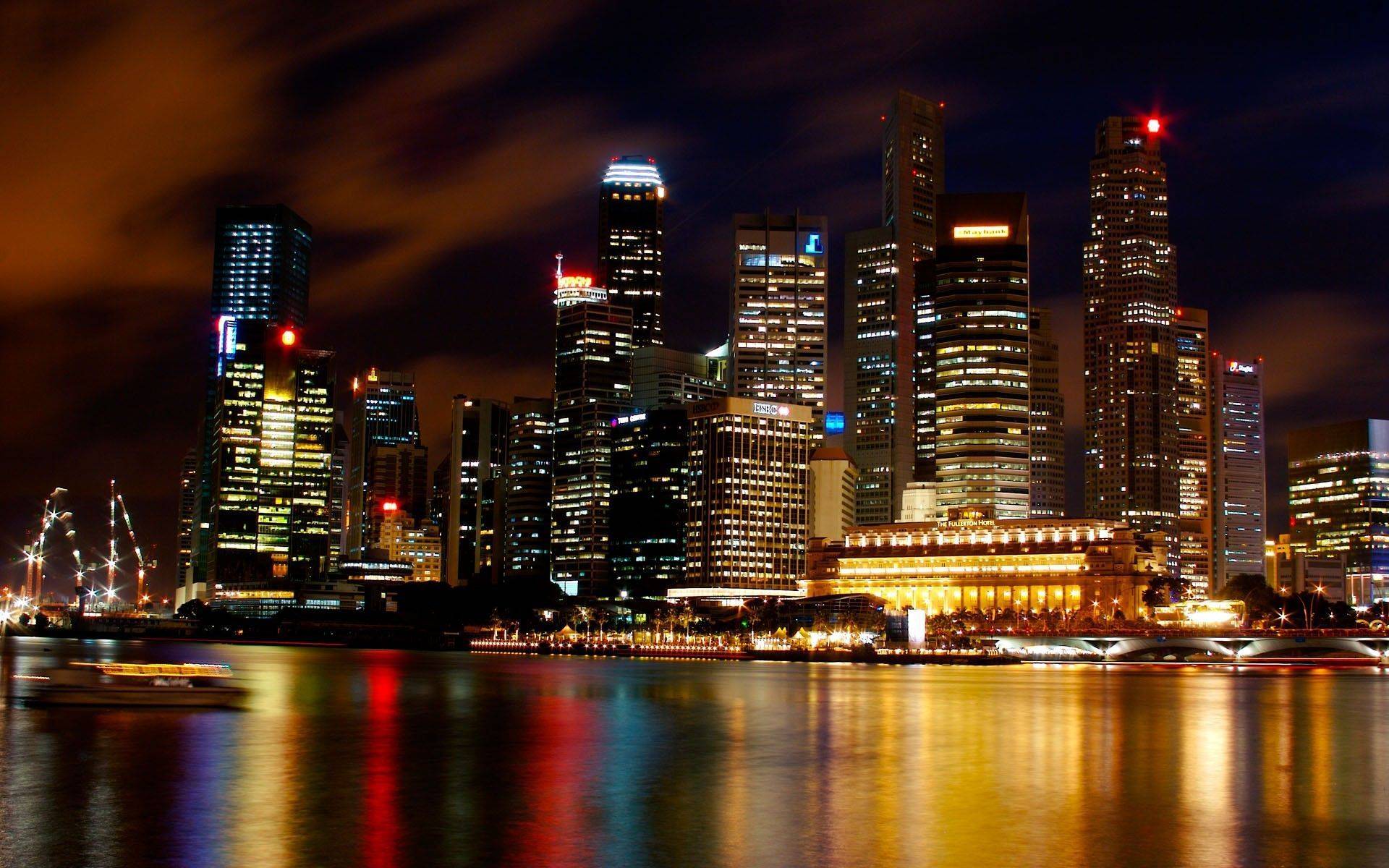}
			\label{fig:nas:in14}}\hspace{-2.4mm}
		\adjincludegraphics[width=0.04\linewidth, trim={{0.66\width} {0.31\height} {0.25\width} {0.205\height}},clip]{fig/compare_nas/input/14}
		{\includegraphics[width=0.216\linewidth]{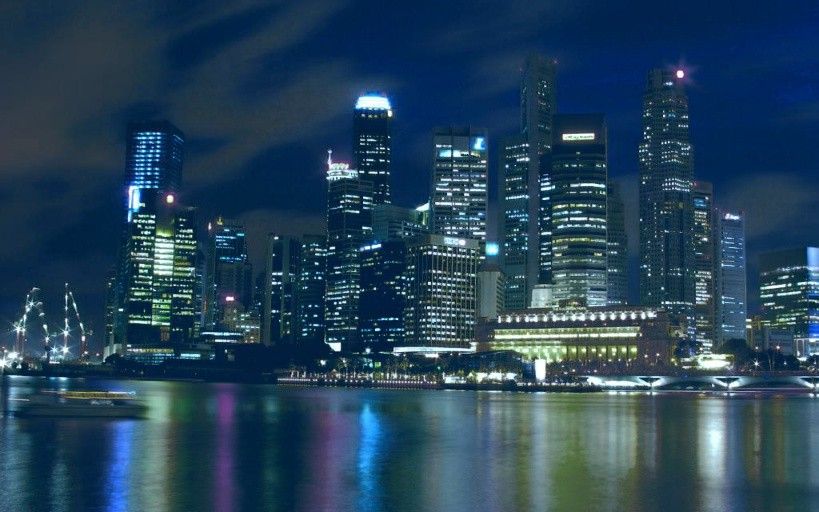}
			\label{fig:nas:res14}}\hspace{-2.4mm}
		\adjincludegraphics[width=0.04\linewidth, trim={{0.66\width} {0.31\height} {0.25\width} {0.205\height}},clip]{fig/compare_nas/nas/14}
		{\includegraphics[width=0.216\linewidth]{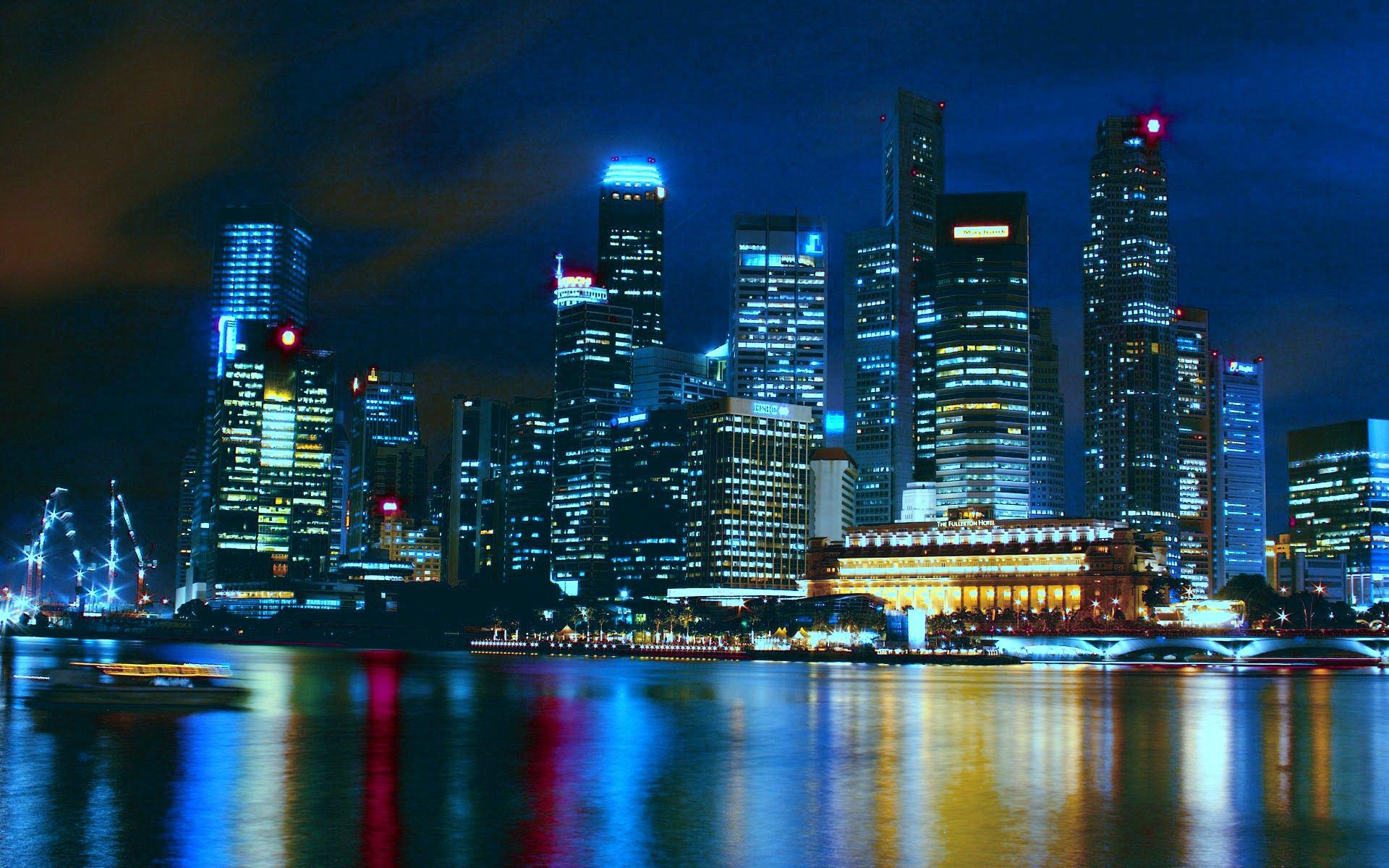}
			\label{fig:ours14}}%\hspace{-2.4mm}
		\adjincludegraphics[width=0.04\linewidth, trim={{0.66\width} {0.31\height} {0.25\width} {0.205\height}},clip]{fig/compare_nas/ours/data3_14}
	\end{minipage}\\
	\begin{minipage}{1\linewidth}
		{\includegraphics[width=0.21\linewidth]{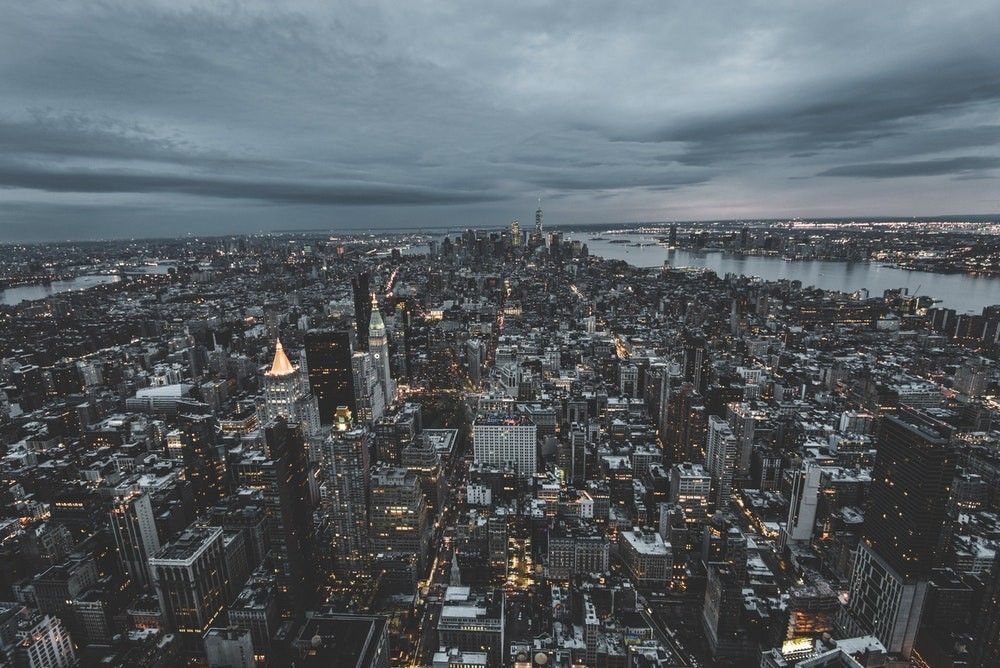}
			\label{fig:nas:tar17}}
		{\includegraphics[width=0.216\linewidth]{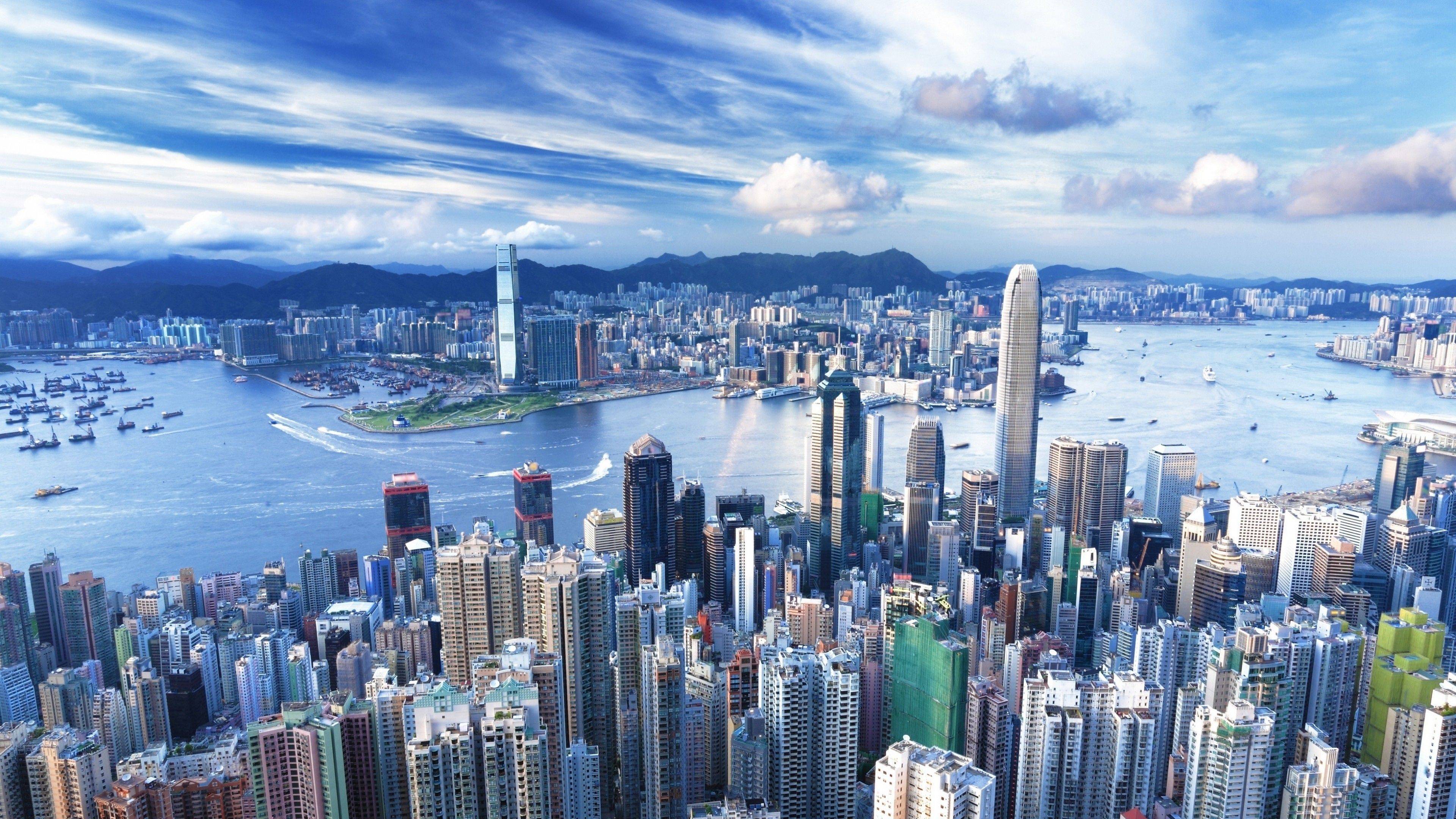}
			\label{fig:nas:in17}}\hspace{-2.4mm}
		\adjincludegraphics[width=0.04\linewidth, trim={{0.603\width} {0.1\height} {0.323\width} {0.5\height}},clip]{fig/compare_nas/input/17}
		{\includegraphics[width=0.216\linewidth]{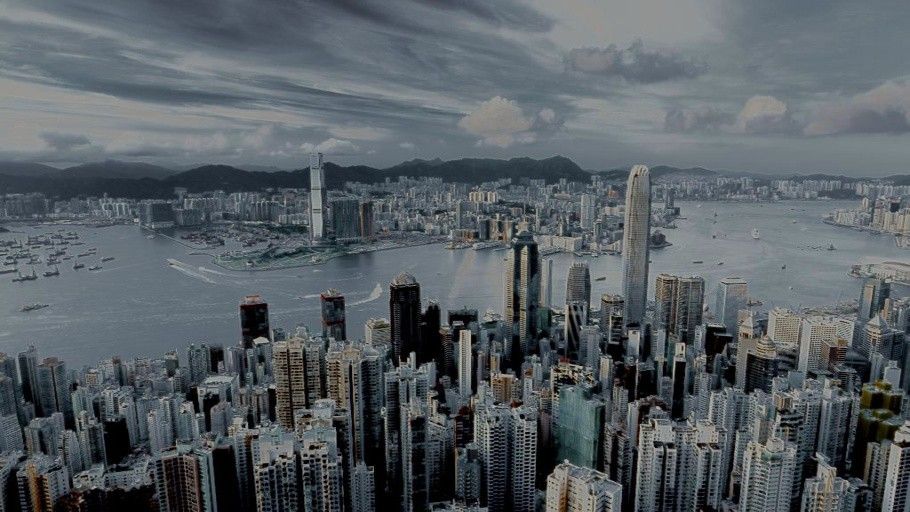}
			\label{fig:nas:res17}}\hspace{-2.4mm}
		\adjincludegraphics[width=0.04\linewidth, trim={{0.603\width} {0.1\height} {0.323\width} {0.5\height}},clip]{fig/compare_nas/nas/17}
		{\includegraphics[width=0.216\linewidth]{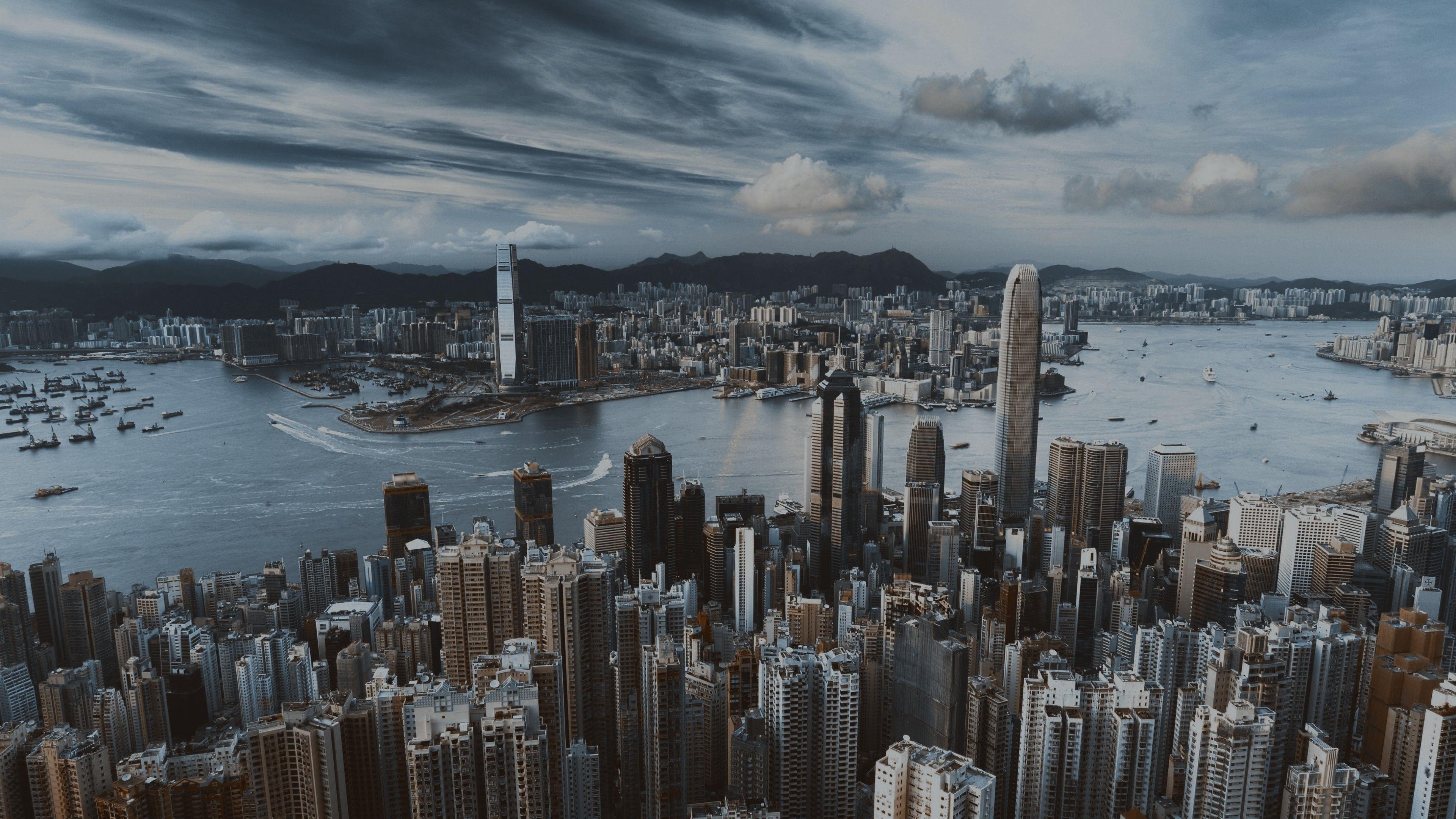}
			\label{fig:ours17}}\hspace{-2.4mm}
		\adjincludegraphics[width=0.04\linewidth, trim={{0.603\width} {0.1\height} {0.323\width} {0.5\height}},clip]{fig/compare_nas/ours/data3_17}
	\end{minipage}\\
	\vspace{1mm}
	\begin{minipage}{1\linewidth}
		{\includegraphics[width=0.21\linewidth]{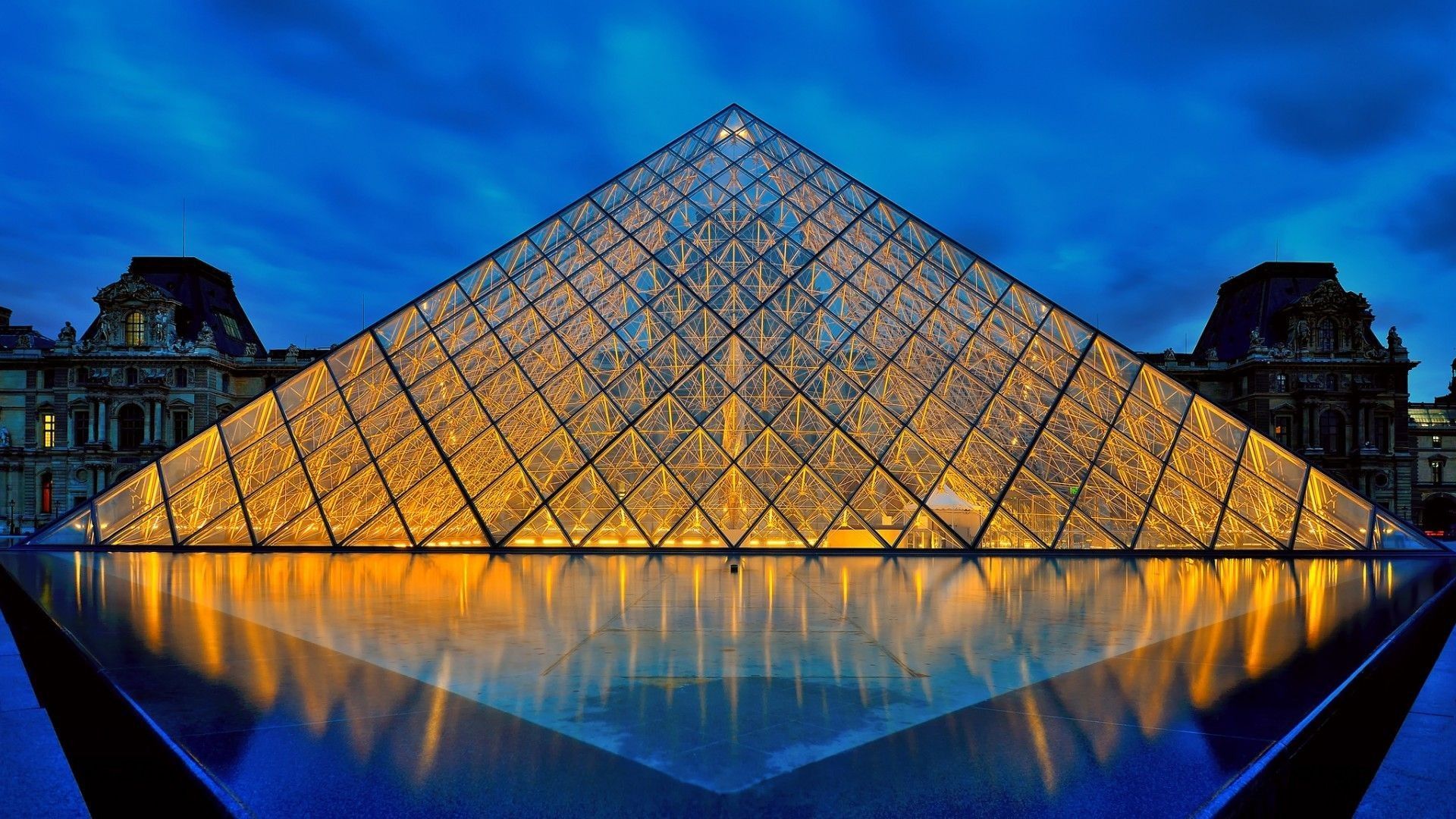}
			\label{fig:nas:tar39}}
		{\includegraphics[width=0.216\linewidth]{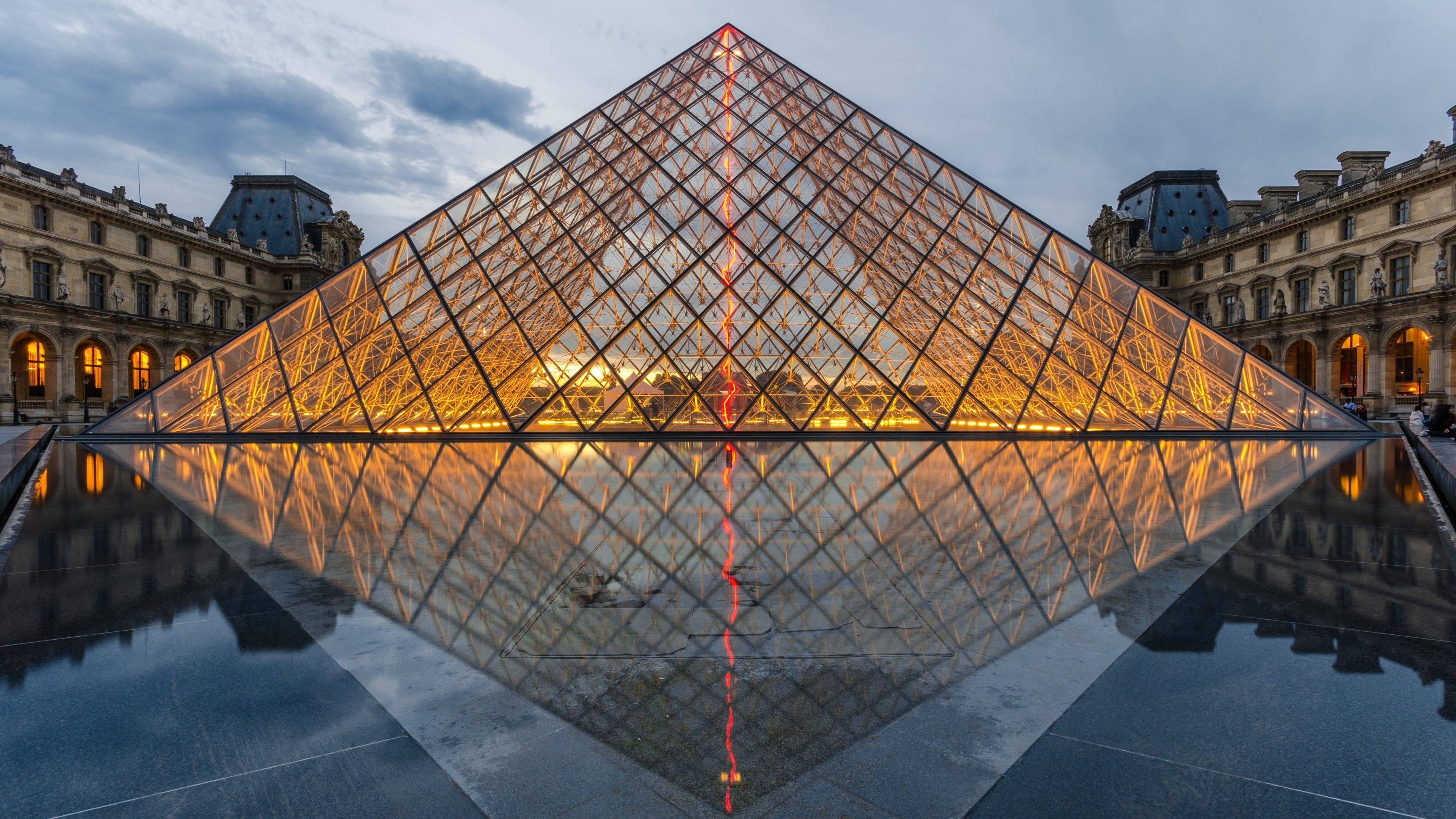}
			\label{fig:nas:in39}}\hspace{-2.4mm}
		\adjincludegraphics[width=0.04\linewidth, trim={{0.17\width} {0.168\height} {0.73\width} {0.292\height}},clip]{fig/compare_nas/input/39}
		{\includegraphics[width=0.216\linewidth]{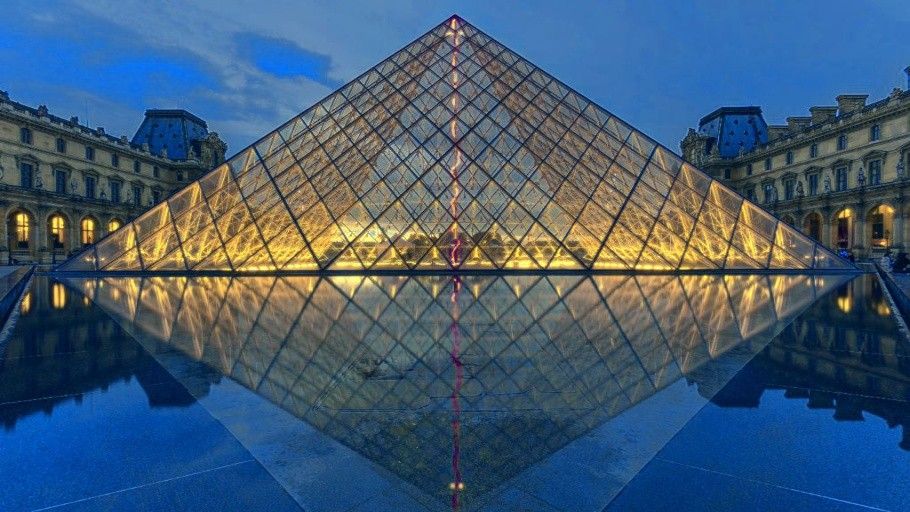}
			\label{fig:nas:res39}}\hspace{-2.4mm}
		\adjincludegraphics[width=0.04\linewidth, trim={{0.17\width} {0.168\height} {0.73\width} {0.292\height}},clip]{fig/compare_nas/nas/39}
		{\includegraphics[width=0.216\linewidth]{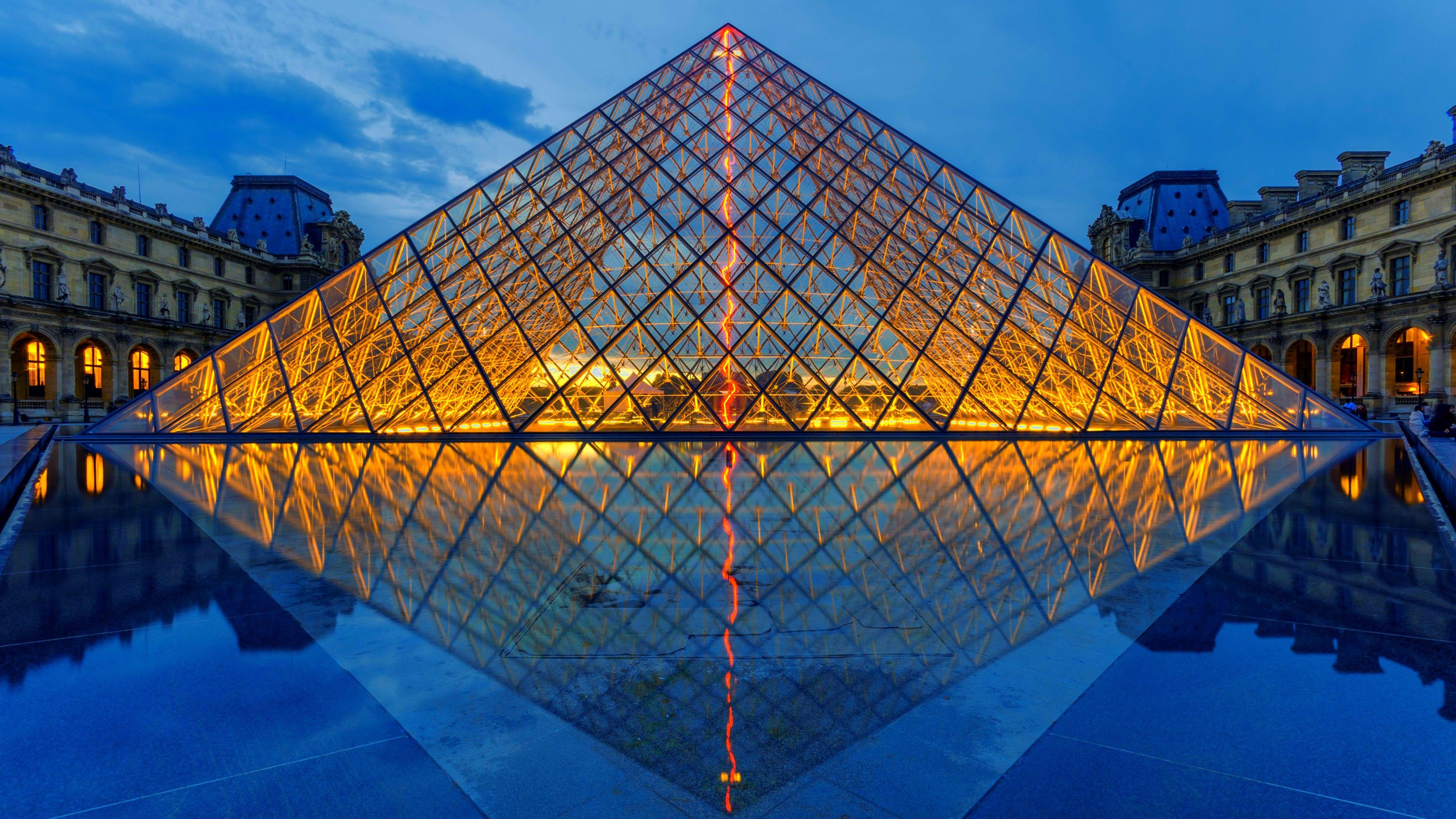}
			\label{fig:ours39}}\hspace{-2.4mm}
		\adjincludegraphics[width=0.04\linewidth, trim={{0.17\width} {0.168\height} {0.73\width} {0.292\height}},clip]{fig/compare_nas/ours/data3_39}
	\end{minipage}
	\caption{Visual comparison with PhotoNAS. 1st column: reference style image. 2nd column: content image. 3rd column: PhotoNAS~\cite{an2020ultrafast}. 4th column: proposed NL-MAT.}
	\label{fig:compare_nas}
\vspace{-5mm}
\end{figure*}

\textcolor{black}{To further evaluate the capability of the proposed scheme in conducting local transfer with global consistency, we conduct experiments on high-resolution images and compare the results with that of the state-of-the-art method PhotoNAS~\cite{an2020ultrafast}. From Fig.~\ref{fig:compare_nas}, we can observe that PhotoNAS is able to preserve the consistency very well in most cases, because it performs the stylization globally on stacked multi-level features extracted from pre-trained models. However, it tends to ignore the context information thus could not transfer dramatic color changes within local areas, as shown in the 3rd column of Fig.~\ref{fig:compare_nas}. On the other hand, the proposed scheme can not only perform  local style transfer according to their contexts, but also preserve the color consistency of the images. Thus it is able to generate more photorealistic images with very fine details.}

\subsection{User Study}
\label{sec:userstudy}
Since the evaluation of photorealistic style transfer tends to be subjective, we conduct two user studies to further validate the proposed method quantitatively. One study asks users to select the result that better carries the style of the reference style image. The other one asks users to select the result that looks more like a real photo without artifacts. We choose 30 images of different scenes from the benchmark dataset offered by Luan~\etal~\cite{luan2017deep} \textcolor{black}{and PhotoNAS~\cite{an2020ultrafast}} and collect responses from Amazon Mechanical Turk (AMT) platform for both studies. The proposed method is compared with photorealistic stylization methods including patch-based (Liao~\etal\cite{liao2017visual} and He~\etal~\cite{he2019progressive}), context-based (Luan~\etal~\cite{luan2017deep}, Li~\etal~\cite{li2018closed}, \textcolor{black}{LST~\cite{li2018learning}, WCT$^2$~\cite{yoo2019photorealistic}),}  and PhotoNAS~\cite{an2020ultrafast}. For each study, there are totally \textcolor{black}{210} questions. For each question, we show the AMT workers a pair of content and style images and the result of our method and one other method. Each question is answered by 30 different workers. Thus the evaluation is based on \textcolor{black}{6,300} responses for each study. The feedback is summarized in Table~\ref{tab:user}. We can observe that, compared to the other photorealistic transfer methods, our method can not only stylize the image well but also generate more photorealistic images. Note that our method only need one pair of data,~\ie, the content and style image without any \textcolor{black}{additional segmentation or classification models}, to generate such results. 

\begin{table}[htb]
\setlength{\abovecaptionskip}{1pt}
\setlength{\belowcaptionskip}{1pt}
	\caption{\textcolor{black}{User study. ($x$\%/$y$\% indicates that for each evaluation, $x$\% users think the other method is better and $y$\% users think the proposed NL-MAT is better.)}}
	\label{tab:user}
	\begin{center}
		\begin{tabular}{l|ll}
			\hline
			Methods&Better Stylization&Photorealistic\\
			\hline
%			Pitie\etal\cite{pitie2005n}/ours&32.17\%/\textbf{67.83\%}&23.5\%/\textbf{76.5\%}\\
			Liao~\etal\cite{liao2017visual}/ours&43.76\%/\textbf{56.24\%}&39.44\%/\textbf{60.56\%}\\
			He~\etal~\cite{he2019progressive}/ours&37.67\%/\textbf{62.33\%}&31.89\%/\textbf{68.11\%}\\
			\hline
			Luan~\etal\cite{luan2017deep}/ours&30.33\%/\textbf{69.67\%}&21.17\%/\textbf{78.83\%}\\
			Li~\cite{li2018closed}/ours&32.83\%/\textbf{67.17\%}&24.0\%/\textbf{76.0\%}\\
			LST~\cite{li2018learning}/ours  & 38.56\%/\textbf{61.44\%} &  31.22\%/\textbf{68.78\%} \\ 
			WCT$^2$~\cite{yoo2019photorealistic}/ours & 25.74\%/\textbf{74.26\%} & 22.22\%/\textbf{77.78\%} \\
			\hline
			PhotoNAS~\cite{an2020ultrafast}/ours  & 26.11\%/\textbf{73.89\%} & 21.56\%/\textbf{78.44\%} \\
			\hline
		\end{tabular}
	\end{center}
\vspace{-8mm}
\end{table}

%TODO:? effect of the representation scheme or ?
\subsection{\textcolor{black}{Effect of Non-local Representation Scheme}}
\label{sec:evaluationdecoupling}
The key to the success of NL-MAT lies in the capability of decoupling the \textcolor{black}{color bases %information (by way of color bases)
 from the matched context-sensitive non-local representations, }%context information (by way of context-sensitive non-local representations) and matching the representations between the image pair. %decoupling-based representations with context-sensitive statistical matching. 
%As discussed in Secs.~\ref{sec:formulate} and~\ref{sec:proposed:context}, we can evaluate the performance of style transfer through the representations. That is, 
%Hence the performance of style transfer is influenced largely by the effectiveness of the proposed representation scheme, 
where the discriminative capacity of the representations reflects the effectiveness of the local style transfer and the correct representation matching indicates the correct color transfer. To better demonstrate the capability of the proposed scheme, we develop a representation visualization mechanism to show the reasoning of the scheme and how the representations are matched after decoupling. %in Fig.~\ref{fig:seg}. 

%The key to the success of NL-MAT lies in the decoupling-based representations with context-sensitive statistical matching. In this section, to further evaluate the proposed method, we show how the representations are matched after decoupling. %in Fig.~\ref{fig:seg}. 

%TODO: reasoning of the scheme, not a good word. 
%TODO: should we use context-sensitivity or discriminative compacity.

\subsubsection{\textcolor{black}{Visualization Mechanism for Representation}}
\label{sec:visualize}
\begin{figure*}[htbp]
\setlength{\abovecaptionskip}{1pt}
\setlength{\belowcaptionskip}{1pt}
	\subfloat[]{\begin{minipage}{0.34\linewidth}
		\centering
		{\includegraphics[width=0.3\linewidth]{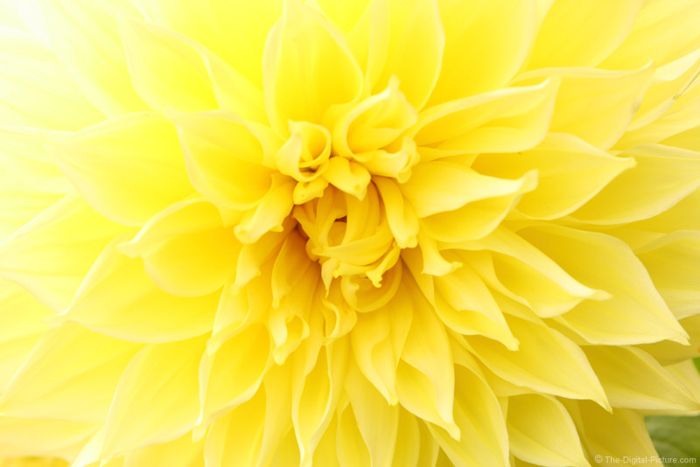}}
		{\includegraphics[width=0.3\linewidth]{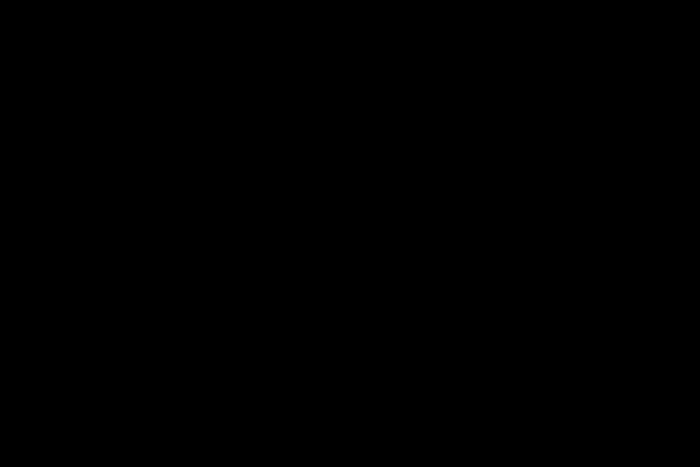}}
		{\includegraphics[width=0.3\linewidth]{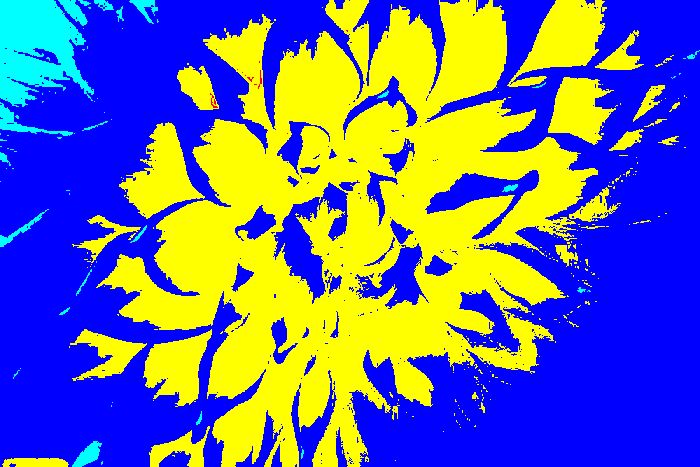}}\\
		{\includegraphics[width=0.3\linewidth]{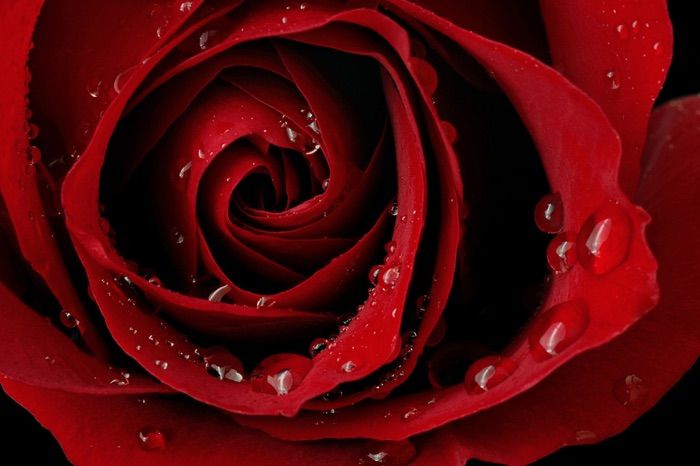}}
		{\includegraphics[width=0.3\linewidth]{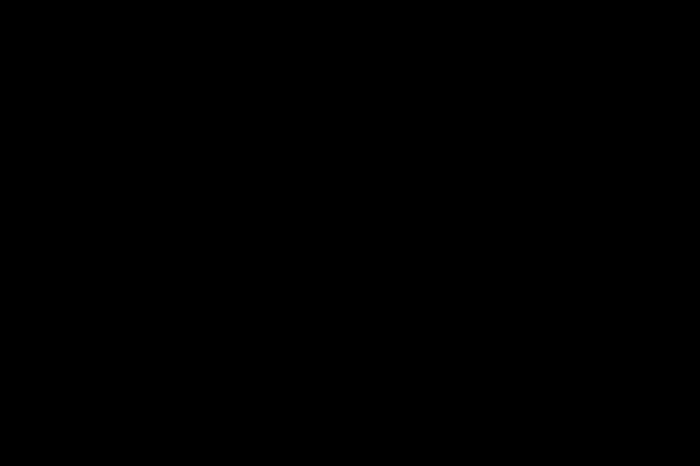}}
		{\includegraphics[width=0.3\linewidth]{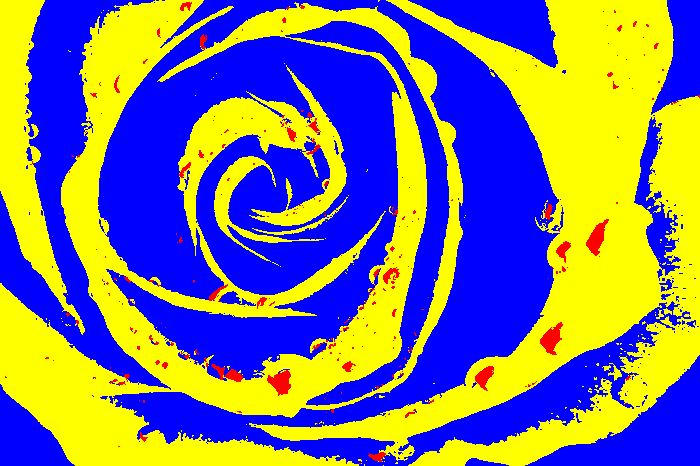}}
	\end{minipage}}%\vspace{1mm}
%	\subfloat[]{\begin{minipage}{0.5\linewidth}
%		\centering
%		{\includegraphics[width=0.3\linewidth]{fig/compare_context/14_0in}}
%		{\includegraphics[width=0.3\linewidth]{fig/seg/in14}}
%		{\includegraphics[width=0.3\linewidth]{fig/seg/14_content_map2}}\\
%		{\includegraphics[width=0.3\linewidth]{fig/compare_context/14_1tar}}
%		{\includegraphics[width=0.3\linewidth]{fig/seg/tar14}}
%		{\includegraphics[width=0.3\linewidth]{fig/seg/14_style_map2}}
%	\end{minipage}}\\\vspace{1mm}
%	\subfloat[]{\begin{minipage}{0.5\linewidth}
%		\centering
%		{\includegraphics[width=0.3\linewidth]{fig/compare_patch/31_0in}}
%		{\includegraphics[width=0.3\linewidth]{fig/seg/in31}}
%		{\includegraphics[width=0.3\linewidth]{fig/seg/31_content_map1}}\\
%		{\includegraphics[width=0.3\linewidth]{fig/compare_patch/31_1tar}}
%		{\includegraphics[width=0.3\linewidth]{fig/seg/tar31}}
%		{\includegraphics[width=0.3\linewidth]{fig/seg/31_style_map1}}
%	\end{minipage}}\\\vspace{1mm}
%	\subfloat[]{\begin{minipage}{0.5\linewidth}
%		\centering
%		{\includegraphics[width=0.3\linewidth]{fig/compare_context/6_0in}}
%		{\includegraphics[width=0.3\linewidth]{fig/seg/in6}}
%		{\includegraphics[width=0.3\linewidth]{fig/seg/6_content_map2}}\\
%		{\includegraphics[width=0.3\linewidth]{fig/compare_context/6_1tar}}
%		{\includegraphics[width=0.3\linewidth]{fig/seg/tar6}}
%		{\includegraphics[width=0.3\linewidth]{fig/seg/6_style_map2}}
%	\end{minipage}}\vspace{1mm}
	\subfloat[]{\begin{minipage}{0.31\linewidth}
		\centering
		{\includegraphics[width=0.3\linewidth]{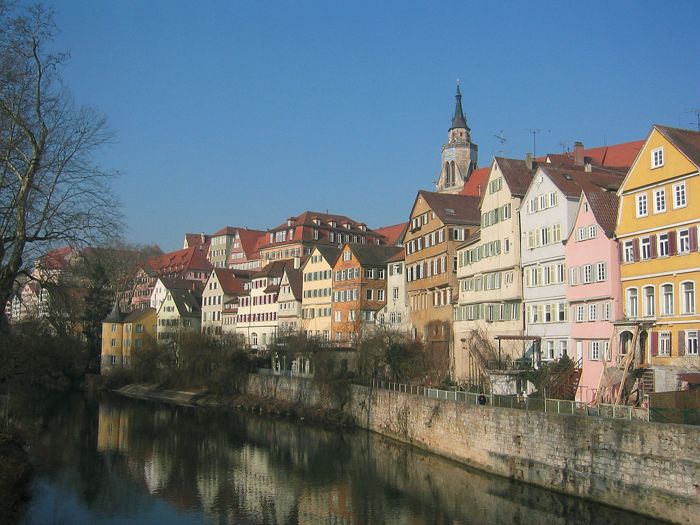}}
		{\includegraphics[width=0.3\linewidth]{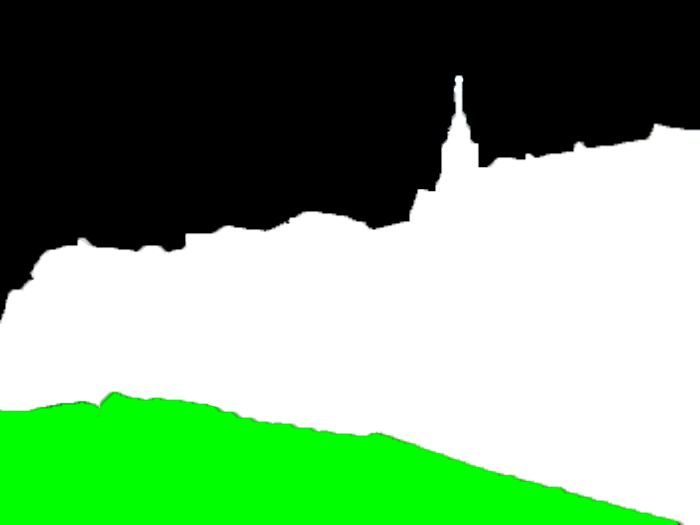}}
		{\includegraphics[width=0.3\linewidth]{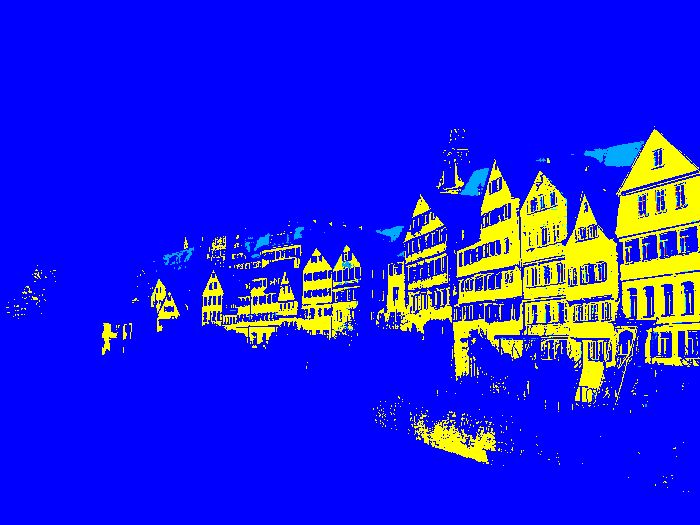}}\\
		{\includegraphics[width=0.3\linewidth]{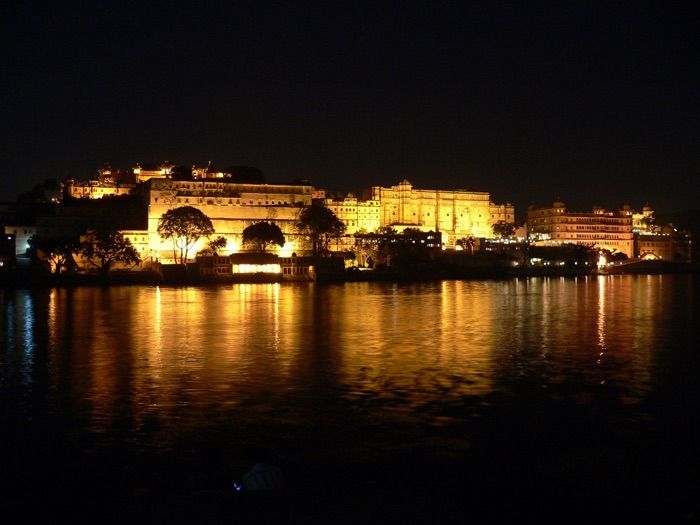}}
		{\includegraphics[width=0.3\linewidth]{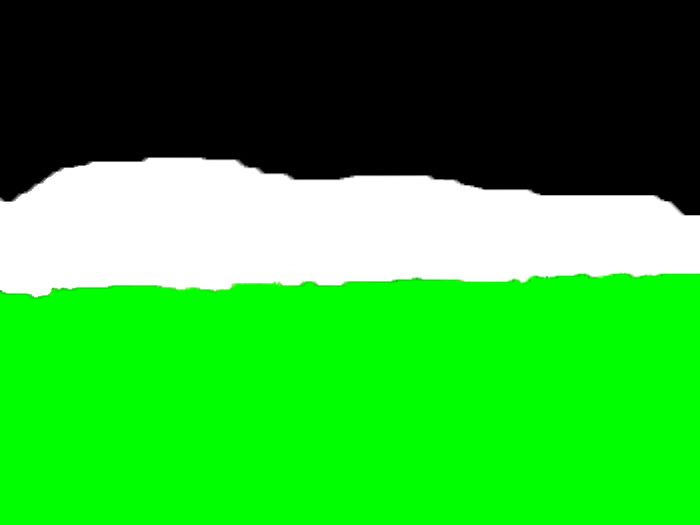}}
		{\includegraphics[width=0.3\linewidth]{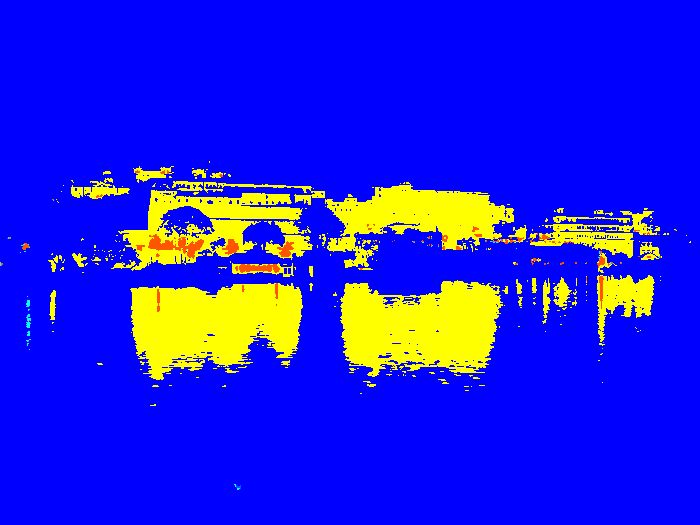}}
	\end{minipage}}%\\\vspace{1mm}
		\subfloat[]{\begin{minipage}{0.36\linewidth}
		\centering
		{\includegraphics[width=0.32\linewidth]{fig/compare_patch/2_0in}}
		{\includegraphics[width=0.32\linewidth]{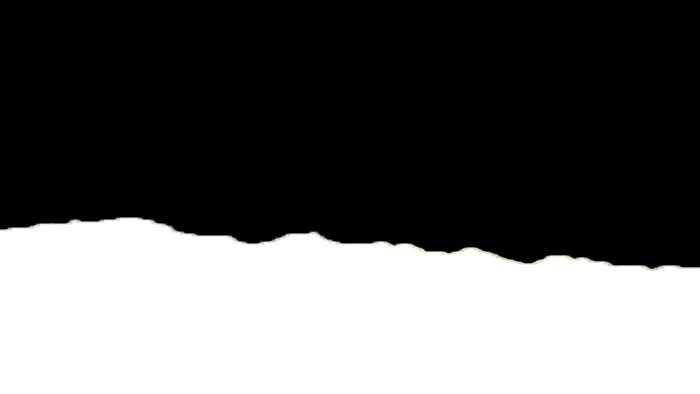}}
		{\includegraphics[width=0.32\linewidth]{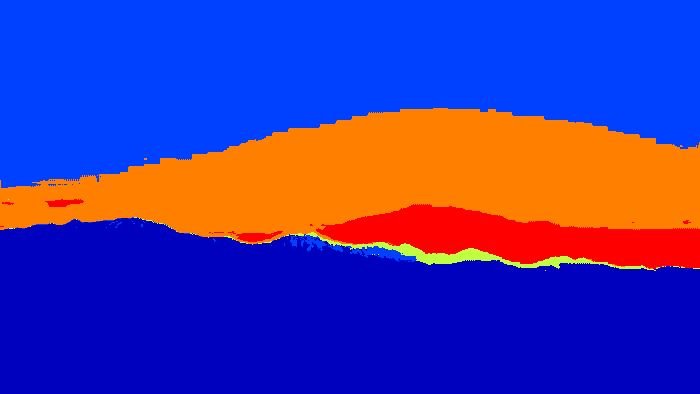}}\\
		{\includegraphics[width=0.32\linewidth]{fig/compare_patch/2_1tar}}
		{\includegraphics[width=0.32\linewidth]{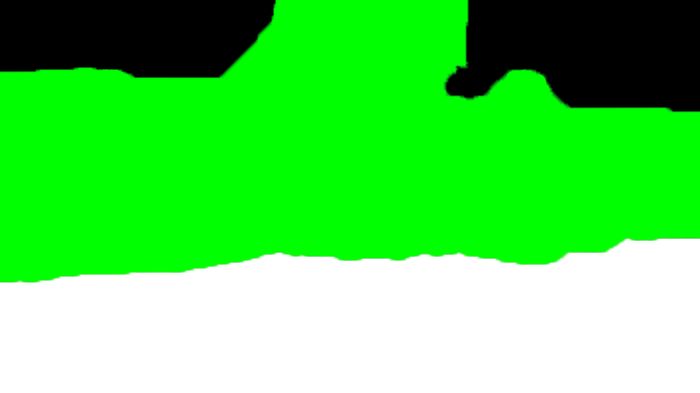}}
		{\includegraphics[width=0.32\linewidth]{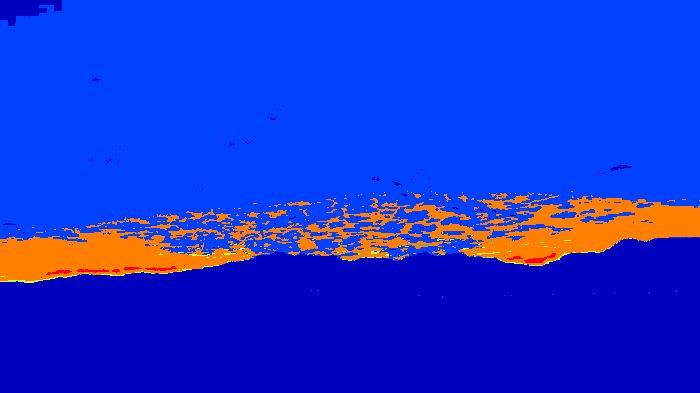}}
	\end{minipage}}%\vspace{1mm}
%	\subfloat[]{\begin{minipage}{0.5\linewidth}
%		\centering
%		{\includegraphics[width=0.3\linewidth]{fig/compare_patch/56_0in}}
%		{\includegraphics[width=0.3\linewidth]{fig/seg/in56}}
%		{\includegraphics[width=0.3\linewidth]{fig/seg/56_content_map2}}\\
%		{\includegraphics[width=0.3\linewidth]{fig/compare_patch/56_1tar}}
%		{\includegraphics[width=0.3\linewidth]{fig/seg/tar56}}
%		{\includegraphics[width=0.3\linewidth]{fig/seg/56_style_map2}}
%	\end{minipage}}\\\vspace{1mm}
	\caption{Major color index map (MCIM) of the decoupled representations from the proposed NL-MAT. The first column of each group shows the content (top) and style (bottom) images, respectively. The second column of each group shows the segmentation maps of the content and style images from the method of Luan~\cite{luan2017deep}. The third column of each group shows the MCIM of the content and style images from the proposed method, respectively. Note that the pseudo-colors in the MCIMs are arbitrarily selected to represent the indices of the largest color basis. Thus the color itself is not important. It is the matching between the content MCIM and the style MCIM that matters.}
	\label{fig:seg}
\vspace{-5mm}
\end{figure*}

As described in Sec.~\ref{sec:proposed}, %due to the stick-breaking structure, 
the representation vector of a single pixel, $\mathbf{s}_\rightarrow = \{s_{i}\}_{1 \leq i \leq k}$ %needs to sum to one, \ie, $\sum_{i=1}^{k}{s_{i}}=1$, where 
indicates the  proportion of each of the $k$ color basis in making up the given pixel. \textcolor{black}{By introducing the sparse constraint, the representations are more discriminative, which allows for diverse local transfer in a global consistent fashion. Since the transfer of each context is mainly determined by dominant color bases that have larger representation values, the visualization mechanism is designed based on such dominant bases, which is}
%To better understand the capability of decoupling, we design a new visualization mechanism, 
referred to as the ``major color index map (MCIM).'' The MCIM of each image is constructed by the ``index'' of the largest representation of each pixel, which indicates the most important color basis for that given pixel. Mathematically, it can be expressed as
\begin{equation}
t =\mathop{\arg\max}_{j}\{s_1,\cdots,s_j,\cdots,s_k\}
\label{equ:max}
\end{equation}
where $k$ is the number of color bases. 
With the indices of the largest representation of all the pixels, we are able to define the MCIM of the entire image.

%For each group, the first column shows the content (top) and style images (bottom), respectively. The second column of each group shows the segmentation maps of the content and style images from the method \textcolor{black}{used by} Luan~\etal~\cite{luan2017deep}. And the third column of each group shows the MCIMs of the content and style images from the proposed method, respectively. 
\textcolor{black}{The MCIMs of toy examples are shown in Fig.~\ref{fig:seg}, where each pseudo-color indicates a single index. Note that, since some objects may have more than one major color basis, the index map only roughly segments the image. Nonetheless, MCIM allows us to perform in-depth visual inspection on how the proposed method works in different scenarios. From Fig.~\ref{fig:seg}, we can observe that, the extracted representations from similar objects or parts in the image pairs are context-sensitive and matched in different scenarios, even if the matched objects/parts are with different colors. This is the main reason why NL-MAT can realize local transfer while preserving global consistency, as shown in~\crefrange{fig:patch}{fig:compare_nas}. It is worth mentioning that, the proposed NL-MAT is unsupervised and does not need any additional models or steps for segmentation to perform photorealistic stylization. As a comparison, even though the segmentation method adopted by Luan~\etal~\cite{luan2017deep} is trained in a supervised way, it may not handle unknown objects or objects with complex components as shown in Fig.~\ref{fig:seg}, which may affect the stylization performance.} 

\subsubsection{\textcolor{black}{Reasoning of Scheme}}
\begin{figure*}[htbp]
\setlength{\abovecaptionskip}{1pt}
\setlength{\belowcaptionskip}{1pt}
	\begin{minipage}{1\linewidth}
		\centering
		\subfloat[Style]{\includegraphics[width=0.16\linewidth]{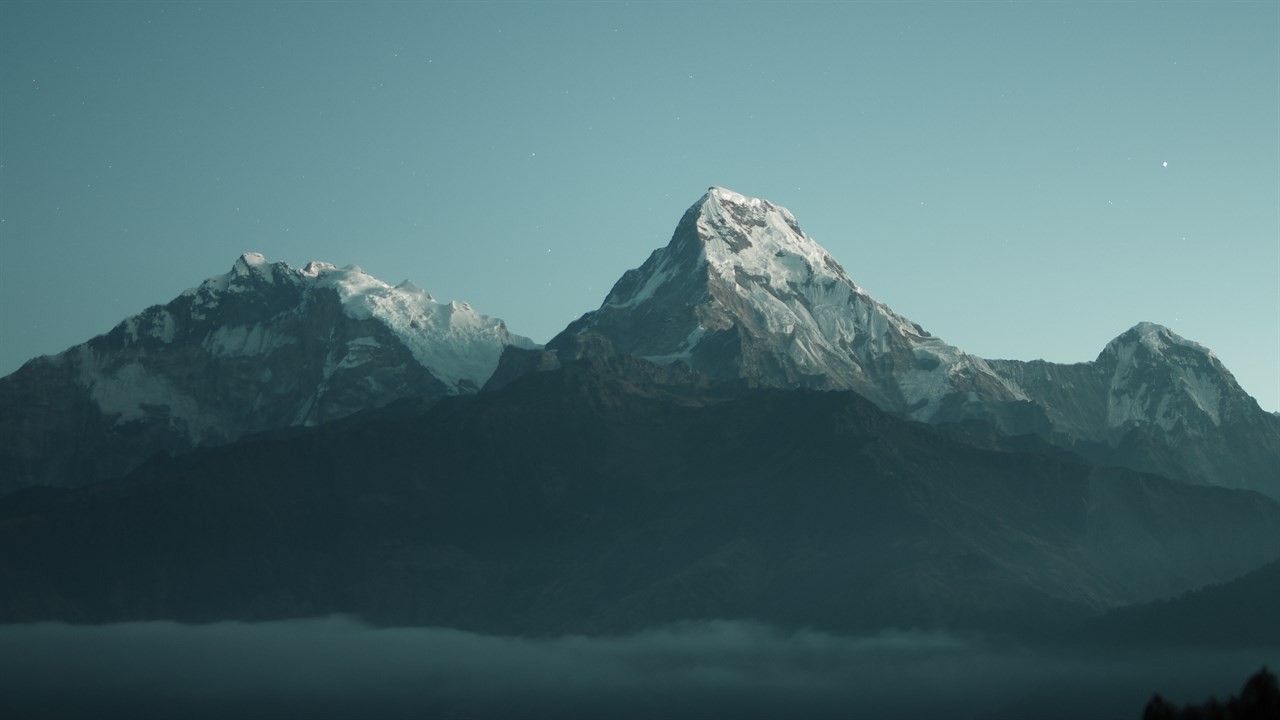}\label{fig:reason:style}}\hspace{0.01mm}
		\subfloat[Style MCIM]{\includegraphics[width=0.16\linewidth]{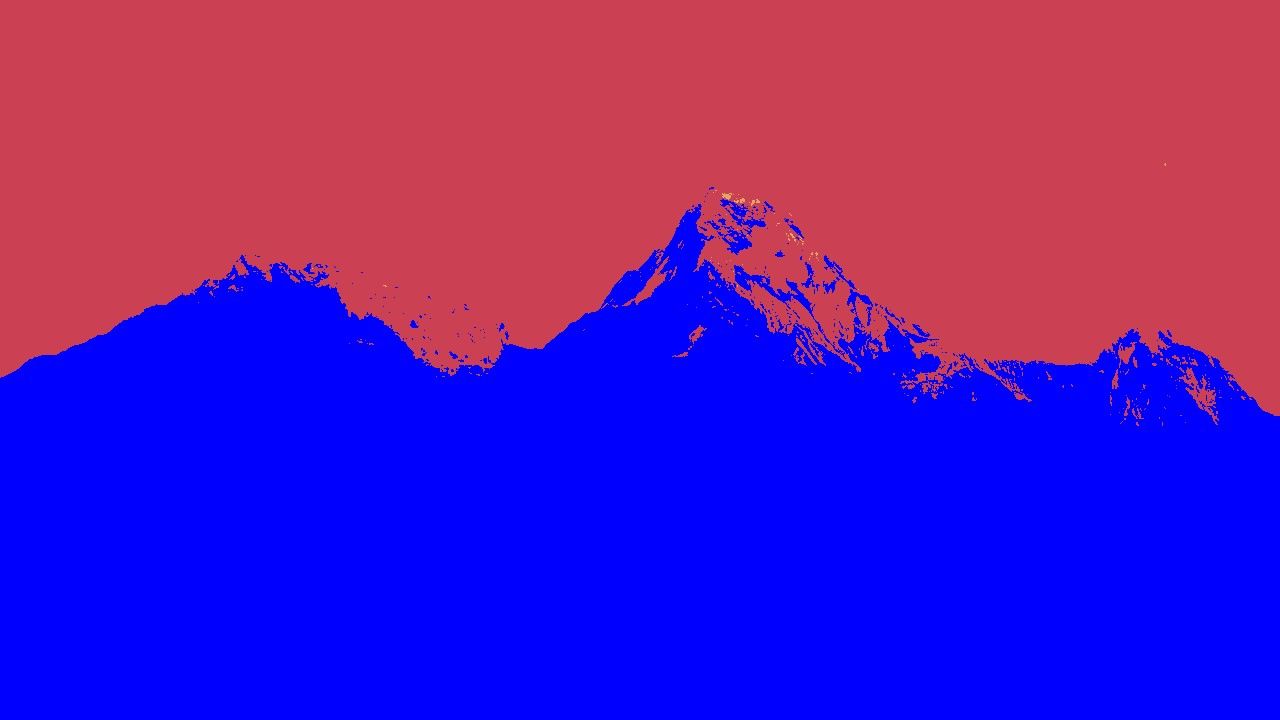}\label{fig:reason:style_seg}}\hspace{0.01mm}
		\subfloat[Content]	{\includegraphics[width=0.16\linewidth]{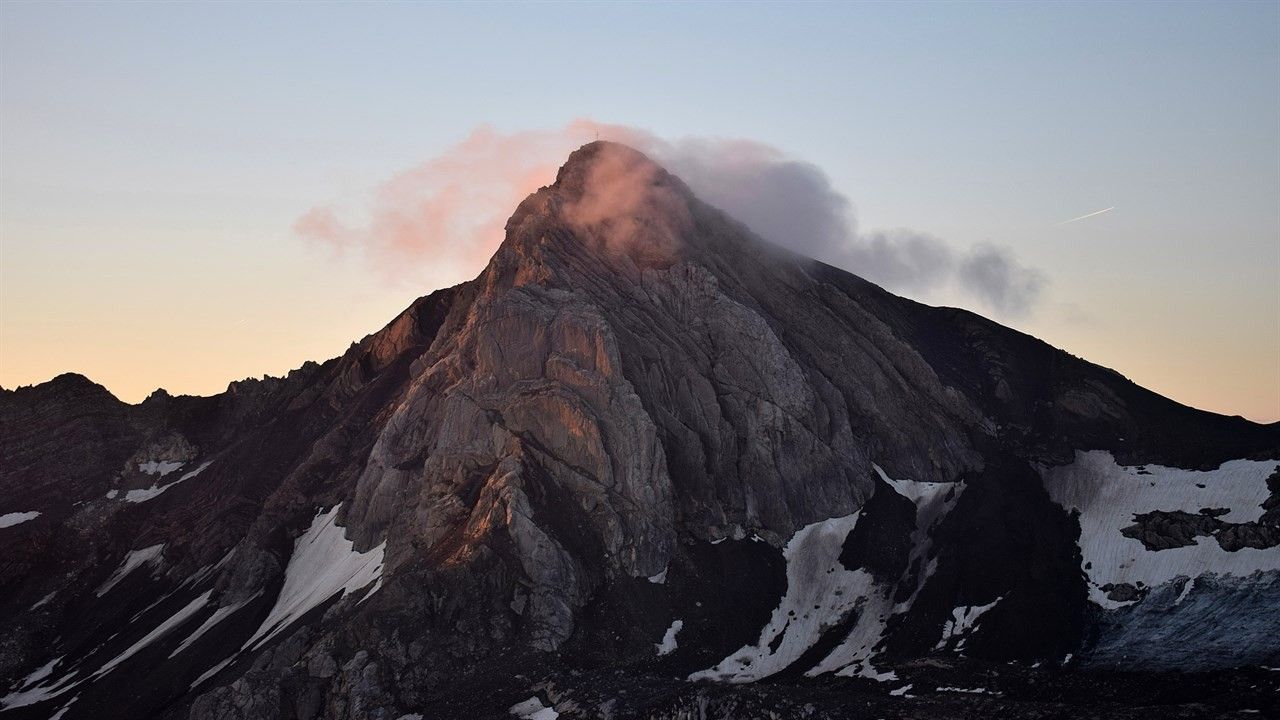}\label{fig:reason:content}}\hspace{0.01mm}
		\subfloat[Content MCIM]{\includegraphics[width=0.16\linewidth]{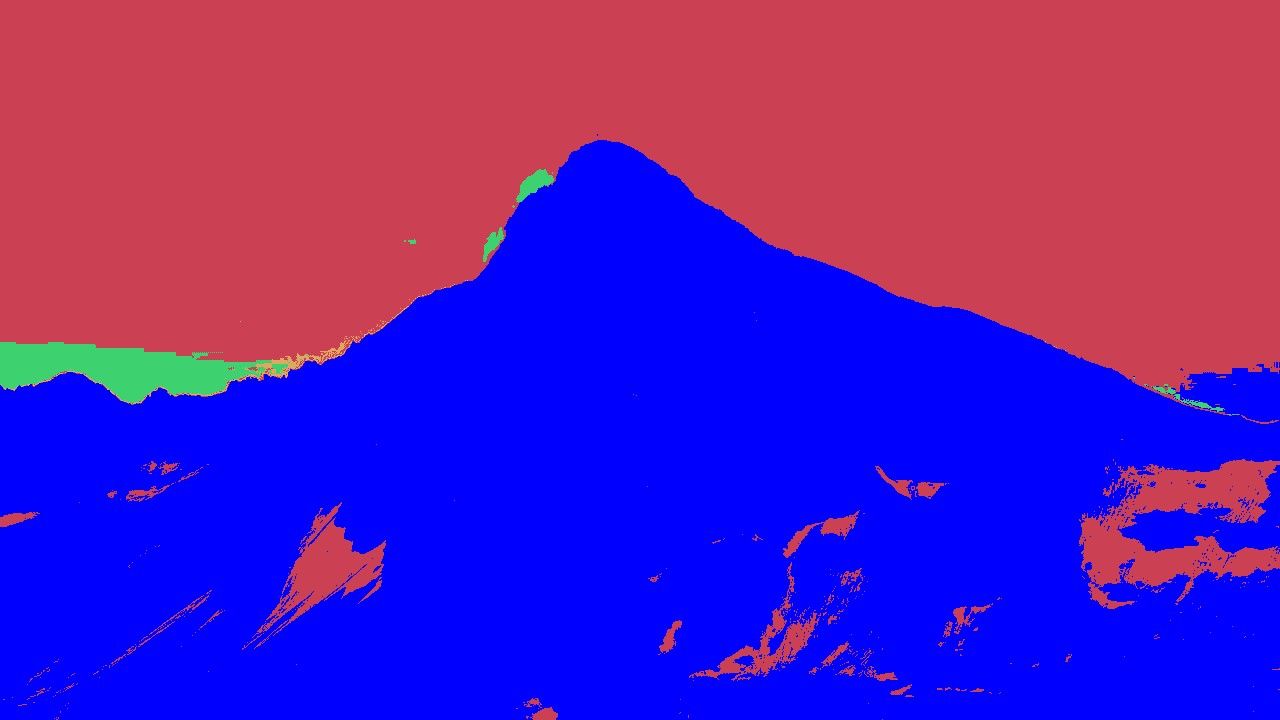}\label{fig:reason:content_seg}}\hspace{0.01mm}
		\subfloat[Output $I_{cs}'$ with affine-transfer]{\includegraphics[width=0.16\linewidth]{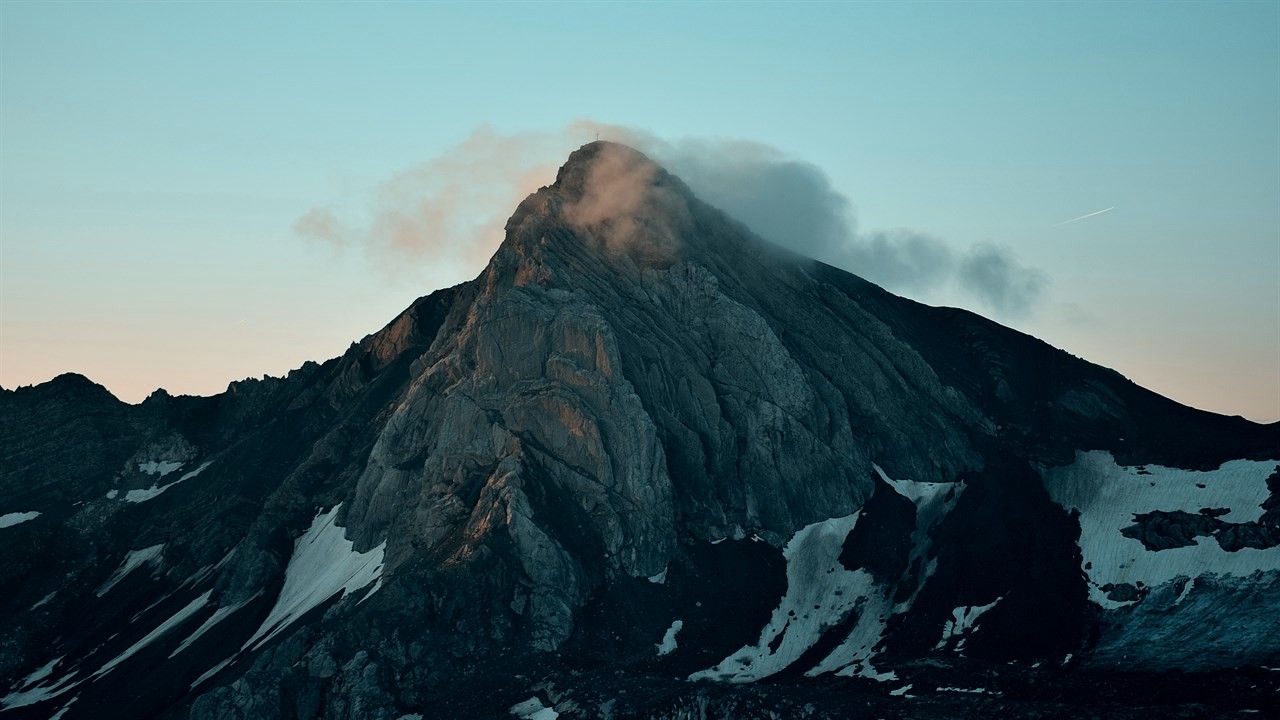}\label{fig:reason:output_affine}}\hspace{0.01mm}
		\subfloat[Output $I_{cs}$ with affine-transfer+WCT]{\includegraphics[width=0.16\linewidth]{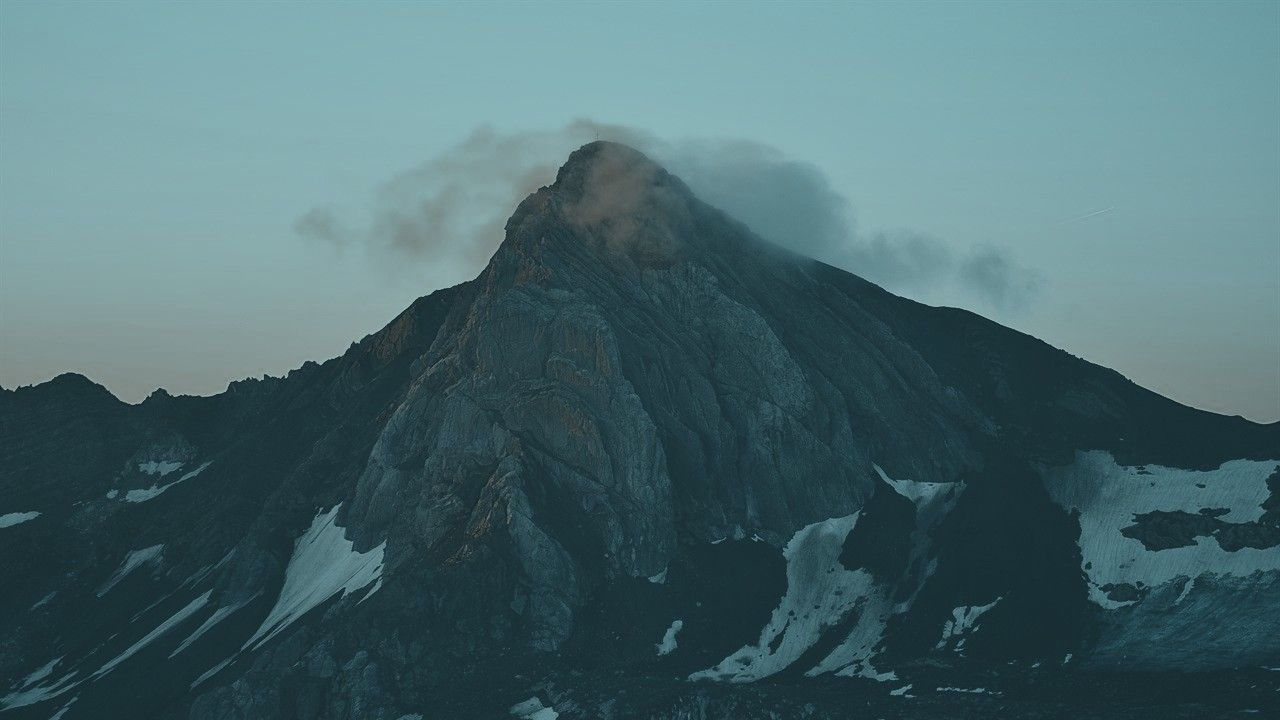}\label{fig:reason:output_wct}}\hfill\\
		\subfloat[Distribution of $I_c$ and $I_s$]{\includegraphics[width=0.19\linewidth]{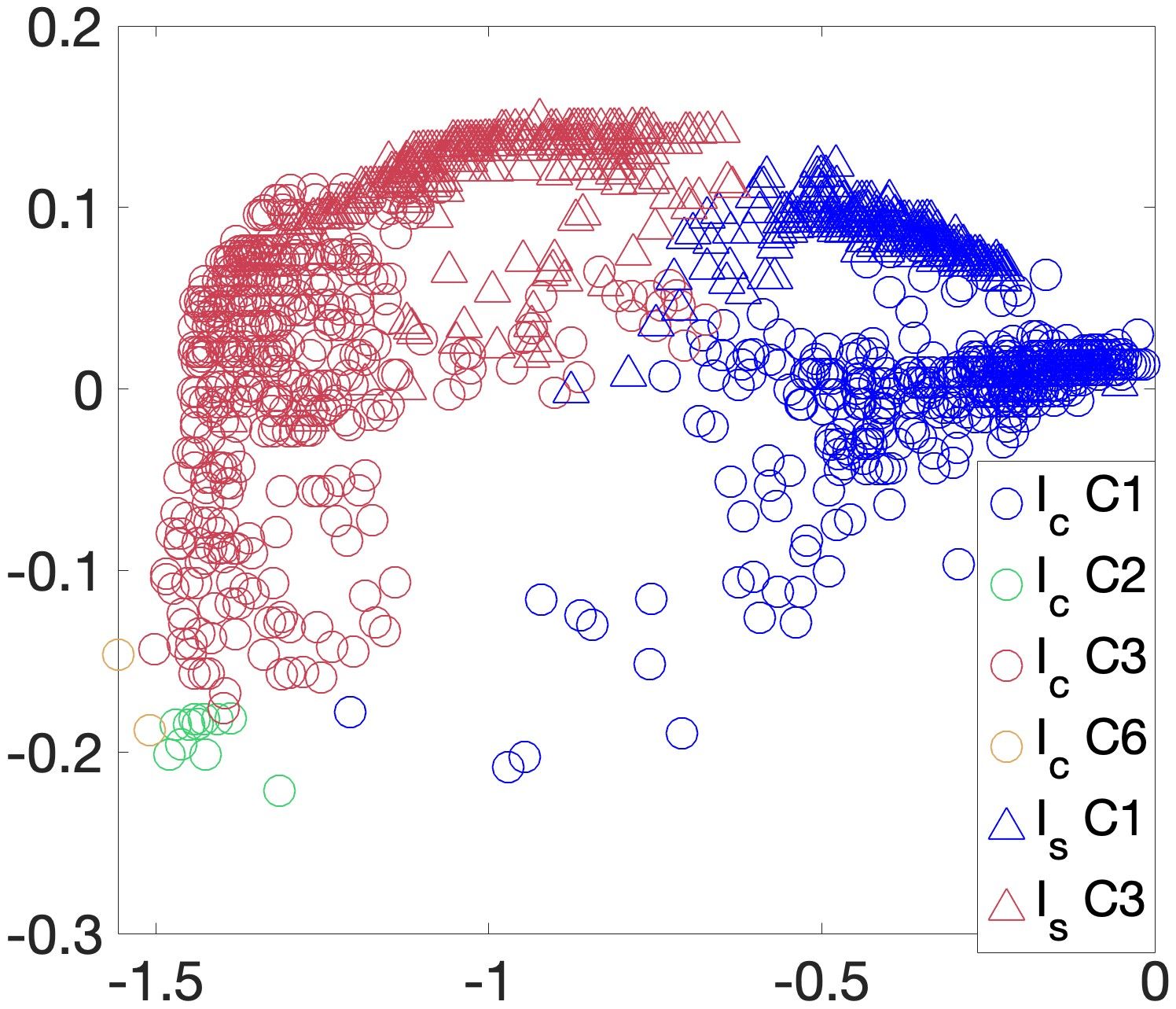}\label{fig:reason:dist_raw}}\hspace{0.01mm}
		\subfloat[Distribution of $S_c$ and $S_s$]{\includegraphics[width=0.19\linewidth]{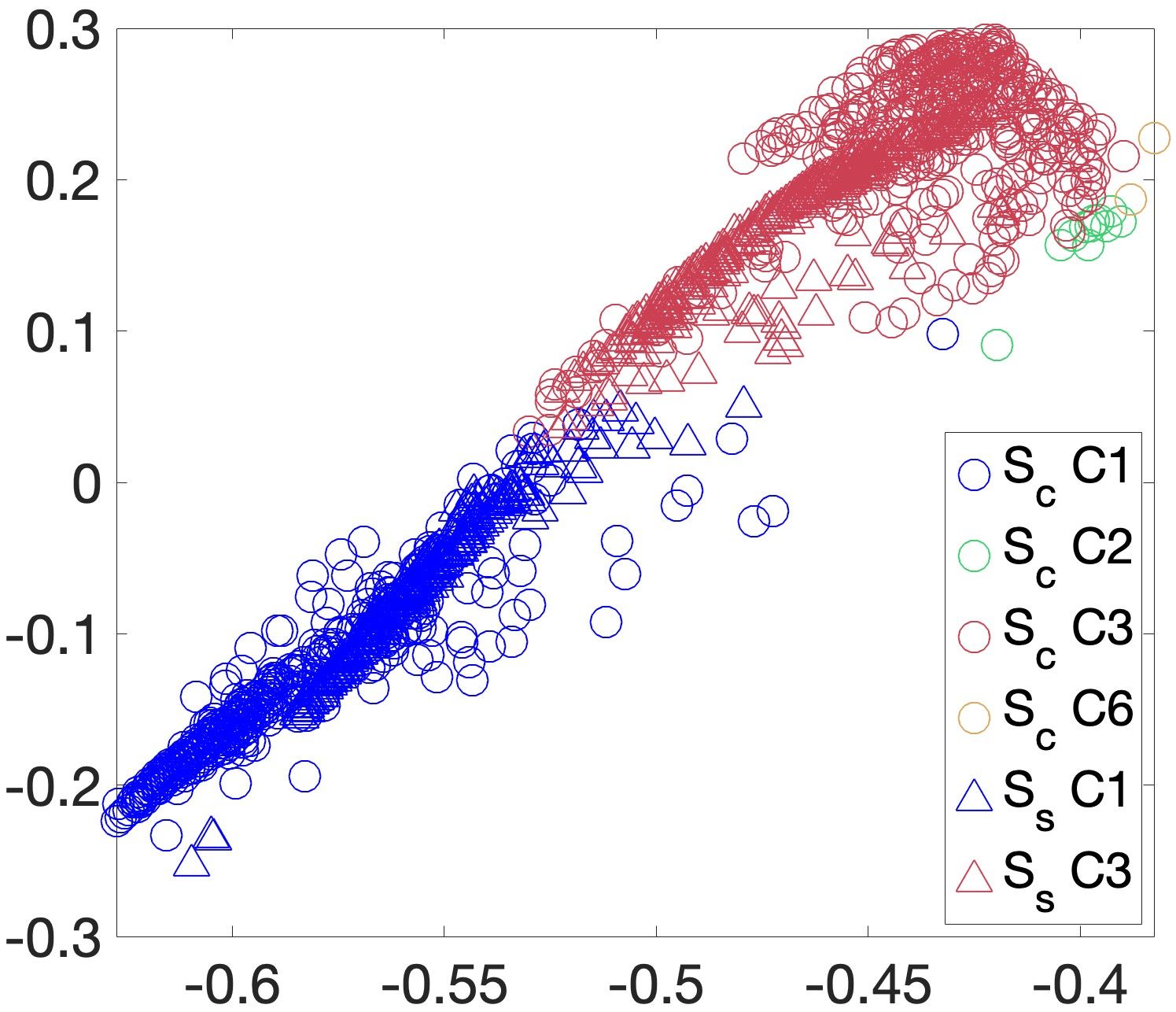}\label{fig:reason:dist_rp}}\hspace{0.01mm}
		\subfloat[Distribution of WCT on $S_c$]	{\includegraphics[width=0.19\linewidth]{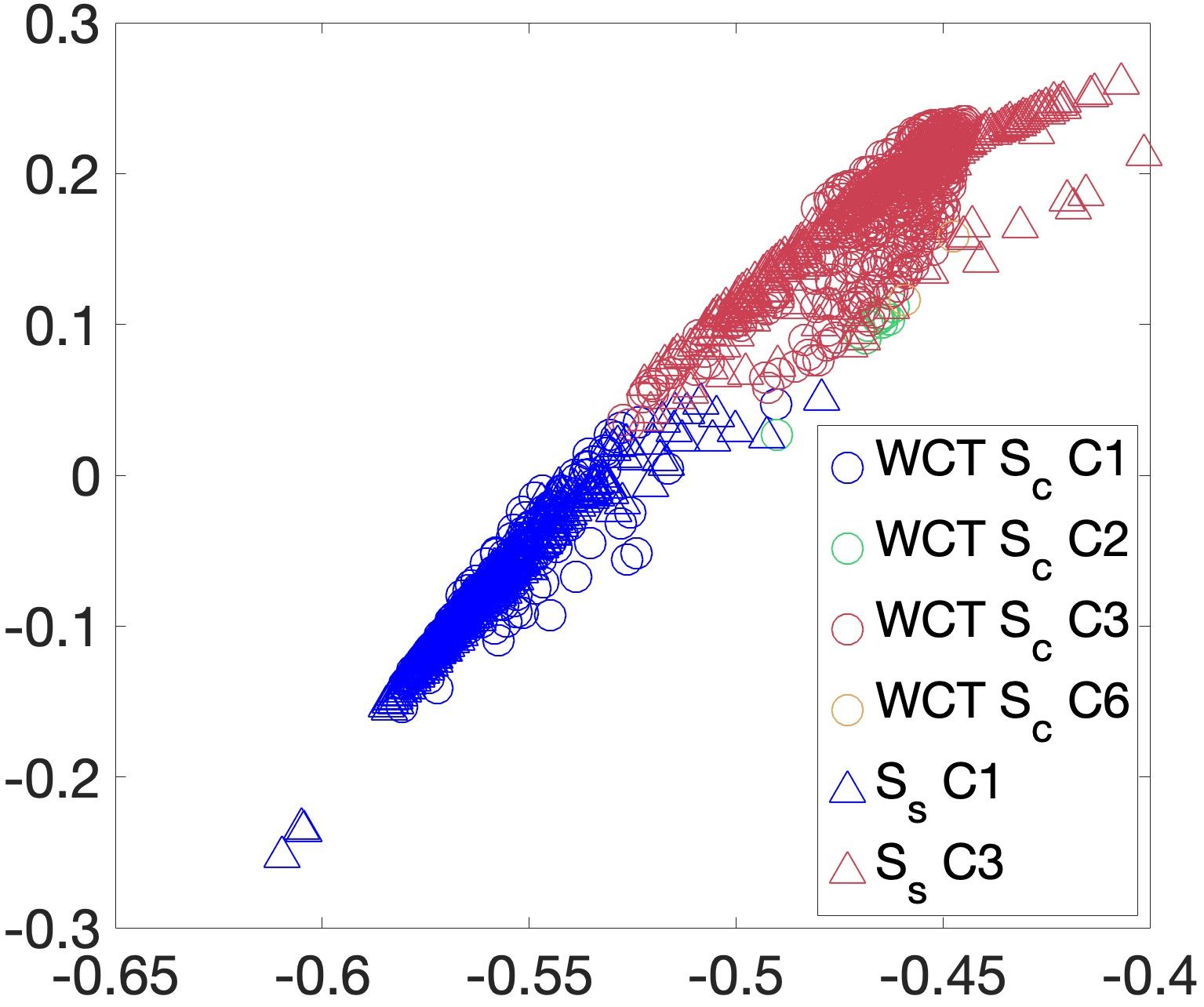}\label{fig:reason:dist_rp_wct}}\hspace{0.01mm}
		\subfloat[Distribution of $I_{cs}'$]{\includegraphics[width=0.19\linewidth]{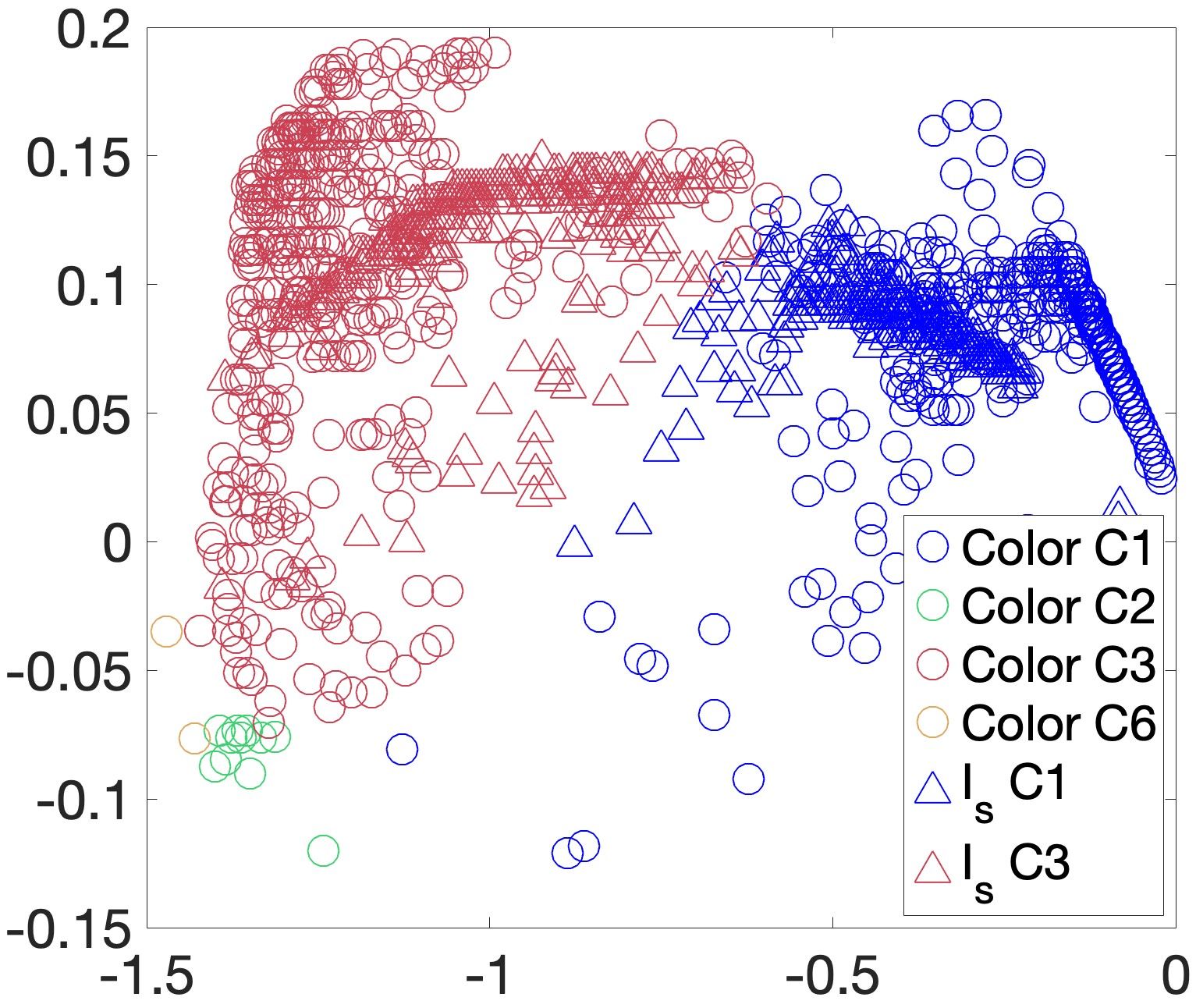}\label{fig:reason:dist_affine}}\hspace{0.01mm}
		\subfloat[Distribution of $I_{cs}$]{\includegraphics[width=0.19\linewidth]{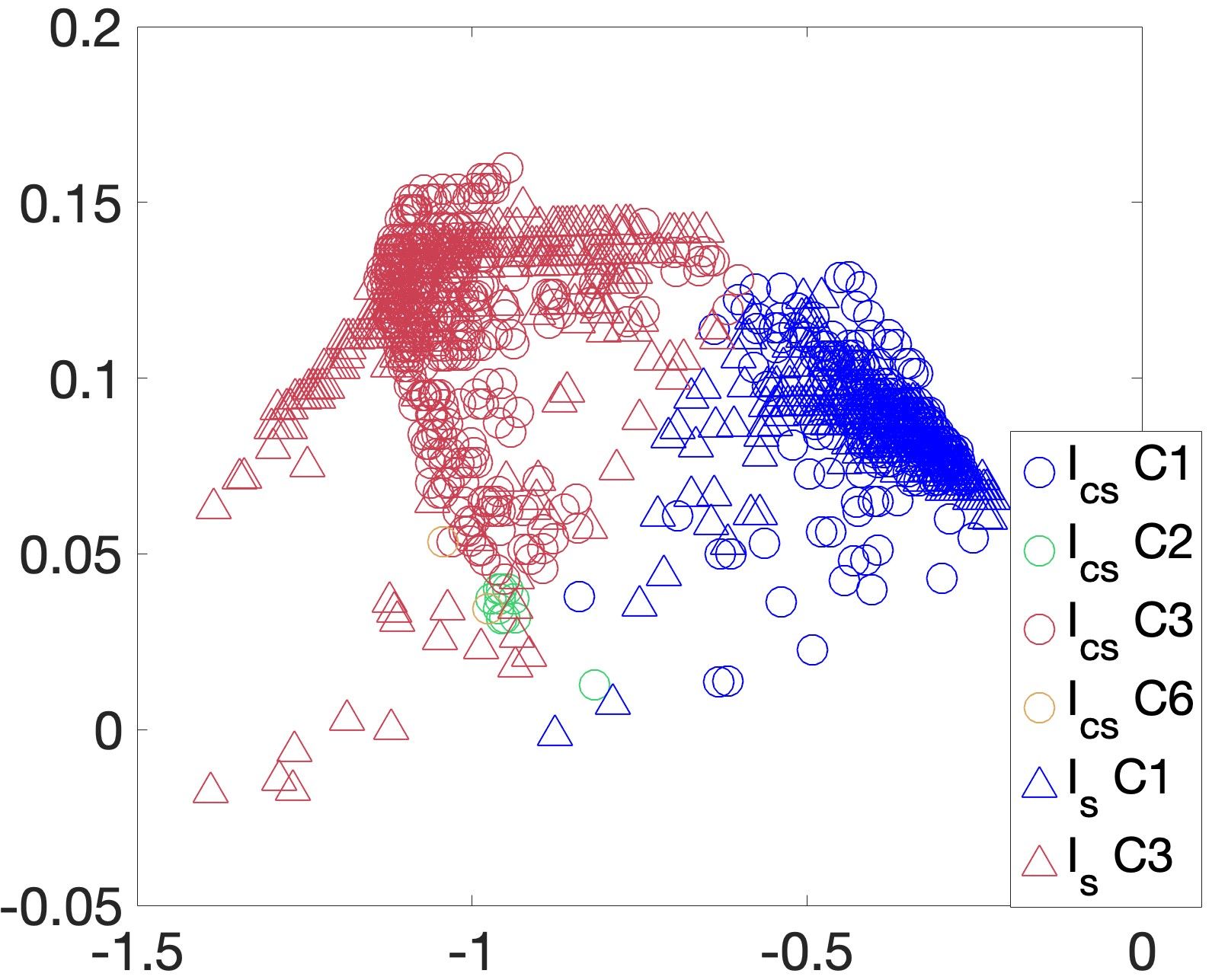}\label{fig:reason:dist_wct}}\hfill
	\end{minipage}\vspace{1mm}\\
	\caption{Reasoning of the proposed NL-MAT using MCIM and the distribution plots masked by MCIM. The raw images ($I_c$, $I_s$) and representations ($S_c$, $S_s$) are projected onto a two-dimensional space using SVD. Different colors indicate different context of the MCIM. C\# denotes the index of the color bases for the corresponding context. Circles and triangles denote the pixels or representations belonging to the content and style images, respectively. The distribution of the stylized image is similar to that of the style image with the affine-transfer, and it became closer with WCT. Note that for each MCIM masked region (or context), we vectorize pixels in that region into a column vector and pick every 1000th element of the vector for display purpose, so that the density change of the distribution is more easily observed.}
	\label{fig:reason}
\vspace{-5mm}
\end{figure*}

\textcolor{black}{Since the style transfer problem can be explained as mapping the distribution of the content image to that of the style image~\cite{pitie2005n}, the stylized result should have similar distribution to that of the style image. To further illustrate why the proposed scheme works, we draw the distributions of intermediate results as well as each context, by way of MCIM, to visualize the effect of stylization in Fig.~\ref{fig:reason}.} 
%We use a toy example to further illustrate, by way of MCIM, why the proposed scheme works with intermediate results as shown in Fig.~\ref{fig:reason}. \textcolor{black}{ To demonstrate the intermediate results more clearly, we draw the distribution of each context masked with MCIM in ~\crefrange{fig:reason:dist_raw}{fig:reason:dist_wct}.}

%Since the resolution of the image pair are quite large, for each MCIM masked region (or context), we vectorize pixels in that region into a column vector and pick every 1000th element of the vector for display purpose, so that the density change of the distribution is more easily observed. 

\textcolor{black}{The distributions are visualized by projecting the raw image ($I_c$, $I_s$) or representations ($S_c$, $S_s$) onto a two-dimensional space using the singular value decomposition (SVD) method. 
Fig.~\ref{fig:reason:dist_raw} shows the distributions of both the raw content and style images before stylization. The projected pixels are colored according to the contexts identified by MCIM, with circles indicating those from the content image and triangles for those from the style image.  For example, the blue and red circles denote the pixels from the mountain area ($I_c$ C1) and the sky area ($I_c$ C3) of the content image, respectively. We can observe from Fig.~\ref{fig:reason:dist_raw} that, the blue circles ($I_c$ C1) and blue triangles ($I_s$ C1), although indicating the same context (\ie, the mountain), belong to the two different blue clusters. This indicates that the distributions of the raw image pair are quite different from each other. Fig.~\ref{fig:reason:dist_rp} presents the distributions of the representations $S_c$ and $S_s$ for the content and style images, respectively. We can observe that the same context of the $S_c$ and $S_s$, \eg, the blue circles ($S_c$ C1) and blue triangles ($S_s$ C1), overlap into one blue cluster. This indicates that when we project the raw image onto the representation space with the stick-breaking encoder, affine-transfer decoder, sparse constraint  and mutual discriminative network, the representations of the same context in the content and style images reveal similar characteristics. This is in consistent with the previous analysis and the matching contexts in the MCIM.}

\textcolor{black}{By feeding the extracted representations $S_c$ to the affine-transfer decoder, we can achieve preliminary stylized result $I_{cs}'$ as shown in Fig.~\ref{fig:reason:output_affine}. From Fig.~\ref{fig:reason:dist_affine}, we can observe that the distribution of $I_{cs}'$ is closer to that of the style image, as compared to that of the content image. Nonetheless, there are still two apparent blue clusters in the distribution plot. To further match the representations, we conduct WCT on $S_c$ and show their distributions in Fig.~\ref{fig:reason:dist_rp_wct}. We observe that the same contexts of $S_c$ move closer to that of the $S_s$, presenting one dense blue cluster.
Therefore, with the WCT on $S_c$, the distribution of the generated result $I_{cs}$ (\eg, the blue circle cluster) is quite similar to that of the style image $I_s$ (\eg, the blue triangle cluster) as shown in Fig.~\ref{fig:reason:dist_wct}. As a result, the transferred image shown in Fig.~\ref{fig:reason:output_wct} carries the style of Fig.~\ref{fig:reason:style} better than $I_{cs}'$.}

%(\eg, the blue cluster formed by the blue circles)
%(\eg, the blue cluster formed by the blue triangles)
%phenomenal
\textcolor{black}{Due to the complexity of real applications, the context of the content and style images may not be matched perfectly. For example, from Fig.~\ref{fig:reason:content_seg}, the content MCIM shows a context area marked in green that does not have any matching areas in the style MCIM. Nonetheless, from Fig.~\ref{fig:reason:dist_raw}, we observe this green area denoted by green circles ($I_c$ C2) is closer to the red circles indicating the sky ($I_c$ C3).  From Fig.~\ref{fig:reason:dist_rp}, we see that the extracted representations of C2 (green circles $S_c$ C2) are also close to that of C3 in the content image (sky red circles). When WCT is performed on the representations of C2, the green cluster stay close to that of the C3 of the style image (red triangle, $I_s$ C3). Hence the C2 area of the stylized image looks still natural as shown in Fig.~\ref{fig:reason:output_wct}.}

%In the ideal case, the same context of the content and style image should have similar proportions with the affine-transfer color bases. However, there may exists differences in real scenario. 

%\subsubsection{\textcolor{black}{Decoupled Representations}}
%\textcolor{black}{As shown in the previous sections, the proposed MCIM is able to visualize the representations, thus could demonstrate the effectiveness of the stylization. }

%\input{affine.tex}
\begin{figure*}
\setlength{\abovecaptionskip}{1pt}
\setlength{\belowcaptionskip}{1pt}
	\begin{minipage}{1\linewidth}
		\centering
		\subfloat[Content]{\includegraphics[width=0.16\linewidth]{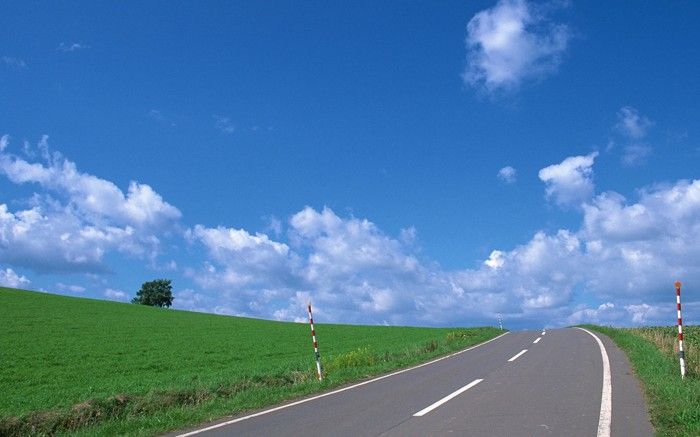}\label{fig:abligation_affine:content}}\hspace{0.01mm}
		\subfloat[Content MCIM with generic decoder]{\includegraphics[width=0.16\linewidth]{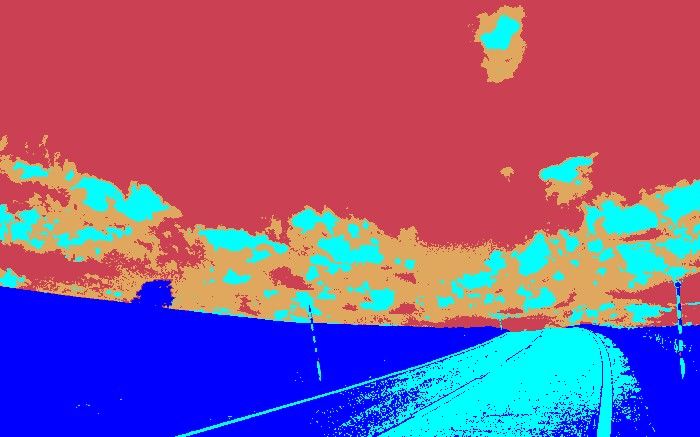}\label{fig:abligation_affine:content_seg_ge}}\hspace{0.01mm}
		\subfloat[Style MCIM with generic decoder]	{\includegraphics[width=0.15\linewidth]{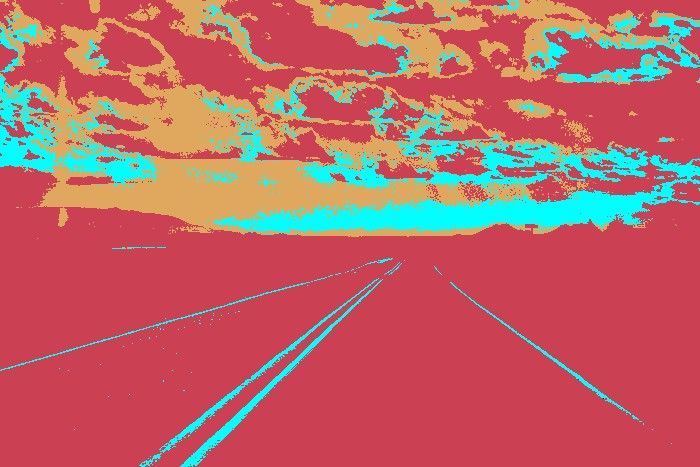}\label{fig:abligation_affine:style_seg_ge}}\hspace{0.01mm}
		\subfloat[Distribution of $S_c$ and $S_s$ with generic decoder]{\includegraphics[width=0.16\linewidth]{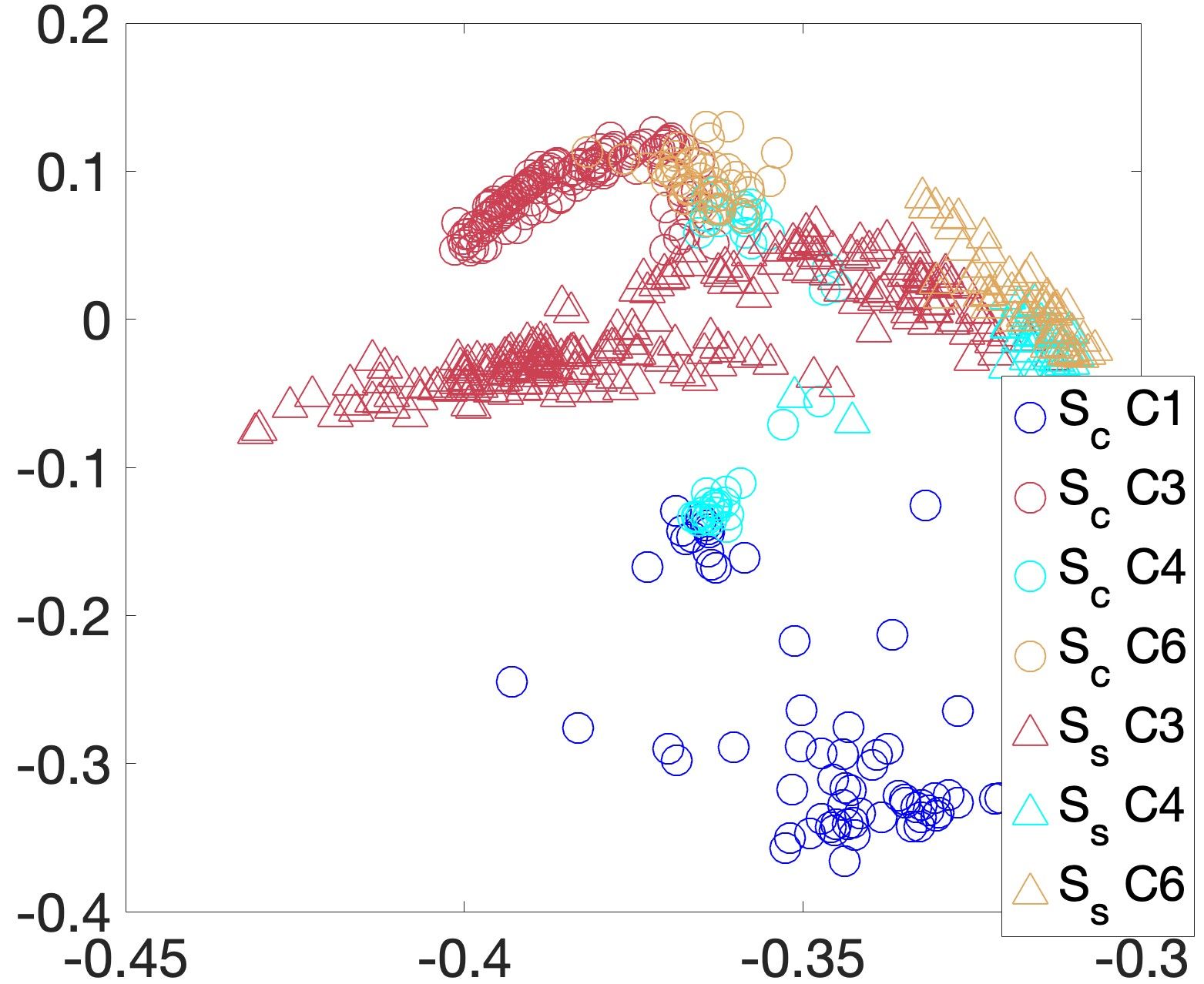}\label{fig:abligation_affine:dist_rp_ge}}\hspace{0.01mm}
		\subfloat[Distribution of WCT on $S_c$ with generic decoder]{\includegraphics[width=0.16\linewidth]{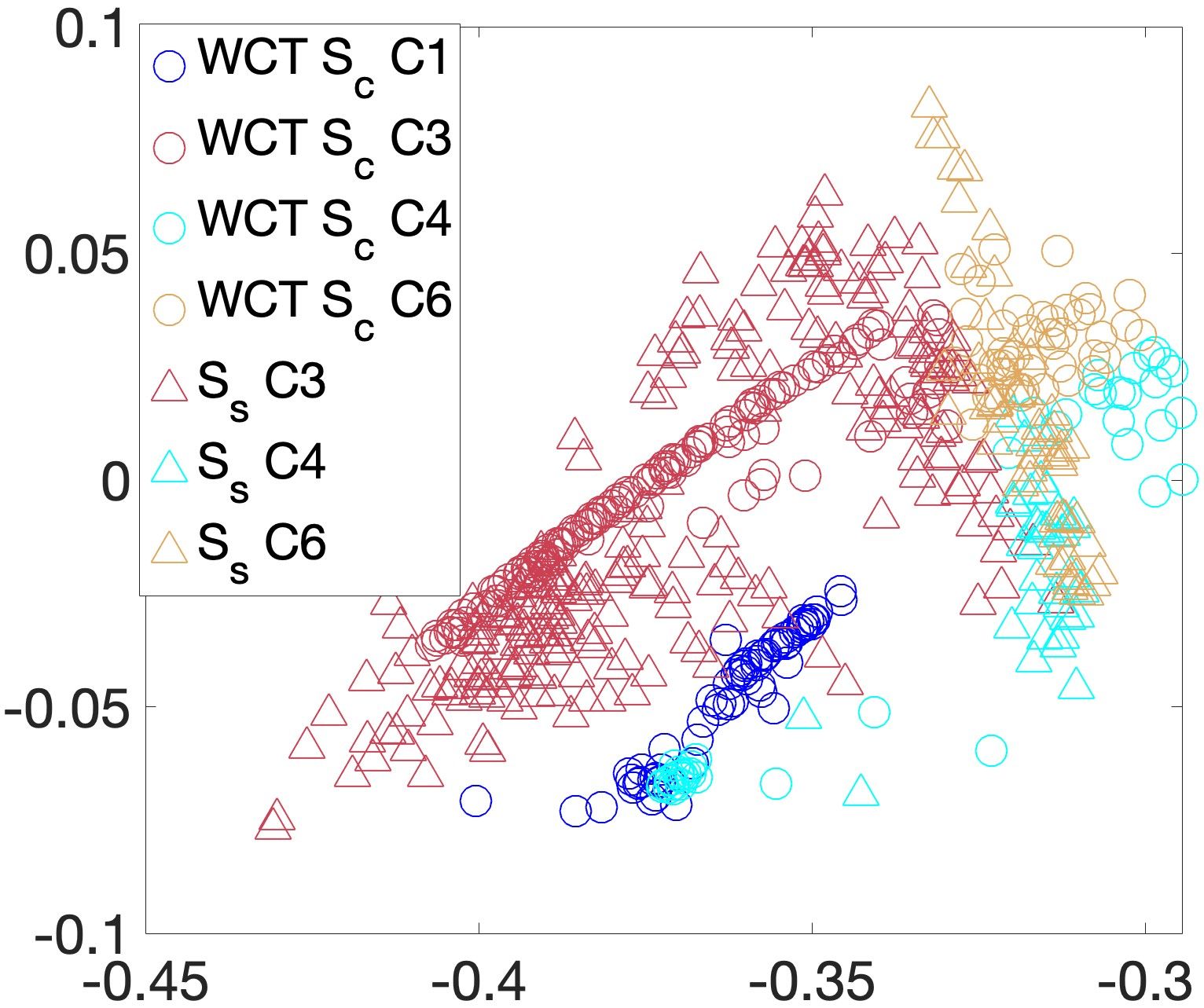}\label{fig:abligation_affine:dist_rp_tf_ge}}\hspace{0.01mm}
		\subfloat[Stylized image with generic decoder+WCT]{\includegraphics[width=0.16\linewidth]{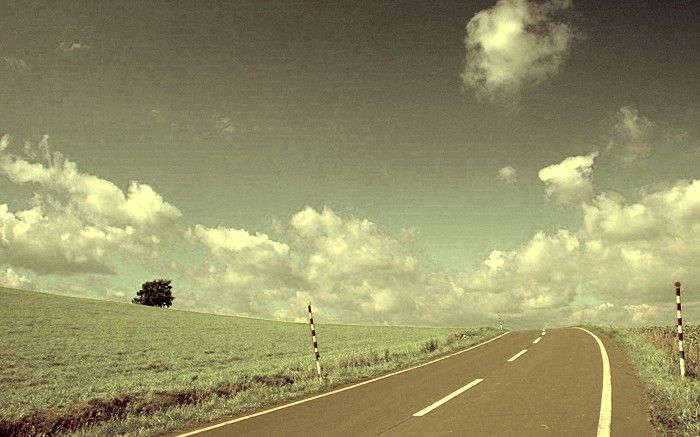}\label{fig:abligation_affine:out_ge}}\\
		\subfloat[Style]{\includegraphics[width=0.16\linewidth]{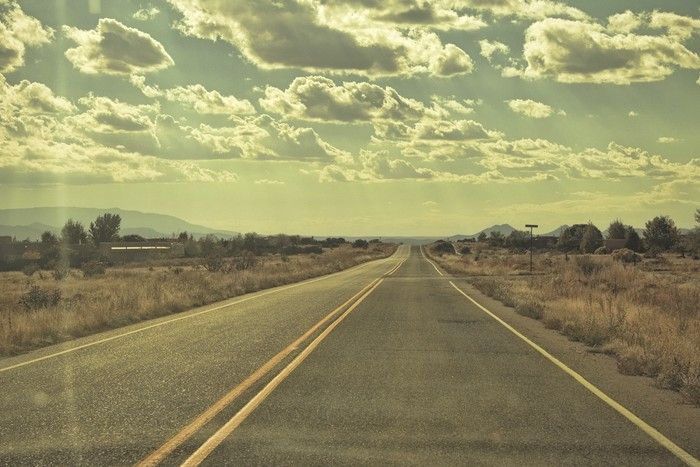}\label{fig:abligation_affine:style}}\hspace{0.01mm}
		\subfloat[Content MCIM with affine-transfer decoder]{\includegraphics[width=0.16\linewidth]{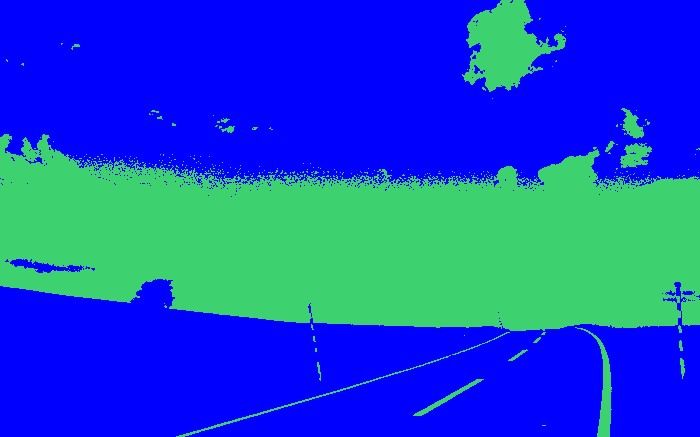}\label{fig:abligation_affine:content_seg}}\hspace{0.01mm}
		\subfloat[Style MCIM with affine-transfer decoder]	{\includegraphics[width=0.15\linewidth]{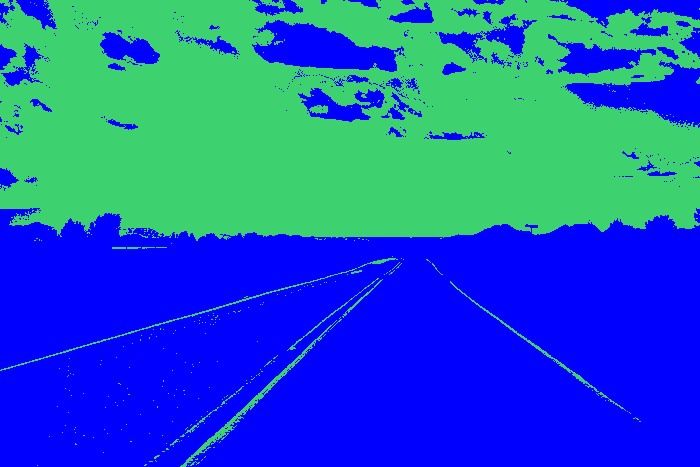}\label{fig:abligation_affine:style_seg}}\hspace{0.01mm}
		\subfloat[Distribution of $S_c$ and $S_s$ with affine-transfer decoder]{\includegraphics[width=0.16\linewidth]{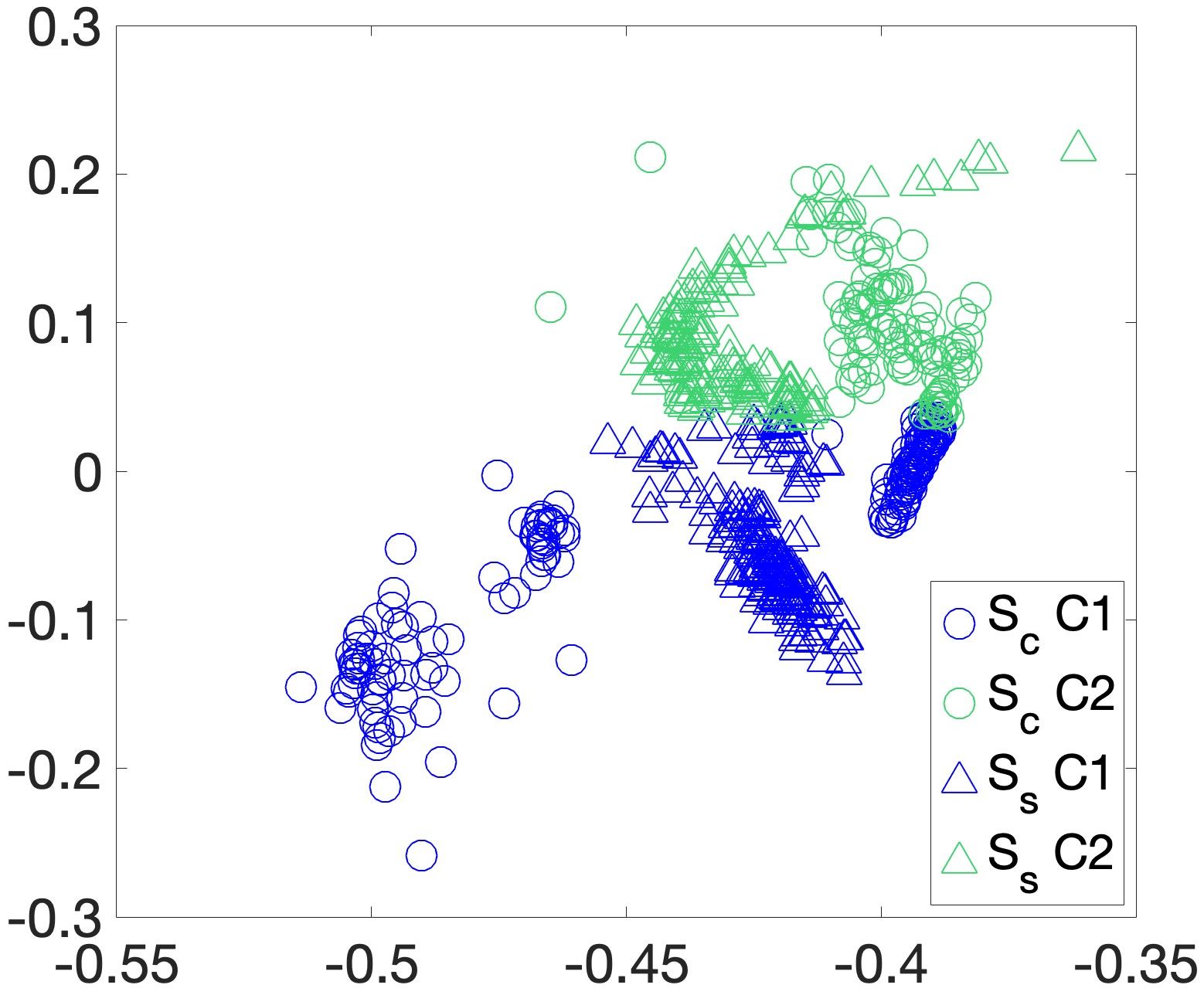}\label{fig:abligation_affine:dist_rp}}\hspace{0.01mm}
		\subfloat[Distribution of WCT on $S_c$ with affine-transfer decoder]{\includegraphics[width=0.16\linewidth]{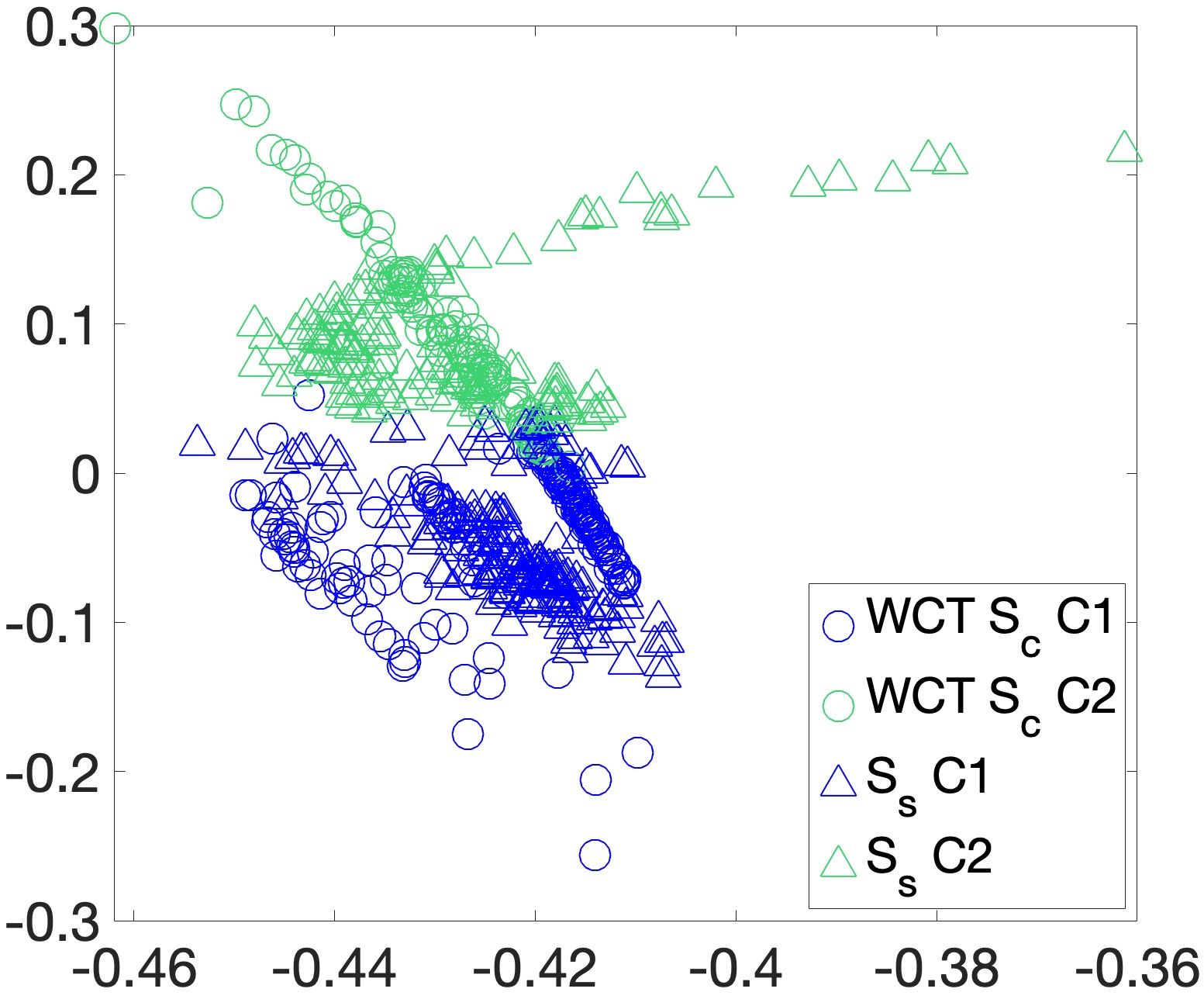}\label{fig:abligation_affine:dist_rp_tf}}\hspace{0.01mm}
		\subfloat[Stylized image with affine-transfer decoder+WCT]{\includegraphics[width=0.16\linewidth]{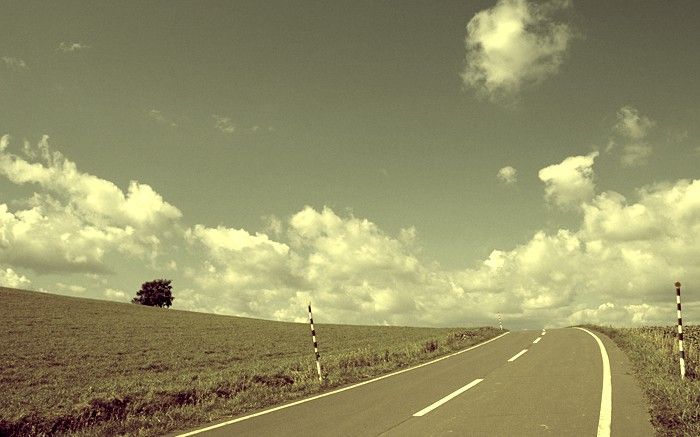}\label{fig:abligation_affine:out_ours}}\hfill\\
	\end{minipage}\vspace{1mm}\\
	\caption{The effect of affine-transfer decoder.}
	\label{fig:abligation_affine}
\vspace{-5mm}
\end{figure*}

\subsection{Ablation Study on Affine-Transfer Decoder}
\label{sec:evaluate_affine}
One of the contributing factors to the effectiveness of the non-local representation scheme is the proposed affine-transfer decoder, which allows the network to learn the color bases of both the content and style images as well as the transfer between them. This decoder, along with the mutual-discriminative network and the sparse constraint, allows the network to match representations of similar contexts regardless of their actual  color content.  

To demonstrate the importance of the affine-transfer decoder, we replace it with a generic fully-connected decoder and show the results in Fig.~\ref{fig:abligation_affine}. We observe that, \textcolor{black}{with a shared generic decoder for both the content and style images}, the network could not match the representations of the \textcolor{black}{image pair} well even with %the mutual-discriminative network and the sparse 
\textcolor{black}{the enforced constraints}. For example, we can observe from Fig.~\ref{fig:abligation_affine:dist_rp_ge} that the \textcolor{black}{extracted representations of the ``grass'' from the content image, blue circles ($S_c$ C1), are far away from that of the same context from the style image, red triangles ($S_s$ C3).}
%the blue circles ($S_c$ C1) denoting the extracted representations of the grass from the content image are far away from the red triangles denoting the \textcolor{black}{representations} of the same context, \ie, grass from the  style image ($S_s$ C3). 
This scattered distribution indicates the extracted representations are not matched well. On the contrary, \textcolor{black}{as shown in Fig.~\ref{fig:abligation_affine:dist_rp}, with the affine transfer decoder, %the network is able to extract matched representations from the content and style images better as shown in %Figs.~\ref{fig:abligation_affine:content_seg},  \ref{fig:abligation_affine:style_seg}, and ~\ref{fig:abligation_affine:dist_rp}. 
 %Figs.~\ref{fig:abligation_affine:content_seg}-\ref{fig:abligation_affine:dist_rp}. Specifically, 
 the representations of the ``grass'' from the content image (blue circles $S_c$ C1) and the style image (blue triangles $S_s$ C1)} are much closer to each other as compared to the distribution in Fig.~\ref{fig:abligation_affine:dist_rp_ge}, showing a better match of similar contexts between the content and style images. Thus, when WCT is applied, the distributions of the representations from the affine-transfer decoder are closer as compared to those from the generic decoder, as shown in Figs.~\ref{fig:abligation_affine:dist_rp_tf} and~\ref{fig:abligation_affine:dist_rp_tf_ge}, respectively. As a result, the proposed method is able to obtain more photorealistic image carrying more natural styles as compared to the one with the generic decoder.

% Specifically, in Fig.~\ref{fig:abligation_affine:dist_rp}, the blue circles ($S_c$ C1) and blue triangles ($S_s$ C1) denoting the \textcolor{black}{representations} of ``grass'' from the content and style images, respectively, are much closer to each other as compared to the distribution in Fig.~\ref{fig:abligation_affine:dist_rp_ge}, showing a better match of similar contexts between the content and style images. Thus, when WCT is applied, the distributions of the representations from the affine-transfer decoder are closer as compared to those from the generic decoder, as shown in Figs.~\ref{fig:abligation_affine:dist_rp_tf} and~\ref{fig:abligation_affine:dist_rp_tf_ge}, respectively. As a result, the proposed method is able to obtain more photorealistic image carrying more natural styles as compared to the one with the generic decoder.}

\subsection{Ablation Study on Context-sensitive Local Color Transfer}
\label{sec:ablation2}
\begin{figure*}
\setlength{\abovecaptionskip}{1pt}
\setlength{\belowcaptionskip}{1pt}
\centering
	\begin{minipage}{0.9\linewidth}
	\centering
		\subfloat[Content and style images]{\begin{minipage}{0.18\linewidth}
			\centering
			{\includegraphics[width=1\linewidth]{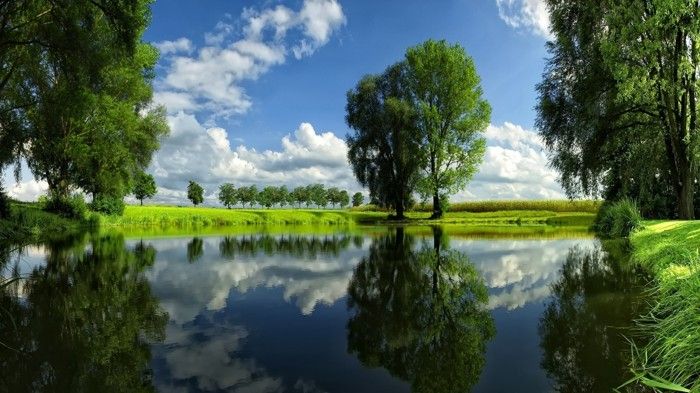}}\vspace{1mm}
			{\includegraphics[width=1\linewidth]{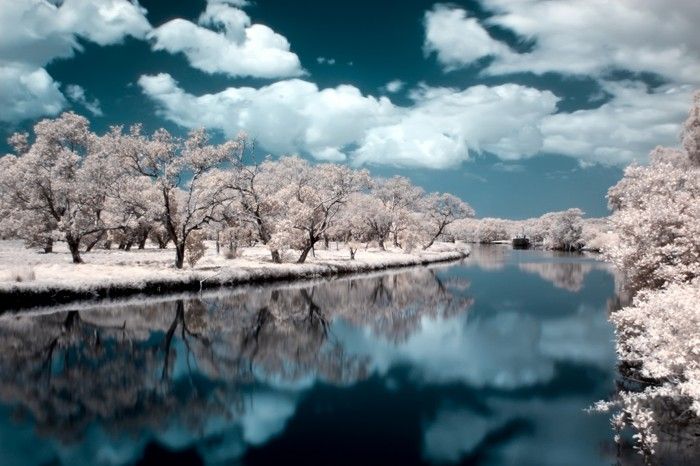}}\vspace{1mm}
			{\includegraphics[width=1\linewidth]{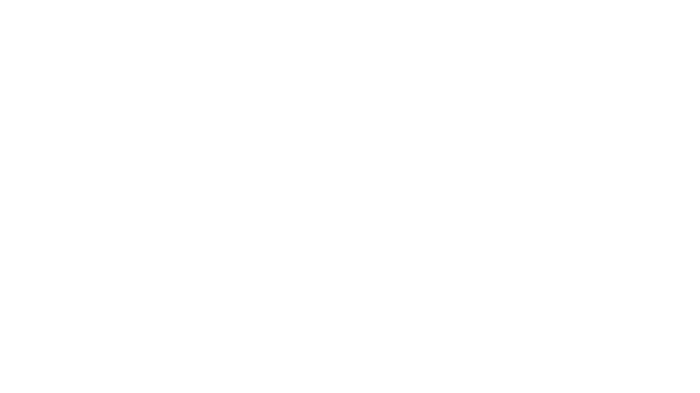}}
		\end{minipage}\label{fig:para:a}}
		\subfloat[$\alpha=0$, $\lambda=0$]{\begin{minipage}{0.18\linewidth}
			\centering
			{\includegraphics[width=1\linewidth]{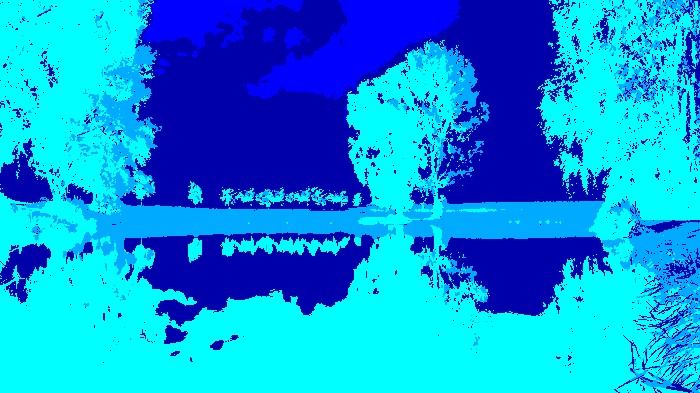}}\vspace{1mm}
			{\includegraphics[width=1\linewidth]{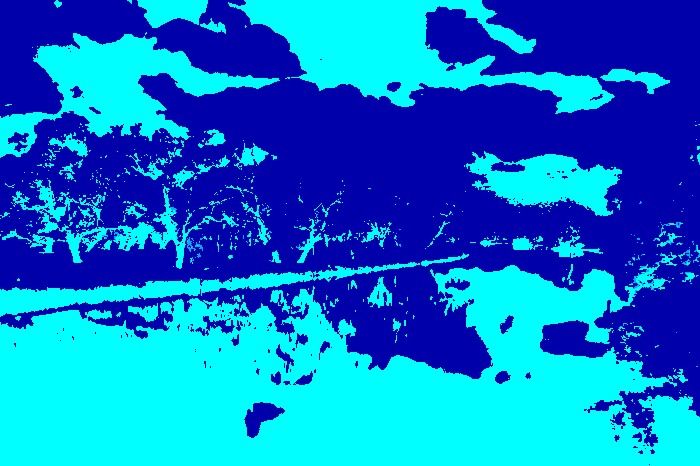}}\vspace{1mm}
			{\includegraphics[width=1\linewidth]{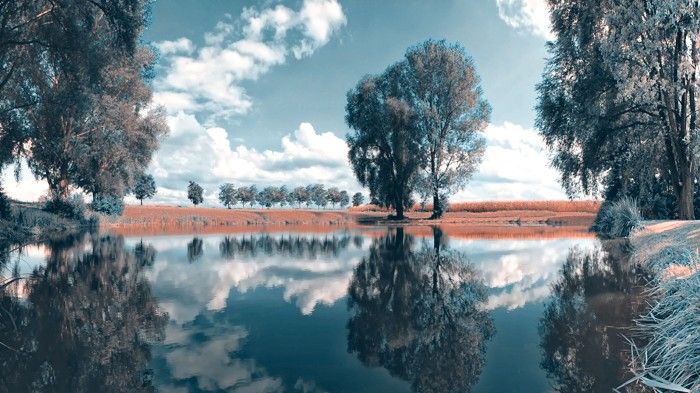}}
		\end{minipage}\label{fig:para:b}}
		\subfloat[$\alpha=0.0001$, $\lambda=0$]{\begin{minipage}{0.18\linewidth}
			\centering
			{\includegraphics[width=1\linewidth]{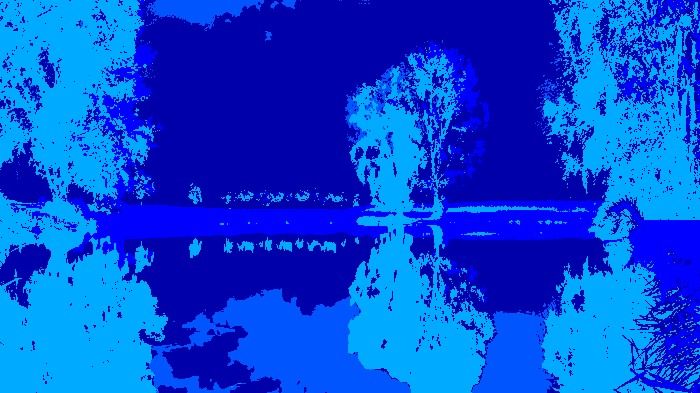}}\vspace{1mm}
			{\includegraphics[width=1\linewidth]{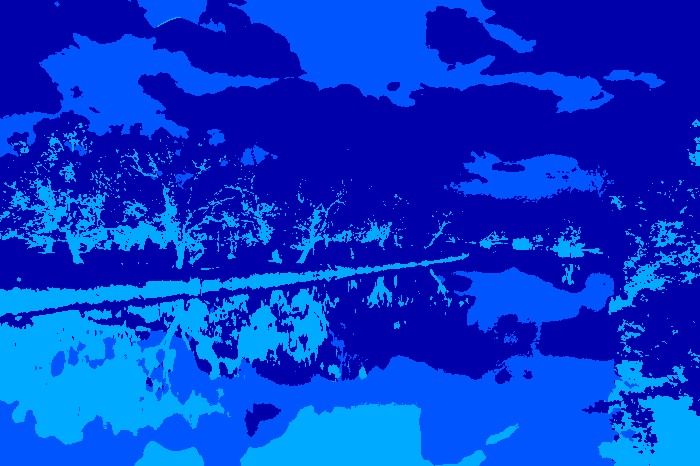}}\vspace{1mm}
			{\includegraphics[width=1\linewidth]{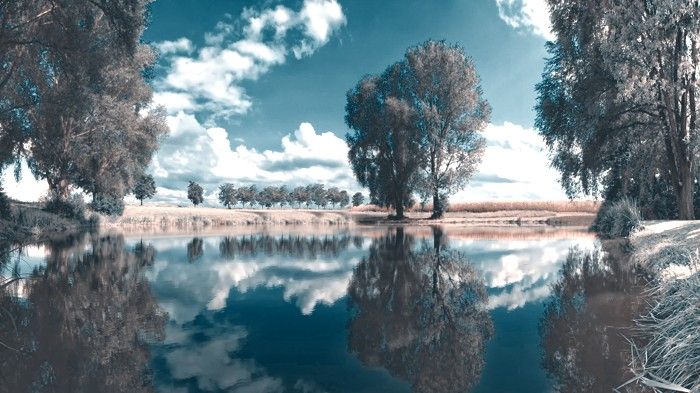}}
		\end{minipage}\label{fig:para:c}}
		\subfloat[$\alpha=0.01$, $\lambda=0$]{\begin{minipage}{0.18\linewidth}
			\centering
			{\includegraphics[width=1\linewidth]{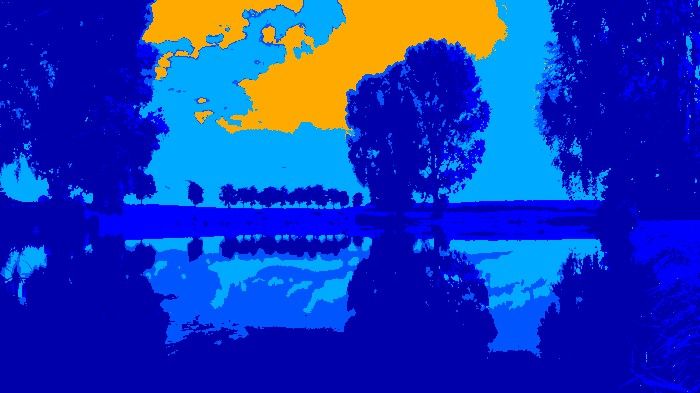}}\vspace{1mm}
			{\includegraphics[width=1\linewidth]{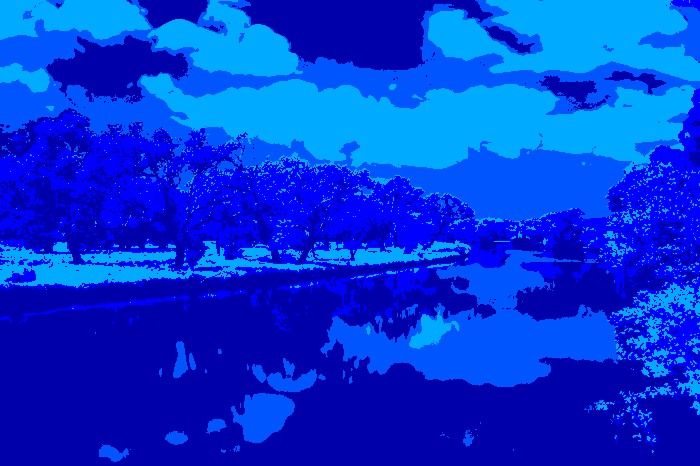}}\vspace{1mm}
			{\includegraphics[width=1\linewidth]{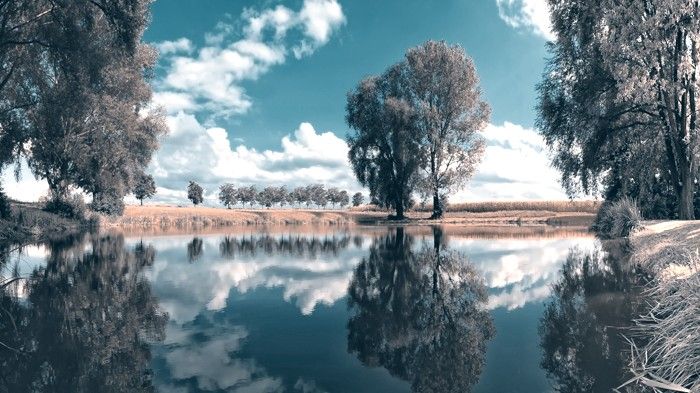}}
		\end{minipage}\label{fig:para:d}}
		\subfloat[$\alpha=0.1$, $\lambda=0$]{\begin{minipage}{0.18\linewidth}
			\centering
			{\includegraphics[width=1\linewidth]{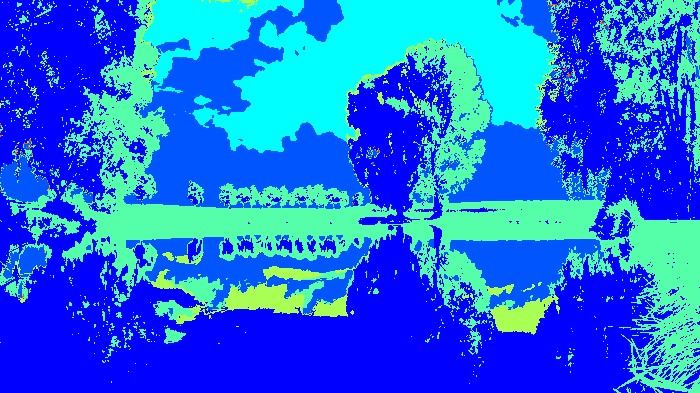}}\vspace{1mm}
			{\includegraphics[width=1\linewidth]{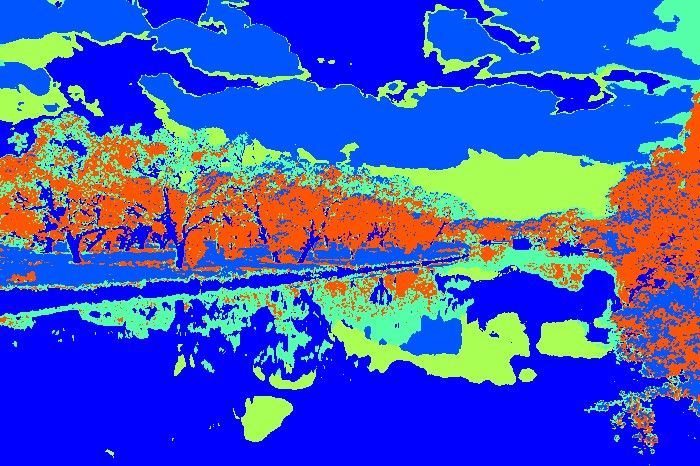}}\vspace{1mm}
			{\includegraphics[width=1\linewidth]{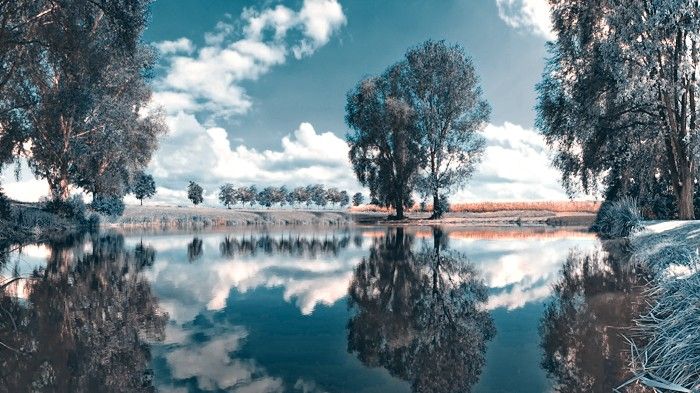}}
		\end{minipage}\label{fig:para:e}}
	\end{minipage}
	\begin{minipage}{0.9\linewidth}
	\centering
		\subfloat[Content and style images]{\begin{minipage}{0.18\linewidth}
			\centering
			{\includegraphics[width=1\linewidth]{fig/ablation_ms/50_0in}}\vspace{1mm}
			{\includegraphics[width=1\linewidth]{fig/ablation_ms/50_1tar}}\vspace{1mm}
			{\includegraphics[width=1\linewidth]{fig/ablation_ms/inbox}\label{fig:para:f}}
		\end{minipage}}
		\subfloat[$\alpha=0.01$, $\lambda=0.0001$]{\begin{minipage}{0.18\linewidth}
			\centering
			{\includegraphics[width=1\linewidth]{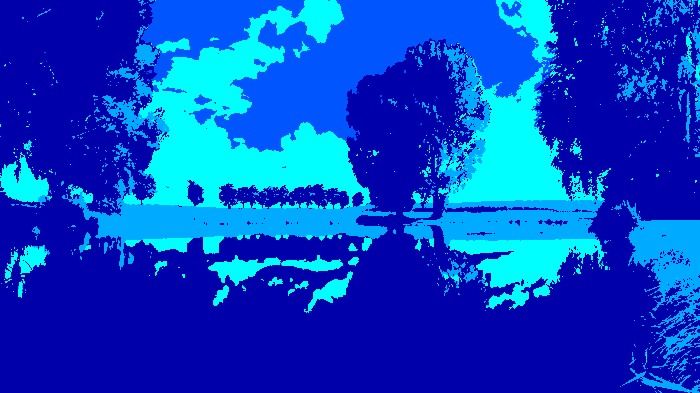}}\vspace{1mm}
			{\includegraphics[width=1\linewidth]{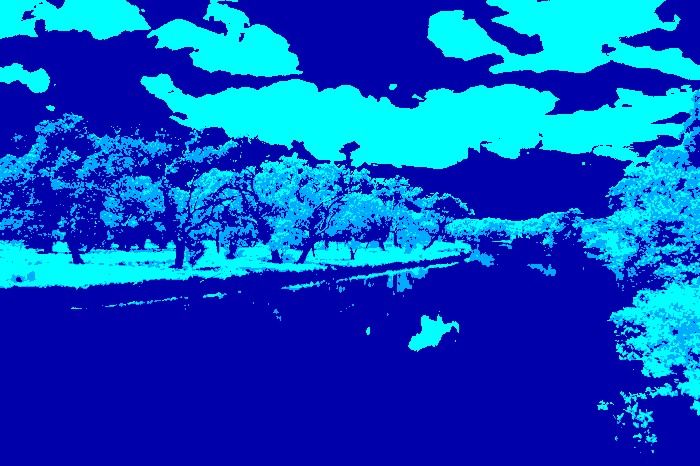}}\vspace{1mm}
			{\includegraphics[width=1\linewidth]{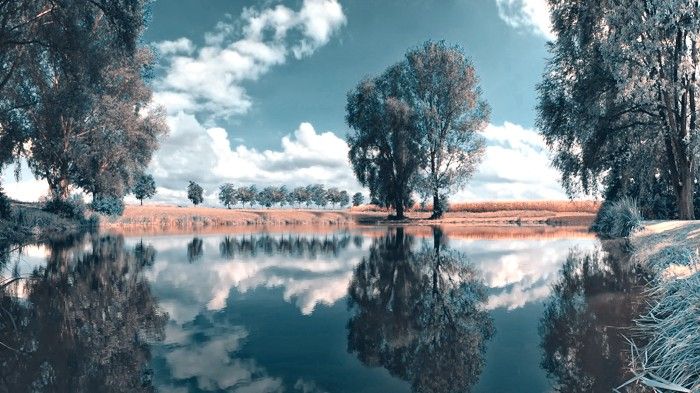}}
		\end{minipage}\label{fig:para:g}}
		\subfloat[$\alpha=0.01$, $\lambda=0.01$]{\begin{minipage}{0.18\linewidth}
			\centering
			{\includegraphics[width=1\linewidth]{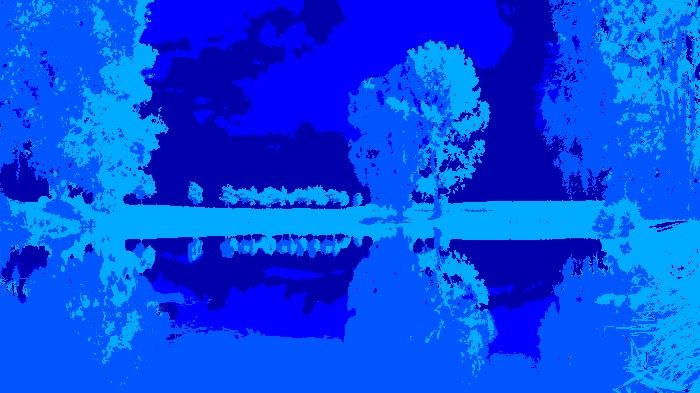}}\vspace{1mm}
			{\includegraphics[width=1\linewidth]{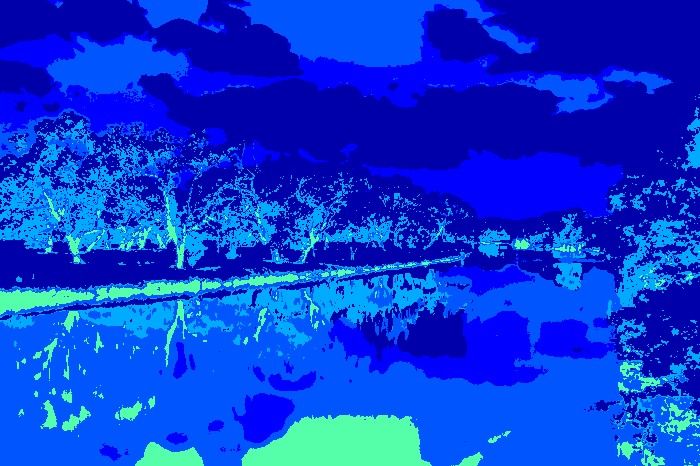}}\vspace{1mm}
			{\includegraphics[width=1\linewidth]{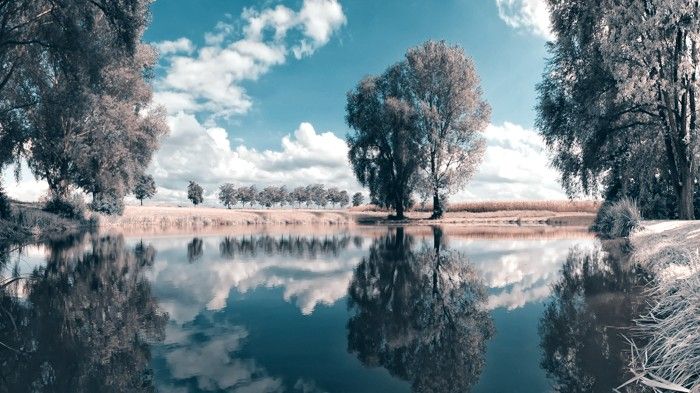}}
		\end{minipage}\label{fig:para:h}}
		\subfloat[$\alpha=0.01$, $\lambda=1$]{\begin{minipage}{0.18\linewidth}
			\centering
			{\includegraphics[width=1\linewidth]{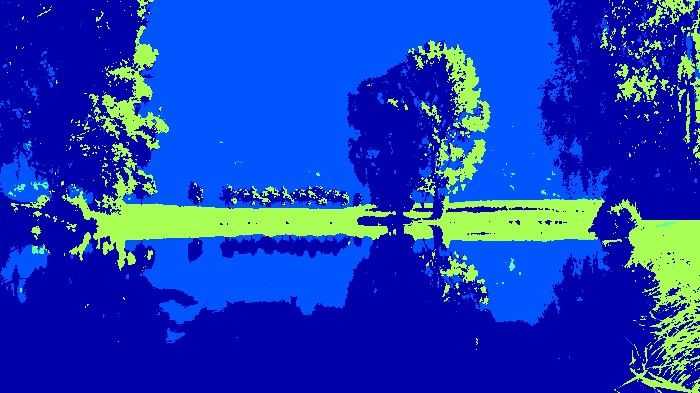}}\vspace{1mm}
			{\includegraphics[width=1\linewidth]{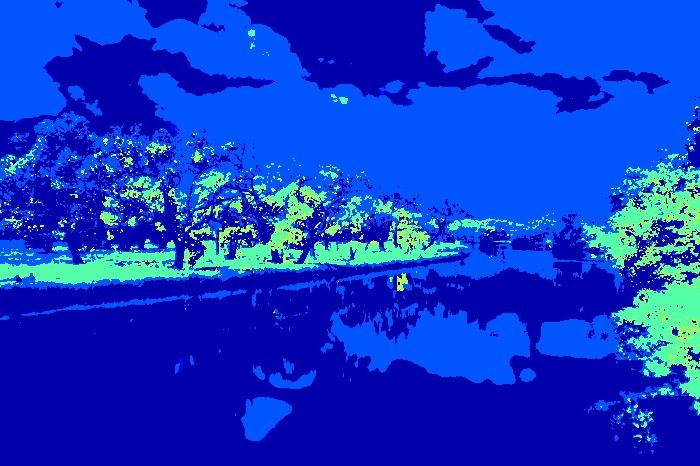}}\vspace{1mm}
			{\includegraphics[width=1\linewidth]{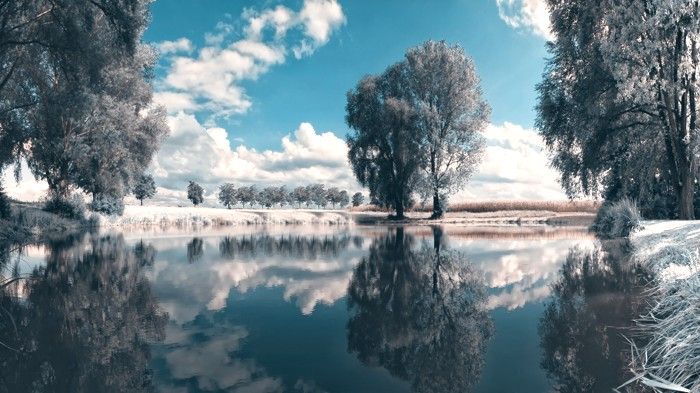}}
		\end{minipage}\label{fig:para:i}}
		\subfloat[$\alpha=0.01$, $\lambda=5$]{\begin{minipage}{0.18\linewidth}
			\centering
			{\includegraphics[width=1\linewidth]{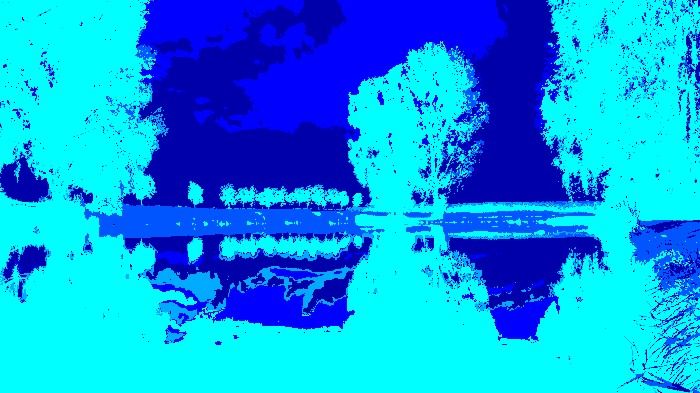}}\vspace{1mm}
			{\includegraphics[width=1\linewidth]{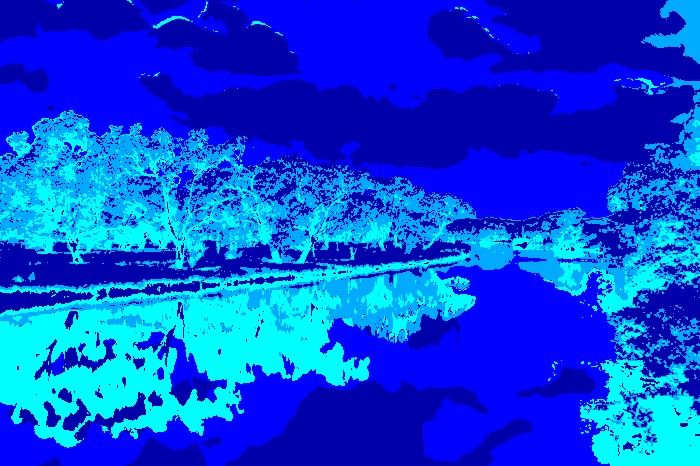}}\vspace{1mm}
			{\includegraphics[width=1\linewidth]{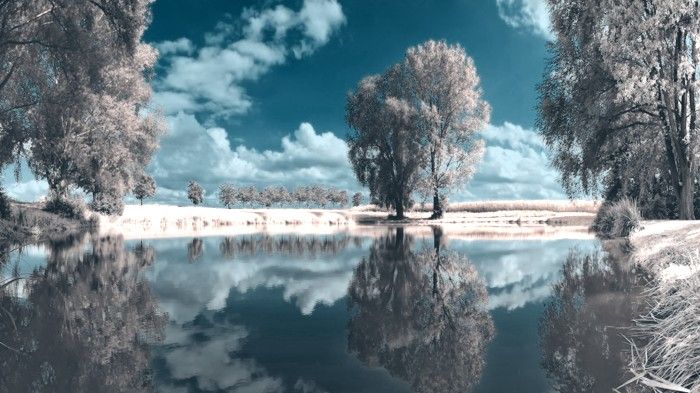}\label{fig:para:j}}
		\end{minipage}}
	\end{minipage}
	\caption{Effects of including the sparsity constraint and the mutual discriminative network by adjusting the parameters $\alpha$  and  $\lambda$, respectively. (b)-(e): stylized images with $\lambda = 0$ and $\alpha$ changed from $\{0, 0.0001, 0.01, 0.1\}$, respectively. 	Top: MCIM of the content image. Middle: MCIM of the style image. Bottom: stylized images. (g)-(j): stylized images with $\alpha = 0.01$ and $\lambda$ changed from $\{0.0001,0.01,1,5\}$. Note mainly the color changes of the tree, sky, and grass in the stylized image as the parameters are adjusted.} %Top  (from left to right): content image, stylized images with $\lambda = 0$ and $\alpha$ changed from $\{0, 0.0001, 0.01, 1\}$, respectively. Bottom (from left to right): style image, stylized images with $\alpha = 1$ and $\lambda$ changed from $\{0.0001,0.001,0.01,1\}$.}
	\label{fig:para}
\vspace{-5mm}
\end{figure*}

In addition to the affine-transfer decoder evaluated in Sec.~\ref{sec:evaluate_affine}, the two other important components of the proposed method are the sparse constraint, used to increase the discriminative capacity of local-color transfer, and the mutual discriminative network, used to enforce the representations of the content and style images to have context correspondence. % With these two factors, the network is able to extract discriminative representations that correspond to each other’s semantic context. %, as shown in Fig.~\ref{fig:context} and~\ref{fig:seg}. 
To evaluate the contribution of these two components, we perform ablation study by changing the weight parameters, $\alpha$ on sparsity constraint and $\lambda$ on mutual information as in Eq.~\ref{equ:optall}. The evaluation results are demonstrated in Fig.~\ref{fig:para}. %\textcolor{red}{Note that, to perform visualize inspection on how the sparsity and correspondence affect the stylized results, we also show the major color index map (MCIM) of each result.}

We can observe that, when both $\alpha$ and $\lambda$ are zero, \ie, no sparsity or mutual information loss is considered, the proposed method could still generate reasonably good stylized images with spatial structure well preserved, showing the effectiveness of the non-local  representation. However, the color was not adequately transferred, as can be seen from the color of the tree and the grass in Fig.~\ref{fig:para:b}. %Because the proposed structure allows more flexible transformation. 
When we gradually increase the sparse parameter $\alpha$, the discriminative capacity is increased as can be observed from the MCIMs (Figs.~\ref{fig:para:c}-\ref{fig:para:e}) with more subtle segments. Also with such sparsity, we observe local colors start to change drastically but with global consistency. However, since the network does not encourage the correspondence between the representations of the content and style images as $\lambda=0$, the color may not be transferred adequately when the representations do not match. When we increase the parameter $\lambda$ to include the mutual information loss, the extracted representations are more correlated with each other, resulting in a more photorealistic local style transfer, as shown in Figs.~\ref{fig:para:g}-\ref{fig:para:j}. We observe that the color of the trees starts showing the snowy effect and the color of grass is mostly white. The color transfer even affects the reflection of the trees in the water, as shown in Figs.~\ref{fig:para:i}-\ref{fig:para:j}. %\textcolor{red}{See mainly the transferred colors of the tree, the sky, and the grass.}

\subsection{\textcolor{black}{How to Choose the Number of Color Bases}}
\label{sec:numofbs}

\begin{figure}[htbp]
\setlength{\abovecaptionskip}{1pt}
\setlength{\belowcaptionskip}{1pt}
	\begin{minipage}{0.9\linewidth}
		\centering
		\subfloat[Content]{\begin{minipage}{0.3\linewidth}
				\centering
				{\includegraphics[width=1\linewidth]{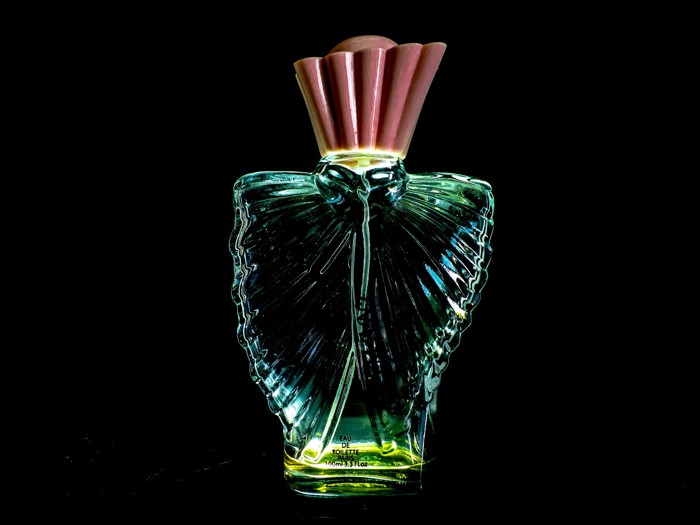}}
			\end{minipage}\label{fig:numofbs:in16}}\hspace{0.01mm}	
		\subfloat[Style]{\begin{minipage}{0.3\linewidth}
				\centering
				\vspace{2.25mm}
				{\includegraphics[width=1\linewidth]{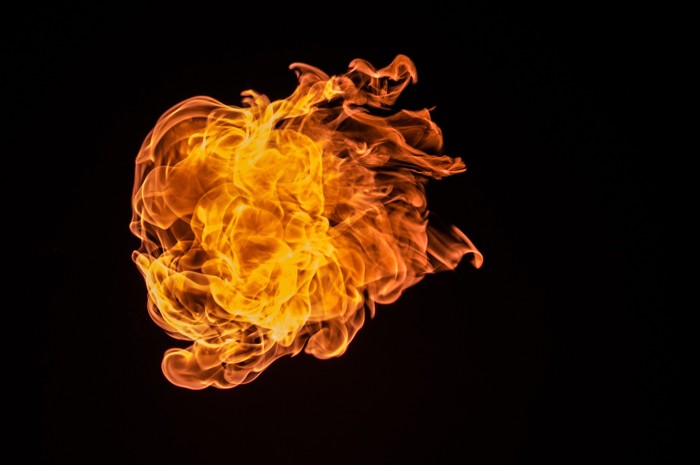}}
			\end{minipage}\label{fig:numofbs:tar16}}
		\subfloat[Global-based~\cite{reinhard2001color}]{\begin{minipage}{0.3\linewidth}
				\centering
				{\includegraphics[width=1\linewidth]{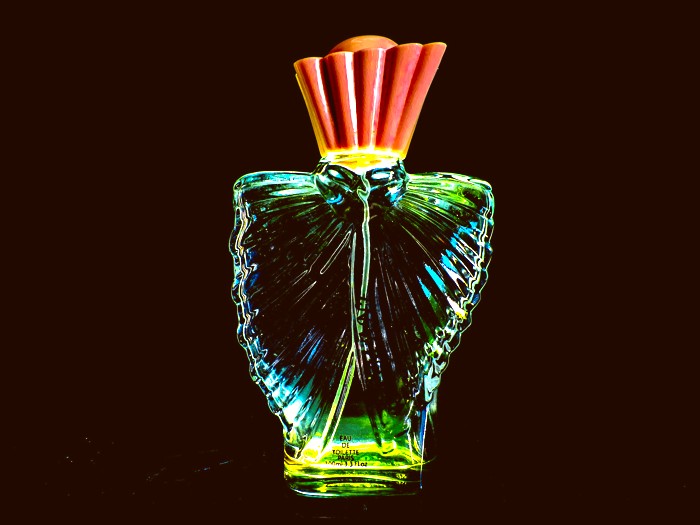}}
			\end{minipage}\label{fig:numofbs:global}}\\
		\subfloat[%The MCIMs of the content (left) and style (middle) image, and the resulting stylized image (right) when 
		$k=5$]	{\begin{minipage}{1\linewidth}
				\centering
				{\includegraphics[width=0.3\linewidth]{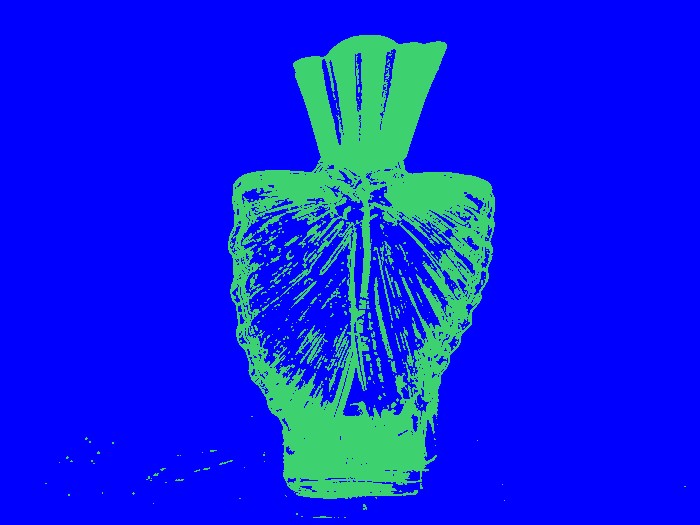}}
				{\includegraphics[width=0.3\linewidth]{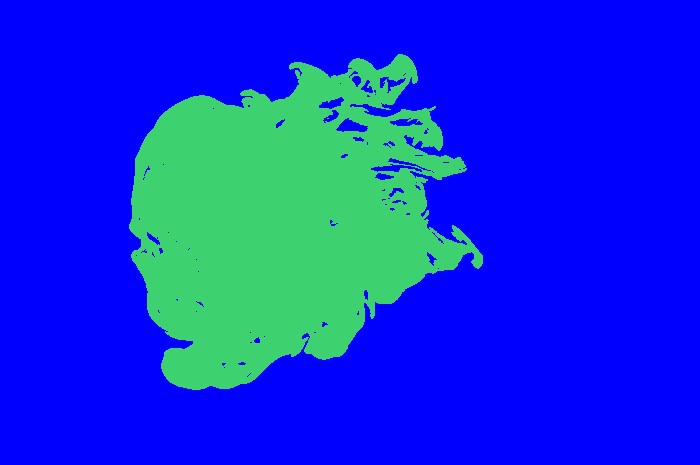}}
				{\includegraphics[width=0.3\linewidth]{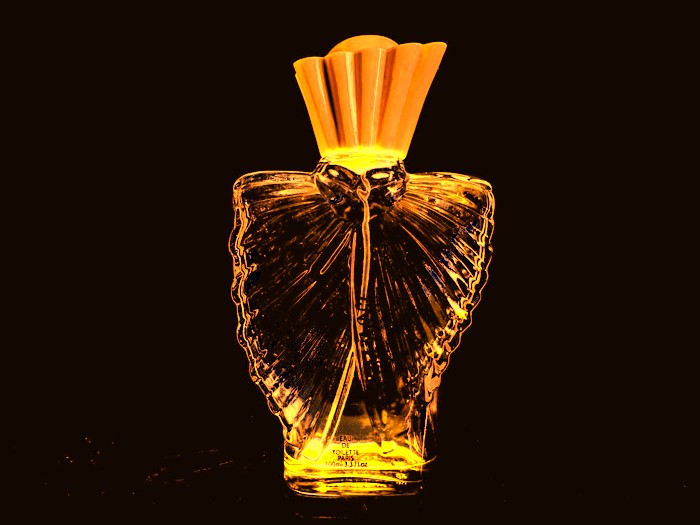}}
			\end{minipage}\label{fig:numofbs:5}}\\
		\subfloat[%The MCIMs of the content, style and stylized images when 
		$k=10$]	{\begin{minipage}{1\linewidth}
				\centering
				{\includegraphics[width=0.3\linewidth]{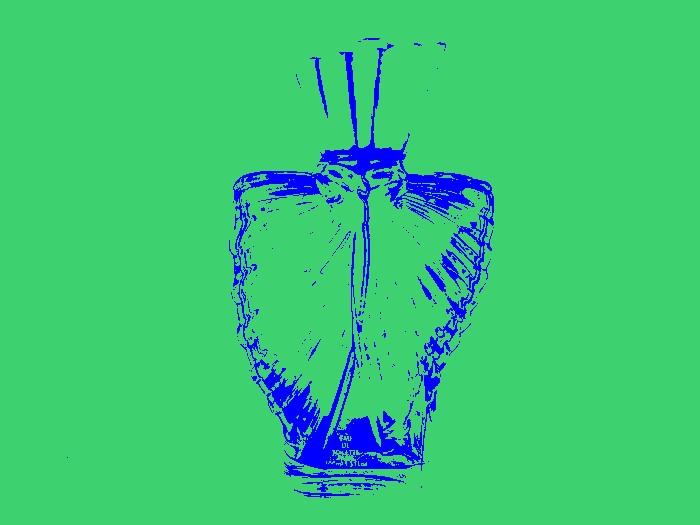}}
				{\includegraphics[width=0.3\linewidth]{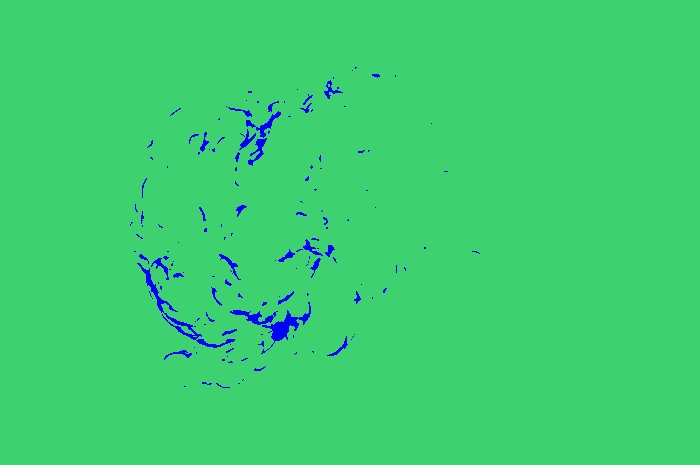}}
				{\includegraphics[width=0.3\linewidth]{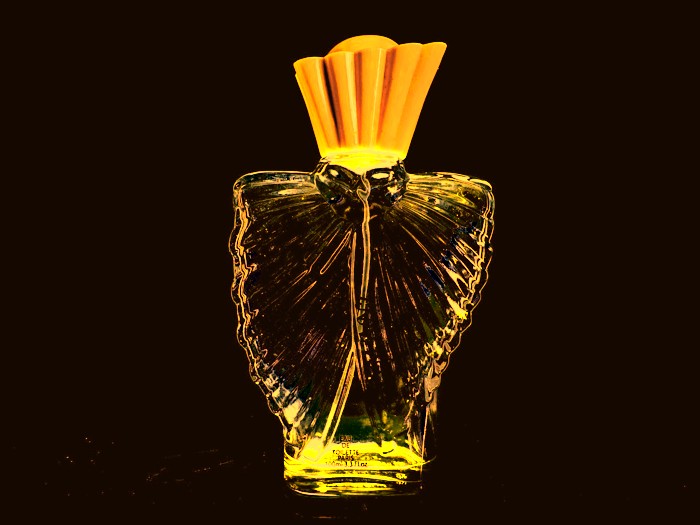}}
			\end{minipage}\label{fig:numofbs:10}}\\
		\subfloat[%The MCIMs of the content, style and stylized images when 
		$k=15$]	{\begin{minipage}{1\linewidth}
				\centering
				{\includegraphics[width=0.3\linewidth]{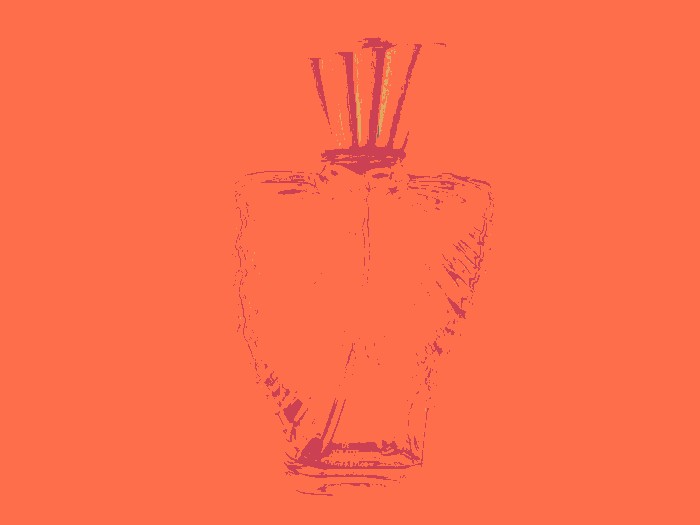}}
				{\includegraphics[width=0.3\linewidth]{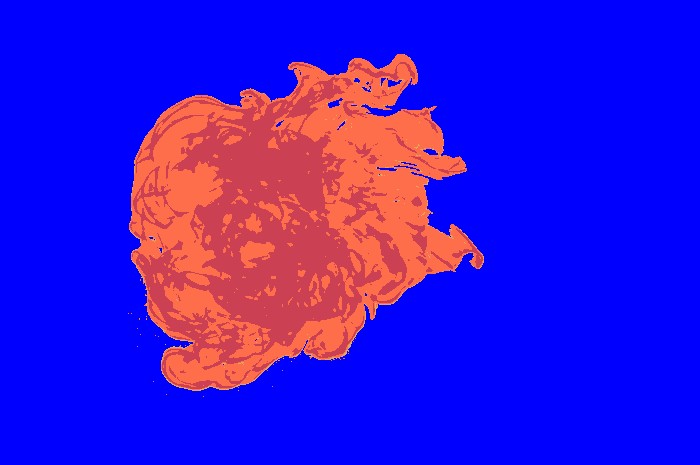}}
				{\includegraphics[width=0.3\linewidth]{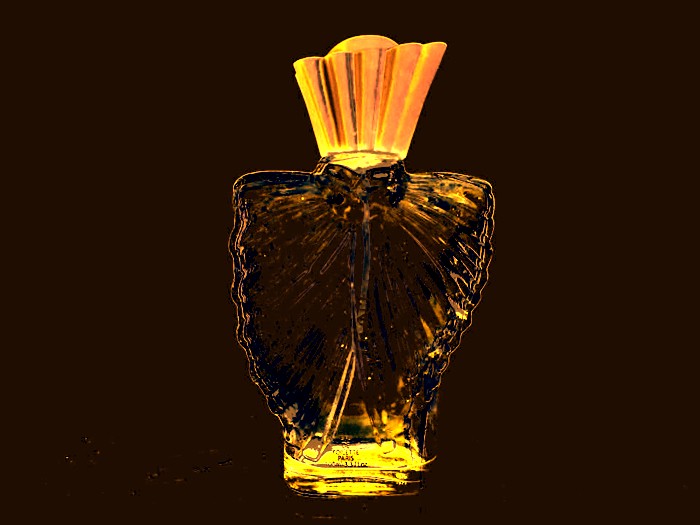}}
			\end{minipage}\label{fig:numofbs:15}}
	\end{minipage}
	\caption{Stylized images with different numbers of color bases $k$ with and small sparse constraint $\alpha=0.001$. In (d), (e), and (f), from left to right: MCIM of the content (left) and style (middle) images, and the resulting stylized image with the corresponding $k$ value.}
	\label{fig:numofbs}
\vspace{-5mm}
\end{figure}
\textcolor{black}{Generally speaking, to extract accurate color bases, the number of bases $k$ should be large enough to encompass the different contexts  so as to reconstruct the RGB images with high fidelity. Large $k$ also allows more flexible local color transfer for different contexts. In general, $k=10$ works well for both datasets adopted in this paper. However, if there is less number of colors in the image pair and the sparse constraint is set to a small value, \eg, $\alpha=0.001$, we find that the stylization is more effective when $k$ is smaller, as shown in Fig.~\ref{fig:numofbs}. In this toy example, the number of colors in the image pair is small, thus, $k=5$ is sufficient to extract a set of effective color bases, as shown in Fig.~\ref{fig:numofbs:5}. However, as we gradually increase the $k$ value with $\alpha$ fixed at 0.001, the extracted representations start losing the discriminative power and the matching between representations of the content and style images start to deteriorate, as shown in Fig.~\ref{fig:numofbs:15}. This is because for images with only a few different colors, if $k$ is set to a large value, without effective sparse constraints, it tends to learn duplicated color bases. Thus, the representations will not be context-sensitive, which would affect the style transfer. Therefore, in this case, we can either increase $\alpha$ or decrease $k$ for an effective transfer. %When we gradually reduce $k$, it is easier to extract context-sensitive representations, as shown in Fig.~\ref{fig:numofbs:5}. 
It is worth mentioning that, even with $k=15$, the stylized results from the proposed method is still more effective than that from the global-based method, as shown in Fig.~\ref{fig:numofbs:global}.}

\section{Conclusion and Future Work}
\label{sec:conclusion}
\textcolor{black}{To tackle the problem of photorealistic style transfer, we proposed a non-local representation scheme realized with a mutual affine-transfer network (NL-MAT).  To the best of our knowledge, this work represents the first attempt to address the photorealistic style transfer problem through a non-local representation model. The proposed scheme successfully decouples the image pairs into non-local representations and color information, with a stick-breaking encoder and an affine-transfer decoder. By enforcing the sparsity  with the entropy function and representation correspondence with the mutual discriminative network, the method is able to extract context-sensitive and matched representations. This largely facilitates context-correspondent local style transfer in a global-consistent fashion. Experimental results demonstrated that the proposed NL-MAT is able to generate photorealistic photos without abrupt color changes or needing any additional models for segmentation or classification.}
% We proposed a \textcolor{black}{non-local representation scheme realized with mutual affine-transfer network} (NL-MAT) to tackle the problem of photorealistic style transfer. To the best of our knowledge, this work represents the first attempt to address this problem through a \textcolor{black}{non-local representation model}. \textcolor{black}{The proposed scheme successfully decouples the image pairs into non-local representations and color information naturally, with a stick-breaking encoder %, for the extraction of representations, 
% and an affine-transfer decoder.} %embedding the color information of both the content and style images, 
% %such that \textcolor{black}{the context information is incorporated naturally 
% By enforcing the sparsity with the entropy function and correspondence of the representations with the mutual discriminative network, the method is able to extract context-sensitive and matched representations. \textcolor{black}{This} largely \textcolor{black}{facilitates} context-correspondence local style transfer in a global-consistent fashion. Experimental results demonstrated that the proposed NL-MAT is able to generate photorealistic photos without abrupt color changes or needing any \textcolor{black}{additional models for segmentation or classification}. 
% %Without adopting any segmentation models

\textcolor{black}{The proposed scheme works well in most scenarios, even when there are some mismatches between the content and style images, as discussed in Sec. V-C. Nonetheless, NL-MAT does have its limitations and may fail on some challenging image pairs. For example, if the distributions of the objects in the content and style images are very different, it could result in large semantic mismatching. Another type of typical failure cases may occur when different semantic contexts in the image actually possess very similar color. The failure cases are further discussed in the supplementary file (Sec. 3). In our future work, we will exploit the usage of prior knowledge serving as additional physical constraints to regulate the learning process and enforce semantic matching.}

\section*{Acknowledgment}
The authors would like to thank all the developers of the evaluated methods who kindly offer their codes or results, \textcolor{black}{and the anonymous reviewers who have helped us greatly in improving the quality of this paper.}

% Can use something like this to put references on a page
% by themselves when using endfloat and the captionsoff option.
%\ifCLASSOPTIONcaptionsoff
%  \newpage
%\fi

% trigger a \newpage just before the given reference
% number - used to balance the columns on the last page
% adjust value as needed - may need to be readjusted if
% the document is modified later
%\IEEEtriggeratref{8}
% The "triggered" command can be changed if desired:
%\IEEEtriggercmd{\enlargethispage{-5in}}

% references section

% The IEEEtran BibTeX style support page is at:
% http://www.michaelshell.org/tex/ieeetran/bibtex/
\bibliographystyle{IEEEtran}
\bibliography{ref}

% Generated by IEEEtran.bst, version: 1.14 (2015/08/26)
\begin{thebibliography}{10}
\providecommand{\url}[1]{#1}
\csname url@samestyle\endcsname
\providecommand{\newblock}{\relax}
\providecommand{\bibinfo}[2]{#2}
\providecommand{\BIBentrySTDinterwordspacing}{\spaceskip=0pt\relax}
\providecommand{\BIBentryALTinterwordstretchfactor}{4}
\providecommand{\BIBentryALTinterwordspacing}{\spaceskip=\fontdimen2\font plus
\BIBentryALTinterwordstretchfactor\fontdimen3\font minus
  \fontdimen4\font\relax}
\providecommand{\BIBforeignlanguage}[2]{{%
\expandafter\ifx\csname l@#1\endcsname\relax
\typeout{** WARNING: IEEEtran.bst: No hyphenation pattern has been}%
\typeout{** loaded for the language `#1'. Using the pattern for}%
\typeout{** the default language instead.}%
\else
\language=\csname l@#1\endcsname
\fi
#2}}
\providecommand{\BIBdecl}{\relax}
\BIBdecl

\bibitem{luan2017deep}
F.~Luan, S.~Paris, E.~Shechtman, and K.~Bala, ``Deep photo style transfer,''
  \emph{Proceedings of the IEEE Conference on Computer Vision and Pattern
  Recognition (CVPR)}, pp. 4990--4998, 2017.

\bibitem{mechrez2017photorealistic}
R.~Mechrez, E.~Shechtman, and L.~Zelnik-Manor, ``Photorealistic style transfer
  with screened poisson equation,'' \emph{arXiv preprint arXiv:1709.09828},
  2017.

\bibitem{li2018closed}
Y.~Li, M.-Y. Liu, X.~Li, M.-H. Yang, and J.~Kautz, ``A closed-form solution to
  photorealistic image stylization,'' \emph{Proceedings of the European
  Conference on Computer Vision (ECCV)}, pp. 453--468, 2018.

\bibitem{gatys2016image}
L.~A. Gatys, A.~S. Ecker, and M.~Bethge, ``Image style transfer using
  convolutional neural networks,'' \emph{Proceedings of the IEEE Conference on
  Computer Vision and Pattern Recognition (CVPR)}, pp. 2414--2423, 2016.

\bibitem{li2017universal}
Y.~Li, C.~Fang, J.~Yang, Z.~Wang, X.~Lu, and M.-H. Yang, ``Universal style
  transfer via feature transforms,'' \emph{Advances in neural information
  processing systems}, pp. 386--396, 2017.

\bibitem{yoo2019photorealistic}
J.~Yoo, Y.~Uh, S.~Chun, B.~Kang, and J.-W. Ha, ``Photorealistic style transfer
  via wavelet transforms,'' \emph{Proceedings of the IEEE International
  Conference on Computer Vision (ICCV)}, pp. 9036--9045, 2019.

\bibitem{reinhard2001color}
E.~Reinhard, M.~Adhikhmin, B.~Gooch, and P.~Shirley, ``Color transfer between
  images,'' \emph{IEEE Computer graphics and applications}, vol.~21, no.~5, pp.
  34--41, 2001.

\bibitem{pitie2005n}
F.~Pitie, A.~C. Kokaram, and R.~Dahyot, ``N-dimensional probability density
  function transfer and its application to color transfer,'' \emph{Tenth IEEE
  International Conference on Computer Vision (ICCV)}, vol.~2, pp. 1434--1439,
  2005.

\bibitem{li2018learning}
X.~Li, S.~Liu, J.~Kautz, and M.-H. Yang, ``Learning linear transformations for
  fast arbitrary style transfer,'' \emph{Proceedings of the IEEE Conference on
  Computer Vision and Pattern Recognition (CVPR)}, 2019.

\bibitem{liao2017visual}
J.~Liao, Y.~Yao, L.~Yuan, G.~Hua, and S.~B. Kang, ``Visual attribute transfer
  through deep image analogy,'' \emph{arXiv preprint arXiv:1705.01088}, 2017.

\bibitem{he2019progressive}
M.~He, J.~Liao, D.~Chen, L.~Yuan, and P.~V. Sander, ``Progressive color
  transfer with dense semantic correspondences,'' \emph{ACM Transactions on
  Graphics (TOG)}, vol.~38, no.~2, p.~13, 2019.

\bibitem{an2020ultrafast}
J.~An, H.~Xiong, J.~Huan, and J.~Luo, ``Ultrafast photorealistic style transfer
  via neural architecture search,'' \emph{Proceedings of the AAAI Conference on
  Artificial Intelligence}, vol.~34, no.~07, pp. 10\,443--10\,450, 2020.

\bibitem{bae2006two}
S.~Bae, S.~Paris, and F.~Durand, ``Two-scale tone management for photographic
  look,'' vol.~25, no.~3, pp. 637--645, 2006.

\bibitem{pitie2007automated}
F.~Piti{\'e}, A.~C. Kokaram, and R.~Dahyot, ``Automated colour grading using
  colour distribution transfer,'' \emph{Computer Vision and Image
  Understanding}, vol. 107, no. 1-2, pp. 123--137, 2007.

\bibitem{freedman2010object}
D.~Freedman and P.~Kisilev, ``Object-to-object color transfer: Optimal flows
  and smsp transformations,'' \emph{2010 IEEE Computer Society Conference on
  Computer Vision and Pattern Recognition}, pp. 287--294, 2010.

\bibitem{farbman2008edge}
Z.~Farbman, R.~Fattal, D.~Lischinski, and R.~Szeliski, ``Edge-preserving
  decompositions for multi-scale tone and detail manipulation,'' vol.~27,
  no.~3, p.~67, 2008.

\bibitem{chen2017deeplab}
L.-C. Chen, G.~Papandreou, I.~Kokkinos, K.~Murphy, and A.~L. Yuille, ``Deeplab:
  Semantic image segmentation with deep convolutional nets, atrous convolution,
  and fully connected crfs,'' \emph{IEEE Transactions on Pattern Analysis and
  Machine Intelligence}, vol.~40, no.~4, pp. 834--848, 2017.

\bibitem{omer2004color}
I.~Omer and M.~Werman, ``Color lines: image specific color representation,''
  \emph{Proceedings of IEEE computer society conference on Computer vision and
  pattern recognition (CVPR)}, pp. 946--953, 2004.

\bibitem{laffont2012coherent}
P.-Y. Laffont, A.~Bousseau, S.~Paris, F.~Durand, and G.~Drettakis, ``Coherent
  intrinsic images from photo collections,'' \emph{ACM Transactions on
  Graphics}, vol.~31, no.~6, 2012.

\bibitem{2010Non}
J.~Mairal, F.~Bach, J.~Ponce, G.~Sapiro, and A.~Zisserman, ``Non-local sparse
  models for image restoration,'' \emph{2009 IEEE 12th International Conference
  on Computer Vision (ICCV)}, 2010.

\bibitem{sethuraman1994constructive}
J.~Sethuraman, ``A constructive definition of dirichlet priors,''
  \emph{Statistica sinica}, pp. 639--650, 1994.

\bibitem{nalisnick2016deep}
E.~Nalisnick and P.~Smyth, ``Deep generative models with stick-breaking
  priors,'' \emph{ICML}, 2017.

\bibitem{qu2018unsupervised}
Y.~Qu, H.~Qi, and C.~Kwan, ``Unsupervised sparse dirichlet-net for
  hyperspectral image super-resolution,'' \emph{The IEEE Conference on Computer
  Vision and Pattern Recognition (CVPR)}, pp. 2511--2520, 2018.

\bibitem{kumaraswamy1980generalized}
P.~Kumaraswamy, ``A generalized probability density function for double-bounded
  random processes,'' \emph{Journal of Hydrology}, vol.~46, no. 1-2, pp.
  79--88, 1980.

\bibitem{dugas2001incorporating}
C.~Dugas, Y.~Bengio, F.~B{\'e}lisle, C.~Nadeau, and R.~Garcia, ``Incorporating
  second-order functional knowledge for better option pricing,'' \emph{Advances
  in neural information processing systems}, pp. 472--478, 2001.

\bibitem{han1995influence}
J.~Han and C.~Moraga, ``The influence of the sigmoid function parameters on the
  speed of backpropagation learning,'' \emph{From Natural to Artificial Neural
  Computation}, pp. 195--201, 1995.

\bibitem{Goodfellow-et-al-2016}
I.~Goodfellow, Y.~Bengio, and A.~Courville, \emph{Deep Learning}.\hskip 1em
  plus 0.5em minus 0.4em\relax MIT Press, 2016,
  \url{http://www.deeplearningbook.org}.

\bibitem{huang2017sparse}
S.~Huang and T.~D. Tran, ``Sparse signal recovery via generalized entropy
  functions minimization,'' \emph{arXiv preprint arXiv:1703.10556}, 2017.

\bibitem{zitova2003image}
B.~Zitova and J.~Flusser, ``Image registration methods: a survey,'' \emph{Image
  and vision computing}, vol.~21, no.~11, pp. 977--1000, 2003.

\bibitem{woo2015multimodal}
J.~Woo, M.~Stone, and J.~L. Prince, ``Multimodal registration via mutual
  information incorporating geometric and spatial context,'' \emph{IEEE
  Transactions on Image Processing}, vol.~24, no.~2, pp. 757--769, 2015.

\bibitem{belghazi2018mine}
I.~Belghazi, S.~Rajeswar, A.~Baratin, R.~D. Hjelm, and A.~Courville, ``Mine:
  mutual information neural estimation,'' \emph{arXiv preprint
  arXiv:1801.04062}, 2018.

\bibitem{donsker1983asymptotic}
M.~D. Donsker and S.~S. Varadhan, ``Asymptotic evaluation of certain markov
  process expectations for large time. iv,'' \emph{Communications on Pure and
  Applied Mathematics}, vol.~36, no.~2, pp. 183--212, 1983.

\bibitem{hjelm2018learning}
R.~D. Hjelm, A.~Fedorov, S.~Lavoie-Marchildon, K.~Grewal, A.~Trischler, and
  Y.~Bengio, ``Learning deep representations by mutual information estimation
  and maximization,'' \emph{arXiv preprint arXiv:1808.06670}, 2018.

\bibitem{nie2010efficient}
F.~Nie, H.~Huang, X.~Cai, and C.~H. Ding, ``Efficient and robust feature
  selection via joint ℓ2, 1-norms minimization,'' \emph{Advances in neural
  information processing systems}, pp. 1813--1821, 2010.

\end{thebibliography}

% biography section
% 
% If you have an EPS/PDF photo (graphicx package needed) extra braces are
% needed around the contents of the optional argument to biography to prevent
% the LaTeX parser from getting confused when it sees the complicated
% \includegraphics command within an optional argument. (You could create
% your own custom macro containing the \includegraphics command to make things
% simpler here.)
\begin{IEEEbiography}[{\includegraphics[width=1in,keepaspectratio]{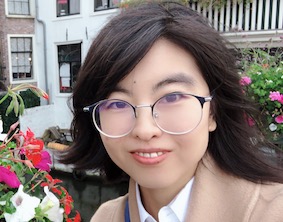}}]{Ying Qu} (IEEE Member since 2016)
	received the B.S. degree in automatics and M.S. degree in pattern recognition \& artificial intelligence from Northeastern University, Shenyang, China in 2008 and 2010, respectively, and the Ph.D. degree in computer engineering from the University of Tennessee, Knoxville. She is currently working as a research associate in the Department of Electrical Engineering and Computer Science at the University of Tennessee, Knoxville. She was the recipient of the IEEE MIKIO Takagi Student Prize (Best Student Paper Awards) at the International Geoscience and Remote Sensing Symposium (IGARSS) in 2016. Her research interests include computer vision, remote sensing and artificial intelligence.
\end{IEEEbiography}

\begin{IEEEbiography}[{\includegraphics[width=1in,keepaspectratio]{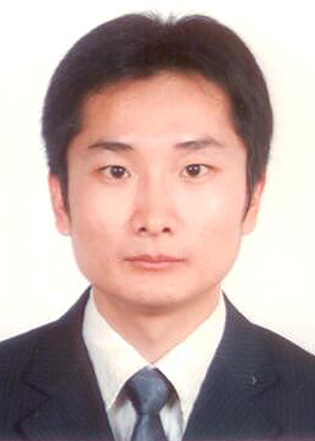}}]{Zhenzhou Shao} (IEEE Member since 2010)
	received the B.E. degree and M.E. degree in the Department of Information Engineering at Northeastern University, China, in 2007 and 2009, respectively, and the Ph.D. degree in mechanical engineering %The Department of Mechanical, Aerospace, and Biomedical Engineering 
	at the University of Tennessee, Knoxville, in 2013. He is currently the associate professor with the College of Information Engineering at Capital Normal University, China. His research interests include computer vision, machine learning and human-robot interaction.
\end{IEEEbiography}

\begin{IEEEbiography}[{\includegraphics[width=1in,keepaspectratio]{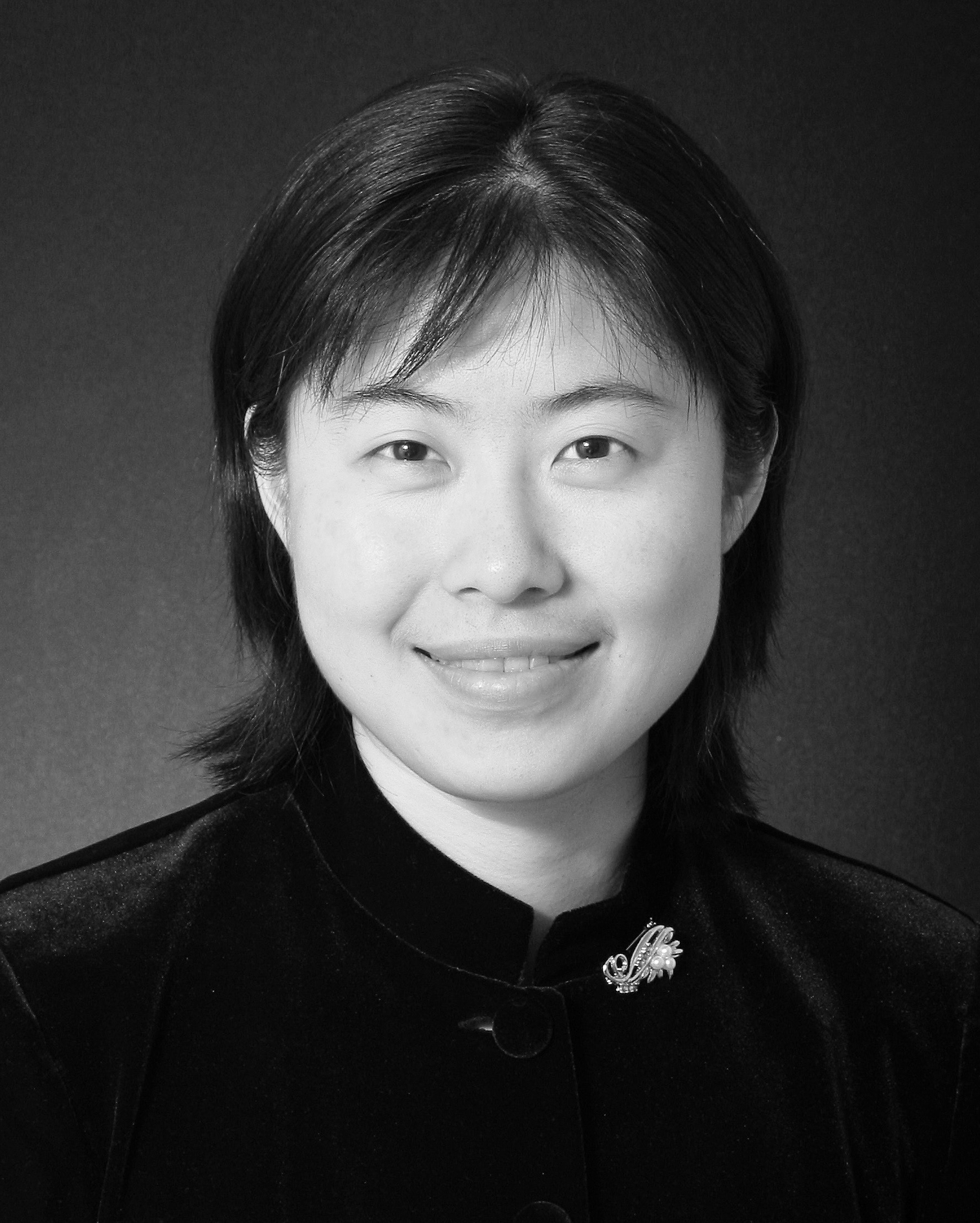}}]{Hairong Qi} (IEEE Fellow since 2017) received the B.S. and M.S. degrees in computer science from Northern JiaoTong University, Beijing, China in 1992 and 1995, respectively, and the Ph.D. degree in computer engineering from North Carolina State University, Raleigh, in 1999. She is currently the Gonzalez Family Professor with the Department of Electrical Engineering and Computer Science at the University of Tennessee, Knoxville. Her current research interests are in advanced imaging and collaborative processing in resource-constrained distributed environment, hyperspectral image analysis, and automatic target recognition. Dr. Qi's research is supported by National Science Foundation (NSF), DARPA, Office of Naval Research (ONR), Department of Homeland Security (DHS), U.S. Army Space and Missile Defense Command, and U.S. Army Medical Research and Materiel Command. Dr. Qi is the recipient of the NSF CAREER Award. She also received the Best Paper Awards at the 18th International Conference on Pattern Recognition (ICPR) in 2006, the 3rd ACM/IEEE International Conference on Distributed Smart Cameras (ICDSC) in 2009, and IEEE Workshop on Hyperspectral Image and Signal Processing: Evolution in Remote Sensor (WHISPERS) in 2015. She is awarded the Highest Impact Paper from the IEEE Geoscience and Remote Sensing Society in 2012.
\end{IEEEbiography}

%\begin{IEEEbiography}[{\includegraphics[width=1in,height=1.25in,clip,keepaspectratio]{mshell}}]{Michael Shell}
% or if you just want to reserve a space for a photo:

%\begin{IEEEbiography}{Michael Shell}
%Biography text here.
%\end{IEEEbiography}
%
%% if you will not have a photo at all:
%\begin{IEEEbiographynophoto}{John Doe}
%Biography text here.
%\end{IEEEbiographynophoto}

% insert where needed to balance the two columns on the last page with
% biographies
%\newpage

%\begin{IEEEbiographynophoto}{Jane Doe}
%Biography text here.
%\end{IEEEbiographynophoto}

%\vfill

% Can be used to pull up biographies so that the bottom of the last one
% is flush with the other column.
%\enlargethispage{-5in}
\end{document}